\documentclass[10pt,twocolumn,letterpaper]{article}
\usepackage[pagenumbers]{cvpr}

\makeatletter
\providecommand{\@LN@col}[1]{}
\makeatother

\usepackage{xspace}
\ifdefined\DeclareUnicodeCharacter
  \DeclareUnicodeCharacter{00B1}{\ensuremath{\pm}}
\fi
\usepackage{multirow}
\usepackage{placeins}
\newcommand{\method}{PhysDec\xspace}
\newcommand{\chamfer}{CD-$L_1$\xspace}

\makeatletter\@namedef{ver@everyshi.sty}{}\makeatother
\usepackage{tikz}
\usetikzlibrary{patterns,arrows.meta,calc}
\usepackage{pifont}
\newcommand{\cmark}{\ding{51}}
\newcommand{\xmark}{\ding{55}}
\definecolor{inkc}{RGB}{40,44,52}
\definecolor{partc}{RGB}{150,160,176}
\definecolor{passc}{RGB}{35,120,74}
\definecolor{failc}{RGB}{190,52,40}

\makeatletter
\renewcommand\LenToUnit[1]{\strip@pt\dimexpr#1\relax}
\makeatother

\usepackage[T1]{fontenc} 
\usepackage{textcomp}
\usepackage{listings}
\usepackage{longtable}
\usepackage{needspace}
\usepackage{array}
\usepackage{float}

\definecolor{annotc}{RGB}{20,90,170}
\definecolor{oldc}{RGB}{200,70,0}
\definecolor{newc}{RGB}{0,110,150}
\definecolor{evoc}{RGB}{112,72,160}
\definecolor{libc}{RGB}{22,128,112}
\definecolor{promptbg}{RGB}{247,247,249}
\newcommand{\oldw}[1]{{\color{oldc}\bfseries #1}}
\newcommand{\neww}[1]{{\color{newc}\bfseries #1}}

\lstdefinestyle{prompt}{
  basicstyle=\ttfamily\fontsize{7.5}{9.1}\selectfont,
  columns=fixed, basewidth=0.525em, keepspaces=true, upquote=true, showstringspaces=false,
  breaklines=true, breakatwhitespace=true, breakindent=0pt,
  frame=single, framerule=0pt, framesep=3pt, backgroundcolor=\color{promptbg},
  xleftmargin=3pt, xrightmargin=3pt, aboveskip=6pt, belowskip=8pt,
  captionpos=t, abovecaptionskip=3pt, belowcaptionskip=3pt,
  moredelim=[is][\color{annotc}\itshape]{@@}{@@},
}
\pretocmd{\lstinputlisting}{\Needspace{5\baselineskip}}{}{}

\AtBeginDocument{}
\newcolumntype{L}[1]{>{\raggedright\arraybackslash}p{#1}}

\AtBeginDocument{%
\crefname{section}{Sec.}{Secs.}
\Crefname{section}{Section}{Sections}
\crefname{table}{Tab.}{Tabs.}
\Crefname{table}{Table}{Tables}
\crefname{listing}{Prompt}{Prompts}
\Crefname{listing}{Prompt}{Prompts}}

\usepackage{microtype}
\definecolor{cvprblue}{rgb}{0.21,0.49,0.74}
\usepackage[pagebackref,breaklinks,colorlinks,allcolors=cvprblue]{hyperref}
\hypersetup{pdfauthor={},pdftitle={Decompose, Verify, Evolve: Physically Grounded Shape Decomposition}}
\def\confName{CVPR}
\def\confYear{2027}
\title{Decompose, Verify, Evolve:\\Physically Grounded Shape Decomposition}
\newcommand{\physdecauthors}{%
Wenchao Ma\textsuperscript{1,*}\quad
Rui Yu\textsuperscript{2,*}\quad
Fan Zhu\textsuperscript{2}\quad
Jiachen Liu\textsuperscript{1}\quad
Sharon X. Huang\textsuperscript{1}\\[4pt]
\textsuperscript{1}Pennsylvania State University\quad
\textsuperscript{2}University of Louisville\\[3pt]
{\small \textsuperscript{*}Equal contribution.}\\[3pt]
{\small Project page: \url{https://wenchao-m.github.io/PhysDec/}}
}
\author{\physdecauthors}
\hypersetup{pdfauthor={Wenchao Ma, Rui Yu, Fan Zhu, Jiachen Liu, Sharon X. Huang}}

\def\physdeccombined{1}
\ifdefined\physdeccombined
\newcommand{\mainloop}{\ref{sec:loop}}
\newcommand{\mainverifiers}{\ref{sec:verifiers}}
\newcommand{\mainevolve}{\ref{sec:evolve}}
\newcommand{\mainscore}{\ref{eq:score}}
\newcommand{\mainbaselines}{\ref{tab:baselines}}
\newcommand{\mainevolution}{\ref{tab:evolution}}
\newcommand{\mainablation}{\ref{tab:ablation}}

\newcommand{\mainuserstudy}{\ref{tab:userstudy}}
\newcommand{\mainunseen}{\ref{tab:unseen}}
\else
\newcommand{\mainloop}{3.2}
\newcommand{\mainverifiers}{3.3}
\newcommand{\mainevolve}{3.4}
\newcommand{\mainscore}{2}
\newcommand{\mainbaselines}{1}
\newcommand{\mainevolution}{2}
\newcommand{\mainablation}{5}

\newcommand{\mainuserstudy}{6}
\newcommand{\mainunseen}{3}
\fi

\begin{document}
\twocolumn[{%
\renewcommand\twocolumn[1][]{#1}%
\maketitle

\begin{center}
  \centering
{\setlength{\tabcolsep}{0pt}\renewcommand{\arraystretch}{0}%
\begin{tabular}{@{}c*{3}{@{\hspace{0.005\textwidth}}c}@{\hspace{0.01\textwidth}{\color{inkc!30}\vrule width 0.4pt}\hspace{0.01\textwidth}}c*{3}{@{\hspace{0.005\textwidth}}c}@{}}
  \scriptsize Input & \scriptsize SuperDec~\cite{fedele2025superdec} & \scriptsize SuperFlex~\cite{tavernini2026superflex} & \scriptsize \textbf{Ours} & \scriptsize Input & \scriptsize SuperDec~\cite{fedele2025superdec} & \scriptsize SuperFlex~\cite{tavernini2026superflex} & \scriptsize \textbf{Ours} \\[3pt]
  \makebox[0.0976\textwidth]{\includegraphics[height=0.1300\textwidth]{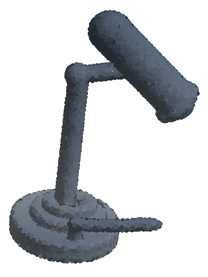}} & \makebox[0.1811\textwidth]{\includegraphics[height=0.1300\textwidth]{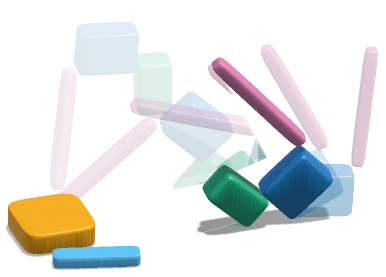}} & \makebox[0.0947\textwidth]{\includegraphics[height=0.1300\textwidth]{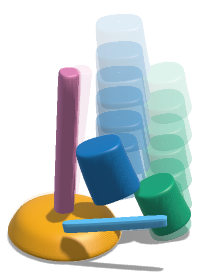}} & \makebox[0.0991\textwidth]{\includegraphics[height=0.1300\textwidth]{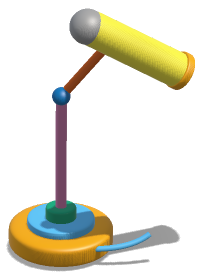}} & \makebox[0.0976\textwidth]{\includegraphics[height=0.1300\textwidth]{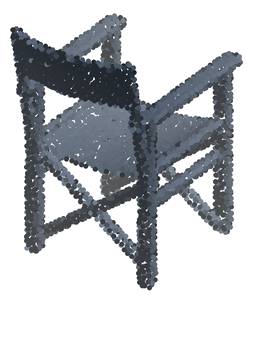}} & \makebox[0.1811\textwidth]{\includegraphics[height=0.1300\textwidth]{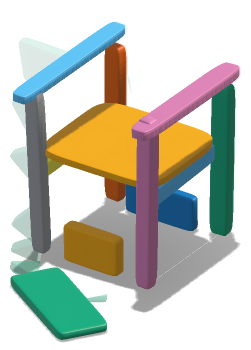}} & \makebox[0.0947\textwidth]{\includegraphics[height=0.1300\textwidth]{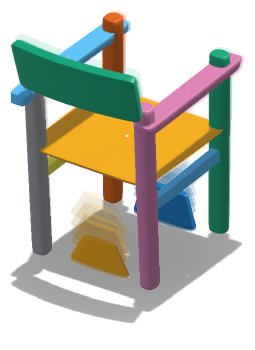}} & \makebox[0.0991\textwidth]{\includegraphics[height=0.1300\textwidth]{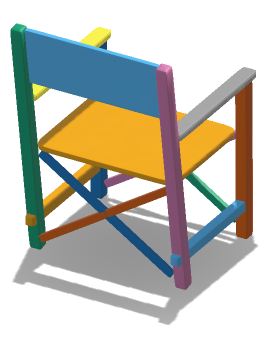}} \\[3pt]
\end{tabular}\par\nointerlineskip%
\begin{tabular}{@{}c*{3}{@{\hspace{0.005\textwidth}}c}@{\hspace{0.01\textwidth}{\color{inkc!30}\vrule width 0.4pt}\hspace{0.01\textwidth}}c*{3}{@{\hspace{0.005\textwidth}}c}@{}}
  \makebox[0.1162\textwidth]{\includegraphics[height=0.1367\textwidth]{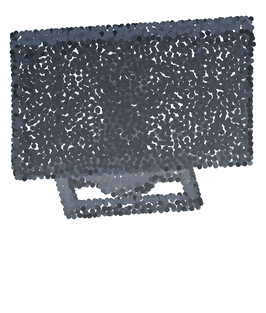}} & \makebox[0.1139\textwidth]{\includegraphics[height=0.1367\textwidth]{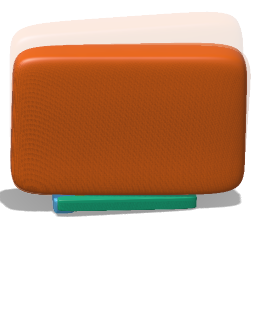}} & \makebox[0.1219\textwidth]{\includegraphics[height=0.1367\textwidth]{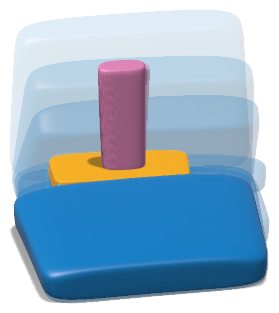}} & \makebox[0.1205\textwidth]{\includegraphics[height=0.1367\textwidth]{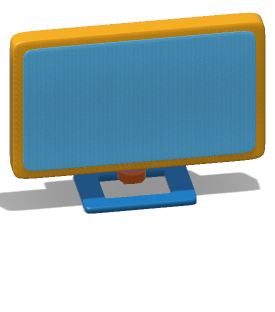}} & \makebox[0.1162\textwidth]{\includegraphics[height=0.1367\textwidth]{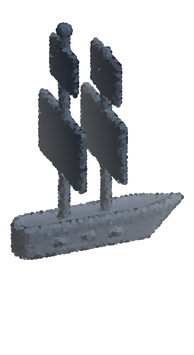}} & \makebox[0.1139\textwidth]{\includegraphics[height=0.1367\textwidth]{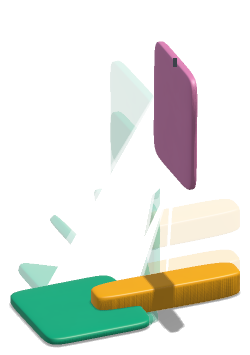}} & \makebox[0.1219\textwidth]{\includegraphics[height=0.1367\textwidth]{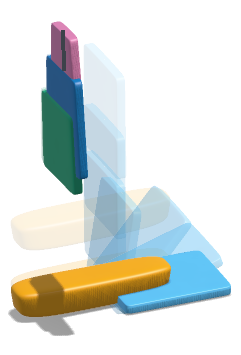}} & \makebox[0.1205\textwidth]{\includegraphics[height=0.1367\textwidth]{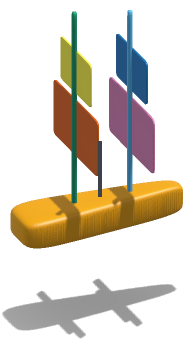}} \\[3pt]
\end{tabular}\par\nointerlineskip%
\begin{tabular}{@{}c*{3}{@{\hspace{0.005\textwidth}}c}@{\hspace{0.01\textwidth}{\color{inkc!30}\vrule width 0.4pt}\hspace{0.01\textwidth}}c*{3}{@{\hspace{0.005\textwidth}}c}@{}}
  \makebox[0.1175\textwidth]{\includegraphics[height=0.0914\textwidth]{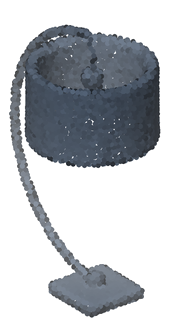}} & \makebox[0.1191\textwidth]{\includegraphics[height=0.0914\textwidth]{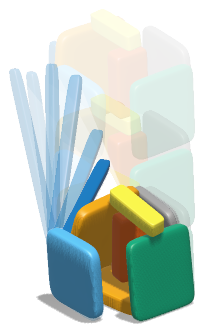}} & \makebox[0.1212\textwidth]{\includegraphics[height=0.0914\textwidth]{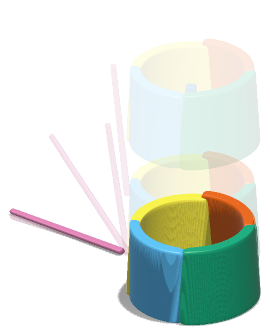}} & \makebox[0.1147\textwidth]{\includegraphics[height=0.0914\textwidth]{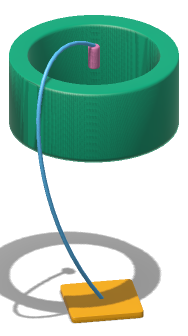}} & \makebox[0.1175\textwidth]{\includegraphics[height=0.0914\textwidth]{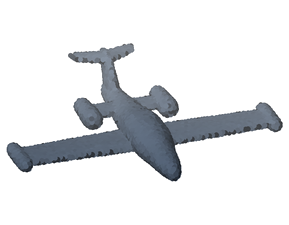}} & \makebox[0.1191\textwidth]{\includegraphics[height=0.0914\textwidth]{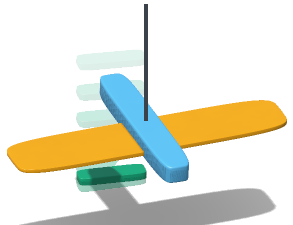}} & \makebox[0.1212\textwidth]{\includegraphics[height=0.0914\textwidth]{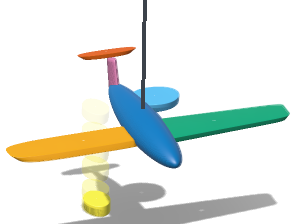}} & \makebox[0.1147\textwidth]{\includegraphics[height=0.0914\textwidth]{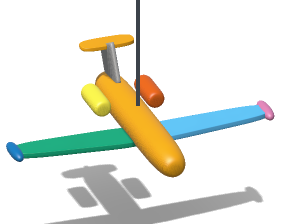}} \\
\end{tabular}}%

  \par
  \captionsetup{hypcap=false}
  \captionof{figure}{We present \textbf{PhysDec}, an agentic framework for physically grounded shape decomposition from images. Compared with prior geometry-focused methods~\cite{fedele2025superdec,tavernini2026superflex}, PhysDec produces physically plausible decompositions whose parts remain connected and stable in simulation. In the images, faint overlays trace motion over $4$\,s in a physics simulator.}
  \label{fig:teaser}
  \captionsetup{hypcap=true}

\end{center}%
}]
\begin{abstract}
Shape decomposition represents 3D objects as a compact set of geometric primitives for interpretable modeling and simulation.
However, existing decomposition methods are trained primarily for geometric fidelity, without explicit constraints on physical plausibility, and can therefore produce disconnected parts or physically unstable reconstructions.
We present \method, an agentic framework that draws on a pretrained vision-language model's semantic understanding and commonsense reasoning to predict material-aware primitive decompositions from a few images.
Within this framework, PhysDec-Loop is an agentic pipeline that reasons about part structure and materials, using a reference 3D point cloud for geometric feedback and physical verifiers to guide iterative correction.
Through self-evolution, we further develop another agentic pipeline, PhysDec-Evolved, which learns from prior corrections to predict primitive decompositions from images without point-cloud feedback or a correction loop at inference.
Experiments show that PhysDec-Loop improves geometric fidelity, physical plausibility and visual quality over geometry-focused baselines, while PhysDec-Evolved approaches its performance at substantially lower inference cost.
\end{abstract}

\section{Introduction}
\label{sec:intro}

Shape decomposition represents a complex object with a small set of interpretable geometric parts, enabling compact storage, editing and simulation.
Recent learned methods predict these representations from images or 3D observations using cuboids, superquadrics and autoregressive primitive representations~\cite{tulsiani2017primitives,paschalidou2019superquadrics,yu2024dpa,fedele2025superdec,ye2025primitiveanything}.
However, their training objectives emphasize geometric fidelity without explicitly enforcing physical plausibility.
As shown in \cref{fig:teaser}, their predictions can resemble the target shape yet contain disconnected parts or become unstable in simulation.

A part-based representation should explain both an object's shape and its physical behavior~\cite{liu2018physical}.
Parts must connect and provide the support needed to keep the object stable, while their geometry and materials determine mass distribution.
For example, a chair leg may closely match the observed surface yet fail to support the seat if their connection is missing.
Therefore, we need to consider part structure, material properties and physical stability together with geometric fidelity.

Pretrained vision-language models (VLMs) provide semantic understanding and commonsense knowledge for reasoning about these coupled structural and material requirements.
Recent work has applied these capabilities to 3D scene creation and image-based reconstruction, using code generation, execution and visual feedback~\cite{hu2024scenecraft,yin2026viga,he2026seig}.
Such workflows can incorporate explicit task constraints through instructions and tools, then revise their outputs in response to execution feedback~\cite{pun2026brickwright}.
These capabilities motivate us to couple VLM reasoning with geometric tools and physical verifiers, so that primitive decomposition is guided by both geometric fidelity and physical plausibility.

We present \method, an agentic framework for physically grounded shape decomposition from images.
We equip a pretrained VLM with tools for generating and rendering material-aware primitive decompositions. An agentic pipeline,
PhysDec-Loop, uses a reference 3D point cloud to measure geometric errors and physical verifiers to assess physical plausibility.
Guided by this feedback, the agent iteratively revises part structure and material choices to satisfy geometric and physical constraints.

Further, we develop another agentic pipeline, PhysDec-Evolved, from PhysDec-Loop to reuse its correction experience.
We first run PhysDec-Loop on training objects with known 3D shapes, collecting verified revisions and successful decompositions.
Self-evolution uses this experience to improve the instructions~\cite{agrawal2026gepa} and build a library of examples and reusable insights~\cite{zhao2023expel}.
PhysDec-Evolved predicts a decomposition directly from images, without point-cloud feedback or an online correction loop.

Experiments show that PhysDec-Loop improves geometric fidelity, physical plausibility and visual quality over geometry-focused baselines. Self-evolution eliminates repeated correction, substantially reducing inference cost while retaining quality comparable to PhysDec-Loop.
Our contributions are:
\begin{itemize}
    \item PhysDec-Loop, an agentic pipeline that couples VLM reasoning with geometric tools and physical verifiers to jointly refine part structure and materials, yielding faithful decompositions whose parts remain connected and stable under simulation.
    \item PhysDec-Evolved, which combines prompts iteratively evolved using verifier feedback with a retrieval library built from PhysDec-Loop's verified refinements. This self-evolution process removes online correction and reduces inference cost without updating VLM weights.
    \item Experiments showing that our method improves geometric fidelity, physical plausibility and visual quality over geometry-focused baselines, with applications to scene reconstruction and robot manipulation.
\end{itemize}

\section{Related Work}
\label{sec:related}

\paragraph{Shape decomposition and programs.}
Cuboid and superquadric decompositions provide compact geometric representations~\cite{barr1981superquadrics,barr1984global,tulsiani2017primitives,paschalidou2019superquadrics}.
Hierarchical prediction~\cite{paschalidou2020hierarchical} and DPA-Net~\cite{yu2024dpa} extend decomposition to single and sparse-view images.
EMS, Marching-Primitives and Light-SQ improve geometric fitting~\cite{liu2022ems,liu2023marching,wang2025lightsq}; SweepNet and SuperFrusta expand primitive expressiveness~\cite{zhao2024sweepnet,ganeshan2025superfrusta}; SuperDec and SuperFlex learn compact decompositions~\cite{fedele2025superdec,tavernini2026superflex}.
Kobsik \etal~\cite{kobsik2026prompting} use masks from pretrained visual models to guide fitting.
Complementing these geometric representations, CSGNet and ShapeAssembly encode Boolean composition and attachment relationships~\cite{sharma2017csgnet,jones2020shapeassembly}, while PrimitiveAnything learns autoregressive assembly from human-crafted decompositions~\cite{ye2025primitiveanything}.
ShapeMOD discovers reusable macros and ShapeLib designs semantic modeling functions with LLMs~\cite{jones2021shapemod,jones2025shapelib}.
Physical Primitive Decomposition jointly estimates geometric and physical attributes~\cite{liu2018physical}, while Gen2Physics recovers material-based mesh components with assigned densities~\cite{comi2026gen2physics}.
SNAP3D uses simulation feedback to refine connections between generated parts~\cite{tuan2026snap3d}.
\method brings geometric and physical verification into a shared refinement loop for material-aware primitive decomposition.

\paragraph{Agentic 3D modeling and experience reuse.}
SceneCraft and Holodeck combine language-based planning with code generation or asset layout optimization~\cite{hu2024scenecraft,yang2023holodeck}.
VIGA, SEIG, aDSL and AgentSTAR use executable representations and revision for reconstruction, structured modeling and tracking~\cite{yin2026viga,he2026seig,wang2026adsl,mazur2026agentstar}, with related benchmarks covering scene authoring and dynamic inverse graphics~\cite{ye2026code4scene,shen2026codebench}.
HARMONY combines hierarchical agentic reasoning with point-cloud-guided geometric refinement for monocular scene reconstruction~\cite{sun2026harmony}.
Physical requirements extend this modeling problem beyond appearance: Make It Stand optimizes balance~\cite{prevost2013stand}, PhyScene constrains scene layouts~\cite{yang2024physcene}, and SceneSmith, Articulate-Anything, CoDimRecon and WorldAgent target simulation-ready scenes and assets~\cite{pfaff2026scenesmith,le2025articulate,xie2026codimrecon,wang2026worldagent}.
PhysAg3D and DiagGen use physical tools and simulation feedback to refine interactive and deformable assets~\cite{du2026physag3d,chen2026diaggen}.
BrickWright and BrickBench address buildability~\cite{pun2026brickwright,kulits2026brickbench}, while FunCAD-Harness tests mechanism functionality~\cite{zhang2026funcad}.

Self-Refine and Reflexion use feedback and reflection~\cite{madaan2023selfrefine,shinn2023reflexion}; ExpeL and Voyager retain transferable experience~\cite{zhao2023expel,wang2023voyager}.
GEPA evolves prompts and Agentic Context Engineering maintains structured playbooks~\cite{agrawal2026gepa,zhang2026ace}.
Text2Sim retrieves distilled simulation expertise during execution-based repair~\cite{xiong2026text2sim}.
SceneEvolve evolves a 3D reconstruction harness with fixed model weights, while PhySGA distills physical trial-and-error into reusable skills~\cite{yao2026sceneevolve,lyu2026physga}.
\method reuses corrections verified by geometric and physical checks as improved prompts, examples and insights. This experience enables direct primitive prediction at inference with fixed VLM weights.

\section{Method}
\label{sec:method}

Given one to four images, \method predicts a compact set of primitives, each with geometry, pose and material attributes.
PhysDec-Loop revises proposals using a reference 3D point cloud for geometric feedback, together with physical checks and visual review.
We run this loop on training objects and reuse its correction experience to improve prompts and build a retrieval library.
The resulting PhysDec-Evolved uses these prompts and retrieved experience to predict decompositions directly from images, without online verification or correction.


\begin{figure}[!htbp]
  \centering
  \resizebox{\linewidth}{!}{\input{figures/primitives_col}}
  \caption{\textbf{Primitive vocabulary.} (a) Supported shapes and geometric operators. (b) An office-chair decomposition, colored by primitive type.}
  \label{fig:primitives}

  \par\vspace{5pt}
  \resizebox{\linewidth}{!}{\input{figures/materials_col}}
  \caption{\textbf{Material-aware mass assignment.} (a) Available materials and their effective densities. (b) The same lamp reconstruction simulated with uniform and material-aware densities. Assigning materials to primitives enables material-dependent mass distributions and physically plausible simulation: with a uniform density the lamp falls over, whereas with the assigned materials it stands.}
  \label{fig:materials}
\end{figure}

\subsection{Representation}
\label{sec:repr}

\paragraph{Primitives.}
\Cref{fig:primitives} summarizes our primitive vocabulary. Following SuperDec~\cite{fedele2025superdec}, a superellipsoid uses eleven parameters: half-extents $\mathbf{s}\in\mathbb{R}^3_{>0}$, shape exponents $(\epsilon_1,\epsilon_2)\in[0.1,1.9]^2$, a rotation $\mathbf{R}$ (intrinsic $XYZ$ Euler angles) and a translation $\mathbf{t}$.
In local coordinates, its interior satisfies
\begin{equation}
F(\mathbf{x}) = \left( \left|\tfrac{x}{s_1}\right|^{\frac{2}{\epsilon_2}} + \left|\tfrac{y}{s_2}\right|^{\frac{2}{\epsilon_2}} \right)^{\frac{\epsilon_2}{\epsilon_1}} + \left|\tfrac{z}{s_3}\right|^{\frac{2}{\epsilon_1}} \le 1 .
\label{eq:sq}
\end{equation}
We also support toroids with a hole ratio and one- or two-sheet hyperboloids with a waist ratio~\cite{barr1981superquadrics}.
All families are normalized so that $\mathbf{s}$ denotes bounding half-extents.
Extensions include tapering and bending~\cite{barr1984global}, tubes with a superellipse cross-section swept along a cubic B\'ezier spine with linear taper, and radial arrays of $n$ copies about a vertical axis. An array occupies one entry in the primitive program.

Each primitive also has a construction form describing its assumed internal structure: \emph{solid} for a filled part, \emph{tube} for a hollow pole or frame member, and \emph{shell} for a thin-walled casing or shade.
This attribute affects mass estimation.

\paragraph{Materials.}
The VLM selects each part's material from a fixed menu of eleven materials (\cref{fig:materials}). No ground-truth material annotations are used.
Each material--construction pair has a fixed effective density that accounts for hollow interiors. We compute each part's simulation mass by multiplying this density by its primitive volume.

\begin{figure*}[!t]
  \centering
  \input{figures/loop}
  \caption{\textbf{PhysDec-Loop.} \textbf{Analyse} identifies the object's structure and materials from images, and \textbf{Propose} generates a primitive decomposition. \textbf{Verify} assesses it using geometric and physical checks and a VLM review. \textbf{Critique} turns these outputs into feedback, which \textbf{Propose} uses to generate the next decomposition. The highest-scoring decomposition that passes the physical checks is selected. Blue marks VLM reasoning; orange marks verification with tools and VLM review.}
  \label{fig:loop}

  \par\vspace{5pt}
  \centering
\setlength{\tabcolsep}{0pt}%
\begin{minipage}[t]{0.235\textwidth}\centering
  {\footnotesize\textbf{One piece} \textnormal{\scriptsize(all)}}\\[2pt]
  \parbox[b][0.47\linewidth][b]{\linewidth}{\centering\begin{tabular}[b]{@{}c@{\hspace{0.0619\linewidth}}c@{}}
    \includegraphics[width=0.3574\linewidth]{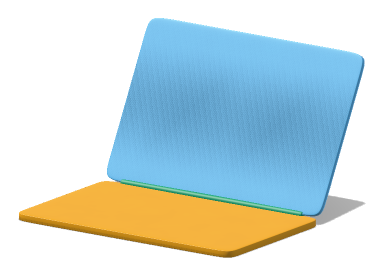} & \includegraphics[width=0.3555\linewidth]{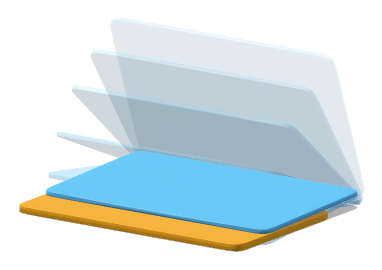} \\
    \scriptsize \textcolor{passc}{\cmark\,one piece} & \scriptsize \textcolor{failc}{\xmark\,screen falls}
  \end{tabular}}
\end{minipage}%
\hfill
\begin{minipage}[t]{0.235\textwidth}\centering
  {\footnotesize\textbf{Stand} \textnormal{\scriptsize(free-standing)}}\\[2pt]
  \parbox[b][0.47\linewidth][b]{\linewidth}{\centering\begin{tabular}[b]{@{}c@{\hspace{0.0619\linewidth}}c@{}}
    \includegraphics[width=0.1223\linewidth]{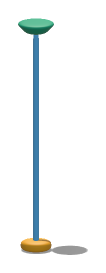} & \includegraphics[width=0.3034\linewidth]{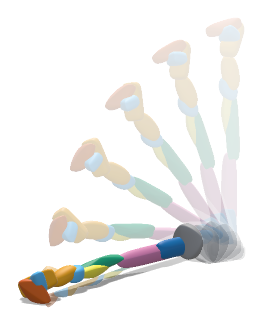} \\
    \scriptsize \textcolor{passc}{\cmark\,stands} & \scriptsize \textcolor{failc}{\xmark\,falls}
  \end{tabular}}
\end{minipage}%
\hfill
\begin{minipage}[t]{0.235\textwidth}\centering
  {\footnotesize\textbf{Wall} \textnormal{\scriptsize(wall-mounted)}}\\[2pt]
  \parbox[b][0.47\linewidth][b]{\linewidth}{\centering\begin{tabular}[b]{@{}c@{\hspace{0.0619\linewidth}}c@{}}
    \includegraphics[width=0.2277\linewidth]{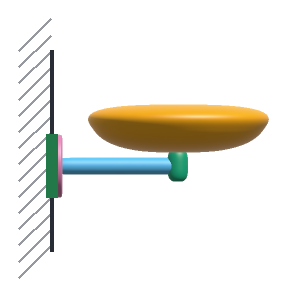} & \includegraphics[width=0.2316\linewidth]{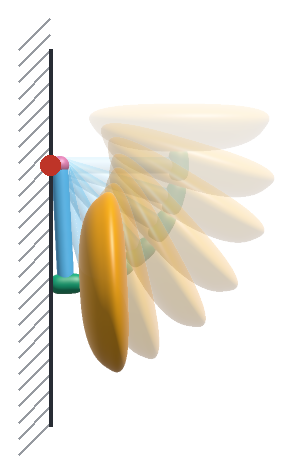} \\
    \scriptsize \textcolor{passc}{\cmark\,a face} & \scriptsize \textcolor{failc}{\xmark\,a point}
  \end{tabular}}
\end{minipage}%
\hfill
\begin{minipage}[t]{0.235\textwidth}\centering
  {\footnotesize\textbf{Hang} \textnormal{\scriptsize(airborne or afloat)}}\\[2pt]
  \parbox[b][0.47\linewidth][b]{\linewidth}{\centering\begin{tabular}[b]{@{}c@{\hspace{0.0619\linewidth}}c@{}}
    \includegraphics[width=0.3063\linewidth]{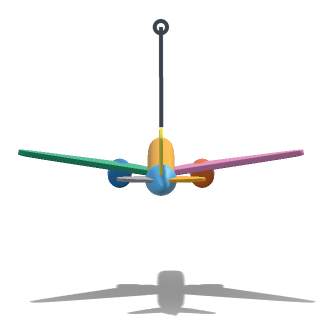} & \includegraphics[width=0.2769\linewidth]{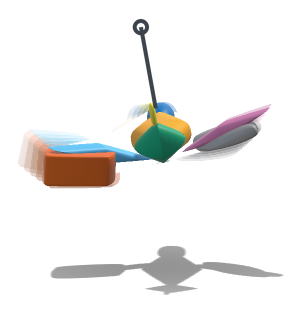} \\
    \scriptsize \textcolor{passc}{\cmark\,level} & \scriptsize \textcolor{failc}{\xmark\,hangs $10^\circ$}
  \end{tabular}}
\end{minipage}

  \caption{\textbf{Physical verification.} The predicted parts must remain connected and satisfy checks for their support type: standing stability, wall contact or suspended balance. Green and red mark passing and failing examples. Wall contact is checked geometrically; the perturbed failure is shown with hinge motion for illustration.}
  \label{fig:physics_checks}
\end{figure*}

\subsection{The self-correction loop}
\label{sec:loop}

PhysDec-Loop follows four stages (\cref{fig:loop}).
\textbf{Analyse} identifies the object's category, support type, symmetry, proportions and parts with their materials.
\textbf{Propose} outputs a primitive decomposition in JSON format.
\textbf{Verify} evaluates the proposal using geometric checks, physical checks and a separate VLM review (\cref{sec:verifiers}).
\textbf{Critique} examines the images and verifier outputs to identify errors and provide revision guidance.
\textbf{Propose} then uses this feedback to generate an updated decomposition, which is verified again.
We repeat this cycle for up to $T{=}3$ correction rounds, with optional early stopping (see supplementary material for full prompts and example feedback).

Each round $r$ receives
\begin{equation}
\begin{split}
S_r = {} & -\tfrac{1}{4}\,\mathrm{CD}_r - w_o \max(0, o_r - 0.05) \\
         & - w_c \max(0, c_r - 1) + w_v \tfrac{v_r - 75}{100},
\end{split}
\label{eq:score}
\end{equation}
where $\mathrm{CD}_r$ is reference-surface Chamfer distance, $o_r$ volumetric overlap, $c_r$ the connected-component count (one for a connected object) and $v_r$ the VLM review score.
We use $w_o{=}0.15$, $w_c{=}0.05$ and $w_v{=}2$.
We select the highest-scoring decomposition that passes the connectivity and physical checks; if none passes, the highest-scoring one.

\subsection{Verifiers \& Critique}
\label{sec:verifiers}

Verification combines geometric measurements, physical checks and visual review to assess each proposal.
We measure Chamfer distance between sampled primitive surfaces and the reference 3D point cloud, and render reference points colored by their distance to the predicted surface to expose missing regions and local shape errors.
Physical checks follow the support type inferred during \textbf{Analyse} (\cref{fig:physics_checks}), since objects that stand, rest on a surface, attach to a wall or hang require different tests.
With material-dependent part masses, MuJoCo~\cite{todorov2012mujoco} tests stability in standing or resting poses and balance under suspension, recording motion and tilt to diagnose unstable or unsupported parts.
Wall-mounted objects are checked geometrically for a rear contact face.
Suspension also provides a balance test for airborne and floating objects.

A separate VLM compares proposal renders with the input images, assessing proportions, part placement and completeness, individual part shape, object identity, structural coherence and material plausibility.
It provides an overall visual score together with criterion-level explanations that identify the relevant parts and views (see supplementary material for the scoring rubric).
\textbf{Critique} combines this review with geometric errors, contact and support diagnostics, rendered error maps and simulation snapshots to identify failures and provide part-specific revision guidance to \textbf{Propose}.

\begin{figure*}[!t]
  \centering
  \input{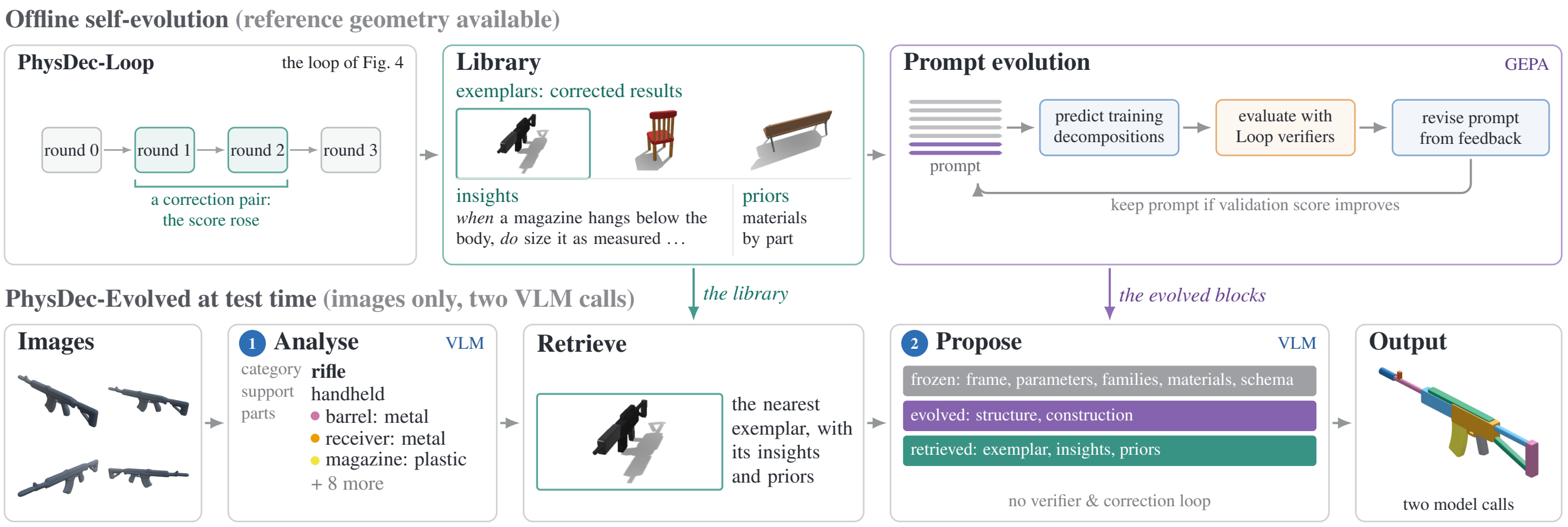}
  \caption{\textbf{From PhysDec-Loop to PhysDec-Evolved.} Offline, PhysDec-Loop provides verified corrections on training objects to build a retrieval library and guide prompt evolution with GEPA. Candidate prompts are evaluated using the same verifiers. At test time, PhysDec-Evolved reuses \textbf{Analyse} and \textbf{Propose} with the evolved prompts and retrieved experience, predicting in two VLM calls without point-cloud feedback or an online correction loop. VLM weights remain fixed throughout.}
  \label{fig:evolution}
\end{figure*}

\subsection{Self-evolution}
\label{sec:evolve}

Self-evolution converts experience from PhysDec-Loop into reusable guidance for PhysDec-Evolved (\cref{fig:evolution}).
Both use the same VLM and primitive representation; PhysDec-Evolved replaces per-object correction with evolved prompts and a retrieval library.
To construct this guidance offline, we run PhysDec-Loop on training objects with reference geometry and record the decompositions and feedback from successive rounds, keeping VLM weights fixed.
When a revision improves $S_r$, we retain the before-and-after pair and describe the changes that led to the improvement.

The library stores this experience in three forms.
\emph{Exemplars} are physically valid decompositions with good geometric fit, providing concrete examples of how to decompose an object.
\emph{Insights} summarize successful corrections as general advice: when a change is useful, what to change and why it helps.
\emph{Material priors} summarize typical material and construction choices for each object category and part role (see supplementary material for library examples and retrieval details).

Alongside the library, we improve the instructions that guide part structure and primitive construction.
We use GEPA~\cite{agrawal2026gepa} to repeat a simple process: predict decompositions for training objects, inspect their verifier feedback, and ask the VLM to rewrite the instructions to address recurring errors.
Each candidate prompt produces decompositions without online correction.
We evaluate these predictions using the same scoring rule and verifiers as PhysDec-Loop (\cref{sec:loop,sec:verifiers}), and retain changes that improve performance on validation objects.
The coordinate frame, primitive definitions, material menu and output schema remain fixed (see supplementary material for prompt blocks before and after evolution).

At deployment, PhysDec-Evolved uses \textbf{Analyse} to identify the object's parts and category in one VLM call.
We filter exemplars by this category and use DINOv2~\cite{oquab2024dinov2} image similarity to retrieve a relevant example, together with applicable insights and material priors.
\textbf{Propose} then predicts the decomposition in a second VLM call using the improved instructions and retrieved context.
The \textbf{Verify}--\textbf{Critique} loop is omitted: no point-cloud feedback, proposal rendering or simulation is required at deployment.

\FloatBarrier
\section{Experiments}
\label{sec:experiments}

We evaluate \method against baseline methods on ShapeNet objects (\cref{sec:exp_baselines,sec:exp_evolution}), partial observations from ASE (\cref{sec:exp_partial}) and Replica scene scans (\cref{sec:scenes}), and test robot manipulation of the results (\cref{sec:robot}).
We also conduct ablation studies to assess the contribution of each component, and a user study to validate the VLM score (\cref{sec:analysis}).

\subsection{Setup}
\label{sec:setup}

\paragraph{Comparisons and metrics.}
We compare against SQ~\cite{paschalidou2019superquadrics}, EMS~\cite{liu2022ems}, Marching-Primitives~\cite{liu2023marching}, SuperDec~\cite{fedele2025superdec} and SuperFlex~\cite{tavernini2026superflex}.
We report Chamfer-$L_1$ ($\times100$), primitive count, physical-check pass rate and VLM visual score (0--100).
For a fairer geometric comparison with baselines that use point clouds or SDFs, we refine the scales and poses of parts predicted from images against reference point clouds before Chamfer evaluation (see supplementary material for details).
Physical validity requires the decomposed object to remain connected and pass its support-specific check (\cref{sec:verifiers}).
Claude Opus 5.5 performs \textbf{Analyse}, \textbf{Propose} and \textbf{Critique} in PhysDec-Loop, \textbf{Analyse} and \textbf{Propose} in PhysDec-Evolved, and prompt rewriting during self-evolution.
Claude Sonnet 5.5 provides the VLM review in \textbf{Verify} and the reported visual scores; GPT-6.1 Sol provides an additional independent visual score in \cref{tab:baselines}.
MuJoCo~\cite{todorov2012mujoco} performs the physical simulations.

\begin{figure*}[!t]
  \centering
  {
\setlength{\tabcolsep}{0pt}\renewcommand{\arraystretch}{0}
\def\gw{\dimexpr(\linewidth-7\dimexpr0.16em\relax)/8\relax}
\begin{tabular}{@{}c*{7}{@{\hspace{0.16em}}c}@{}}
\parbox[b]{\gw}{\centering\fontsize{8}{9}\selectfont\renewcommand{\arraystretch}{1}Input} & \parbox[b]{\gw}{\centering\fontsize{8}{9}\selectfont\renewcommand{\arraystretch}{1}SQ} & \parbox[b]{\gw}{\centering\fontsize{8}{9}\selectfont\renewcommand{\arraystretch}{1}EMS} & \parbox[b]{\gw}{\centering\fontsize{8}{9}\selectfont\renewcommand{\arraystretch}{1}Marching-\\Prim.} & \parbox[b]{\gw}{\centering\fontsize{8}{9}\selectfont\renewcommand{\arraystretch}{1}SuperDec} & \parbox[b]{\gw}{\centering\fontsize{8}{9}\selectfont\renewcommand{\arraystretch}{1}SuperFlex} & \parbox[b]{\gw}{\centering\fontsize{8}{9}\selectfont\renewcommand{\arraystretch}{1}PhysDec\\(Evolved)} & \parbox[b]{\gw}{\centering\fontsize{8}{9}\selectfont\renewcommand{\arraystretch}{1}PhysDec\\(Loop)} \\[0.5ex]
\includegraphics[width=\gw]{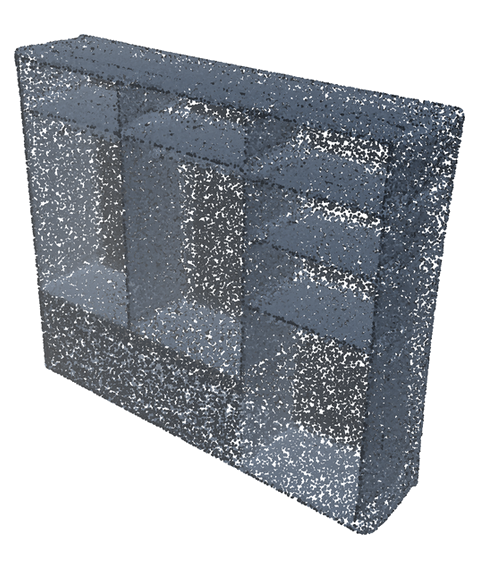} & \includegraphics[width=\gw]{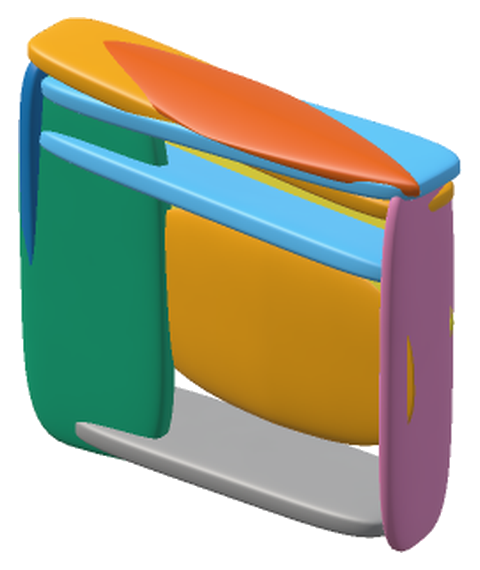} & \includegraphics[width=\gw]{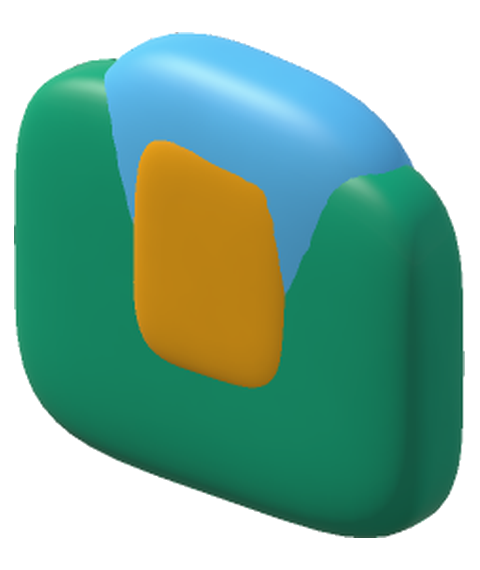} & \includegraphics[width=\gw]{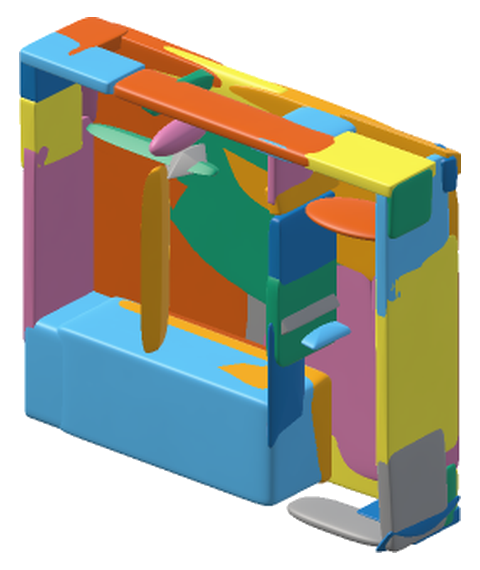} & \includegraphics[width=\gw]{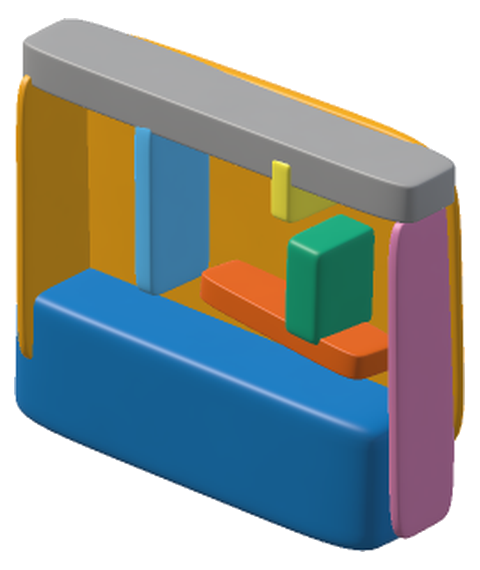} & \includegraphics[width=\gw]{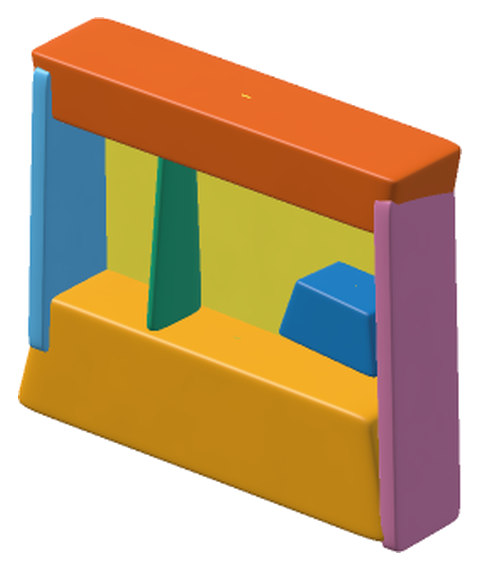} & \includegraphics[width=\gw]{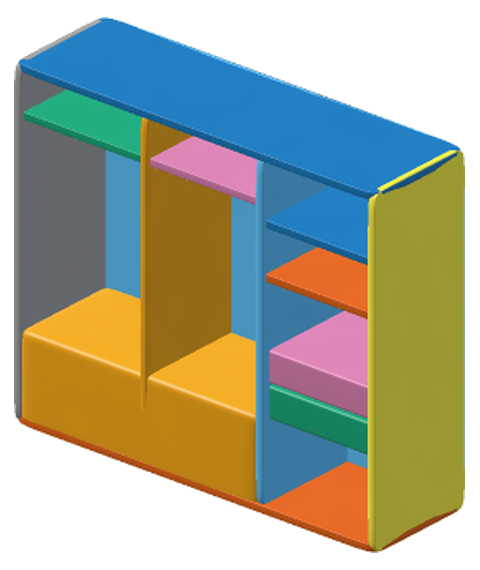} & \includegraphics[width=\gw]{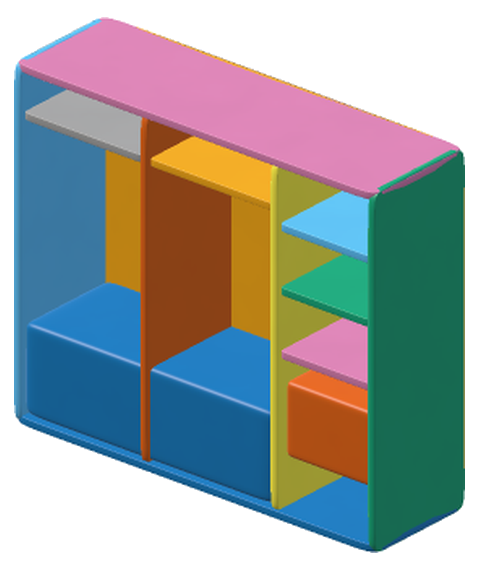} \\[0.16em]
\includegraphics[width=\gw]{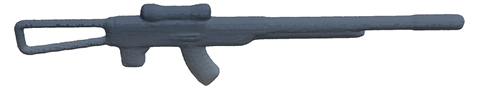} & \includegraphics[width=\gw]{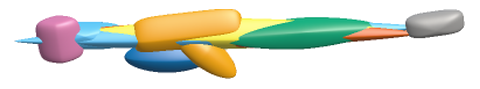} & \includegraphics[width=\gw]{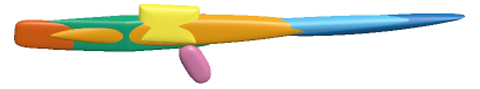} & \includegraphics[width=\gw]{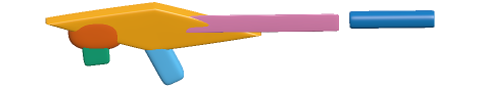} & \includegraphics[width=\gw]{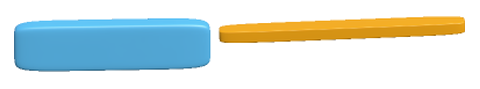} & \includegraphics[width=\gw]{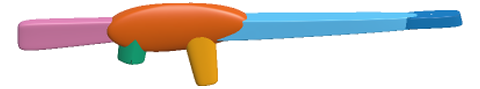} & \includegraphics[width=\gw]{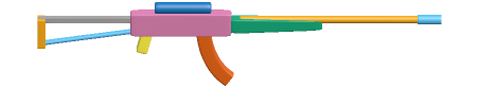} & \includegraphics[width=\gw]{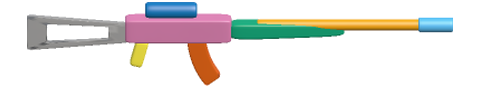} \\[0.16em]
\includegraphics[width=\gw]{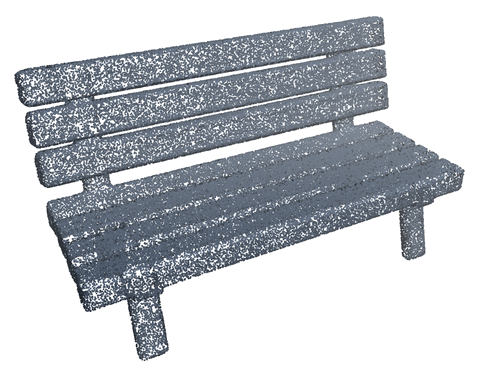} & \includegraphics[width=\gw]{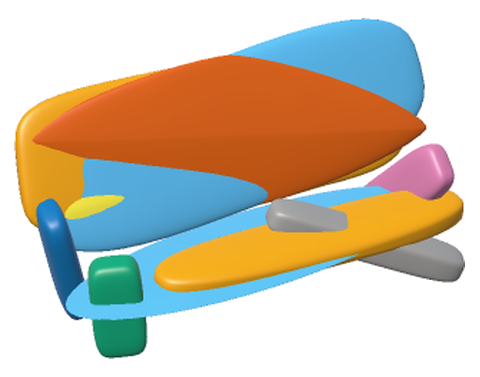} & \includegraphics[width=\gw]{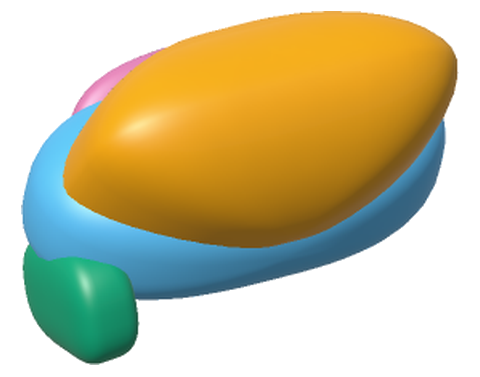} & \includegraphics[width=\gw]{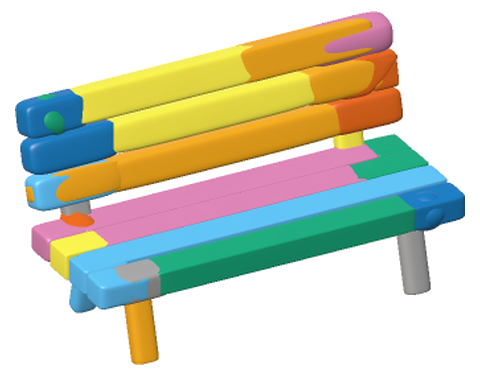} & \includegraphics[width=\gw]{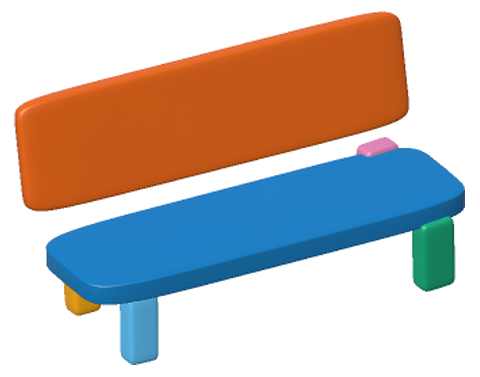} & \includegraphics[width=\gw]{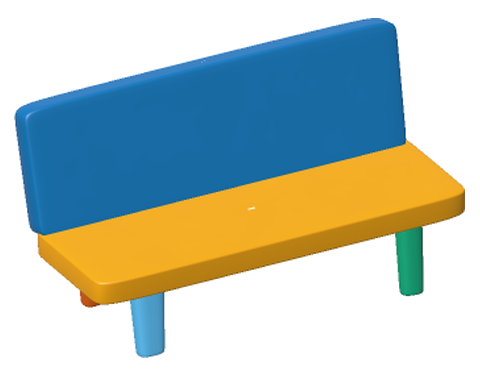} & \includegraphics[width=\gw]{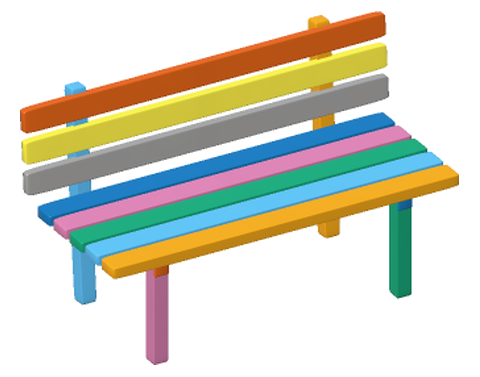} & \includegraphics[width=\gw]{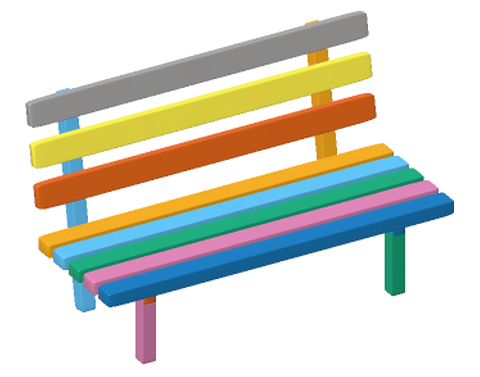} \\[0.16em]
\includegraphics[width=\gw]{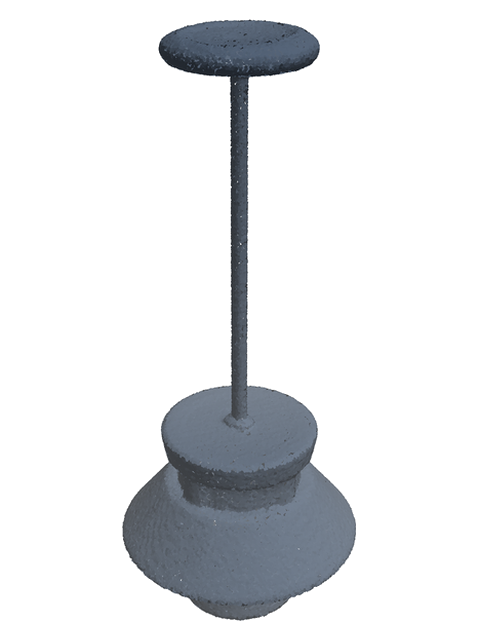} & \includegraphics[width=\gw]{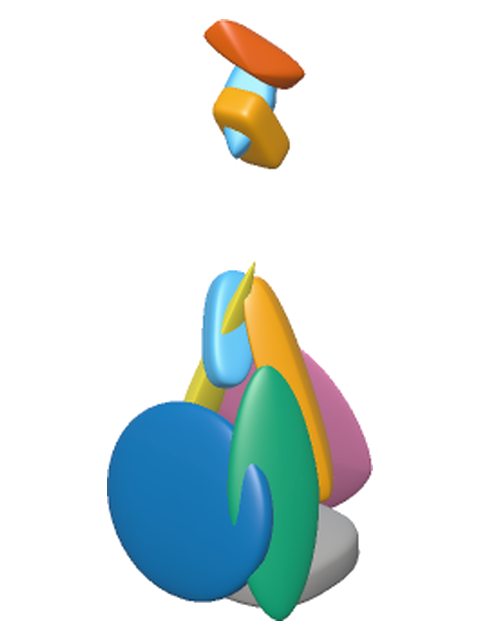} & \includegraphics[width=\gw]{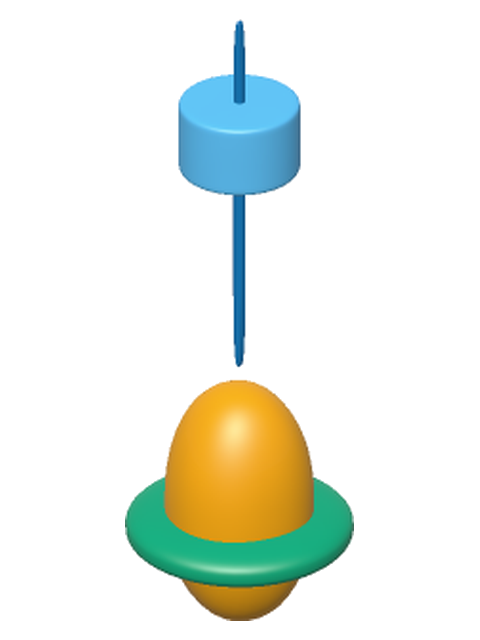} & \includegraphics[width=\gw]{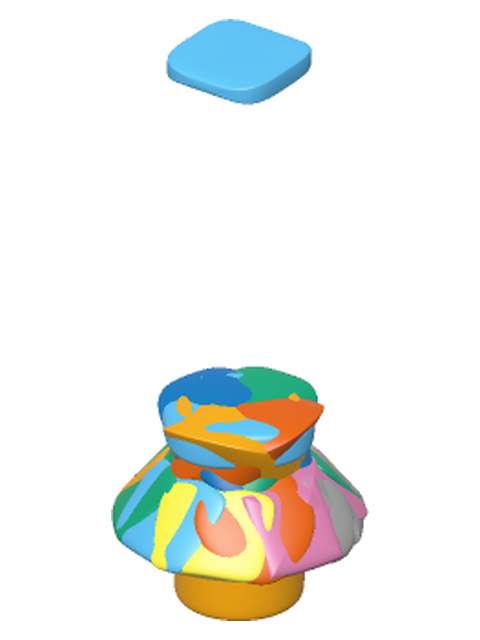} & \includegraphics[width=\gw]{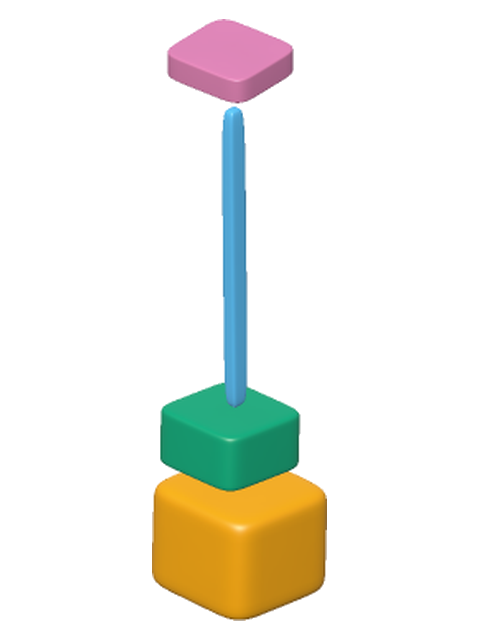} & \includegraphics[width=\gw]{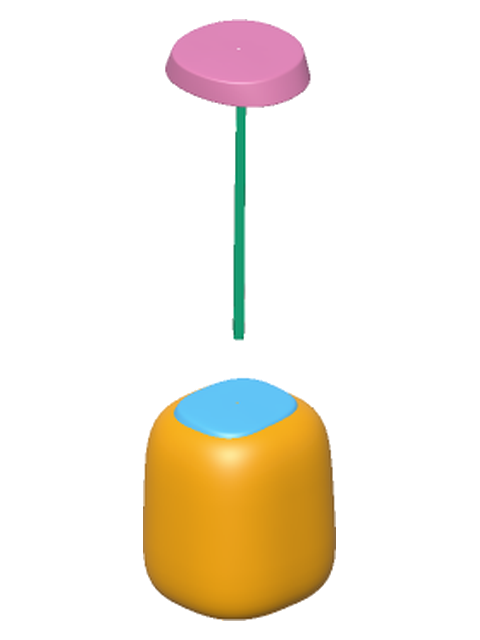} & \includegraphics[width=\gw]{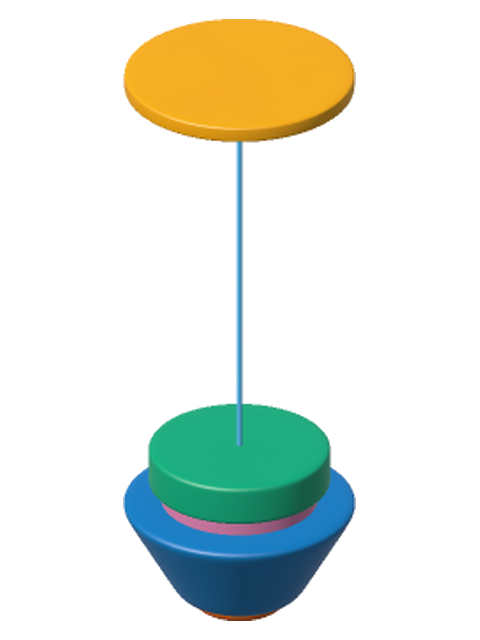} & \includegraphics[width=\gw]{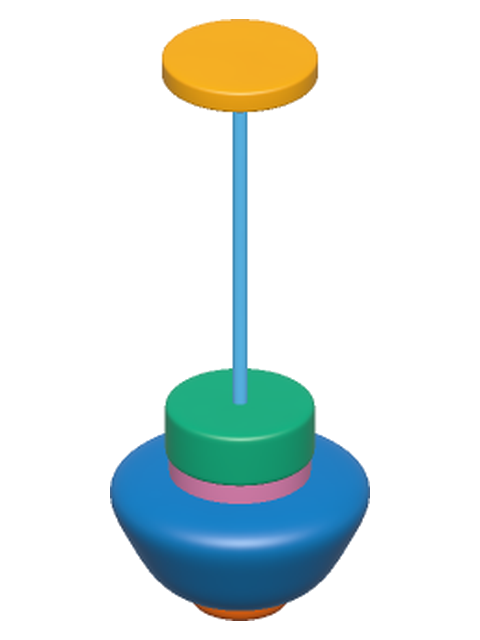} \\[0.16em]
\includegraphics[width=\gw]{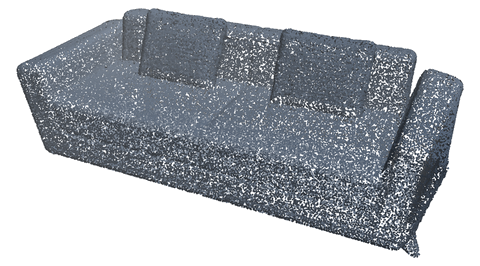} & \includegraphics[width=\gw]{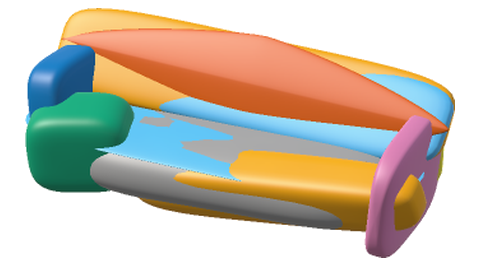} & \includegraphics[width=\gw]{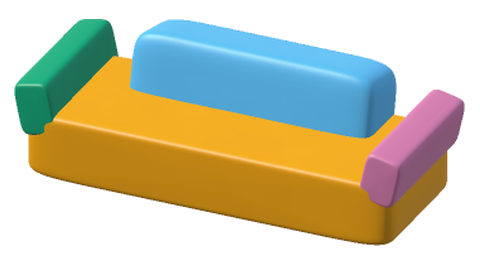} & \includegraphics[width=\gw]{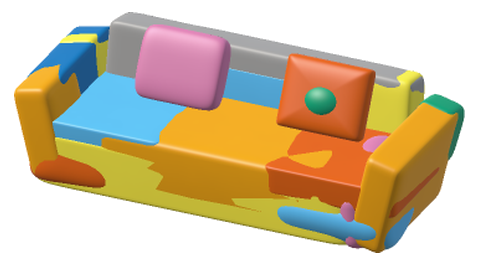} & \includegraphics[width=\gw]{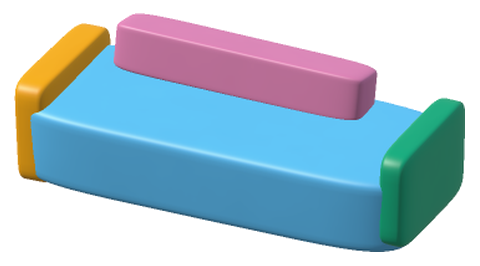} & \includegraphics[width=\gw]{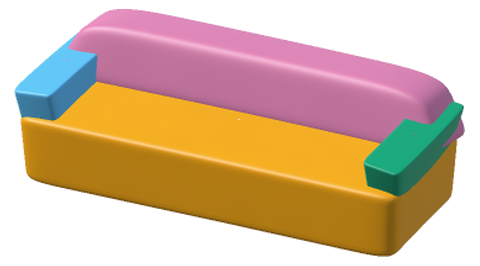} & \includegraphics[width=\gw]{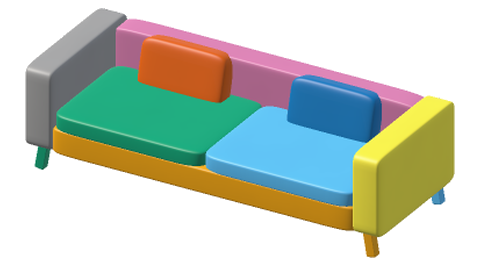} & \includegraphics[width=\gw]{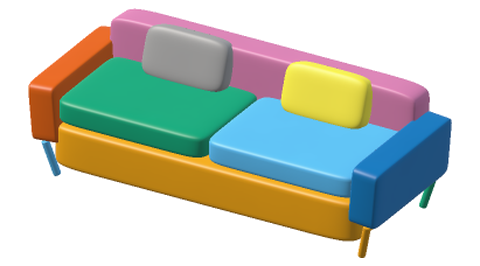} \\[0.16em]
\end{tabular}
}
  \caption{\textbf{Visual comparison.} Primitive decompositions from the baselines, PhysDec-Evolved and PhysDec-Loop.}
  \label{fig:gallery_visual}
\end{figure*}

\begin{figure*}[!t]
  \centering
  {
\setlength{\tabcolsep}{0pt}\renewcommand{\arraystretch}{0}
\def\gw{\dimexpr(\linewidth-7\dimexpr0.16em\relax)/8\relax}
\begin{tabular}{@{}c*{7}{@{\hspace{0.16em}}c}@{}}
\parbox[b]{\gw}{\centering\fontsize{8}{9}\selectfont\renewcommand{\arraystretch}{1}Input} & \parbox[b]{\gw}{\centering\fontsize{8}{9}\selectfont\renewcommand{\arraystretch}{1}SQ} & \parbox[b]{\gw}{\centering\fontsize{8}{9}\selectfont\renewcommand{\arraystretch}{1}EMS} & \parbox[b]{\gw}{\centering\fontsize{8}{9}\selectfont\renewcommand{\arraystretch}{1}Marching-\\Prim.} & \parbox[b]{\gw}{\centering\fontsize{8}{9}\selectfont\renewcommand{\arraystretch}{1}SuperDec} & \parbox[b]{\gw}{\centering\fontsize{8}{9}\selectfont\renewcommand{\arraystretch}{1}SuperFlex} & \parbox[b]{\gw}{\centering\fontsize{8}{9}\selectfont\renewcommand{\arraystretch}{1}PhysDec\\(Evolved)} & \parbox[b]{\gw}{\centering\fontsize{8}{9}\selectfont\renewcommand{\arraystretch}{1}PhysDec\\(Loop)} \\[0.5ex]
\includegraphics[width=\gw]{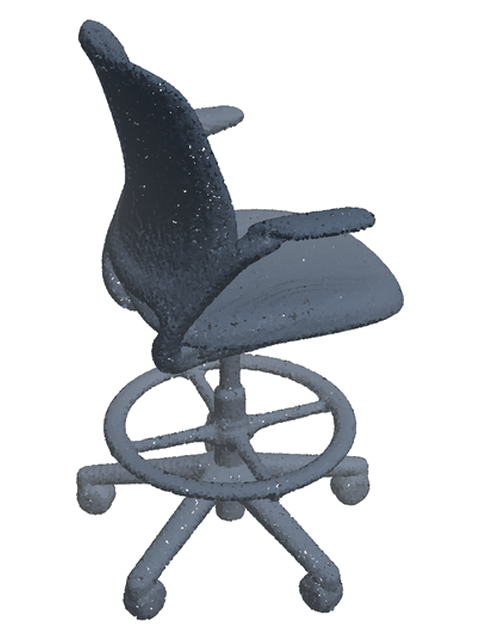} & \includegraphics[width=\gw]{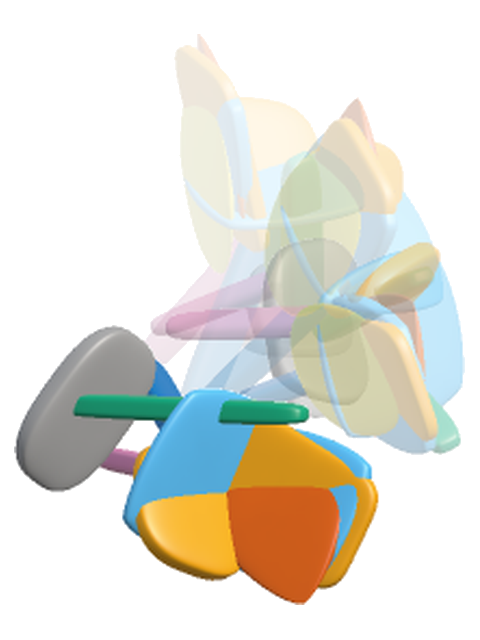} & \includegraphics[width=\gw]{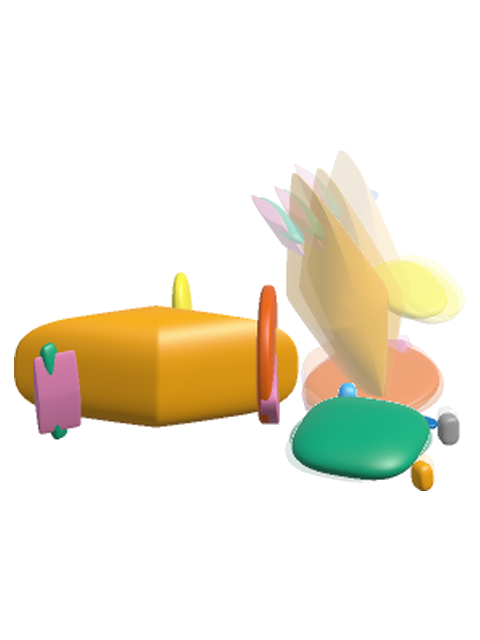} & \includegraphics[width=\gw]{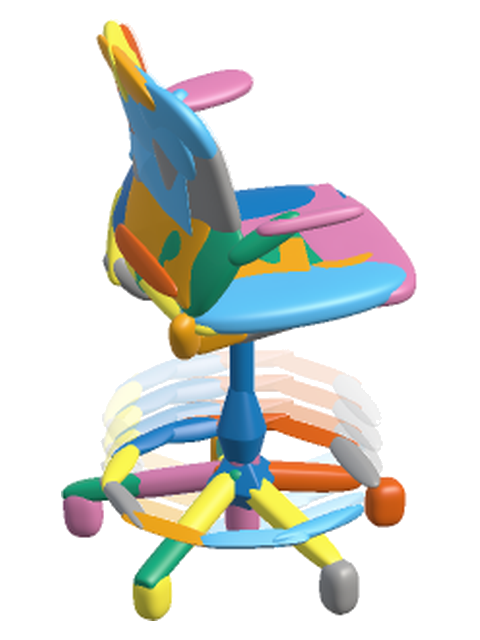} & \includegraphics[width=\gw]{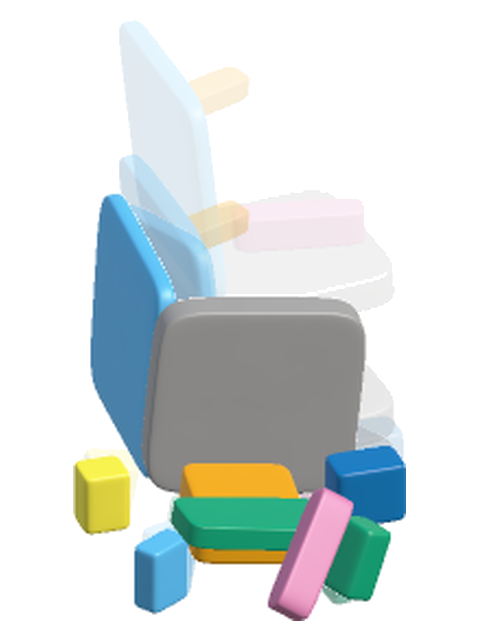} & \includegraphics[width=\gw]{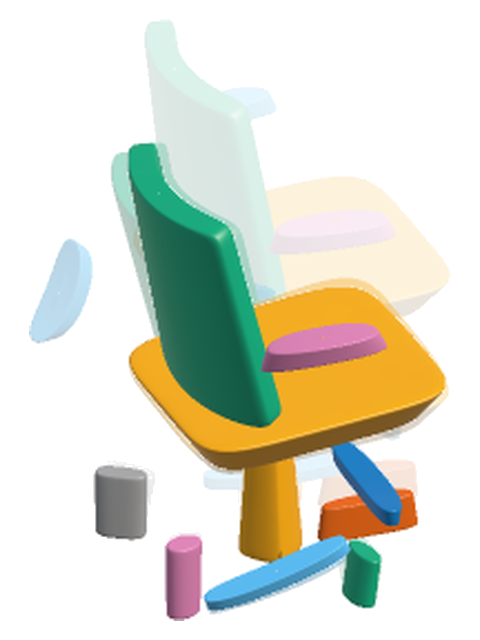} & \includegraphics[width=\gw]{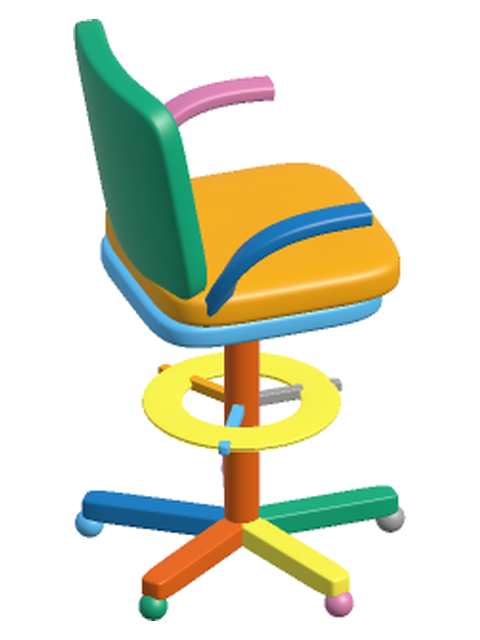} & \includegraphics[width=\gw]{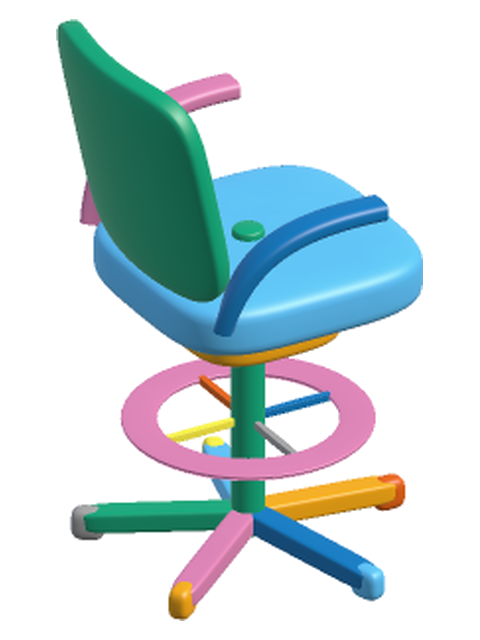} \\[0.16em]
\includegraphics[width=\gw]{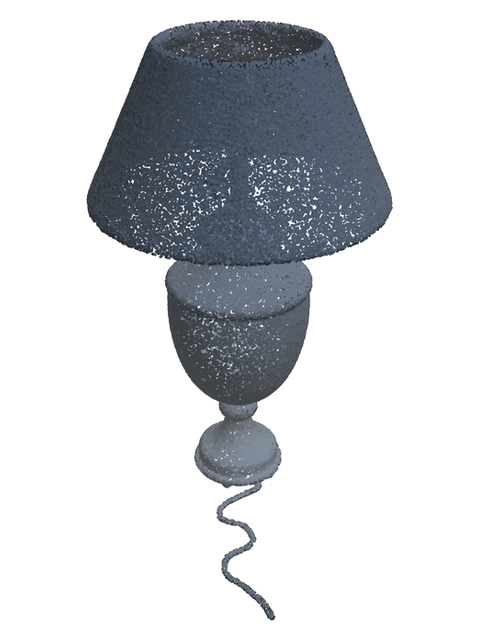} & \includegraphics[width=\gw]{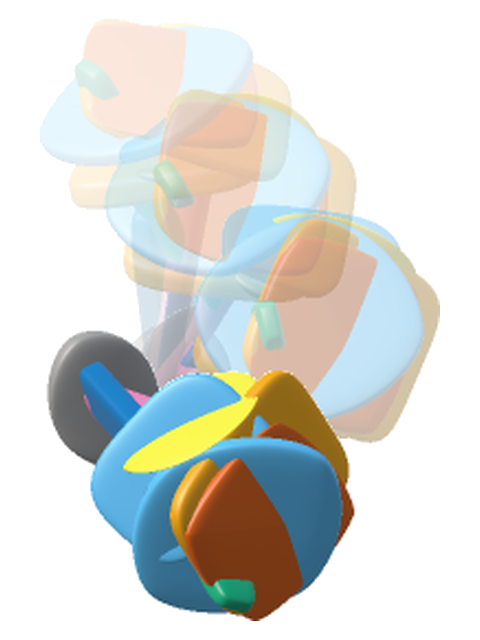} & \includegraphics[width=\gw]{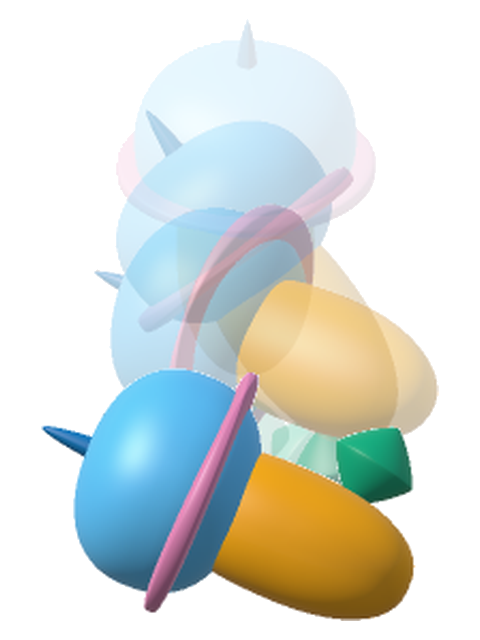} & \includegraphics[width=\gw]{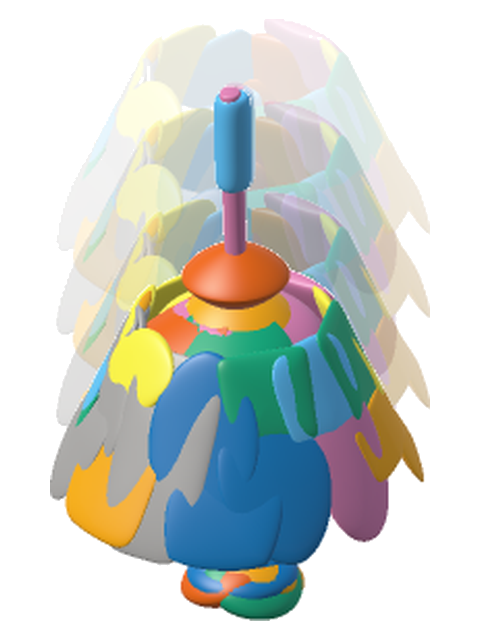} & \includegraphics[width=\gw]{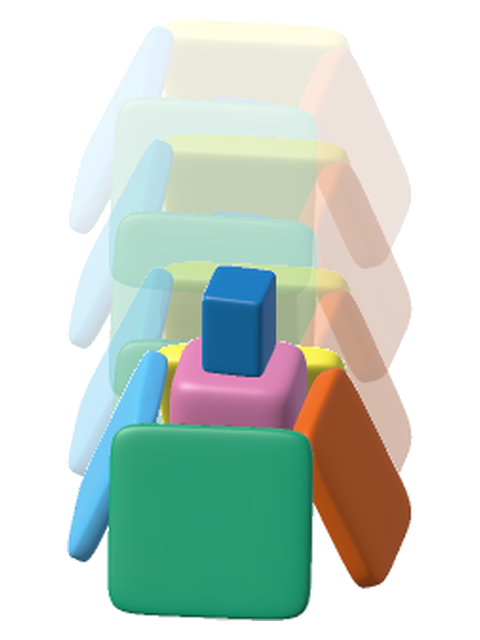} & \includegraphics[width=\gw]{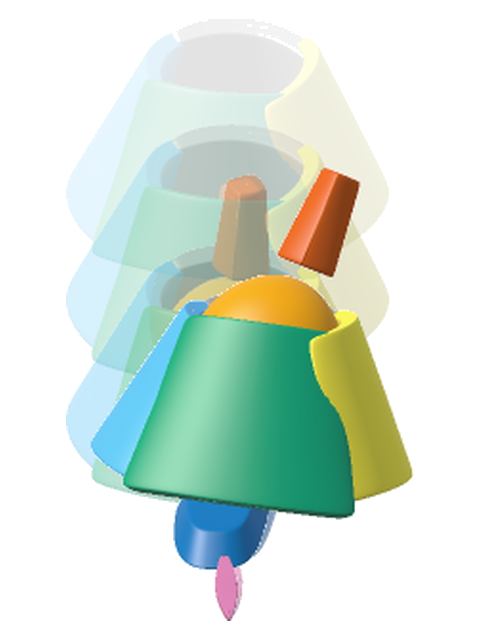} & \includegraphics[width=\gw]{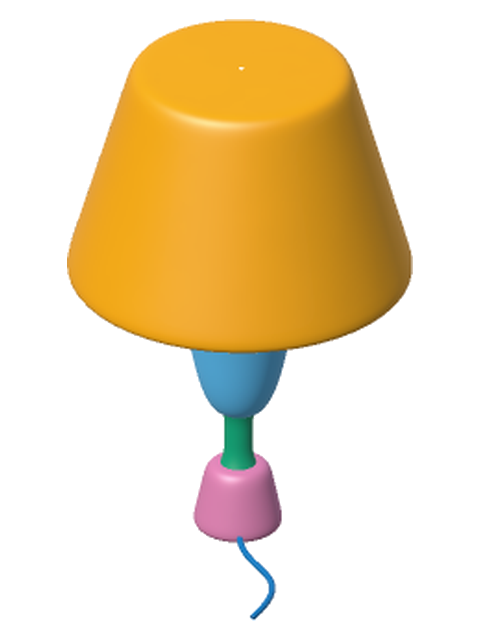} & \includegraphics[width=\gw]{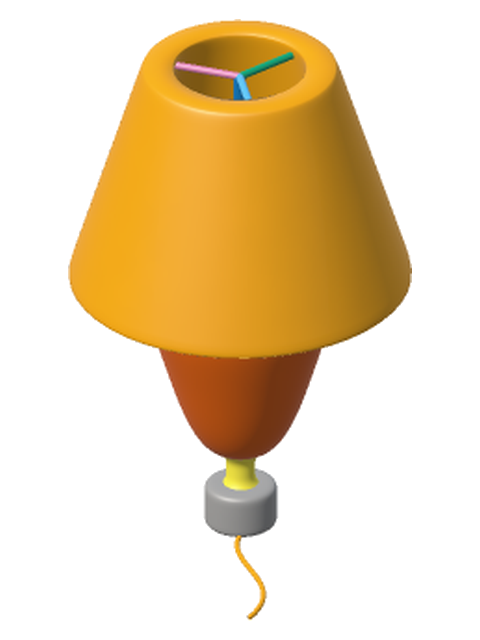} \\[0.16em]
\includegraphics[width=\gw]{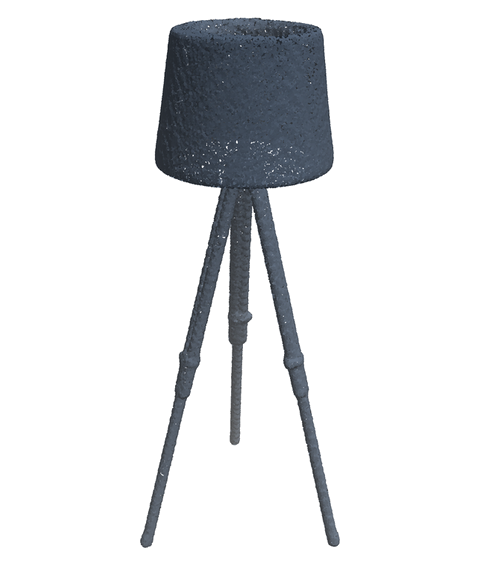} & \includegraphics[width=\gw]{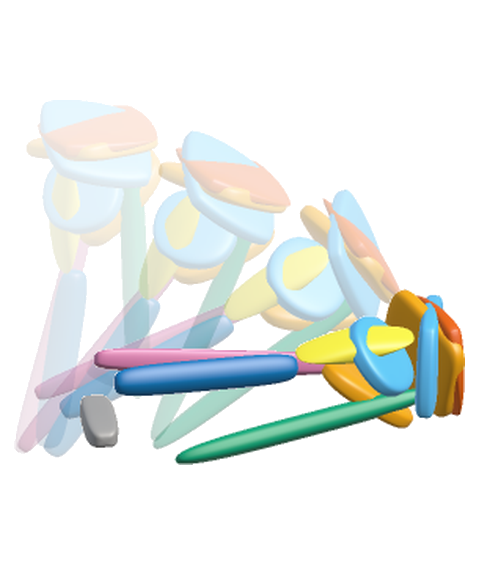} & \includegraphics[width=\gw]{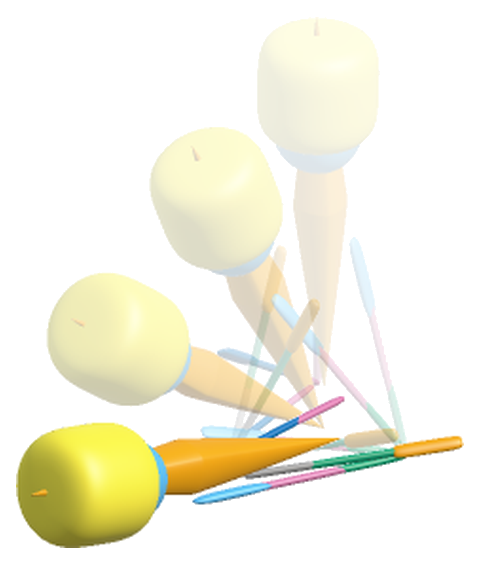} & \includegraphics[width=\gw]{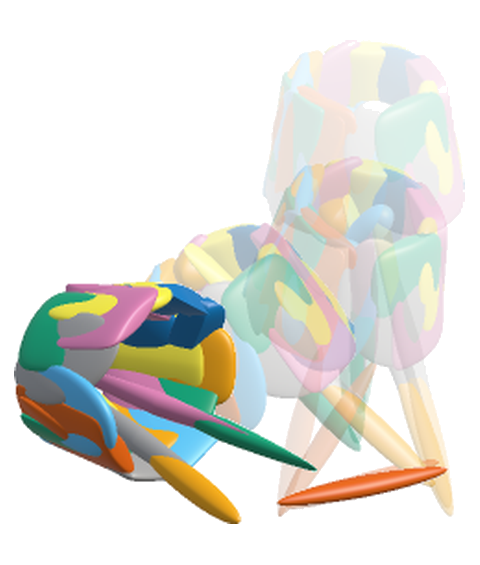} & \includegraphics[width=\gw]{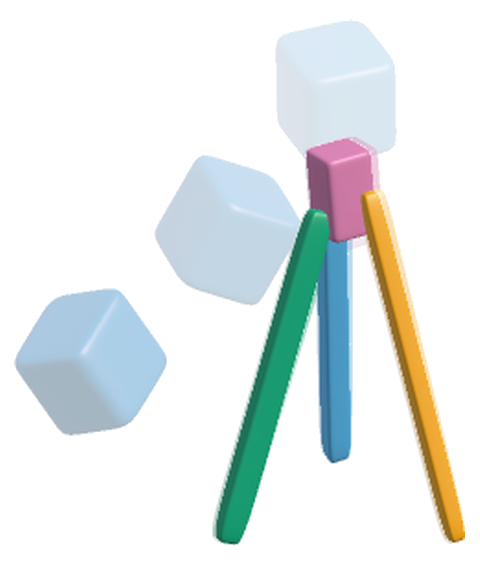} & \includegraphics[width=\gw]{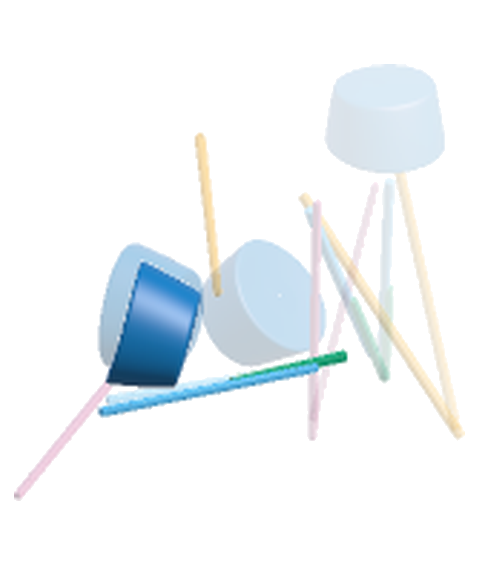} & \includegraphics[width=\gw]{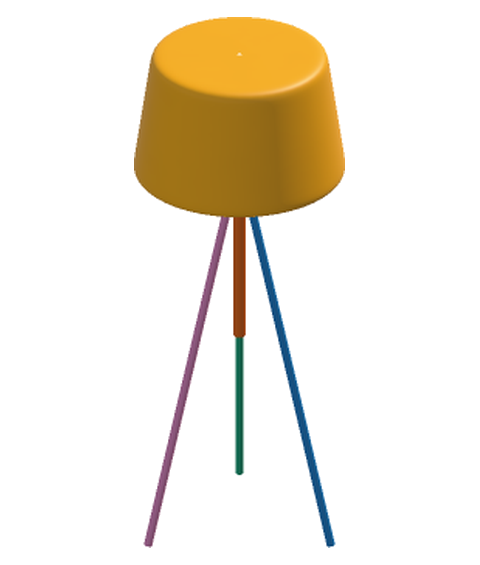} & \includegraphics[width=\gw]{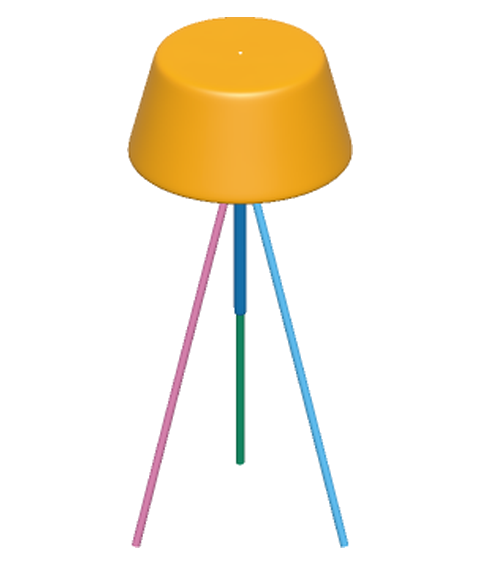} \\[0.16em]
\includegraphics[width=\gw]{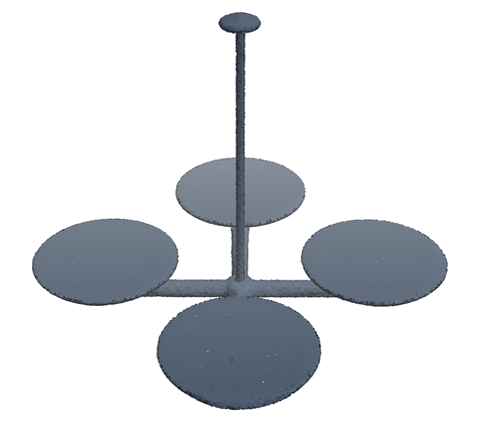} & \includegraphics[width=\gw]{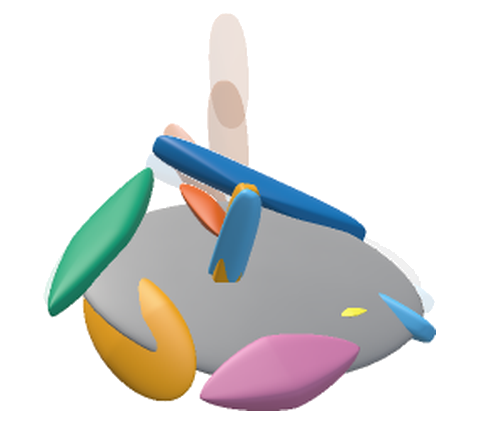} & \includegraphics[width=\gw]{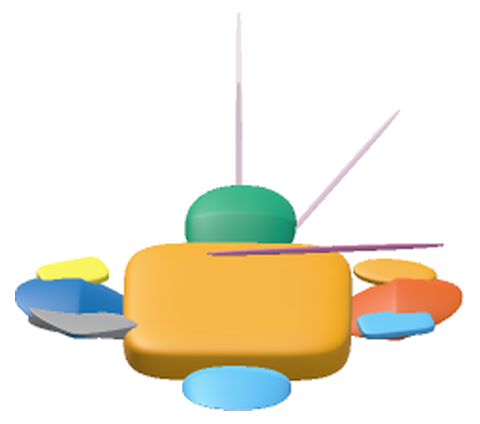} & \includegraphics[width=\gw]{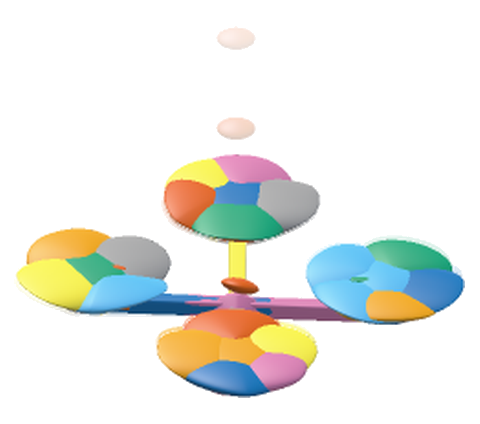} & \includegraphics[width=\gw]{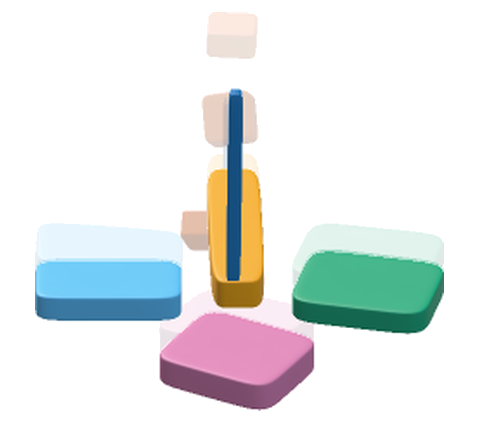} & \includegraphics[width=\gw]{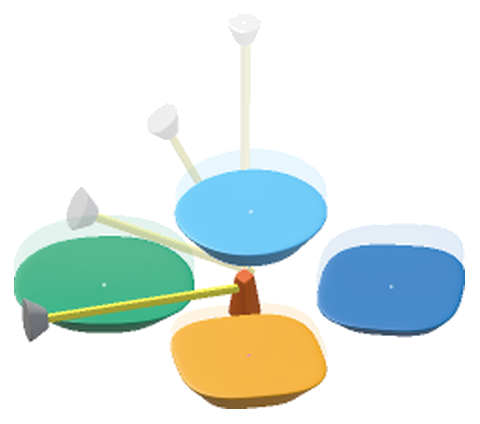} & \includegraphics[width=\gw]{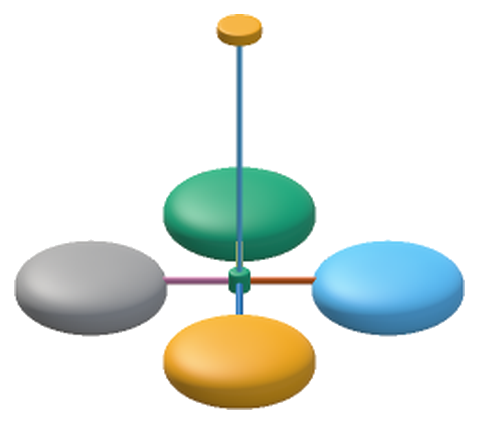} & \includegraphics[width=\gw]{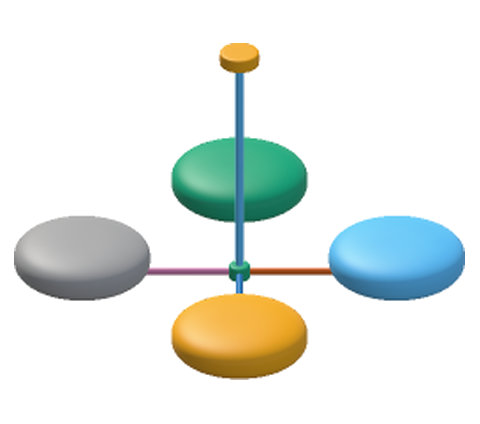} \\[0.16em]
\end{tabular}
}
  \caption{\textbf{Physical plausibility.} Selected reconstructions from the baselines and PhysDec after $4$\,s of simulation.}
  \label{fig:gallery}
\end{figure*}

\subsection{PhysDec-Loop}
\label{sec:exp_baselines}
\label{sec:qualitative}

We evaluate on 585 ShapeNet~\cite{chang2015shapenet} objects, 45 from each of the 13 categories used by SuperDec and SuperFlex~\cite{fedele2025superdec,tavernini2026superflex}.
We use the data preprocessed by Occupancy Networks~\cite{mescheder2019occupancy}. Our agent receives four $137{\times}137$ RGB images and their camera parameters from this dataset.
The baselines use the corresponding ground-truth 3D point clouds as input (converted to an SDF for Marching-Primitives~\cite{liu2023marching}).
Each object's support type is labelled once by a VLM from its four views; every method is checked against the same labels.

\begin{table}[!htbp]
  \centering\footnotesize
  \captionsetup{position=top,skip=6pt}
  \caption{\textbf{Primitive-fitting comparison} on ShapeNet objects. VLM scores are graded by two judges, Claude Sonnet 5.5 and GPT-6.1 Sol, on the same renders. $^\dagger$Receives an SDF of the reference shape instead of a point cloud.}
  \label{tab:baselines}
  \setlength{\tabcolsep}{2.6pt}
  \resizebox{\linewidth}{!}{%
\begin{tabular*}{\linewidth}{@{\extracolsep{\fill}}lccccc@{}}
    \toprule
    Method & \chamfer$\downarrow$ & \#Prim. & \begin{tabular}[b]{@{}c@{}}Phys.\\pass\\(\%)$\uparrow$\end{tabular} & \begin{tabular}[b]{@{}c@{}}VLM$\uparrow$\\(Sonnet\\5.5)\end{tabular} & \begin{tabular}[b]{@{}c@{}}VLM$\uparrow$\\(GPT-6.1\\Sol)\end{tabular} \\
    \midrule
    SQ~\cite{paschalidou2019superquadrics} & 2.205 & 10.0 & 72.6 & 38.8 & 37.8 \\
    EMS~\cite{liu2022ems} & 2.620 & 6.1 & 60.5 & 38.6 & 38.4 \\
    Marching-Primitives$^\dagger$~\cite{liu2023marching} & 1.829 & 31.3 & 77.3 & 64.1 & 66.0 \\
    SuperDec~\cite{fedele2025superdec} & 1.755 & 5.1 & 56.2 & 49.3 & 49.9 \\
    SuperFlex~\cite{tavernini2026superflex} & 1.543 & 4.9 & 81.7 & 58.4 & 58.1 \\
    PhysDec-Loop & 1.448 & 7.4 & 99.3 & 78.0 & 77.8 \\
    \bottomrule
  \end{tabular*}%
  }

\end{table}

With the 3D point cloud available in the verifier, PhysDec-Loop achieves the lowest Chamfer distance, the highest VLM score and the highest physical-check pass rate, 99.3\% (\cref{tab:baselines}).
Most baseline failures involve disconnected parts, despite plausible surface fits. Selected comparisons (\cref{fig:gallery_visual,fig:gallery}) show that PhysDec recovers chair supports and lamp arms that keep the reconstructed objects connected during simulation.

\subsection{PhysDec-Evolved}
\label{sec:exp_evolution}

Self-evolution uses only 20 training objects per category (260 in total), and all variants are evaluated on the same 585 test objects.
We compare PhysDec-Evolved with a base predictor that uses the same VLM with the original prompts: one call analyzes the input images, and a second proposes a decomposition without retrieval or online correction.

\begin{table}[!htbp]
  \centering\footnotesize
  \captionsetup{position=top,skip=6pt}
  \caption{\textbf{Self-evolution.} Base and PhysDec-Evolved use no correction loop. Calls count only the reported run, including VLM review for PhysDec-Loop.}
  \label{tab:evolution}
  \setlength{\tabcolsep}{1.2pt}
  \resizebox{\linewidth}{!}{%
\begin{tabular*}{\linewidth}{@{\extracolsep{\fill}}lccccc@{}}
    \toprule
     & Calls & \chamfer$\downarrow$ & \#Prim. & \begin{tabular}[b]{@{}c@{}}Phys.\ pass\\(\%)$\uparrow$\end{tabular} & VLM$\uparrow$ \\
    \midrule
    Base (original prompt) & 2 & 1.526 & 7.0 & 88.4 & 73.5 \\
    PhysDec-Evolved & 2.00 & 1.528 & 8.2 & 99.0 & 75.8 \\
    \midrule
    \shortstack[l]{PhysDec-Loop\\(3 rounds)} & 12.9 & 1.448 & 7.4 & 99.3 & 78.0 \\
    \bottomrule
  \end{tabular*}%
  }

\end{table}

As shown in \cref{tab:evolution}, self-evolution raises physical validity from 88.4\% to 99.0\% relative to the base predictor. This is close to PhysDec-Loop's 99.3\% pass rate, while model calls drop from 12.9 to 2 per prediction. The VLM score improves from 73.5 to 75.8, approaching the performance of PhysDec-Loop.

\begin{table}[!htbp]
  \centering\footnotesize
  \captionsetup{position=top,skip=6pt}
  \caption{\textbf{Unseen categories.} Comparison on 43 Google Scanned Objects from six categories unseen during self-evolution. Img: four images; PC: ground-truth point cloud.}
  \label{tab:unseen}
  \setlength{\tabcolsep}{1.2pt}
  \begingroup
  \resizebox{\linewidth}{!}{%
\begin{tabular*}{\linewidth}{@{\extracolsep{\fill}}lcccccc@{}}
    \toprule
     & Input & Calls & \chamfer$\downarrow$ & \#Prim. & \begin{tabular}[b]{@{}c@{}}Phys.\ pass\\(\%)$\uparrow$\end{tabular} & VLM$\uparrow$ \\
    \midrule
    SuperDec~\cite{fedele2025superdec} & PC & -- & 2.384 & 5.7 & 55.8 & 39.8 \\
    SuperFlex~\cite{tavernini2026superflex} & PC & -- & 2.198 & 5.0 & 72.1 & 47.6 \\
    \midrule
    \shortstack[l]{Base\\(original prompt)} & Img & 2 & 2.127 & 5.2 & 86.0 & 69.2 \\
    PhysDec-Evolved & Img & 2 & 2.123 & 5.9 & 100.0 & 71.7 \\
    \midrule
    \shortstack[l]{PhysDec-Loop\\(3 rounds)} & \shortstack{Img\\+\,PC} & 13.2 & 1.894 & 5.8 & 97.7 & 73.3 \\
    \bottomrule
  \end{tabular*}%
  }

  \endgroup
\end{table}

\begin{figure}[!htbp]
  \centering
  \resizebox{\linewidth}{!}{
\setlength{\tabcolsep}{0pt}\renewcommand{\arraystretch}{0}
\def\gw{\dimexpr(\linewidth-5\dimexpr0.25em\relax)/6\relax}
\begin{tabular}{@{}c*{5}{@{\hspace{0.25em}}c}@{}}
\parbox[b]{\gw}{\centering\scriptsize\renewcommand{\arraystretch}{1}Input} & \parbox[b]{\gw}{\centering\scriptsize\renewcommand{\arraystretch}{1}SuperDec} & \parbox[b]{\gw}{\centering\scriptsize\renewcommand{\arraystretch}{1}SuperFlex} & \parbox[b]{\gw}{\centering\scriptsize\renewcommand{\arraystretch}{1}Base} & \parbox[b]{\gw}{\centering\scriptsize\renewcommand{\arraystretch}{1}PhysDec-\\Evolved} & \parbox[b]{\gw}{\centering\scriptsize\renewcommand{\arraystretch}{1}PhysDec-\\Loop} \\[0.5ex]
\includegraphics[width=\gw]{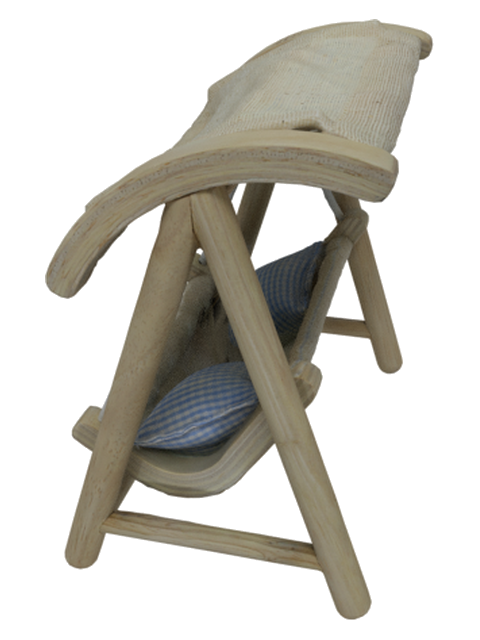} & \includegraphics[width=\gw]{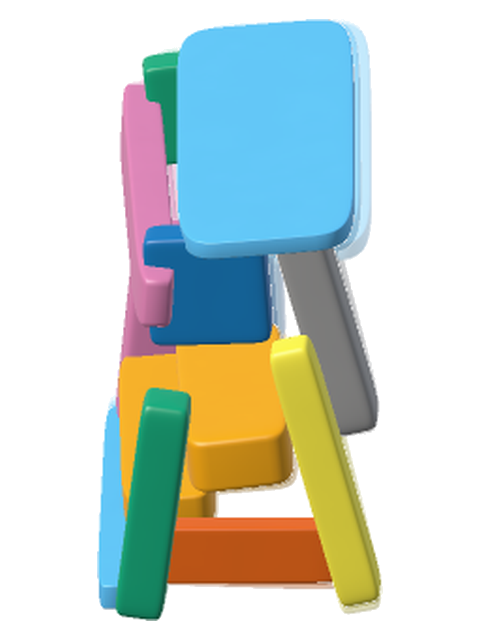} & \includegraphics[width=\gw]{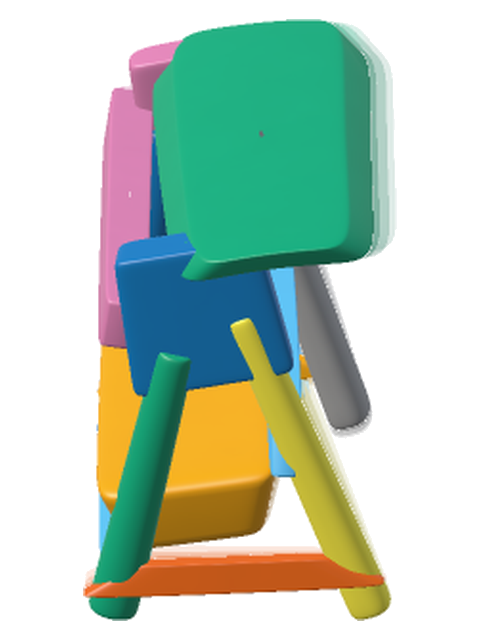} & \includegraphics[width=\gw]{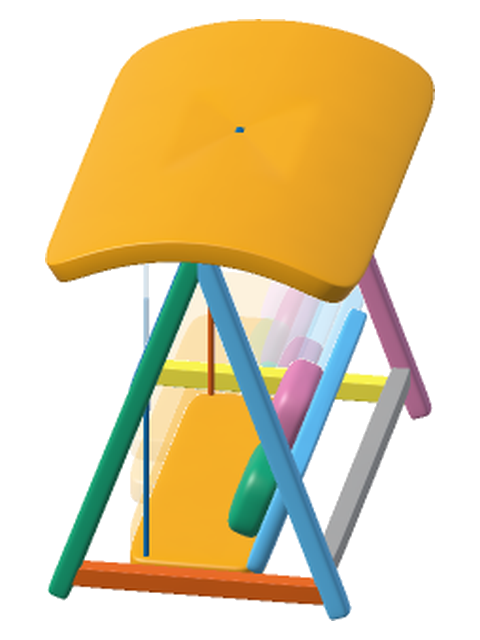} & \includegraphics[width=\gw]{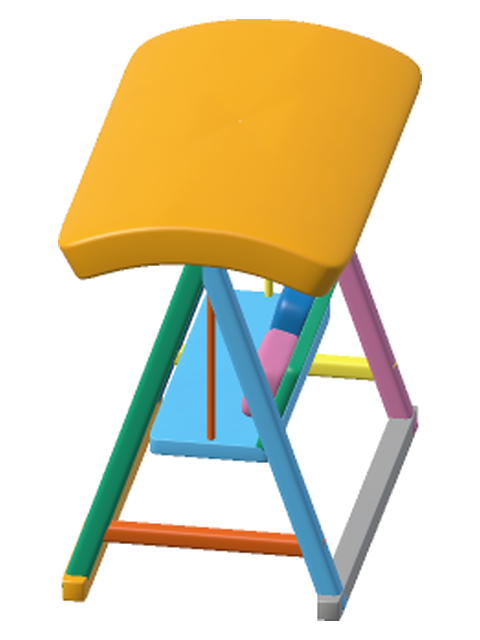} & \includegraphics[width=\gw]{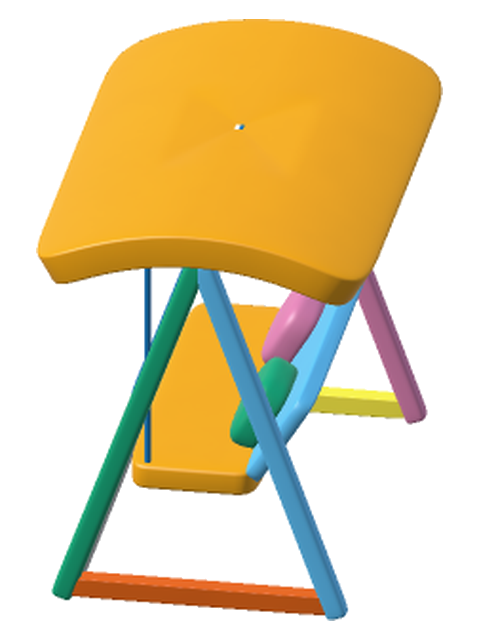} \\[0.4ex]
\parbox[t]{\gw}{\centering\scriptsize\renewcommand{\arraystretch}{1}garden swing} & \parbox[t]{\gw}{\centering\scriptsize\renewcommand{\arraystretch}{1}\textcolor{passc}{stands}} & \parbox[t]{\gw}{\centering\scriptsize\renewcommand{\arraystretch}{1}\textcolor{passc}{stands}} & \parbox[t]{\gw}{\centering\scriptsize\renewcommand{\arraystretch}{1}\textcolor{failc}{2 pieces}} & \parbox[t]{\gw}{\centering\scriptsize\renewcommand{\arraystretch}{1}\textcolor{passc}{stands}} & \parbox[t]{\gw}{\centering\scriptsize\renewcommand{\arraystretch}{1}\textcolor{passc}{stands}} \\[1.2ex]
\includegraphics[width=\gw]{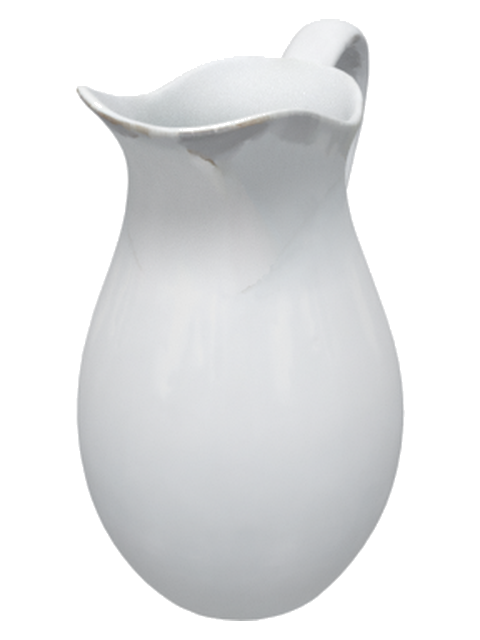} & \includegraphics[width=\gw]{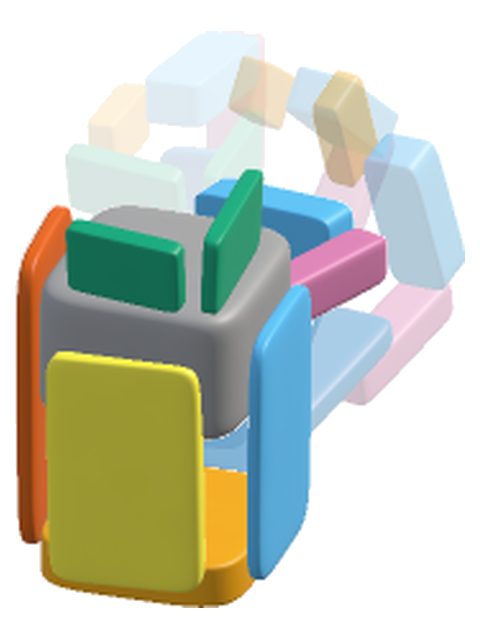} & \includegraphics[width=\gw]{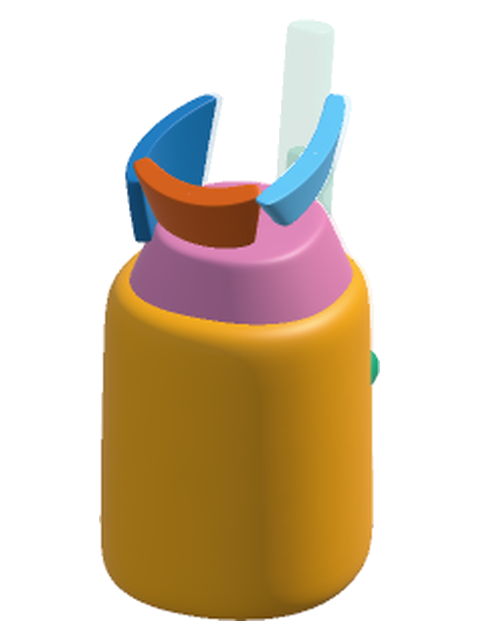} & \includegraphics[width=\gw]{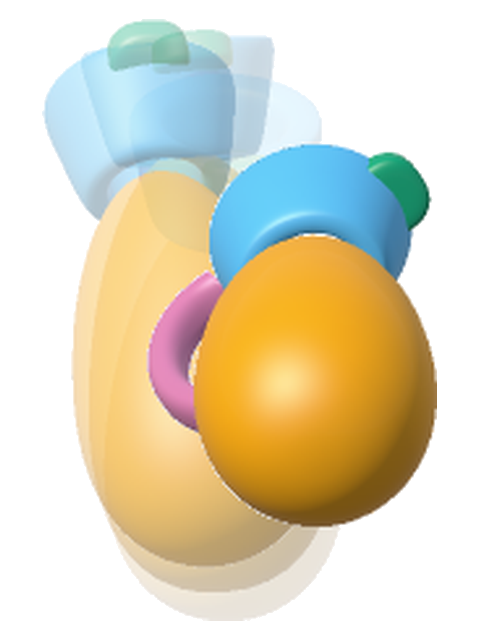} & \includegraphics[width=\gw]{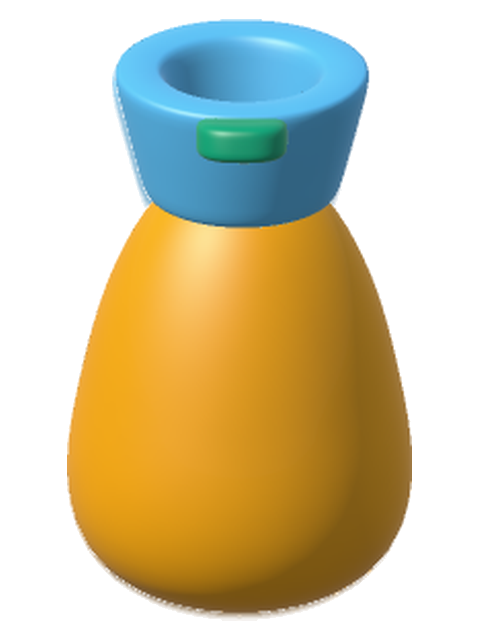} & \includegraphics[width=\gw]{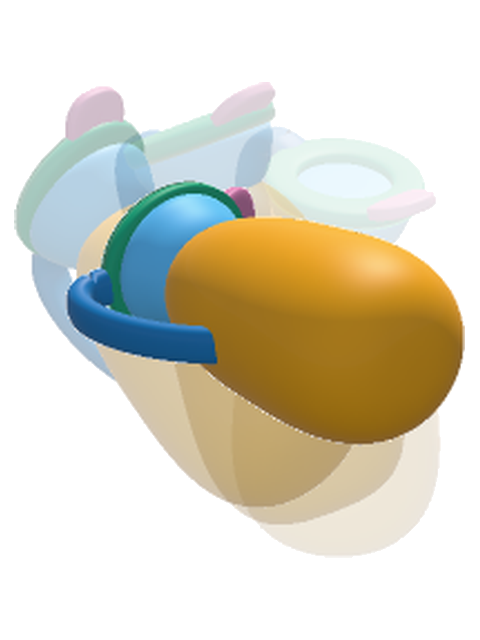} \\[0.4ex]
\parbox[t]{\gw}{\centering\scriptsize\renewcommand{\arraystretch}{1}pitcher} & \parbox[t]{\gw}{\centering\scriptsize\renewcommand{\arraystretch}{1}\textcolor{failc}{4 pieces, falls}} & \parbox[t]{\gw}{\centering\scriptsize\renewcommand{\arraystretch}{1}\textcolor{failc}{2 pieces, falls}} & \parbox[t]{\gw}{\centering\scriptsize\renewcommand{\arraystretch}{1}\textcolor{failc}{falls}} & \parbox[t]{\gw}{\centering\scriptsize\renewcommand{\arraystretch}{1}\textcolor{passc}{stands}} & \parbox[t]{\gw}{\centering\scriptsize\renewcommand{\arraystretch}{1}\textcolor{failc}{falls}} \\[1.2ex]
\includegraphics[width=\gw]{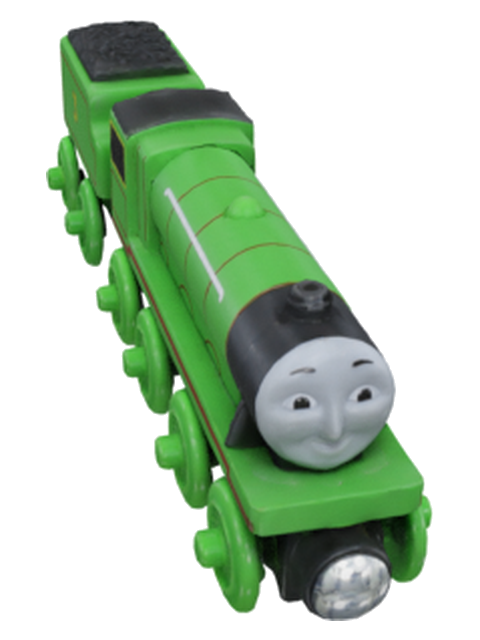} & \includegraphics[width=\gw]{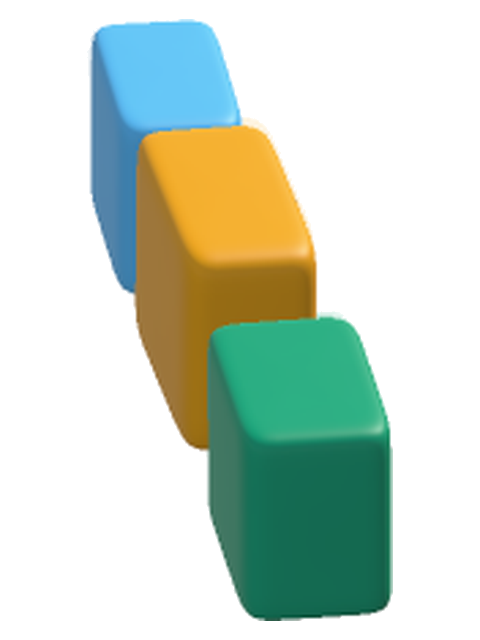} & \includegraphics[width=\gw]{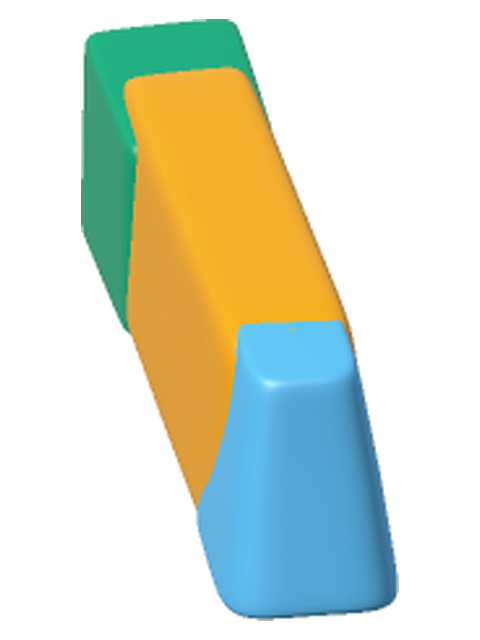} & \includegraphics[width=\gw]{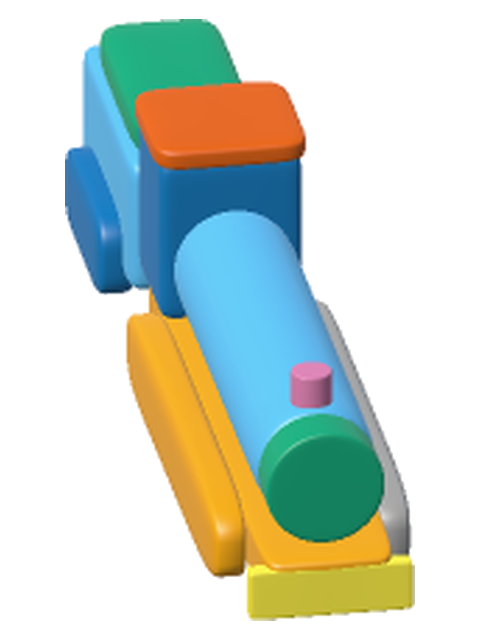} & \includegraphics[width=\gw]{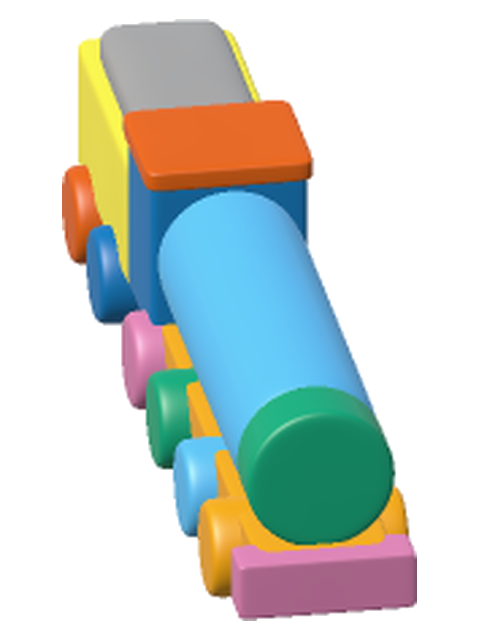} & \includegraphics[width=\gw]{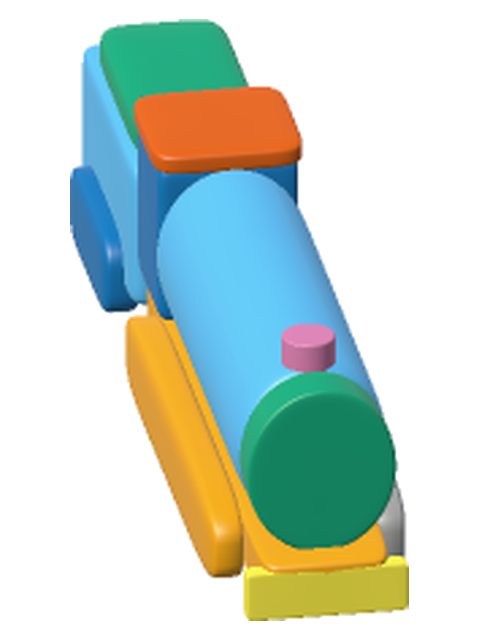} \\[0.4ex]
\parbox[t]{\gw}{\centering\scriptsize\renewcommand{\arraystretch}{1}toy engine} & \parbox[t]{\gw}{\centering\scriptsize\renewcommand{\arraystretch}{1}\textcolor{failc}{3 pieces}} & \parbox[t]{\gw}{\centering\scriptsize\renewcommand{\arraystretch}{1}\textcolor{passc}{stands}} & \parbox[t]{\gw}{\centering\scriptsize\renewcommand{\arraystretch}{1}\textcolor{failc}{2 pieces}} & \parbox[t]{\gw}{\centering\scriptsize\renewcommand{\arraystretch}{1}\textcolor{passc}{stands}} & \parbox[t]{\gw}{\centering\scriptsize\renewcommand{\arraystretch}{1}\textcolor{passc}{stands}} \\[1.2ex]
\includegraphics[width=\gw]{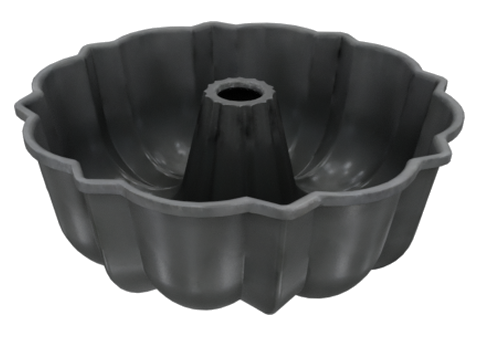} & \includegraphics[width=\gw]{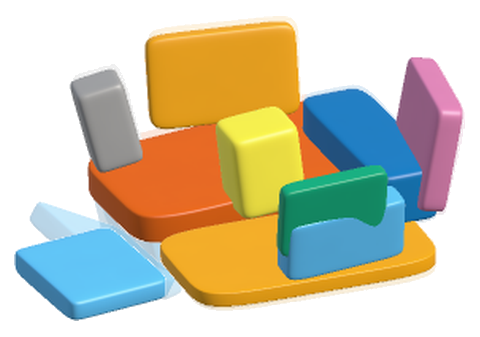} & \includegraphics[width=\gw]{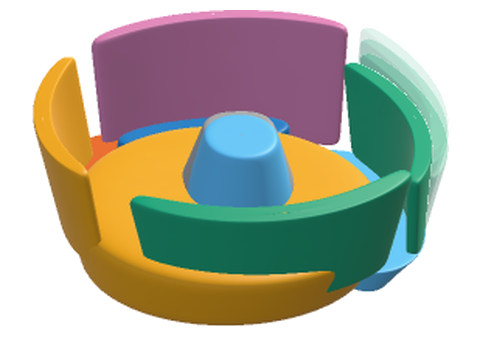} & \includegraphics[width=\gw]{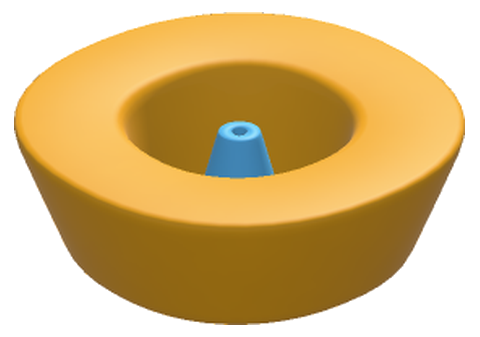} & \includegraphics[width=\gw]{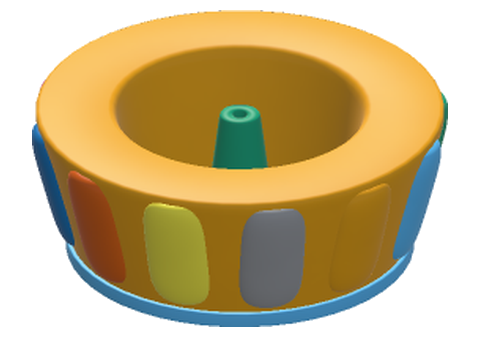} & \includegraphics[width=\gw]{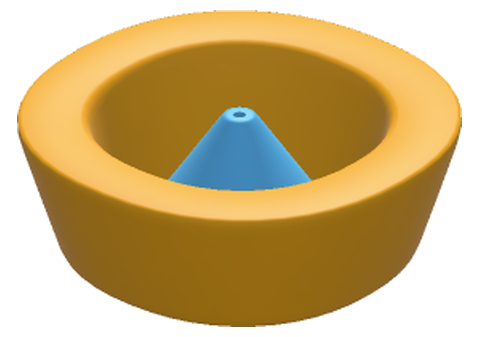} \\[0.4ex]
\parbox[t]{\gw}{\centering\scriptsize\renewcommand{\arraystretch}{1}bundt pan} & \parbox[t]{\gw}{\centering\scriptsize\renewcommand{\arraystretch}{1}\textcolor{failc}{2 pieces, falls}} & \parbox[t]{\gw}{\centering\scriptsize\renewcommand{\arraystretch}{1}\textcolor{failc}{2 pieces, falls}} & \parbox[t]{\gw}{\centering\scriptsize\renewcommand{\arraystretch}{1}\textcolor{failc}{2 pieces}} & \parbox[t]{\gw}{\centering\scriptsize\renewcommand{\arraystretch}{1}\textcolor{passc}{stands}} & \parbox[t]{\gw}{\centering\scriptsize\renewcommand{\arraystretch}{1}\textcolor{passc}{stands}} \\[1.2ex]
\end{tabular}
}
  \caption{\textbf{Physical plausibility on unseen categories.} Reconstructions from the baselines and PhysDec on Google Scanned Objects after $4$\,s of simulation.}
  \label{fig:unseen}
\end{figure}

\paragraph{Unseen categories.}
We evaluate on 43 Google Scanned Objects~\cite{downs2022gso} from six categories unseen during self-evolution: bags, baskets, appliances, kitchenware, hand tools and toys.
PhysDec-Evolved uses the evolved prompts and category-independent insights, without category-specific exemplars or material priors.
It raises physical validity from 86.0\% to 100\% over the base predictor and improves visual quality with two calls per object (\cref{tab:unseen,fig:unseen}).
With point-cloud feedback, PhysDec-Loop also outperforms SuperDec and SuperFlex in geometric fidelity, physical validity and visual quality.

\subsection{Partial-object reconstruction}
\label{sec:exp_partial}

\begin{figure}[!htbp]
  \centering\scriptsize
  \setlength{\tabcolsep}{0pt}
  
  \resizebox{\linewidth}{!}{%
\begin{tabular}{@{}cccc@{}}
    Observed points & SuperDec & SuperFlex (ASE) & PhysDec-Loop \\
    \includegraphics[trim=0 36 0 15,clip,width=0.245\linewidth]{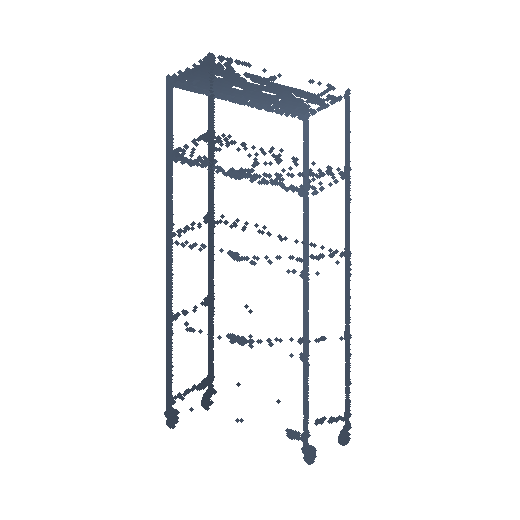} &
    \includegraphics[trim=0 36 0 15,clip,width=0.245\linewidth]{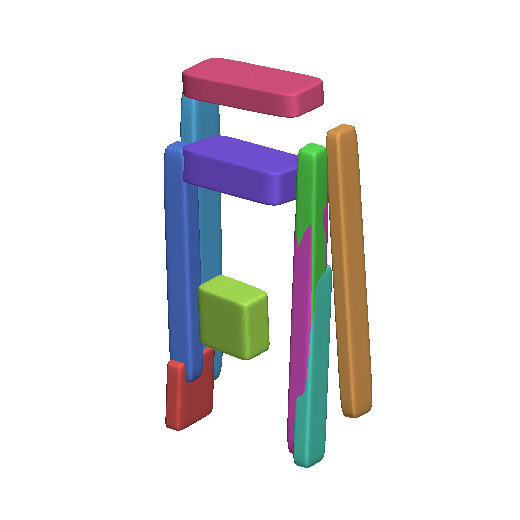} &
    \includegraphics[trim=0 36 0 15,clip,width=0.245\linewidth]{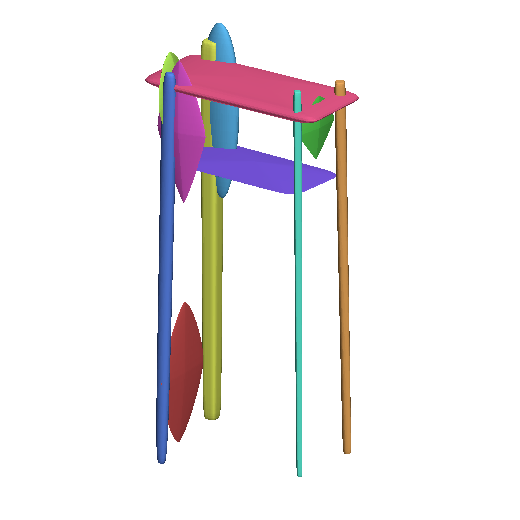} &
    \includegraphics[trim=0 36 0 15,clip,width=0.245\linewidth]{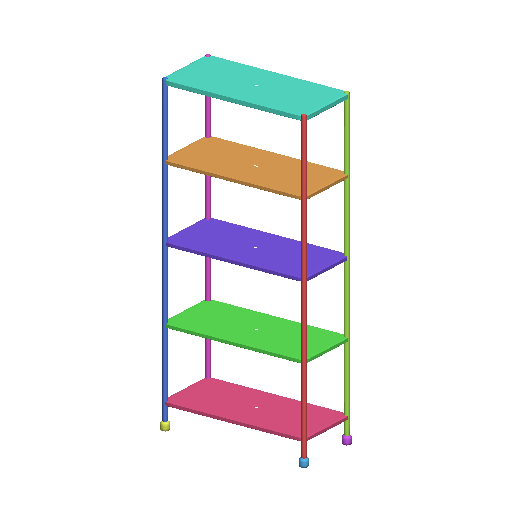} \\
    \includegraphics[trim=0 20 0 52,clip,width=0.245\linewidth]{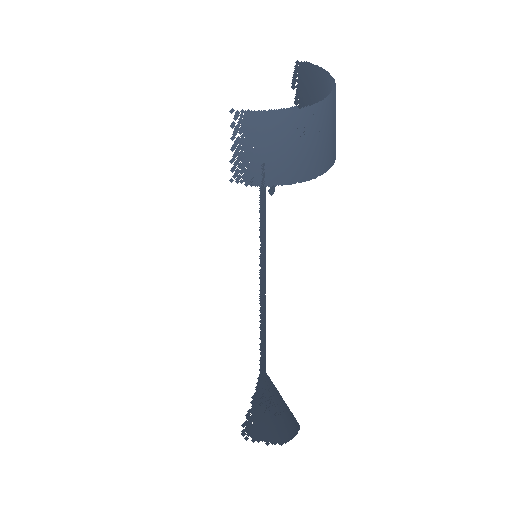} &
    \includegraphics[trim=0 20 0 52,clip,width=0.245\linewidth]{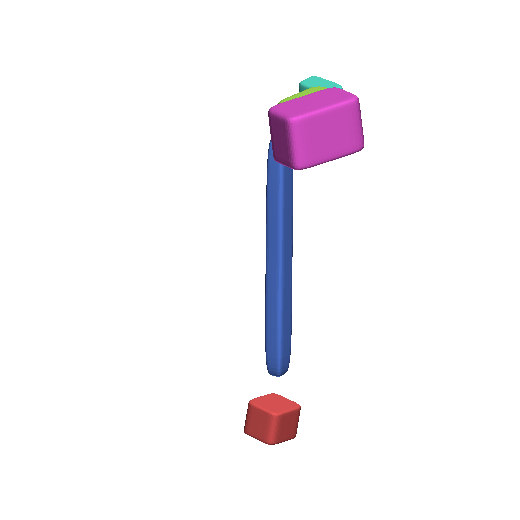} &
    \includegraphics[trim=0 20 0 52,clip,width=0.245\linewidth]{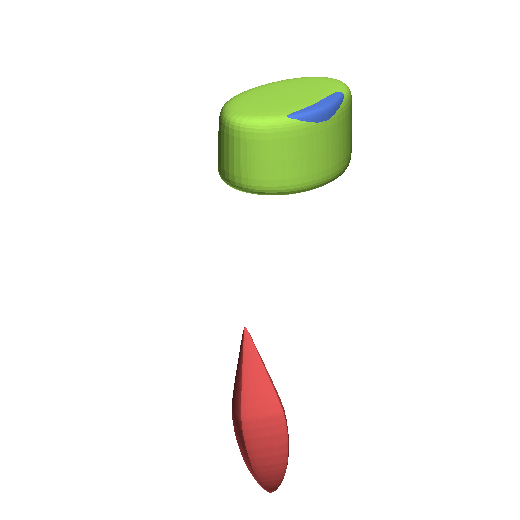} &
    \includegraphics[trim=0 20 0 52,clip,width=0.245\linewidth]{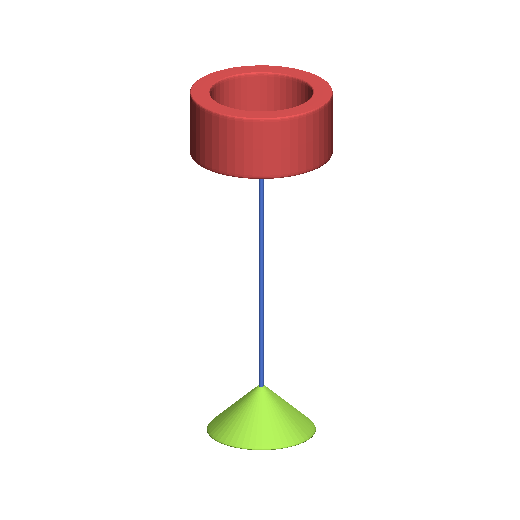} \\
    \includegraphics[trim=57 114 56 111,clip,width=0.245\linewidth]{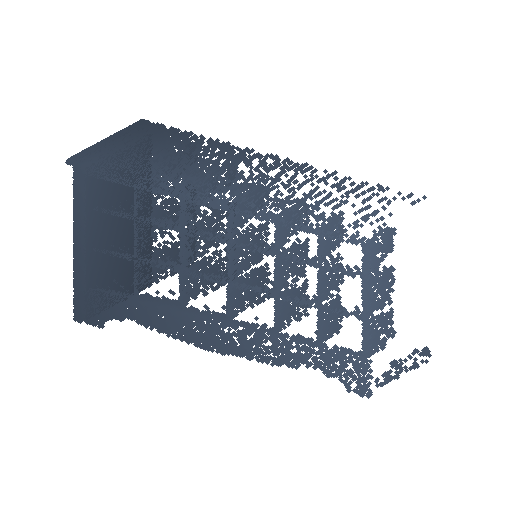} &
    \includegraphics[trim=57 114 56 111,clip,width=0.245\linewidth]{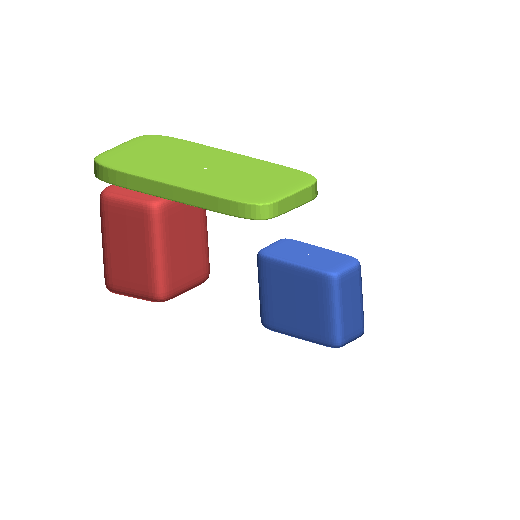} &
    \includegraphics[trim=57 114 56 111,clip,width=0.245\linewidth]{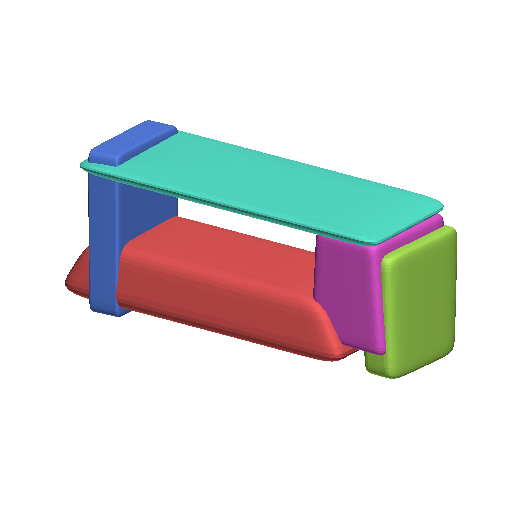} &
    \includegraphics[trim=57 114 56 111,clip,width=0.245\linewidth]{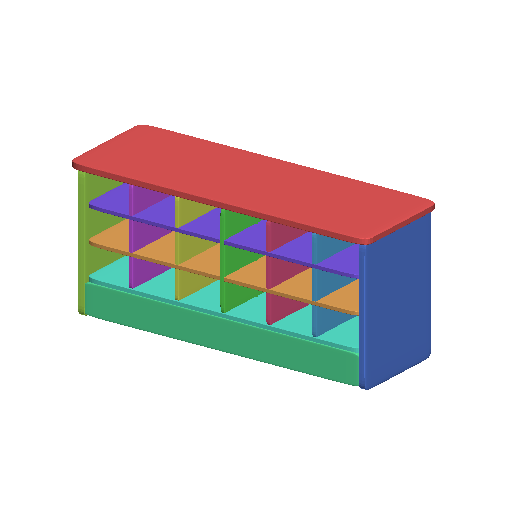} \\
    \includegraphics[trim=52 121 33 124,clip,width=0.245\linewidth]{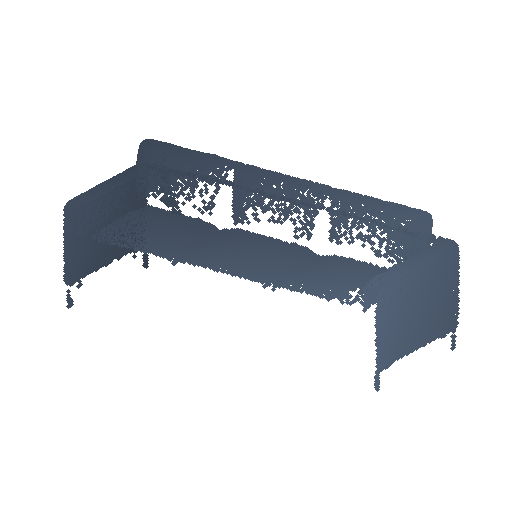} &
    \includegraphics[trim=52 121 33 124,clip,width=0.245\linewidth]{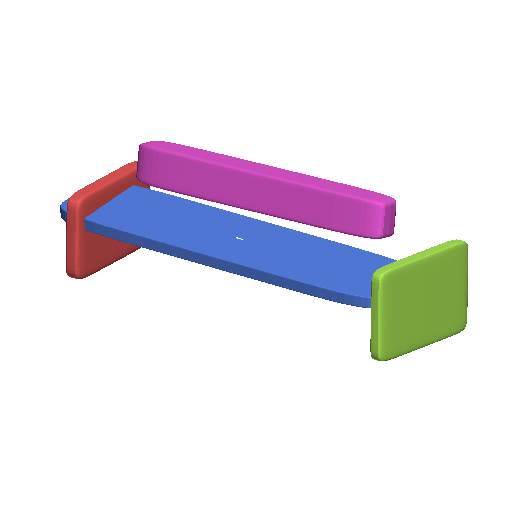} &
    \includegraphics[trim=52 121 33 124,clip,width=0.245\linewidth]{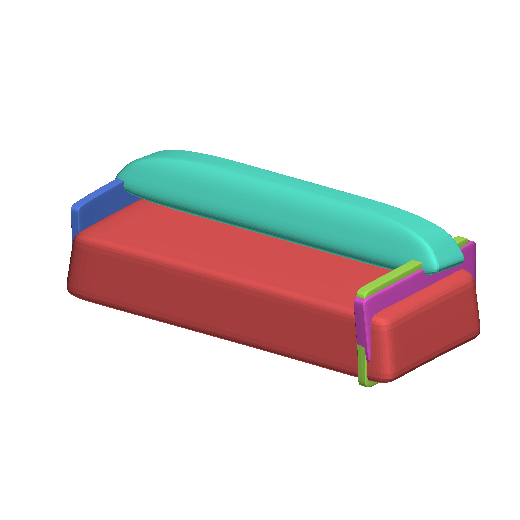} &
    \includegraphics[trim=52 121 33 124,clip,width=0.245\linewidth]{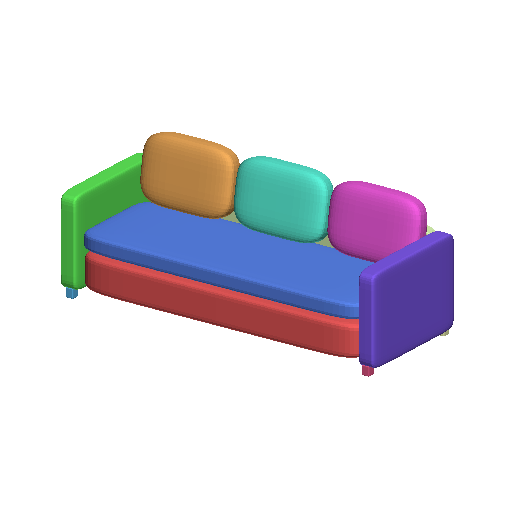} \\
  \end{tabular}%
  }
  \caption{\textbf{Reconstruction from partial observations.} Observed points and reconstructions by SuperDec, ASE-tuned SuperFlex and PhysDec-Loop.}
  \label{fig:partial}
\end{figure}

Following SuperFlex~\cite{tavernini2026superflex}, we evaluate partial reconstruction on Aria Synthetic Environments (ASE)~\cite{avetisyan2024scenescript}.
We use 52 floor-furniture objects from 11 scenes, with meshes and object boxes from EFM3D~\cite{straub2024efm3d}. Four masked depth views yield a normalized 4096-point partial cloud in the object's reference orientation. Baselines receive the partial point cloud; our pipeline uses its rendered views as input and the same points for geometric feedback. As shown in \cref{tab:partial,fig:partial}, our pipeline also improves both geometric fidelity and physical validity when reconstructing partially observed objects.

\begin{table}[!htbp]
  \centering\footnotesize
  \captionsetup{position=top,skip=6pt}
  \caption{\textbf{Partial-object reconstruction} on objects from the ASE dataset. SuperFlex (ASE-tuned) is fine-tuned on partial ASE clouds.}
  \label{tab:partial}
  \setlength{\tabcolsep}{2.4pt}
  \resizebox{\linewidth}{!}{%
\begin{tabular*}{\linewidth}{@{\extracolsep{\fill}}lcccc@{}}
    \toprule
    Method & \chamfer$\downarrow$ & \#Prim. & Phys.\ pass (\%)$\uparrow$ & VLM$\uparrow$ \\
    \midrule
    SuperDec~\cite{fedele2025superdec} & 4.18 & 5.5 & 11.5 & 35.7 \\
    SuperFlex~\cite{tavernini2026superflex} & 4.37 & 5.5 & 19.2 & 41.8 \\
    SuperFlex (ASE-tuned) & 2.59 & 4.2 & 65.4 & 52.4 \\
    PhysDec-Loop & 2.22 & 7.5 & 100.0 & 77.4 \\
    \bottomrule
  \end{tabular*}%
  }

\end{table}

\begin{figure*}[!t]
  \centering\scriptsize
  \setlength{\tabcolsep}{0pt}
  \def\scenew{\dimexpr(\textwidth-1.5em)/7\relax}

  \resizebox{\textwidth}{!}{%
\begin{tabular}{@{}c@{\hspace{0.4em}}c@{\hspace{0.1em}}c@{\hspace{0.4em}}c@{\hspace{0.1em}}c@{\hspace{0.4em}}c@{\hspace{0.1em}}c@{}}
    The scan & \multicolumn{2}{c}{SuperDec~\cite{fedele2025superdec}} & \multicolumn{2}{c}{SuperFlex~\cite{tavernini2026superflex}} & \multicolumn{2}{c}{Ours (VLM and solver)} \\
    & Reconstruction & Simulation & Reconstruction & Simulation & Reconstruction & Simulation \\[1pt]
    \includegraphics[width=\scenew]{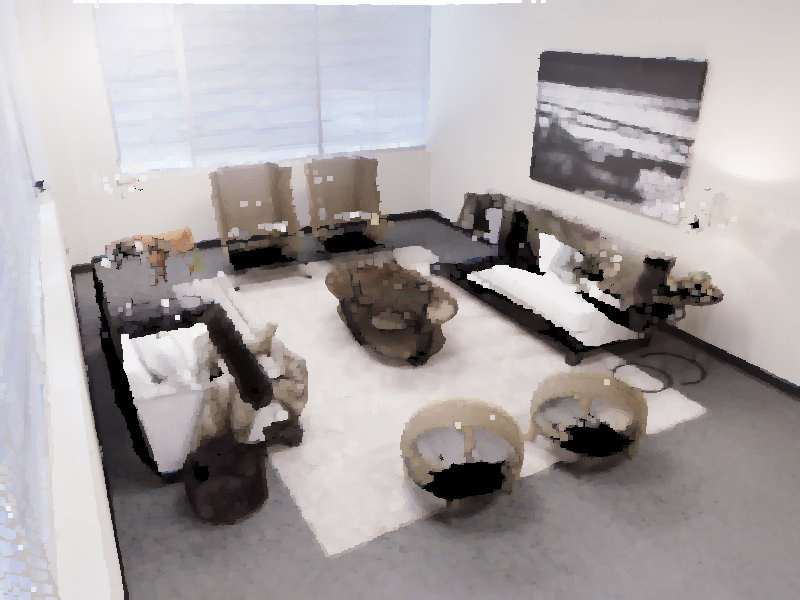} &
    \includegraphics[width=\scenew]{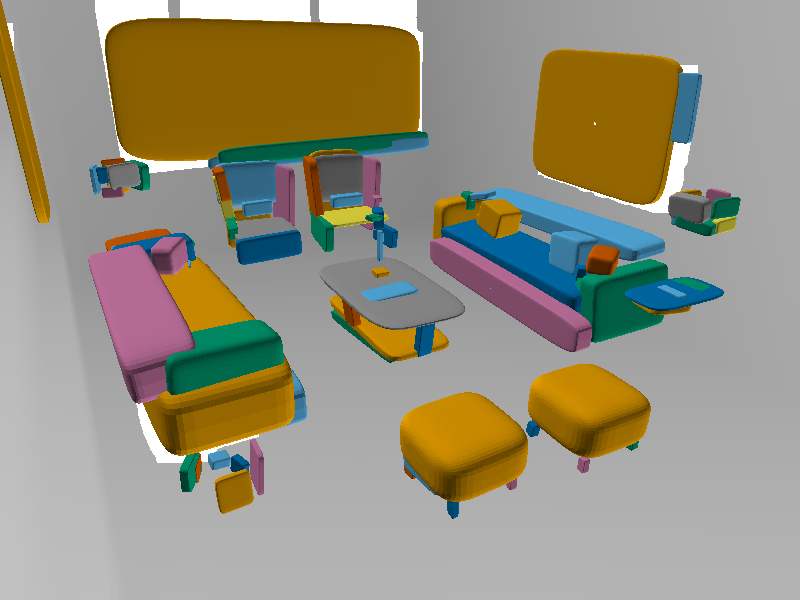} & \includegraphics[width=\scenew]{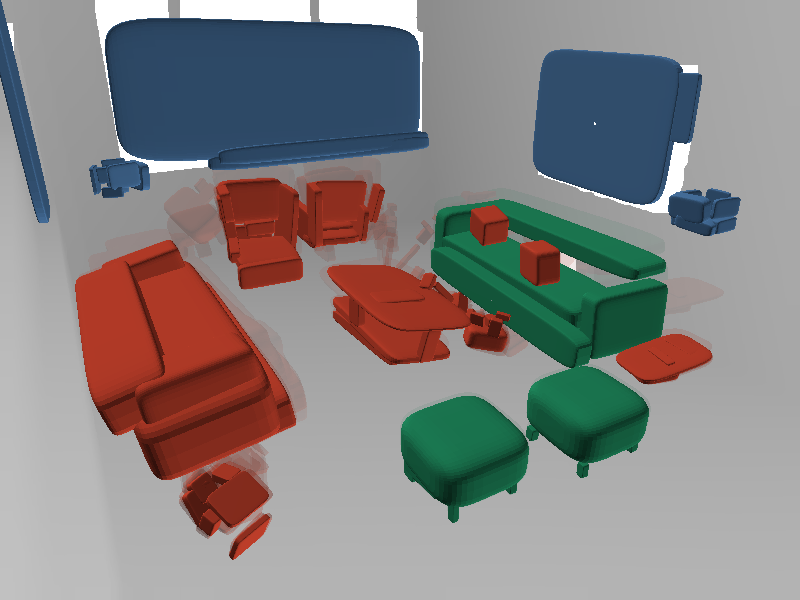} &
    \includegraphics[width=\scenew]{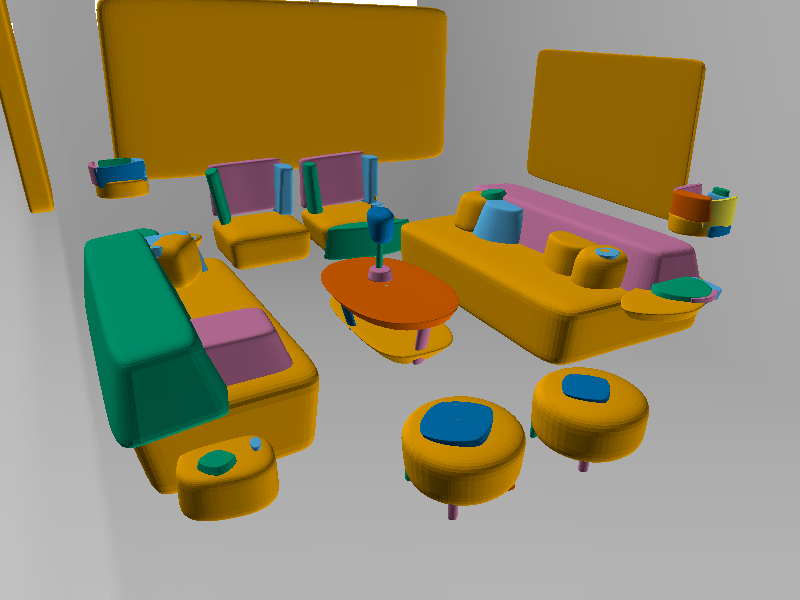} & \includegraphics[width=\scenew]{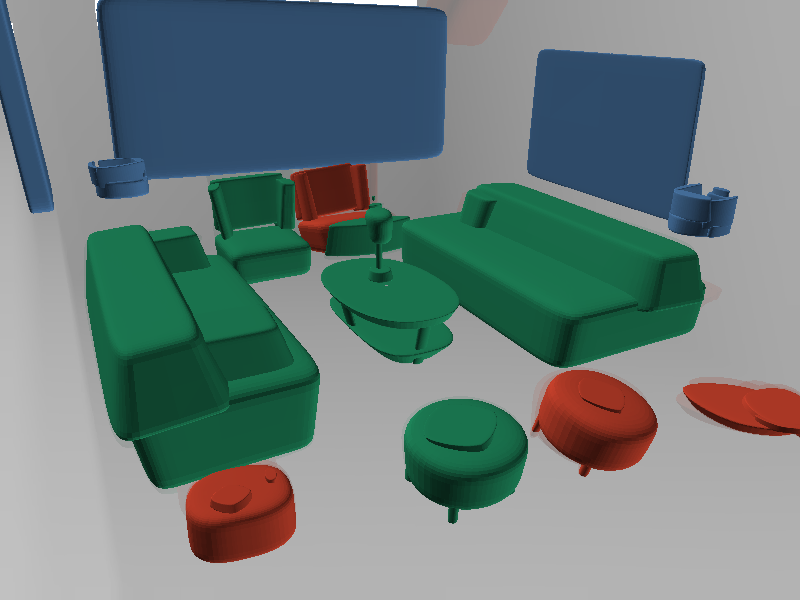} &
    \includegraphics[width=\scenew]{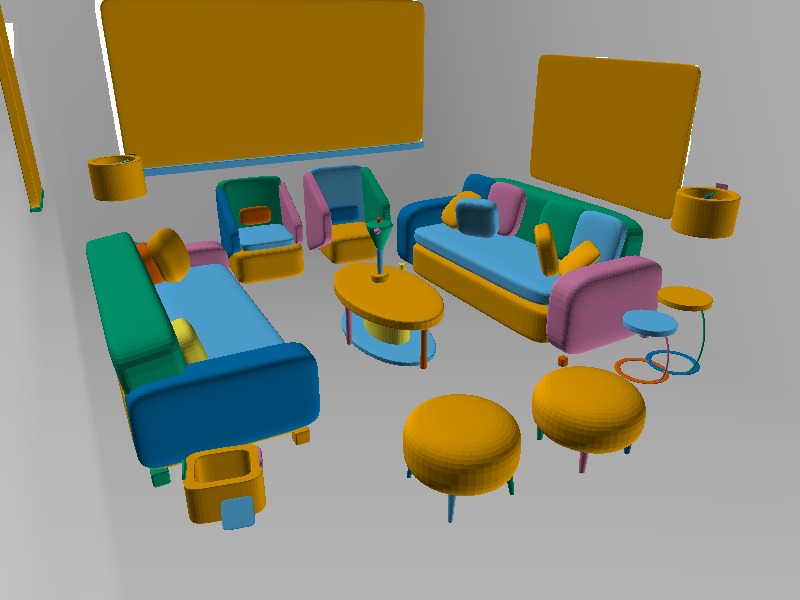} & \includegraphics[width=\scenew]{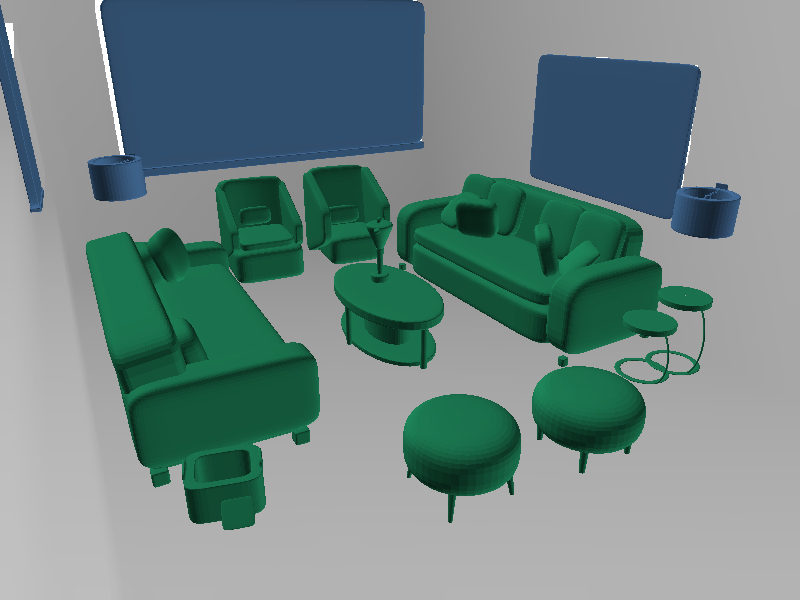} \\[0.1em]
    \includegraphics[width=\scenew]{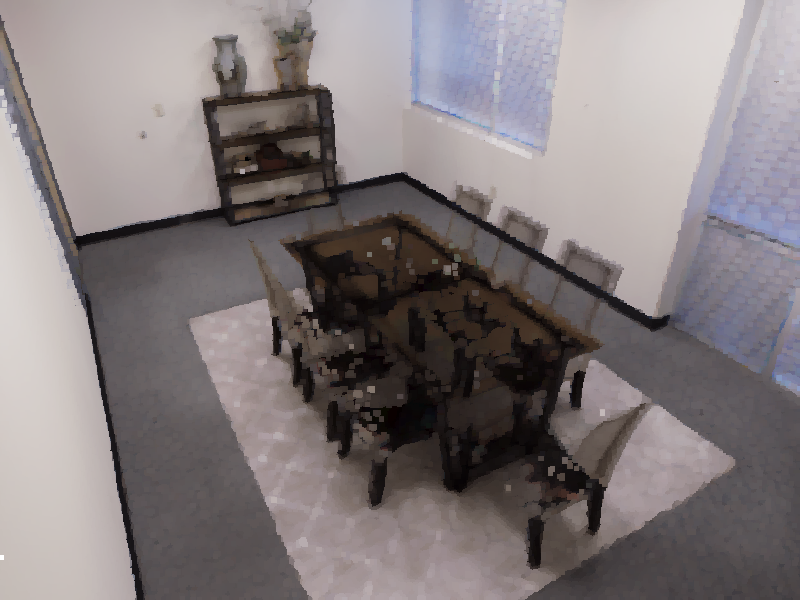} &
    \includegraphics[width=\scenew]{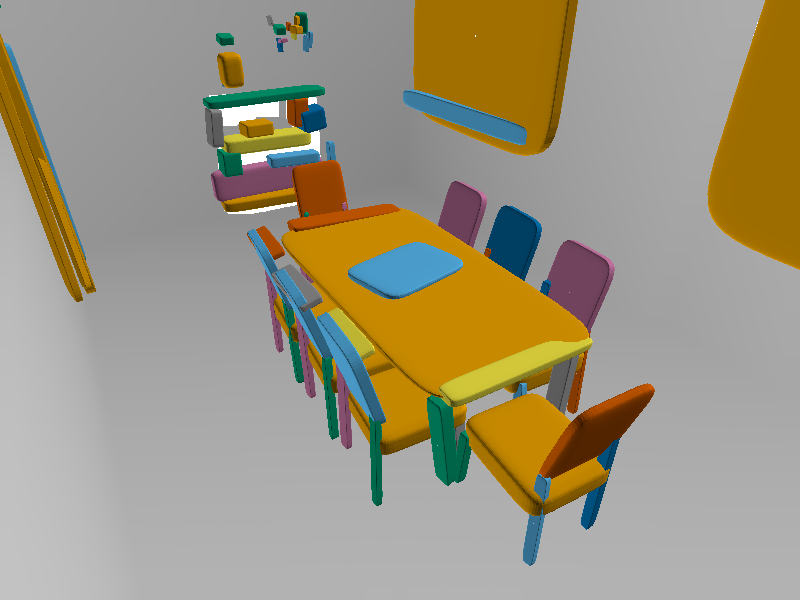} & \includegraphics[width=\scenew]{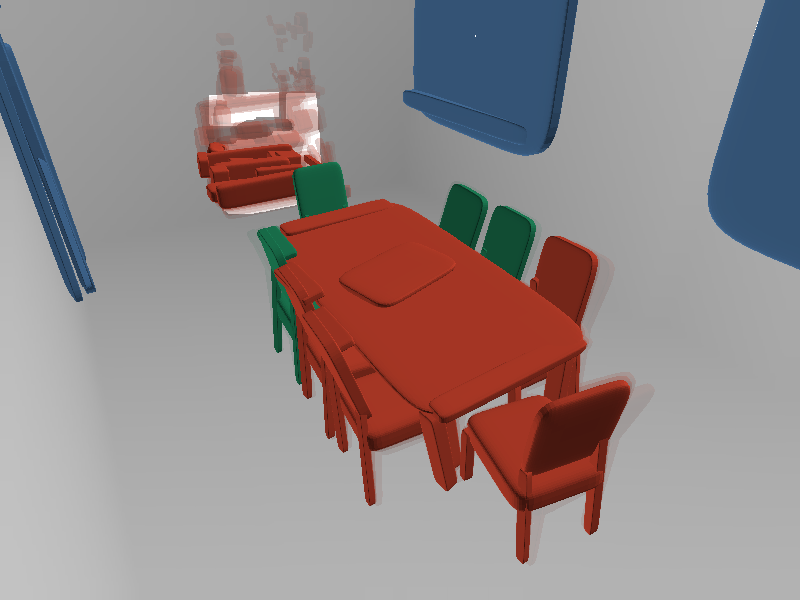} &
    \includegraphics[width=\scenew]{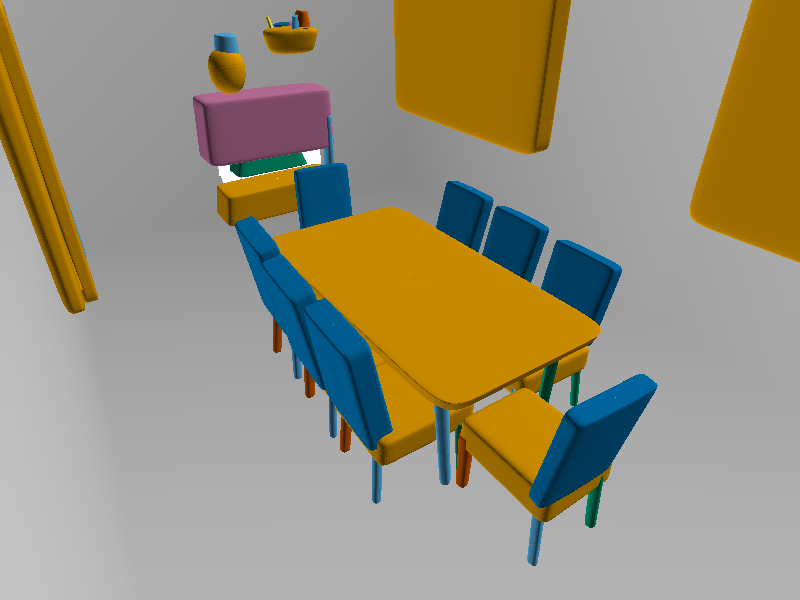} & \includegraphics[width=\scenew]{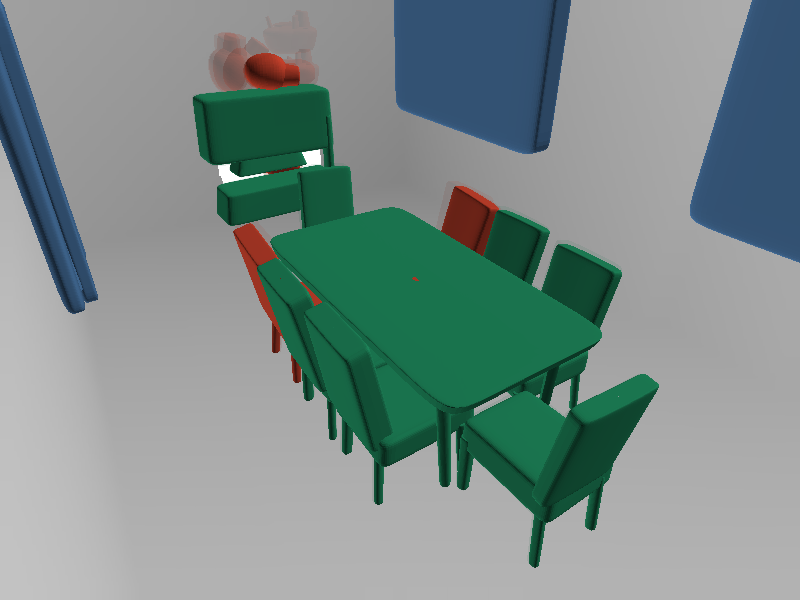} &
    \includegraphics[width=\scenew]{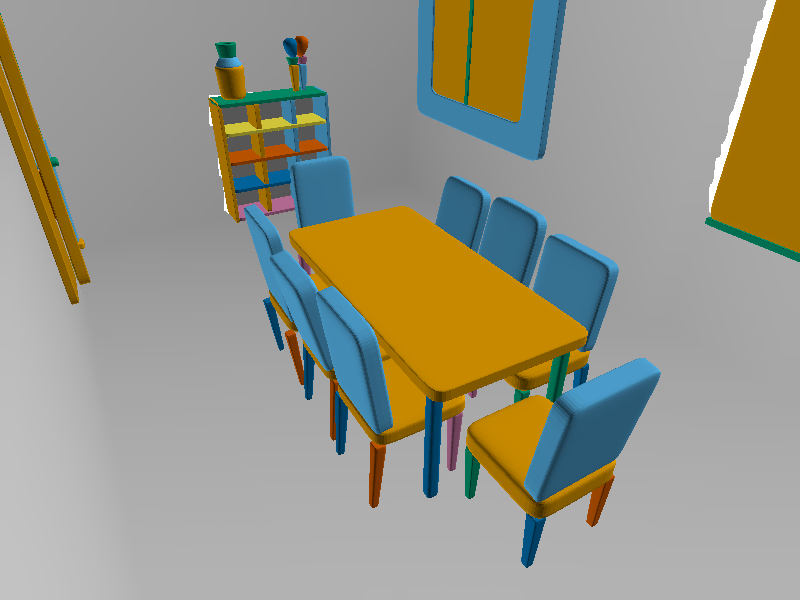} & \includegraphics[width=\scenew]{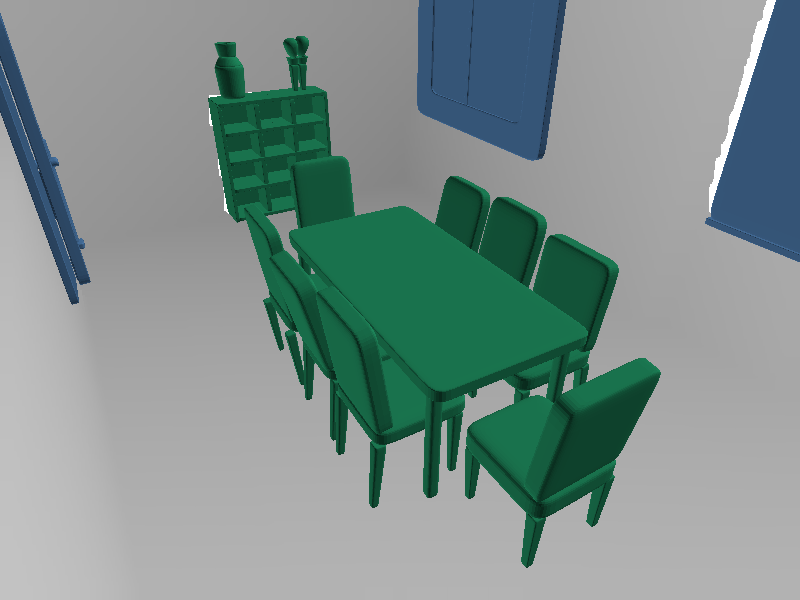} \\
  \end{tabular}%
  }
  \caption{\textbf{Scene reconstruction.} Reconstructions and simulation results for two Replica scenes. Reconstruction colors distinguish parts. In simulation, green denotes resting objects, red unstable objects and blue objects fixed by inferred support constraints shared across methods. Faint overlays trace motion.}
  \label{fig:scenes}
\end{figure*}

\subsection{Scene-level reconstruction}
\label{sec:scenes}

Following SuperDec~\cite{fedele2025superdec}, we reconstruct Replica~\cite{straub2019replica} scenes object by object using Mask3D~\cite{schult2023mask3d} instances. PhysDec-Loop uses images rendered from each instance's scanned point cloud, with the same points providing geometric feedback. Baselines receive 4096 sampled points per instance. The reconstructions are then placed back in the scene.
We then use a scene-level VLM to infer support relations from room images and simulation feedback. A physics solver resolves inter-object penetrations and adjusts object placements to these supports without changing part geometry. We iteratively refine the support relations through simulation feedback and retain the scene with the most fixed or stable objects. \Cref{fig:scenes} shows two scenes, counting an object as resting when its translation and rotation stay below 3\,cm and $5^\circ$.
Compared with the baselines, our method produces more complete and visually faithful scene reconstructions with improved physical stability.


\begin{figure*}[!t]
  \centering
  \begin{minipage}[t]{0.485\textwidth}
    \vspace{0pt}
    \def\robotfigscale{1}

  \centering\scriptsize
  \setlength{\tabcolsep}{0pt}
  \def\robotw{\dimexpr(\linewidth-1.1em)/3\relax}
  
  \providecommand{\robotfigscale}{1}
  \resizebox{\robotfigscale\linewidth}{!}{%
\begin{tabular}{@{}c@{\hspace{0.25em}}c@{}c@{}c@{}}
    & Ours & SuperDec~\cite{fedele2025superdec} & SuperFlex~\cite{tavernini2026superflex} \\[1pt]
    \rotatebox{90}{\parbox{1.3cm}{\centering (a) Grip}} &
    \includegraphics[width=\robotw]{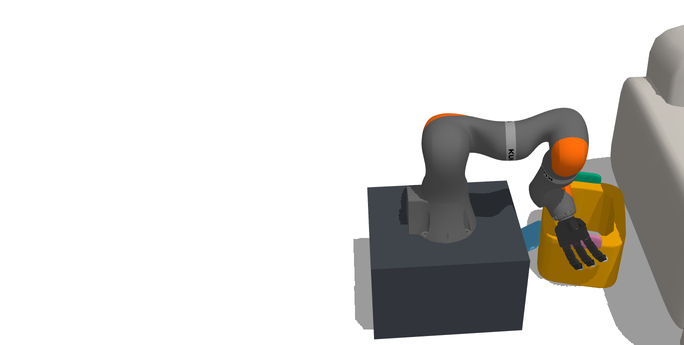} &
    \includegraphics[width=\robotw]{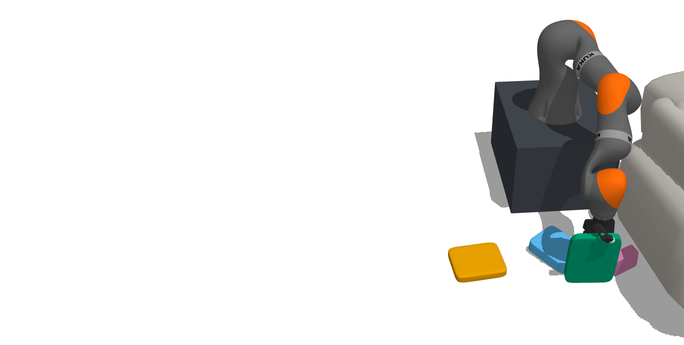} &
    \includegraphics[width=\robotw]{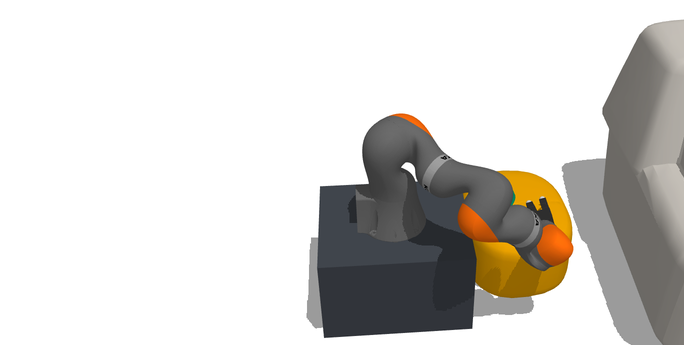} \\
    \rotatebox{90}{\parbox{1.8cm}{\centering Set down}} &
    \includegraphics[width=\robotw]{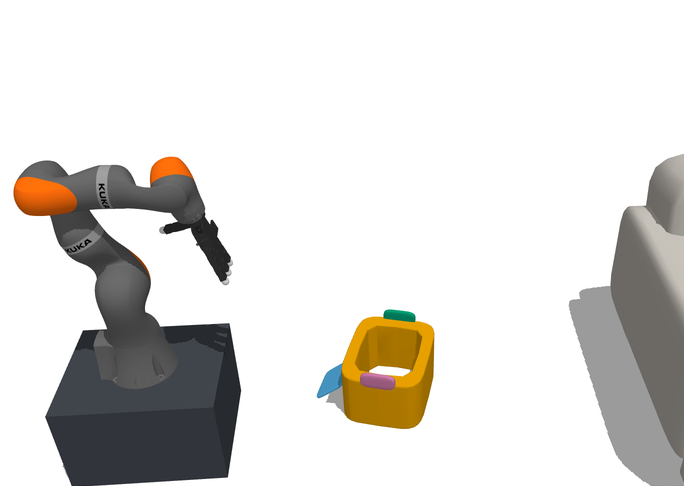} &
    \includegraphics[width=\robotw]{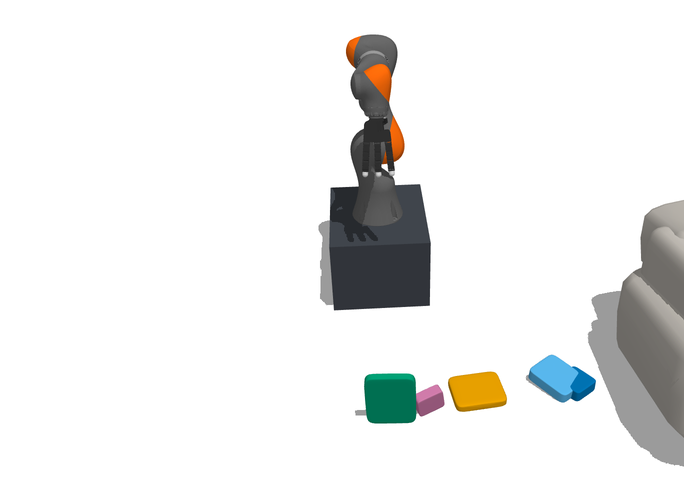} &
    \includegraphics[width=\robotw]{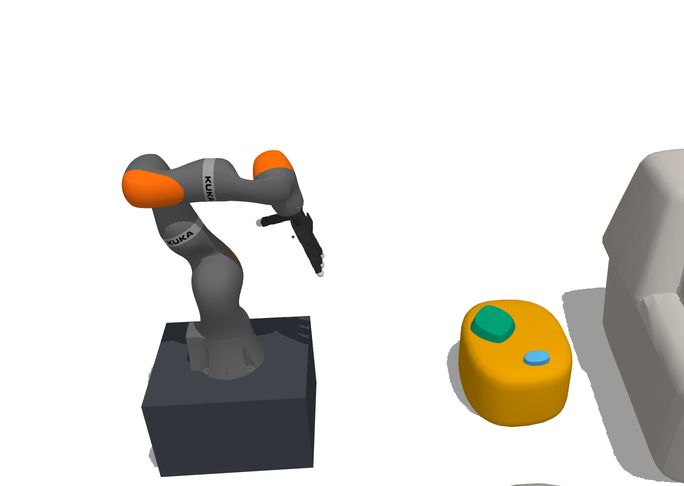} \\
    & \textcolor{passc}{placed} & \textcolor{failc}{falls apart, lifts $36\%$} & \textcolor{failc}{no grasp} \\[3pt]
    \rotatebox{90}{\parbox{1.5cm}{\centering (b) Carry}} &
    \includegraphics[width=\robotw]{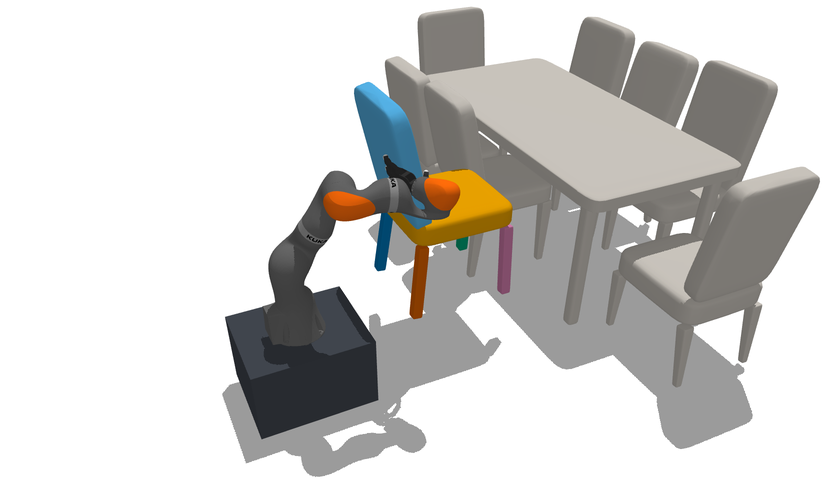} &
    \includegraphics[width=\robotw]{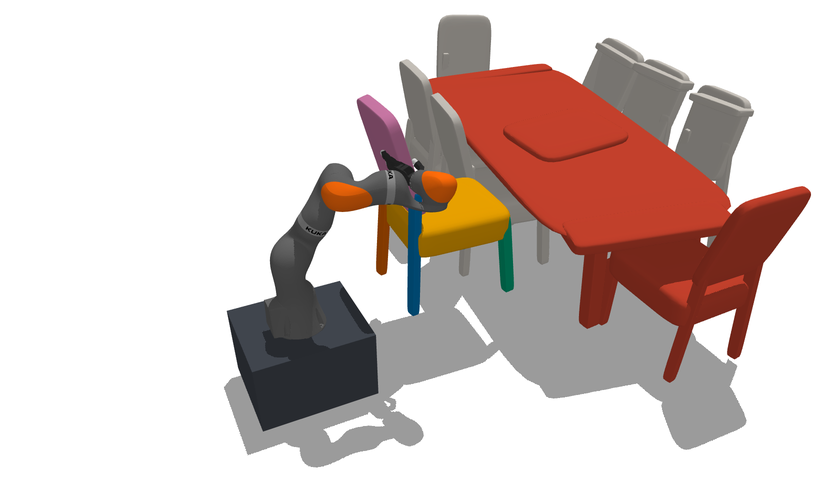} &
    \includegraphics[width=\robotw]{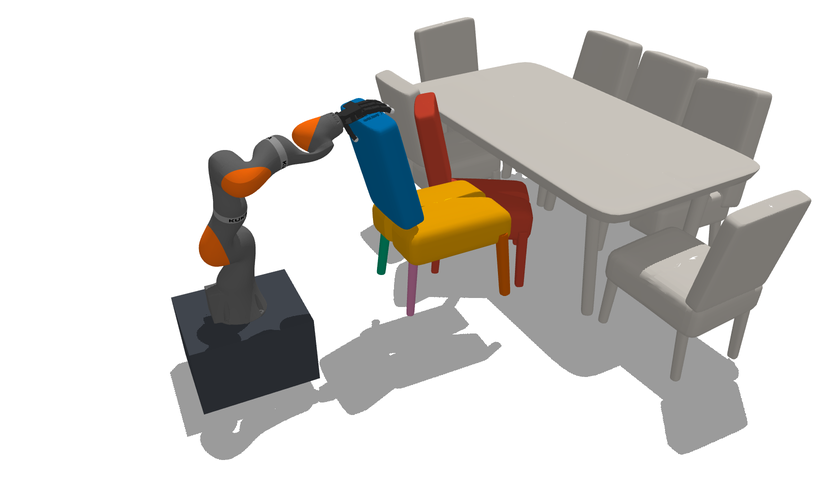} \\
    \rotatebox{90}{\parbox{1.6cm}{\centering Set down}} &
    \includegraphics[width=\robotw]{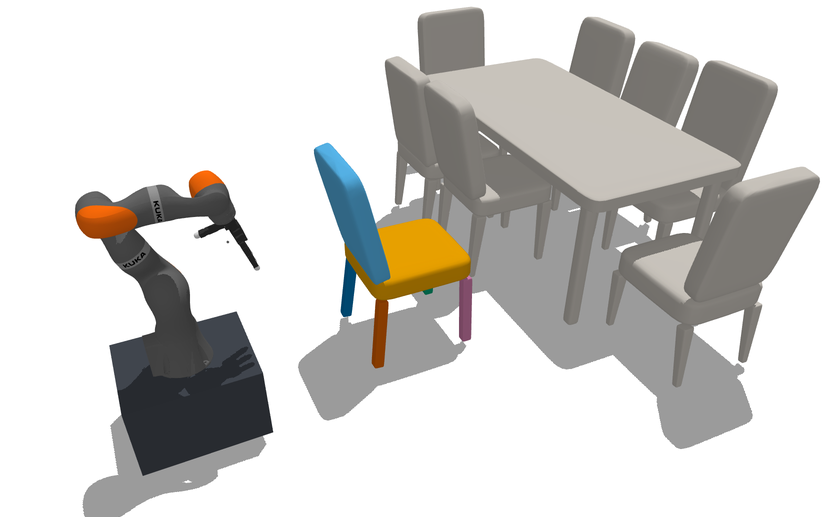} &
    \includegraphics[width=\robotw]{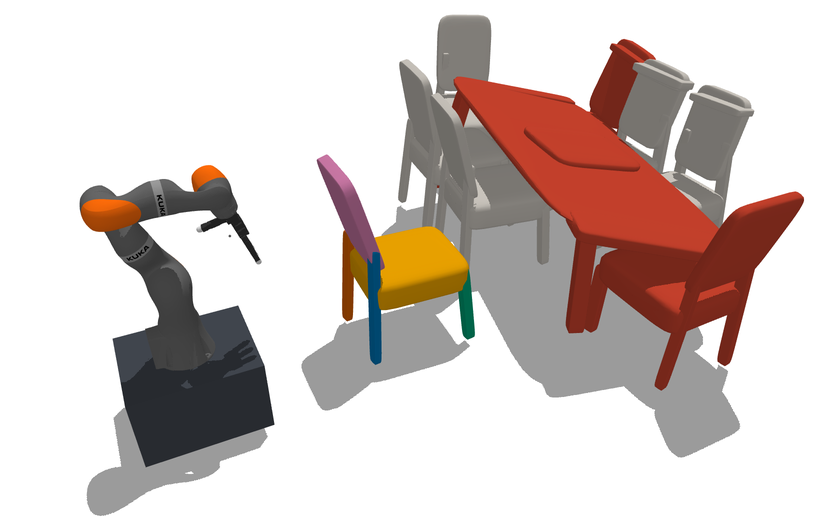} &
    \includegraphics[width=\robotw]{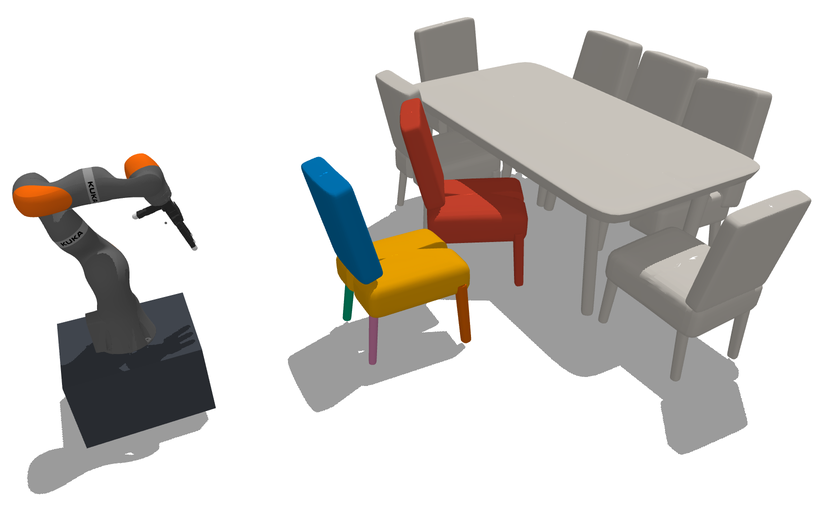} \\
    & \textcolor{passc}{placed} & \textcolor{failc}{table collapses} & \textcolor{failc}{hits a chair} \\
  \end{tabular}%
  }
  \caption{\textbf{Manipulation in reconstructed scenes.} Selected bin and chair examples during grasping or carrying and after placement.}
  \providecommand{\robotlabel}{fig:robot}
  \label{\robotlabel}

  \end{minipage}\hfill
  \begin{minipage}[t]{0.485\textwidth}
    \vspace{0pt}
    \def\robotfigscale{1}

  \centering\scriptsize
  \setlength{\tabcolsep}{0pt}
  \def\planew{\dimexpr(\linewidth-1.1em)/3\relax}
  
  \providecommand{\robotfigscale}{1}
  \resizebox{\robotfigscale\linewidth}{!}{%
\begin{tabular}{@{}c@{\hspace{0.25em}}c@{}c@{}c@{}}
    & Ours & SuperDec~\cite{fedele2025superdec} & SuperFlex~\cite{tavernini2026superflex} \\[1pt]
    \rotatebox{90}{\parbox{1.88cm}{\centering (a) Lift}} &
    \includegraphics[width=\planew]{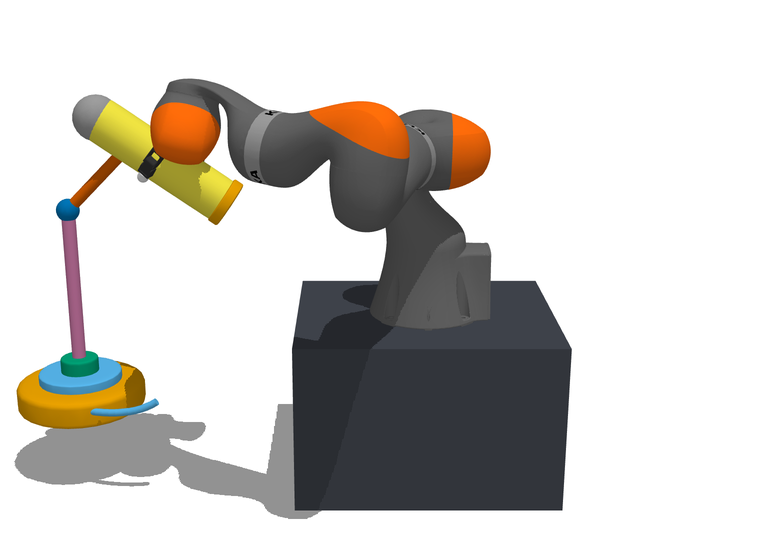} &
    \includegraphics[width=\planew]{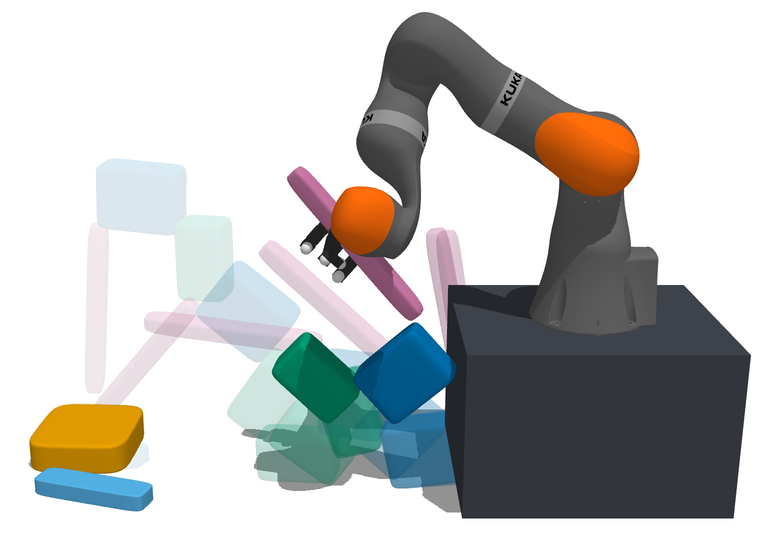} &
    \includegraphics[width=\planew]{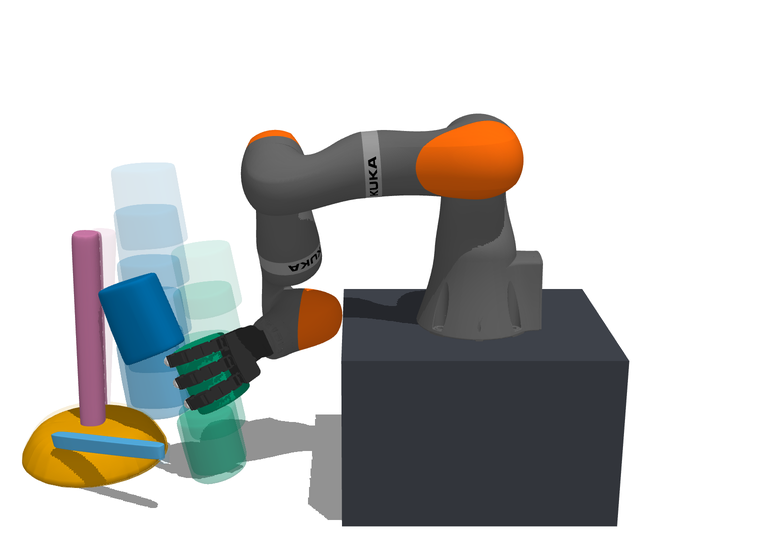} \\
    \rotatebox{90}{\parbox{1.44cm}{\centering Set down}} &
    \includegraphics[width=\planew]{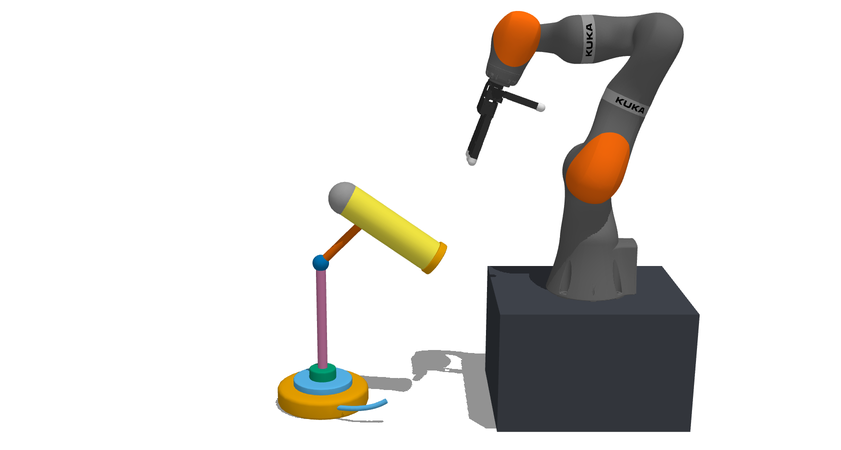} &
    \includegraphics[width=\planew]{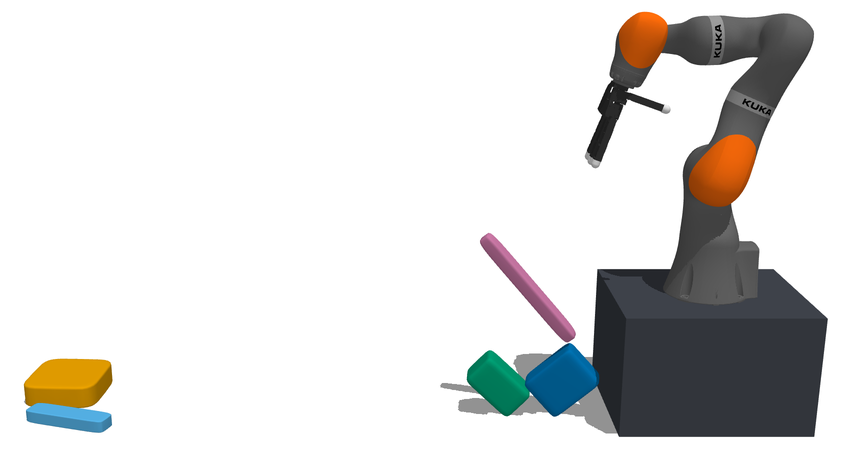} &
    \includegraphics[width=\planew]{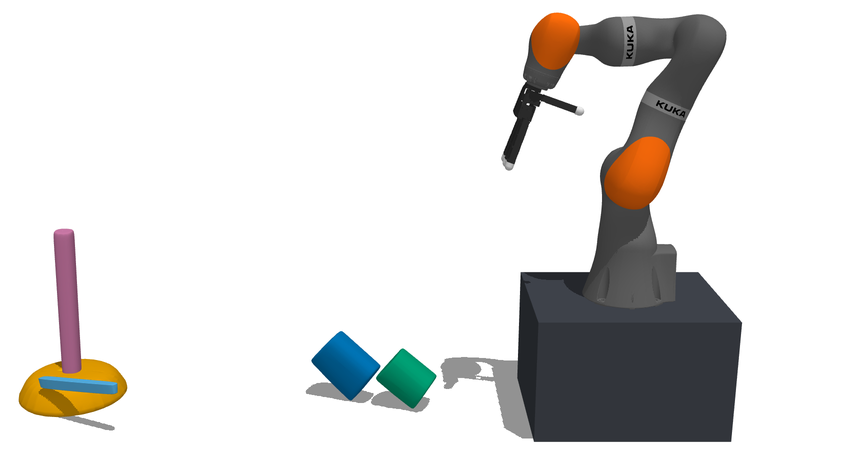} \\
    & \textcolor{passc}{placed} & \textcolor{failc}{falls apart, lifts $52\%$} & \textcolor{failc}{falls apart, lifts $34\%$} \\[3pt]
    \rotatebox{90}{\parbox{1.64cm}{\centering (b) Lift}} &
    \includegraphics[width=\planew]{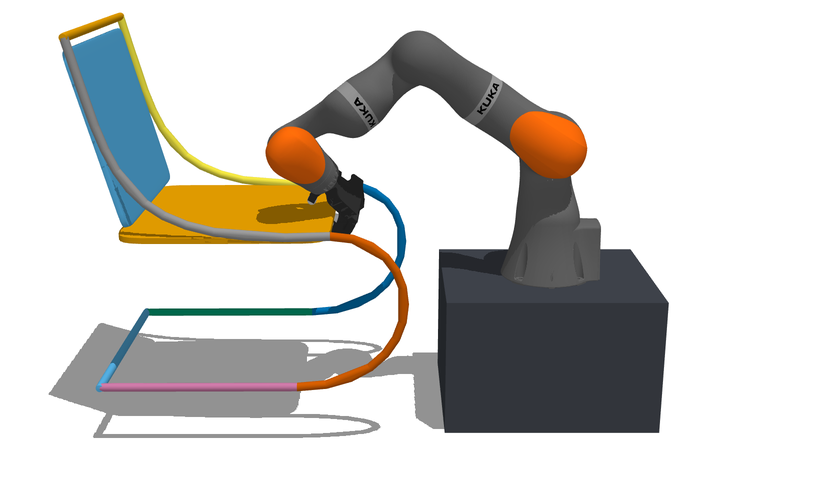} &
    \includegraphics[width=\planew]{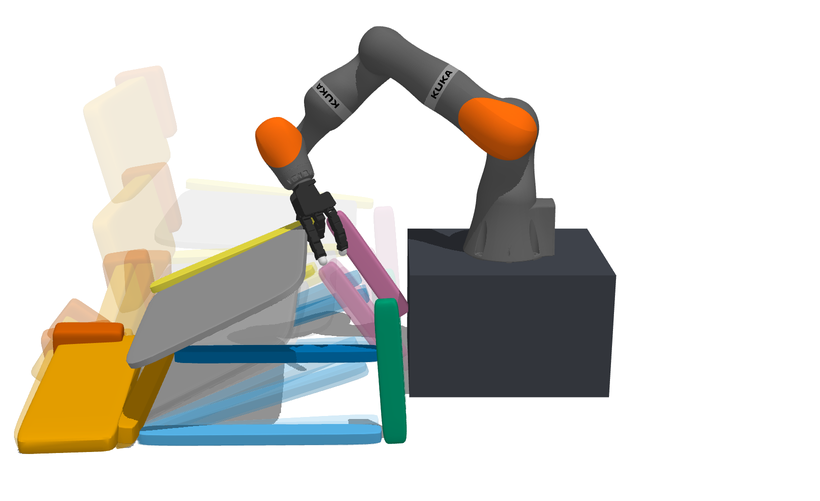} &
    \includegraphics[width=\planew]{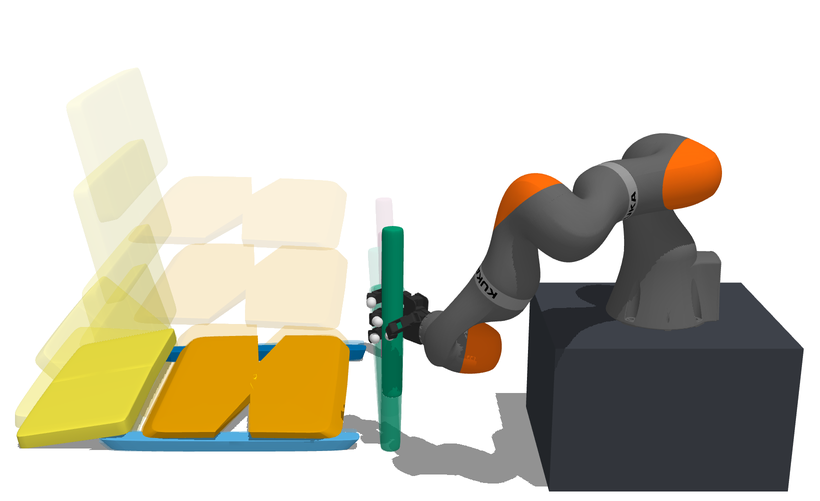} \\
    \rotatebox{90}{\parbox{1.31cm}{\centering Set down}} &
    \includegraphics[width=\planew]{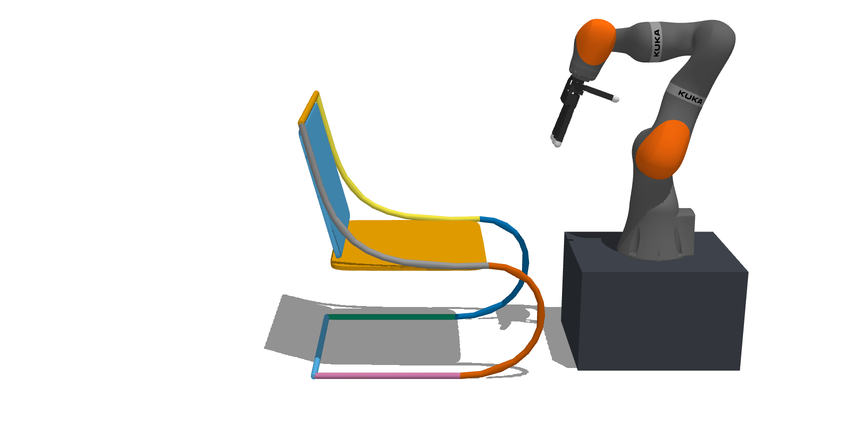} &
    \includegraphics[width=\planew]{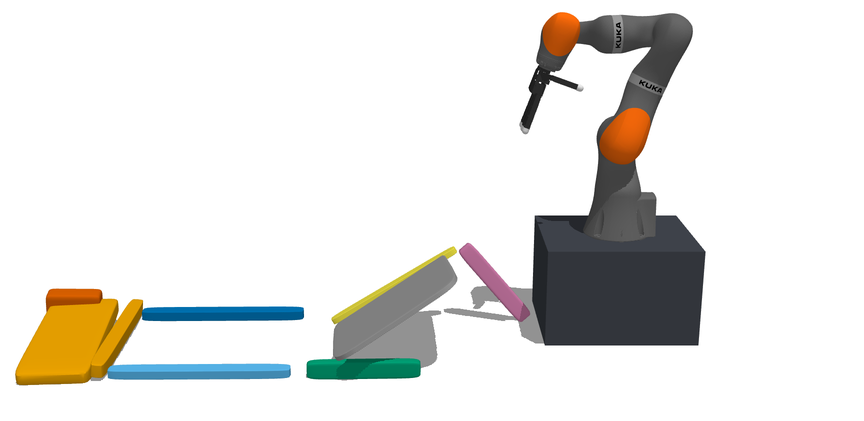} &
    \includegraphics[width=\planew]{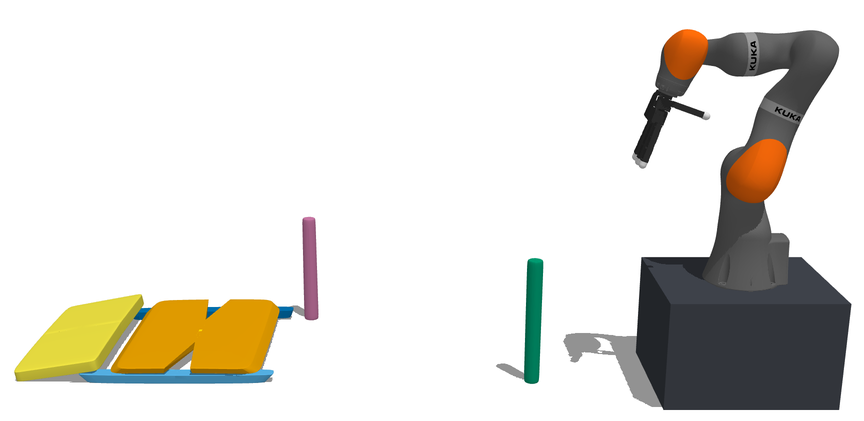} \\
    & \textcolor{passc}{placed} & \textcolor{failc}{falls apart, lifts $32\%$} & \textcolor{failc}{falls apart, lifts $4\%$} \\
  \end{tabular}%
  }
  \caption{\textbf{Manipulation on a ground plane.} Selected ShapeNet objects during lifting and after placement. Percentages indicate the fraction of object mass lifted.}
  \providecommand{\robotplanelabel}{fig:robot_plane}
  \label{\robotplanelabel}

  \end{minipage}
\end{figure*}

\subsection{Robot manipulation}
\label{sec:robot}

We evaluate pick-and-place manipulation with a KUKA LBR iiwa 14 and an Allegro hand in MuJoCo~\cite{menagerie2022github}, using scenes from \cref{sec:scenes} and ShapeNet objects from \cref{sec:exp_baselines}.
All methods use the same superquadric grasp planner~\cite{vezzani2017grasping}, with matched object masses.
PhysDec supports successful scene manipulation (\cref{fig:robot}), while baseline failures include disconnected parts, missing grasps and collisions.
On a ground plane (\cref{fig:robot_plane}), our lamp and chair remain connected and are successfully placed in all five tested carry directions; baselines allow only fragments to be lifted.
Additional trial details are provided in the supplementary material.

\subsection{Ablation and user studies}
\label{sec:analysis}

\begin{table}[!htbp]
  \centering\footnotesize
  \captionsetup{position=top,skip=6pt}
  \caption{Ablations on 195 ShapeNet test objects, with one run per variant and no resampling. Ablated verifiers are removed from feedback, scoring and selection. The second view is about $90^\circ$ from the single view. Model configurations use GPT-6.1 Sol with Sonnet 5.5 review, or Qwen3.8-Flash-Next (FP8) with Qwen3.8-27B review. Sonnet 5.5 grades every row.}
  \label{tab:ablation}
  \setlength{\tabcolsep}{2.6pt}
  \resizebox{\linewidth}{!}{%
\begin{tabular*}{\linewidth}{@{\extracolsep{\fill}}lcccc@{}}
    \toprule
    Variant & \chamfer$\downarrow$ & \#Prim. & Phys.\ pass (\%)$\uparrow$ & VLM$\uparrow$ \\
    \midrule
    PhysDec-Loop & 1.421 & 7.2 & 99.0 & 78.2 \\
    \multicolumn{5}{@{}l}{\emph{Verifiers}} \\
    \quad w/o physics & 1.415 & 7.4 & 93.3 & 79.0 \\
    \quad w/o VLM review & 1.417 & 7.1 & 97.9 & 75.1 \\
    \quad w/o point cloud & 1.479 & 7.4 & 98.5 & 80.0 \\
    \multicolumn{5}{@{}l}{\emph{Input}} \\
    \quad Single view & 1.601 & 7.0 & 97.9 & 72.9 \\
    \quad Two views & 1.394 & 7.0 & 98.5 & 75.6 \\
    \multicolumn{5}{@{}l}{\emph{Model configuration}} \\
    \quad GPT-6.1 Sol & 1.329 & 9.3 & 99.0 & 82.0 \\
    \quad Qwen3.8-Flash-Next & 1.663 & 7.0 & 98.5 & 69.5 \\
    \bottomrule
  \end{tabular*}%
  }
\end{table}

Removing physical checks reduces physical validity, while removing VLM review lowers visual quality (\cref{tab:ablation}).
Point-cloud feedback mainly lowers Chamfer distance; physical validity and visual quality remain comparable without it.
Using a single view reduces geometric and visual quality. Adding a second view about $90^\circ$ from it recovers geometric fidelity, and visual quality improves further with all four views, highlighting the benefit of complementary views.

Replacing the task VLM with GPT-6.1 Sol improves geometric and visual quality while maintaining physical validity.
The cheaper, open-source Qwen configuration retains a high physical-check pass rate of 98.5\%, using Qwen models for both proposal generation and review.

\begin{table}[!htbp]
  \centering\footnotesize
  \captionsetup{position=top,skip=6pt}
  \caption{\textbf{Agreement with human judgments.} Human ratings are linearly rescaled from 1--5 to 20--100. Pairwise agreement compares methods on the same object, excluding ties. See supplementary material for the study protocol.}
  \label{tab:userstudy}
  \setlength{\tabcolsep}{3pt}
  \resizebox{\linewidth}{!}{%
\begin{tabular*}{\linewidth}{@{\extracolsep{\fill}}lccc@{}}
    \toprule
     & User & Sonnet 5.5 & GPT-6.1 Sol \\
    \midrule
    \multicolumn{4}{@{}l}{\emph{Mean score}} \\
    \quad SuperDec~\cite{fedele2025superdec} & 47.1 & 51.4 & 50.8 \\
    \quad SuperFlex~\cite{tavernini2026superflex} & 59.3 & 57.2 & 56.6 \\
    \quad PhysDec-Loop & 87.9 & 79.2 & 78.6 \\
    \multicolumn{4}{@{}l}{\emph{Agreement with users}} \\
    \quad Spearman $\rho\uparrow$ & -- & 0.81 & 0.81 \\
    \quad Pairwise agreement (\%)$\uparrow$ & -- & 86.7 & 86.4 \\
    \bottomrule
  \end{tabular*}%
  }

\end{table}

\paragraph{User study.}
To assess agreement between VLM scores and human judgments, five participants rated reconstructions from SuperDec, SuperFlex and PhysDec-Loop on 65 test objects using a five-point scale.
We linearly rescale human ratings from 1--5 to 20--100 for comparison with VLM scores.
The results were presented anonymously alongside the input images.
PhysDec-Loop receives the highest mean human rating, consistent with both VLM judges (\cref{tab:userstudy}).
Both judges correlate strongly with mean human ratings (Spearman $\rho=0.81$) and match over $86\%$ of human pairwise preferences, supporting their use for comparing visual reconstruction quality.

\FloatBarrier
\Needspace{12\baselineskip}
\section{Conclusion}
\label{sec:conclusion}

\method combines VLM reasoning with geometric tools and physical verifiers for physically grounded shape decomposition from a few images.
PhysDec-Loop improves geometric fidelity, physical plausibility and visual quality, while PhysDec-Evolved reuses verified corrections through evolved prompts and retrieval to approach this quality with substantially fewer model calls and without updating VLM weights.
Experiments span complete objects, partial observations and scanned scenes.

\paragraph{Limitations.}
\method currently focuses on objects composed of rigid parts, without modeling joint motion or material deformation.
Our physical checks assess connectivity and stability under prescribed support conditions, including standing, wall attachment and suspension; passing these checks does not establish robustness to arbitrary external forces or interactions.
We primarily use closed-source VLMs. Although the open-source Qwen configuration maintains high physical validity, it trails these models in geometric fidelity and visual quality (\cref{tab:ablation}).
Extending the representation to articulated and deformable objects, broadening physical tests, and improving open-source models for this task are promising directions for future work.

\clearpage
\onecolumn
\raggedbottom
\setcounter{section}{0}
\setcounter{subsection}{0}
\setcounter{figure}{0}
\setcounter{table}{0}
\setcounter{equation}{0}
\setcounter{lstlisting}{0}
\renewcommand{\thesection}{\Alph{section}}
\renewcommand{\thefigure}{S\arabic{figure}}
\renewcommand{\thetable}{S\arabic{table}}
\renewcommand{\theequation}{S\arabic{equation}}
\renewcommand{\theHsection}{supp.\Alph{section}}
\renewcommand{\theHfigure}{supp.\arabic{figure}}
\renewcommand{\theHtable}{supp.\arabic{table}}
\renewcommand{\theHequation}{supp.\arabic{equation}}
\phantomsection
\pdfbookmark[0]{Supplementary Material}{supplement}
\begin{center}
  {\Large\textbf{\thetitle}}\\[0.5em]
  {\large Supplementary Material}
\end{center}
\vspace{0.5em}

This supplement prints the prompts behind the main paper and shows what self-evolution learned.
\Cref{sec:s_correction} gives the prompts used by PhysDec-Loop.
\Cref{sec:s_blocks} shows how self-evolution splits them into blocks, and \cref{sec:s_before} prints the two evolved blocks before and after evolution.
\Cref{sec:s_library} shows the insights and the retrieval library.
\Cref{sec:s_rubric} gives the rubric behind every VLM score.
\Cref{sec:s_chamfer} discusses the limitations of Chamfer distance, its evaluation protocol and an optional optimization tool for the agent.
\Cref{sec:s_evaluation} provides additional evaluation details, and \cref{sec:s_userstudy} compares VLM scores with human ratings.

\textbf{How to read the prompts.}
The prompts are drawn from the code and stored transcripts of the runs behind Tables~\mainbaselines{} and~\mainevolution{}.
\textcolor{annotc}{\itshape Blue italics} are our annotations, not part of a prompt: a \textcolor{annotc}{\itshape<placeholder>} filled in at run time, or a \textcolor{annotc}{\itshape[note]} about the passage after it.

\section{PhysDec-Loop prompts}
\label{sec:s_correction}

\begin{figure}[H]
  \centering
  \begin{minipage}[t]{0.49\linewidth}\centering
    \includegraphics[width=0.245\linewidth]{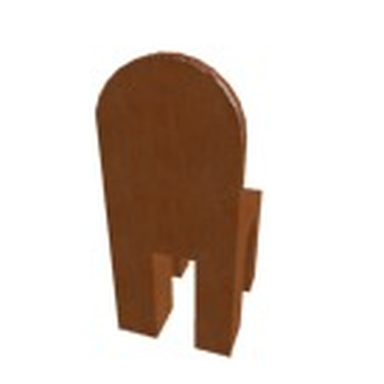}\hfill
    \includegraphics[width=0.245\linewidth]{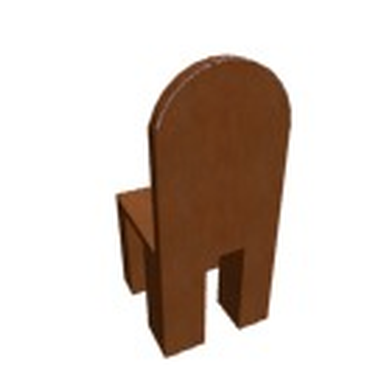}\hfill
    \includegraphics[width=0.245\linewidth]{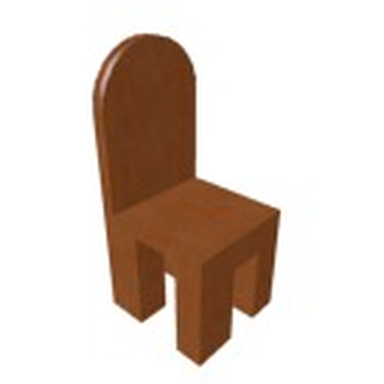}\hfill
    \includegraphics[width=0.245\linewidth]{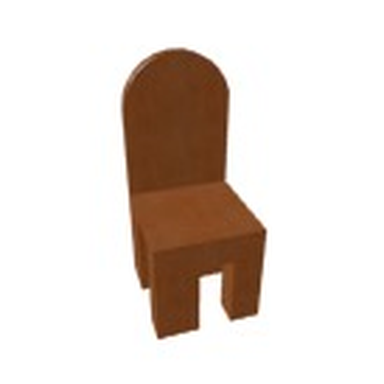}\\[1pt]
    {\scriptsize (a) The four input photos}
  \end{minipage}\hfill
  \begin{minipage}[t]{0.49\linewidth}\centering
    \includegraphics[width=\linewidth]{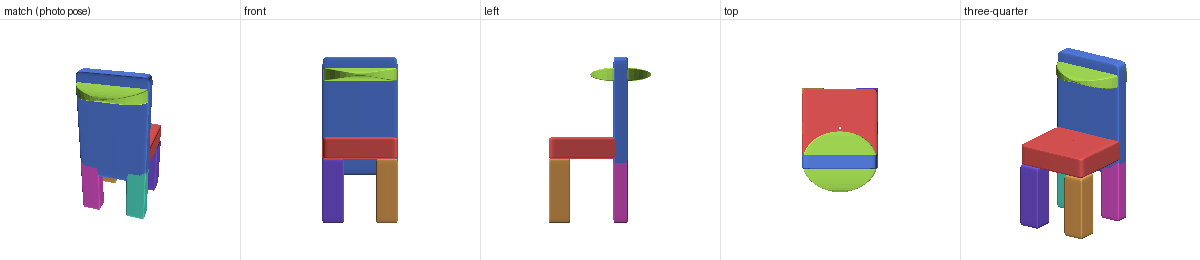}\\[1pt]
    {\scriptsize (b) Current decomposition, a colour per part}
  \end{minipage}\\[6pt]
  \begin{minipage}[t]{0.49\linewidth}\centering
    \includegraphics[width=\linewidth]{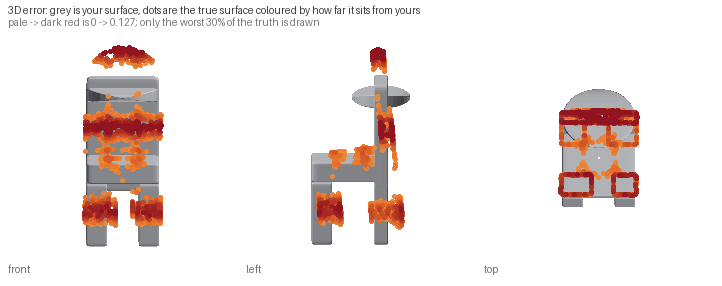}\\[1pt]
    {\scriptsize (c) Error map: true surface coloured by its distance to the result}
  \end{minipage}\hfill
  \begin{minipage}[t]{0.49\linewidth}\centering
    \includegraphics[width=\linewidth]{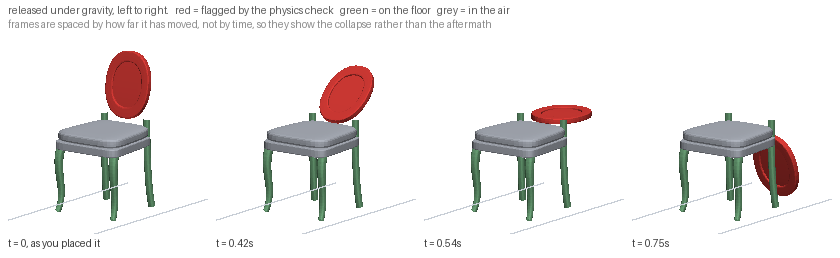}\\[1pt]
    {\scriptsize (d) Physics sheet: as placed, then under gravity}
  \end{minipage}\\[6pt]
  \begin{minipage}[t]{0.49\linewidth}\centering
    \includegraphics[width=\linewidth]{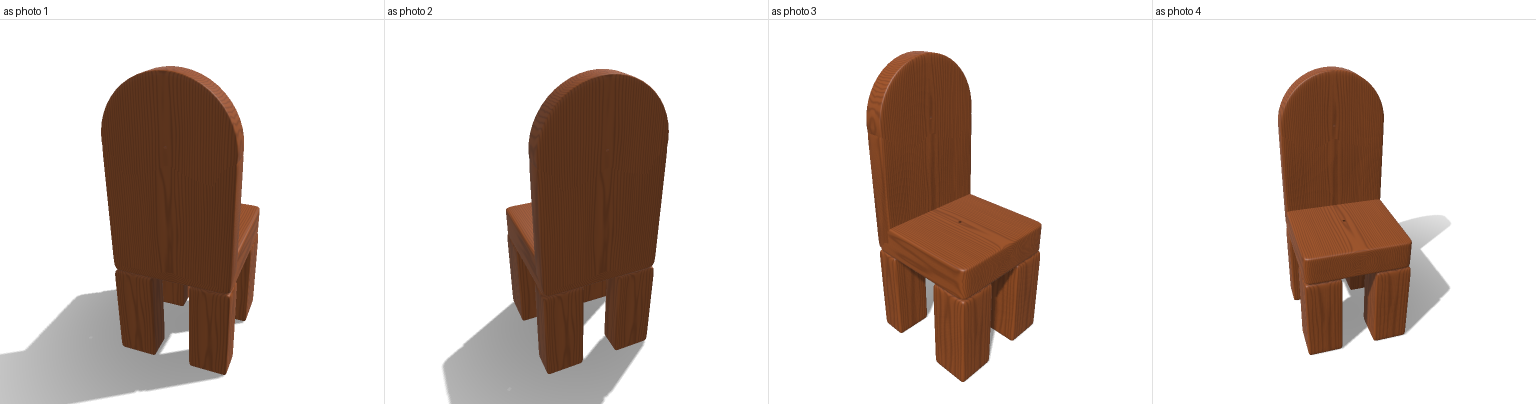}\\[1pt]
    {\scriptsize (e) For the review: the result through each photo's camera}
  \end{minipage}\hfill
  \begin{minipage}[t]{0.49\linewidth}\centering
    \includegraphics[width=0.62\linewidth]{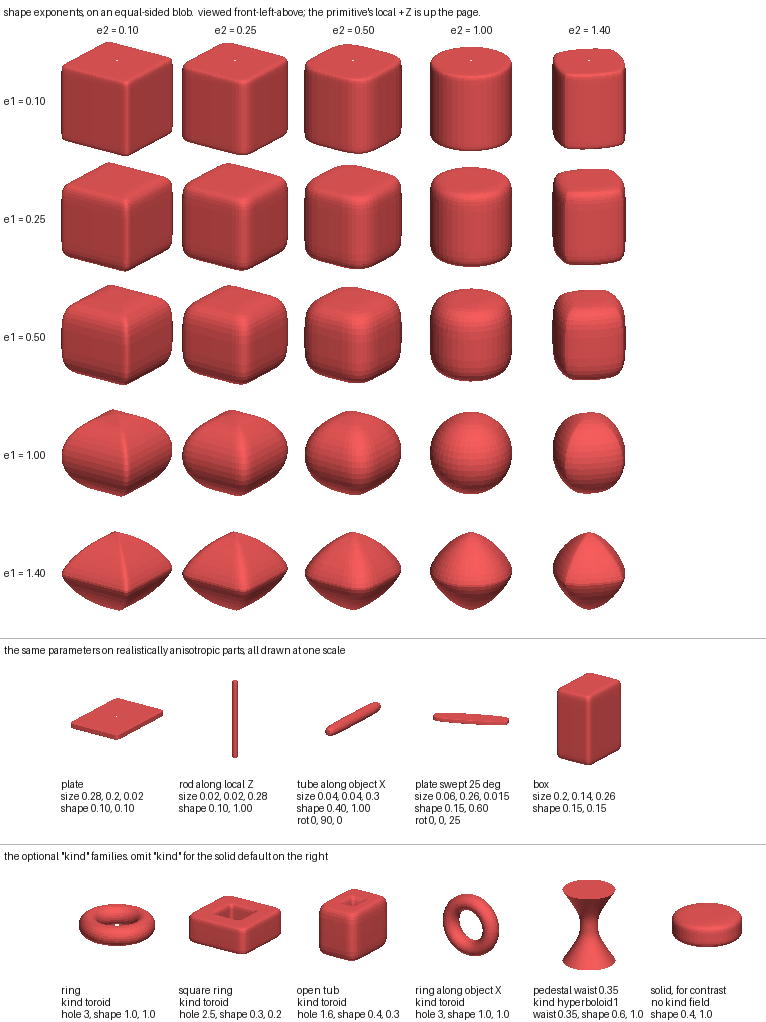}\\[1pt]
    {\scriptsize (f) Exponent reference sheet}
  \end{minipage}
  \caption{\textbf{Images attached to the calls.} Panels (a), (b), (c) and (e) show one test object: its first proposal in (b) and (c), its delivered result in (e). Panel (d) shows another object's first proposal, which falls over.}
  \label{fig:s_images}
\end{figure}

One PhysDec-Loop run (Sec.~\mainloop{} of the main paper) is one conversation with the task model, Claude Opus 5.5, under one system prompt.
\Cref{tab:s_calls} lists its calls and \cref{fig:s_images} the images they attach.
Each object gets one Analyse call, one Propose call and up to $T{=}3$ correction rounds; one call carries out a round's Critique and its next Propose together, so it returns the critique and the full corrected decomposition.
Images from earlier rounds are dropped from the conversation; the photos stay.
The reviewer is a separate model and sees no conversation: it receives only the images and the rubric of \cref{sec:s_rubric}.

\begin{table}[H]
  \centering\footnotesize
  \caption{\textbf{Calls in one PhysDec-Loop run.} The reviewer grades the first proposal and every revision, twice each.}
  \label{tab:s_calls}
  \begin{tabular}{@{}L{1.5cm}L{1.9cm}L{2.0cm}L{6.0cm}L{4.1cm}@{}}
    \toprule
    Call & Model & Prompt & Images attached & Reply \\
    \midrule
    Analyse & Opus 5.5 & \cref{lst:system,lst:analyse} & the 4 input photos & category, support, symmetry, proportions, parts and their materials \\
    Propose & Opus 5.5 & \cref{lst:propose} & the photos, the exponent reference sheet & primitives \\
    Review & Sonnet 5.5, separate calls & \cref{lst:rubric_system,lst:rubric} & the photos, the proposal drawn through each photo's camera & seven graded criteria with evidence \\
    Critique, rounds 1--3 & Opus 5.5 & \cref{lst:critique} & the photos, the current decomposition, the error map, and the physics sheet when the check flags a part on a free-standing object & critique, full corrected primitives, a flag to stop \\
    \bottomrule
  \end{tabular}
\end{table}

\begin{lstlisting}[caption={\textbf{System prompt}, shared by Analyse, Propose and Critique. The extended primitives are on, as in all reported runs.},label=lst:system]
You decompose objects into a small set of superquadric primitives, the representation used by SuperDec (Fedele et al., ICCV 2025). You see one photo or rendering of an object and infer its full 3D shape as primitives.

# Canonical object frame
All coordinates are in the object's own frame, never the camera's.
- +Z is UP (the direction the object's top faces).
- +X is FRONT (the direction the object faces; for a chair, where a sitter's
  knees point; for a car, the windshield side; for a sofa, the open side).
- +Y is the object's LEFT. Right-handed: X cross Y = Z.

# Scale and placement
Work at roughly unit scale: centre the object near the origin and let its
longest dimension span about 1.0, so coordinates land in [-0.5, 0.5]. Only the
RELATIVE proportions of your parts matter -- the pipeline recentres your output
and rescales it into a sphere of radius 0.5 (SuperDec's normalisation) before
rendering, so a uniformly-too-small or off-centre answer costs you nothing.
Getting one part's size wrong RELATIVE to the others does cost you.

# A superquadric
Each primitive has 11 numbers:
- "size": [sx, sy, sz] -- SEMI-axes, i.e. HALF-extents along the primitive's own
  local X/Y/Z before rotation. A plank 0.40 long, 0.30 wide and 0.05 thick is
  size [0.20, 0.15, 0.025]. Halve every dimension you measure. This is the most
  common mistake by a wide margin.
- "shape": [e1, e2] -- roundness exponents in [0.1, 1.9]. Both are defined
  about the primitive's OWN LOCAL AXES, before rotation.
  e2 shapes the cross-section in the primitive's local XY plane:
    0.1 square, 0.5 rounded-square, 1.0 circular, >1.2 pinched/concave.
  e1 shapes the profile along the primitive's local Z:
    0.1 flat top and bottom with straight sides, 0.5 softly rounded edges,
    1.0 elliptical, >1.2 tapering to a point at each end.
  So: [0.1, 0.1] cuboid; [0.1, 1.0] cylinder; [1.0, 1.0] ellipsoid;
  [0.3, 0.3] rounded box; [1.4, 1.0] a spindle / double cone.
  Prefer exponents near 0.1-0.3 for man-made flat-sided parts.
  Watch which exponent owns the outline you care about. For a flat blade --
  a tail fin, a lamp's upright, a chair's splat -- the shape you actually see
  is the one in the chord-by-height plane, and that is governed by e1, not e2.
  A fin at [0.15, 0.6] keeps full chord all the way to the tip; the same fin at
  [0.9, 0.6] tapers to about a third of its chord. Changing e2 there only
  thickens or thins the aerofoil section and does nothing to the profile.
- "rotation_deg": [rx, ry, rz] -- intrinsic X-Y-Z Euler angles in degrees,
  right-handed, applied about the primitive's own centre before translating it.
  Worked examples, so you get the signs right:
    a chair backrest reclined 15 deg away from the sitter is [0, -15, 0];
    a wing swept 25 deg aft on the object's LEFT (+Y) side is [0, 0, 25];
    a tube whose axis must run along the object's X (a fuselage, an axle, a
    horizontal rail) is [0, 90, 0] TOGETHER WITH size [r, r, half_length].
  That last case matters, and it is the easiest way to wreck an answer. A tube
  is always built along its local Z, so its length always goes in size[2], and
  the rotation is what swings that axis onto the object's X. Do not ALSO put
  the length in size[0]: [0.48, 0.06, 0.06] with ry=90 is a long part rotated
  until it points straight DOWN, which is how a fuselage ends up standing on
  its nose. Write [0.06, 0.06, 0.48] with ry=90 instead. The same goes for
  ry=-90 and for rx=+-90 about the object's Y.
  Leave [0, 0, 0] only when the part's local axes really do line up with the
  object's, which is true of most boxy parts.
- "position": [x, y, z] -- the primitive's CENTRE in the object frame.
- "name": a short semantic label, e.g. "seat", "front_left_leg", "left_wing".

# Shape families (optional "kind")
Barr defines four superquadric families. Omit "kind" and you get the default,
which is the right answer for the large majority of parts. Reach for another
family only when the default genuinely cannot express the part.

- "kind": "ellipsoid" (default). Everything above applies to it. Solid, convex,
  no hole. Use it unless one of the cases below is clearly true.
- "kind": "toroid" -- adds "hole": a number from 1 to 8. A ring, a tube open
  end to end, a hollow shell seen through. USE THIS WHEN THE PART HAS A HOLE
  YOU COULD PUT YOUR HAND THROUGH: a boat hull's open cockpit, a mug or basket
  interior, a wheel rim, a handle, an armrest loop, the open frame of a chair
  back. hole = 1 pinches the opening shut, 2 gives an opening about a third of
  the width, 4 makes the wall thin and the opening wide. The hole always runs
  along the primitive's local Z, so rotate it the same way you would a tube.
- "kind": "hyperboloid1" -- adds "waist": 0.05 to 0.95, the middle radius as a
  fraction of the end radius. Narrow in the middle, flared at both ends, along
  local Z. Use for a shape whose sides curve INWARD: a table pedestal, a wine
  glass stem, an hourglass, a spool, a waisted vase. waist 0.8 is a slight
  pinch, 0.3 is a pronounced hourglass. Note an ellipsoid with e1 > 1.2 tapers
  to points at the ends, which is the opposite shape; do not confuse them.
- "kind": "hyperboloid2" -- also takes "waist". Two separate flared caps facing
  away from each other. This is almost never a single object part. Do not use
  it unless the part really is two disjoint opposed shells.

Two rules that do not change between families. "size" is always the half-extent
of the part's bounding box, so a ring 0.40 across and 0.10 tall is still
size [0.20, 0.20, 0.05]. And e1/e2 still control roundness, so a toroid with
e2 = 0.2 is a rectangular ring rather than a round one.

Do not reach for an exotic family to win detail. A hole that you could not put
a finger through is not a hole, it is a dent, and a plain ellipsoid describes
it better and more cheaply.

# Tapering, bending, tubes and rings (optional)
Four extras for parts a plain superquadric fits badly. Leave them out and the
part is exactly as before.
- "taper": [t1, t2], each from -0.9 to 0.9. The part narrows or widens
  linearly along its "deform_axis" (local z unless you say otherwise). At the
  +end of that axis the other two half-extents are multiplied by (1 + t), at
  the -end by (1 - t); "size" is the size at the middle. t1 acts on the first
  of the other two local axes and t2 on the second, in x, y, z order -- for the
  default z axis, t1 on x and t2 on y. A lamp shade 0.32 across at the bottom
  and 0.20 at the top is size [0.13, 0.13, h] with taper [-0.23, -0.23]. A leg
  that thins toward the floor has a POSITIVE taper, because its wide end is
  the +z end. Use it for shades, tapered legs, a car's nose, a boat's bow, a
  backrest that widens upward.
- "bend_deg": 0 to 180, with "bend_dir_deg". The deform axis curves into a
  circular arc that turns through bend_deg over the part's whole length, and
  both ends swing toward the side bend_dir_deg names: an angle in the plane of
  the other two local axes, from the first toward the second (for the default
  z axis, 0 is toward local +x and 90 toward local +y). Use it for a curved
  bench seat or backrest, a bowed arm, a crescent, a banana-shaped part.
- "deform_axis": "x", "y" or "z" (default "z"): the local axis taper and bend
  act along. Use the part's long axis, so a long horizontal seat with size
  [0.10, 0.50, 0.03] that curves around a sitter has "deform_axis": "y".
- A tube: "kind": "tube" with "spine", four points [x, y, z] in the object
  frame -- the start, two control points and the end of a smooth curve (a
  cubic Bezier: it passes through the first and last points and is pulled
  toward the middle two; four points on a straight line make a straight
  tube). "profile": [a, b], the half-thickness of its cross-section (a
  measured upward, or along x for an upright tube; b across), and
  "roundness" from 0.1 to 1.9 (1.0 round, 0.1 square). An optional "taper":
  [t] scales the cross-section by (1 + t) at the end and (1 - t) at the
  start. A tube takes no "size", "shape", "rotation_deg" or "position". Use
  it for a curved or bent slender part -- a lamp's arm, a bent-tube chair
  frame, a handle, a curved rail or armrest, a cable -- as ONE primitive
  instead of a chain of short ones, so it cannot come apart at the joints.
  For a straight slender part a normal primitive is still the simpler answer.
- "repeat": {"count": n, "center": [x, y]} on any part, tubes included: n
  identical copies spaced evenly around the vertical axis through (x, y) in
  the object frame (default [0, 0]). Write copy 1 exactly where it sits; copy
  k is copy 1 turned by (k - 1) x 360/n degrees about that axis. The whole ring
  counts as ONE part toward the budget. Use it for parts repeated around a
  centre -- the arms and candles of a chandelier, the five legs of an office
  chair's base, the blades of a ceiling fan, the feet of a pedestal -- and
  count them in the photos rather than writing fewer to save parts. Copies
  that are not evenly spaced, or not around a vertical axis, are written out
  one by one as usual.

# Symmetry
Most objects are bilaterally symmetric about the XZ plane. To mirror a part
from the object's left to its right:
  negate position y; negate rotation rx and rz; KEEP rotation ry; keep size and
  shape identical.
Negating y alone is wrong for any rotated part -- it would turn a wing swept
aft into one swept forward.

# Materials
Give every primitive a "material" and a "form" from this menu. Each material
lists only the forms it can take, and each form has a fixed weight per volume;
you choose options, you never write a weight.
  "metal"
      "solid"    7800 kg/m^3  cast bases, bars, solid rods, hardware
      "tube"     1200 kg/m^3  poles, arms, tubular legs and frames
      "shell"     200 kg/m^3  metal shades, casings, car bodies, sheet panels
  "wood"
      "solid"     650 kg/m^3  legs, rails, planks, seats, tabletops
      "shell"     120 kg/m^3  cabinet carcasses, drawer boxes: boards around air
  "stone"
      "solid"    2600 kg/m^3  marble, granite, concrete: bases, plinths, tops
  "glass"
      "solid"    2500 kg/m^3  panes, screens, glass tabletops
      "shell"     300 kg/m^3  glass shades, vases, bottles
  "plastic"
      "solid"    1100 kg/m^3  small moulded parts, plastic seats and plates
      "shell"     150 kg/m^3  casings, housings, monitor backs
  "fabric"
      "shell"      20 kg/m^3  lampshades, cloth stretched over a frame
  "upholstery"
      "solid"      60 kg/m^3  fabric-covered foam: cushions, padded arms and backs
  "leather"
      "solid"      90 kg/m^3  leather-covered foam: seats, cushions, padded arms
  "rubber"
      "solid"    1200 kg/m^3  feet, bumpers, grips
      "shell"     300 kg/m^3  tyres
  "ceramic"
      "solid"    2300 kg/m^3  solid ceramic parts
      "shell"     400 kg/m^3  vases, pots, ceramic lamp bodies
  "paper"
      "shell"      10 kg/m^3  paper lampshades, lanterns
- "material": what the part is made of. Read it off the photo and off what
  objects like this are made of: a chrome tube is metal, a turned leg is wood,
  a sofa's cushions are upholstery, a monitor's casing is plastic.
- "form": how that material is built into the part. "solid" if it is solid
  through; "tube" for a pole, an arm or a tubular frame; "shell" for a thin
  wall around air -- a shade, a casing, a cabinet carcass, a car body, a vase.
  It changes what the part weighs, never how it is drawn: every primitive is
  still rendered and scored as a solid.
- "color": the part's dominant colour in the photo as "#rrggbb". Optional.
Mirrored twins share a material and form, like they share a size.
The drop test weighs every part by its material and form, so the same
shapes can stand or fall depending on what they are made of. Choose what
the object is actually made of, never whatever would make it stand.

# Objective
@@[block propose.construction starts]@@
Reproduce the object's geometry with AS FEW primitives as possible; SuperDec
optimises exactly this accuracy/compactness trade-off. Concretely:
- One primitive per functional part. Do not tile a flat surface with several
  boxes when one flatter box will do.
- Do not stack primitives redundantly on the same volume. Interpenetration
  AT JOINTS is expected and not penalised: a wing root belongs inside the
  fuselage, a leg's top belongs inside the seat. Overlap two primitives only
  where the real parts actually meet.
- DO represent parts that are hidden or occluded in the photo. You are
  reconstructing the object, not the visible pixels: a chair photographed from
  the front still has four legs. But do not add parts the object does not have
  -- an airborne aircraft with the gear up has no landing gear.
- Do not fight a taper or a curve with exponents -- raising e1 rounds off both
  ends and narrows the waist. Give a tapered part a "taper" and a curved one a
  "bend_deg", or make a curved slender part a tube.
- Where a moulded shell flows continuously from one part into another (a seat
  into its back, a fuselage into its tail cone) split it at the point of
  maximum curvature, the visual "corner", and let the two primitives
  interpenetrate through the fillet.
@@[block propose.construction ends]@@

# View conventions for the renders you will be shown
Renders are coloured per primitive on a white background. The canonical three
are orthographic; "match" is perspective. These directions are exact -- reason
from them rather than from intuition, and note that "front" shows you a
mirror-image left/right because you are facing the object.
- "front": camera on +X, looking back toward the origin. Screen-right is +Y,
  the object's LEFT. Screen-up is +Z.
- "left": camera on +Y, looking toward the origin; you see the object's left
  flank. Screen-right is -X, the object's BACK -- so the object faces
  screen-LEFT. Screen-up is +Z.
- "top": camera above on +Z, looking down. Screen-right is -Y, the object's
  RIGHT. Screen-up is +X, so the object faces screen-UP.
- "three-quarter": azimuth 35 deg, elevation 20 deg. A front-left-above view:
  the front faces slightly screen-left, the object's left side toward
  screen-right.
- "match": posed to match the input photo using your camera estimate.

# Output
Always reply with a single JSON object and nothing else. No prose outside it,
no markdown fence needed.
\end{lstlisting}

\begin{lstlisting}[caption={\textbf{Analyse.} ShapeNet objects come with cameras, so the last paragraph supplies the first photo's.},label=lst:analyse]
Study the photo. Work out how the object is oriented relative to the camera, then inventory its parts in the canonical object frame.

Decide which way the object FACES before anything else: the azimuth you report is what lets the next step place parts correctly. If the object is bilaterally symmetric, front-left and rear-left views can look alike. Settle it HERE, because nothing downstream can: the render you will be shown is posed using the azimuth you give now, so a front-back flip produces a render that matches the photo just as well and the silhouette score cannot see it. Decide on evidence that survives mirroring -- which way seats face, where an engine intake or a handle sits, which end tapers, how the object would be used -- rather than on the outline.

List every part you would need to rebuild the object, including ones hidden in this view, and merge parts a single primitive can cover.

Treat the photo as a PERSPECTIVE image with a moderate field of view, not an orthographic projection. Near parts are foreshortened less than far ones, so a near leg genuinely projects longer than its identical far twin. Do not read that difference as a real size difference, and do not solve for the pose as if the projection were parallel -- it will overestimate elevation badly.

Numbers: degrees to the nearest 5, proportions to 2 decimals. Do not invent precision you cannot see.

Reply with JSON matching this schema (the values below describe the types to emit, not literal strings):
{
  "category": "string, short noun for the object, e.g. 'chair'",
  "camera": {
    "azimuth_deg": "number. Where the photo was taken from, about the +Z axis. 0 = dead-on the object's front (+X). 90 = looking at the object's LEFT flank (+Y). 180 = behind it. -90 = its RIGHT flank. A front-left-above view like the reference is about +35.",
    "elevation_deg": "number. 0 = level with the object's centre, positive = camera above looking down, negative = below looking up.",
    "roll_deg": "number. In-plane camera tilt, 0 if the object's up axis runs straight up the image. Positive rolls the object clockwise on screen.",
    "confidence": "number 0-1: your probability that azimuth is within 30 deg."
  },
  "up_axis_note": "string, how you identified which way is up in the photo",
  "support": "string, how this object is held up in normal use. One of 'free_standing' (it rests on a floor by itself: a chair, a table, a cabinet, a car), 'wall_mounted' (it is fixed to a wall: a sconce, a picture, a wall-hung screen), 'lies_flat' (it normally rests on its side or back: a phone, a book, a remote), 'handheld' (it is carried or held: a rifle, a bottle, a tool), or 'airborne_or_afloat' (an aircraft in flight, a boat). Answer for the object in general, NOT for the pose in this photo. Only 'free_standing' objects are checked for whether they stand up.",
  "symmetry": "string, one of 'bilateral_xz', 'radial_z', 'none'",
  "proportions": {
    "extent_x": "number, front-to-back depth, relative to the largest of the three",
    "extent_y": "number, left-to-right width, relative to the largest of the three",
    "extent_z": "number, bottom-to-top height, relative to the largest of the three"
  },
  "parts": [
    {
      "name": "string, short label",
      "description": "string, what it looks like and roughly where it sits",
      "count": "integer, how many instances (4 for chair legs). Each instance becomes its own primitive in the next step.",
      "visible": "boolean, false only if EVERY instance is hidden in this view and you are inferring the part entirely",
      "material": "string, one of the material keys in the system prompt",
      "form": "string, 'solid', 'tube' or 'shell', from those its material offers"
    }
  ]
}

# Known camera
The first photo's camera is known from the capture, so do not estimate it: azimuth @@<azimuth>@@ deg, elevation @@<elevation>@@ deg, roll 0 deg, in the terms of the camera field above (azimuth 0 = dead-on the object's front, 90 = its left flank). It is a perspective camera with a @@<fov>@@ deg field of view, @@<distance>@@ times the object's largest dimension from its centre, aimed at the centre. Copy these numbers into "camera", and read the photo from this viewpoint: which parts are at the object's front or back, left or right, follows from it.
\end{lstlisting}

\begin{lstlisting}[caption={\textbf{Propose.}},label=lst:propose]
Here is your own analysis of the object:

@@<the analysis reply, as JSON>@@

Now place the superquadrics. Two images are attached: the input photo again, and a reference sheet rendering what the [e1, e2] exponents look like, so you can pick them by eye. In the reference sheet each tile is viewed from the front-left-above and the primitive's local +Z points up the page.

Budget: at most 14 primitives, and fewer is better. Expand each part with count N from your analysis into N primitives.

Work in the canonical object frame: Z up, +X front, sizes are HALF-extents, object centred near the origin at roughly unit scale.

Before you commit, check three things. That every mirrored pair negates y, negates rx and rz, and keeps ry. That no primitive is twice the size it should be -- halve every dimension you measured. And that any part which is a tube about the object's X or Y axis carries the 90 deg rotation that puts its local Z along that axis.

Carry each part's material and form over from your analysis onto every primitive it becomes.

You will be shown a render of this and get to correct it, so commit to your best reading now rather than hedging.

Reply with JSON matching this schema (the values below describe the types to emit, not literal strings):
{
  "category": "string, short noun for the object",
  "reasoning": "string, 2-4 sentences on how you placed the parts and chose exponents",
  "primitives": [
    {
      "name": "string label for this part",
      "size": "[sx, sy, sz] half-extents",
      "shape": "[e1, e2] in [0.1, 1.9]",
      "rotation_deg": "[rx, ry, rz] degrees",
      "position": "[x, y, z] centre",
      "kind": "OPTIONAL, omit for a normal solid part. One of 'toroid' (then give 'hole': 1-8) for a part with a hole through it, or 'hyperboloid1' (then give 'waist': 0.05-0.95) for a part pinched in the middle, or 'tube' -- a tube is written differently, as in the second entry below.",
      "material": "one of the material keys in the system prompt",
      "form": "'solid', 'tube' or 'shell', one its material offers",
      "color": "OPTIONAL '#rrggbb', the part's colour in the photo",
      "taper": "OPTIONAL [t1, t2] in [-0.9, 0.9]: wider (+) or narrower (-) toward the + end of deform_axis",
      "bend_deg": "OPTIONAL 0-180: curve the deform axis into an arc; then also give bend_dir_deg",
      "bend_dir_deg": "OPTIONAL degrees: which way both ends curve (see the system prompt)",
      "deform_axis": "OPTIONAL 'x', 'y' or 'z' (default 'z'): the axis taper and bend act along",
      "repeat": "OPTIONAL {\"count\": n, \"center\": [x, y]}: n copies evenly around the vertical axis through (x, y); counts as one part"
    },
    {
      "name": "string label",
      "kind": "'tube'",
      "spine": "[[x, y, z] x 4] object frame: start, two control points, end",
      "profile": "[a, b] cross-section half-thickness",
      "roundness": "0.1-1.9",
      "taper": "OPTIONAL [t]: cross-section (1 + t) at the end, (1 - t) at the start",
      "repeat": "OPTIONAL {\"count\": n, \"center\": [x, y]}: n copies evenly around the vertical axis through (x, y); counts as one part",
      "material": "one of the material keys in the system prompt",
      "form": "'solid', 'tube' or 'shell', one its material offers",
      "color": "OPTIONAL '#rrggbb', the part's colour in the photo"
    }
  ]
}
\end{lstlisting}

\begin{lstlisting}[caption={\textbf{Critique.} All conditional passages are shown; the text is the same in rounds 1 to 3 apart from the round number and the progress line.},label=lst:critique]
Refinement round @@<r>@@ of 3.

@@[from round 2 on, one of]@@
Your last round took chamfer_l1 from @@<a>@@ to @@<b>@@, so it IMPROVED the fit to the true surface (@@<b-a>@@, lower is better).
Your last round took chamfer_l1 from @@<a>@@ to @@<b>@@, so it made the fit to the true surface WORSE (@@<b-a>@@, lower is better) -- reconsider those edits before going further in the same direction.
Your last round took chamfer_l1 from @@<a>@@ to @@<b>@@, so it left the fit to the true surface essentially unchanged.

The first image is the original photo. The second is your current decomposition rendered from several angles, each primitive in its own colour. Compare them, using the view conventions from the system prompt -- remember 'front' mirrors left and right, and in 'top' the object faces up the page.

Current primitives:
@@<the current decomposition, as JSON>@@

Measured on the render:
@@<the verifiers' measurements, as JSON; the next prompt listing is an example>@@

Diagnose what is wrong, in this order of importance:
1. Orientation. Does the 'match' view line up with the photo? It is drawn from the photo's TRUE camera, the one you were given, while 'best_pose' is the angle at which your outline fits the photos best. So a large 'azimuth_off_by_deg' means your SHAPE is turned or mirrored in the canonical frame: the parts do not face the way the true camera says they do. Fix the geometry, not the camera.
2. Missing or spurious parts, and parts floating apart from each other. This one is MEASURED, so trust the numbers over the render: a silhouette cannot show you a gap that is hidden behind another part. 'connected_pieces' above 1 means something is attached to nothing, and it is SCORED -- each extra piece costs more than you are likely to win back on chamfer_l1, so never trade connectivity for a closer fit. Reattach the part to whatever should carry it, or delete it; 'widest_gap' is how far it has to move. In 'physics' the parts were also welded where they touch, stood on the ground and released, and anything in 'parts_that_moved' had nothing holding it up -- that one is reported, not scored, because some objects genuinely do not stand up on their own, and objects you marked as not free-standing are not dropped at all. Check 'parts_just_off_the_ground' too: those nearly reach the floor and miss. Nothing needs to fall for that to be wrong -- three good legs will hold an object up while the fourth hangs in the air, and the drop test will report no movement whatsoever. Compare parts that ought to match: a bumper hanging clear is right, one leg of four hanging clear is a length you got wrong.
@@[only for objects not marked free_standing]@@
   Instead of the drop, an object you marked as not free-standing gets the test its support calls for, in 'support_check': laid on a table it must lie still in at least one pose (lies_flat, handheld), its back must offer a face that can sit flat against a wall (wall_mounted), or, hung from a hinge along its length, it must hang level with every part staying on (airborne_or_afloat). 'ok' false means it failed and 'reason' says how. Like the drop test it is reported, not scored.
@@[sent in every round of these runs, because an error map is attached; the physics sheet it describes comes only when the check flags a part on a free-standing object]@@
   A THIRD IMAGE is attached showing this, because the numbers cannot tell you why. Its left panel is your decomposition as you placed it, standing on the ground; the right panel is the same thing three seconds later under gravity. RED parts are the ones the check flagged -- detached from the body, or failing to reach the floor when a part their own size does. GREEN parts are carrying the object. GREY parts are legitimately up in the air. Read the left panel to see WHY: a tilt of 44 degrees is just a number, but four legs bunched on one side with the seat overhanging is a diagnosis, and a red part floating a visible distance from anything that could hold it tells you where to move it. Name what you see there in your critique.
2b. TRUE 3D ERROR, in 'chamfer'. This round you are being measured against the object's actual surface, not just its outline, so trust it over anything the silhouette suggests. Read whichever half is larger: high 'completeness' means real surface you did NOT cover, so something is missing or too small; high 'accuracy' means you put surface where the object has none, so something is too big or misplaced. 'worst_placed_parts' ranks your own parts, and 'uncovered_surface.toward' gives the direction, in YOUR canonical frame, that the missing material lies. A silhouette cannot see depth, and this can: if they disagree, this is right.
@@[when the error map is attached]@@
   AN ERROR MAP is attached for this, in your own canonical views. Grey is the surface you built; the dots are the object's true surface, coloured pale to dark red by how far they lie from yours, and only the worst third is drawn. Dark dots sitting outside the grey shape are material you did not build. Read the PATTERN, which is what the numbers cannot give you: dots in a ring all the way round an outline mean that part is uniformly too small, dots massed on one side mean it is offset or too short in that direction, and dots hugging one component mean that component alone is wrong. Those want different fixes. Say which pattern you see.
2c. VLM REVIEW, in 'vlm_review'. A separate model was shown only pictures -- the photographs, and your result drawn in its materials through each photograph's camera -- and graded it 0 to 4 on proportions, placement, inventory, part_shape, identity, structure and materials, evidence first. 'overall' is their weighted sum out of 100, averaged over its independent passes in 'samples' (the criteria shown are the first pass's), and COUNTS toward your score, 75 being neutral; 'orientation' is whether it thinks your result faces the way the object does. Its 'photo k' is the k-th photograph above. It sees what the numbers cannot: a part missing or extra, an open part built solid, a wrong material, an object that does not read as itself. Act on every criterion below 4 whose evidence names a concrete part. On sizes and depth the measured 3D error is exact and the review is an impression, so where the two disagree, follow the 3D error.
3. Proportions: any part too large, too small or misplaced RELATIVE to the others. Overall scale is normalised for you, so judge ratios.
4. Exponents: flat-sided parts rendering as blobs, round parts as boxes, or a tube pointing the wrong way because its 90 deg rotation is missing.
5. Symmetry: mirrored pairs that disagree.
6. Compactness: primitives you can merge or delete without losing shape. The budget is 14 and using fewer is better.
7. Materials: every primitive made of what the photo shows, in the right form -- poles and arms are tubes, shades and casings are shells -- with colours that match. A new primitive needs a material and form too.

Then output the FULL corrected primitive list -- not a diff. Keep the parts that are already right exactly as they are, and change only what you diagnosed. If the decomposition is already good, say so in 'reasoning', set 'done' to true, and return it unchanged.

Reply with JSON matching this schema:
{
  "critique": "string, what is wrong, concretely, naming parts",
  "done": "boolean, true if no further change is worthwhile",
  "category": "string, short noun for the object",
  "reasoning": "string, 2-4 sentences on how you placed the parts and chose exponents",
  "primitives": [
    {
      "name": "string label for this part",
      "size": "[sx, sy, sz] half-extents",
      "shape": "[e1, e2] in [0.1, 1.9]",
      "rotation_deg": "[rx, ry, rz] degrees",
      "position": "[x, y, z] centre",
      "kind": "OPTIONAL, omit for a normal solid part. One of 'toroid' (then give 'hole': 1-8) for a part with a hole through it, or 'hyperboloid1' (then give 'waist': 0.05-0.95) for a part pinched in the middle, or 'tube' -- a tube is written differently, as in the second entry below.",
      "material": "one of the material keys in the system prompt",
      "form": "'solid', 'tube' or 'shell', one its material offers",
      "color": "OPTIONAL '#rrggbb', the part's colour in the photo",
      "taper": "OPTIONAL [t1, t2] in [-0.9, 0.9]: wider (+) or narrower (-) toward the + end of deform_axis",
      "bend_deg": "OPTIONAL 0-180: curve the deform axis into an arc; then also give bend_dir_deg",
      "bend_dir_deg": "OPTIONAL degrees: which way both ends curve (see the system prompt)",
      "deform_axis": "OPTIONAL 'x', 'y' or 'z' (default 'z'): the axis taper and bend act along",
      "repeat": "OPTIONAL {\"count\": n, \"center\": [x, y]}: n copies evenly around the vertical axis through (x, y); counts as one part"
    },
    {
      "name": "string label",
      "kind": "'tube'",
      "spine": "[[x, y, z] x 4] object frame: start, two control points, end",
      "profile": "[a, b] cross-section half-thickness",
      "roundness": "0.1-1.9",
      "taper": "OPTIONAL [t]: cross-section (1 + t) at the end, (1 - t) at the start",
      "repeat": "OPTIONAL {\"count\": n, \"center\": [x, y]}: n copies evenly around the vertical axis through (x, y); counts as one part",
      "material": "one of the material keys in the system prompt",
      "form": "'solid', 'tube' or 'shell', one its material offers",
      "color": "OPTIONAL '#rrggbb', the part's colour in the photo"
    }
  ]
}
\end{lstlisting}

\Cref{lst:feedback} is the measurement block of \cref{lst:critique} for the object of \cref{fig:s_images}, as its first Critique call received it.
Silhouette IoU is measured but not shown to the model when a reference point cloud is available.

\begin{lstlisting}[caption={\textbf{Verifier feedback}, a real example: a chair's first proposal, before the first Critique call. White space is condensed here.},label=lst:feedback]
{
  "overlap": 0.1285,
  "symmetry_error": 0.0,
  "n_primitives": 7,
  "best_pose": {
    "azimuth_deg": -170.0,
    "elevation_deg": 24.0
  },
  "note": "best_pose: the viewing angle at which your outline fits the photos best. overlap: fraction of solid volume covered twice; the first 0.05 is free so that joints cost nothing, and only the excess is charged. symmetry_error: 0 means every off-centre part has a proper mirror twin. connected_pieces: how many separate lumps the parts form once anything within 2% of the bounding diagonal is treated as touching; anything above 1 means a part is attached to nothing, and each extra piece costs 0.05 of score -- as much as 0.20 of chamfer_l1, so reattaching a floating part is worth more than any closer fit you would give up for it. widest_gap tells you how far you have to move it.",
  "connected_pieces": 1,
  "widest_gap": 0.0,
  "physics": {
    "connected_pieces": 1,
    "widest_gap": 0.0,
    "max_part_movement": 0.0,
    "max_tilt_deg": 0.0,
    "parts_that_moved": [],
    "parts_on_the_ground": ["front_left_leg", "front_right_leg", "rear_left_leg", "rear_right_leg"],
    "parts_just_off_the_ground": [],
    "note": "The parts were welded where their surfaces touch, stood on a ground plane and released under gravity. Distances are fractions of the object's bounding diagonal. connected_pieces > 1 means some part touches nothing else -- it is floating, which no silhouette can show you. A part that moved is one nothing was holding up. parts_just_off_the_ground nearly reach the floor but miss it: decide for yourself which of them were supposed to carry the object -- a bumper hanging clear is correct, one leg of four hanging clear is not. Legs that disagree with each other are the case to look for, since three good legs will hold the object up and nothing will move. Each part weighs what its material and form make it, and heaviest_parts is its share of the total. Whether the object stands depends on where that mass sits over the parts on the ground. If the real object stands and yours falls, check first that each part has the material and form the photo shows -- a pole is a tube and a shade is a shell, not solid lumps -- and only then change geometry. Do not pick a material or form the object does not have to make it stand.",
    "heaviest_parts": [{"name": "seat", "share": 0.32}, {"name": "backrest", "share": 0.32}, {"name": "backrest_top", "share": 0.111}, {"name": "front_left_leg", "share": 0.075}]
  },
  "azimuth_off_by_deg": 8.0,
  "chamfer": {
    "chamfer_l1": 3.254,
    "chamfer_l2": 0.1648,
    "completeness": 3.224,
    "accuracy": 3.283,
    "worst_placed_parts": [{"name": "backrest_top", "accuracy": 5.507}, {"name": "rear_left_leg", "accuracy": 3.676}, {"name": "front_right_leg", "accuracy": 3.656}, {"name": "rear_right_leg", "accuracy": 3.404}, {"name": "front_left_leg", "accuracy": 3.251}],
    "uncovered_surface": {"worst_10pct_distance": 8.902, "toward": "-X", "offset_from_centre": [-0.24, 0.01, 0.1], "nearest_part": "backrest"},
    "note": "Measured against the object's true 3D surface, in units of 1e-2 with the object about 1 across; lower is better. completeness is true surface you did NOT cover -- something is missing or too small. accuracy is surface you placed where the object is not -- something is too big or in the wrong spot. They fail for opposite reasons, so read whichever is larger. worst_placed_parts ranks your parts by their own accuracy. uncovered_surface says which way, in YOUR canonical frame, the missing material lies."
  },
  "vlm_review": {
    "overall": 75.0,
    "orientation": "yes",
    "major_parts": ["backrest with arched top", "seat block", "front legs", "rear legs (extend down from backrest)"],
    "proportions": {"score": 3, "evidence": "The real chair is a slim, tall chair with a wide arched back panel about as wide as the seat. The reconstruction's back is rectangular and narrower in photo 3 and 4, and the overall silhouette is similar in height but the back looks taller and thinner."},
    "placement": {"score": 3, "evidence": "Seat, legs and back are placed roughly right. The back sits at the rear and the legs go to the floor. The reconstruction's rear legs seem split from the back panel and there's an odd extra ledge at the top of the back."},
    "inventory": {"score": 3, "evidence": "missing: none; extra: a half-disc shelf/ledge on top of the backrest that the real chair does not have"},
    "part_shape": {"score": 2, "evidence": "The real back has a smooth semicircular arched top and is a single flat slab. The reconstruction has a flat rectangular top with a protruding half-disc wedge, so the back shape is wrong. Legs and seat are chunky blocks, about right."},
    "identity": {"score": 3, "evidence": "Clearly a chunky wooden chair, but the arched-top back is not captured and the half-disc feature misleads."},
    "structure": {"score": 4, "evidence": "Legs reach the floor, the seat sits on the legs, and the back attaches to the rear. It would stand."},
    "materials": {"score": 4, "evidence": "Wood throughout, matching the brown wooden chair."},
    "samples": [75.0, 75.0]
  },
  "vlm_review_term": 0.0
}
\end{lstlisting}

\section{Prompt blocks for self-evolution}
\label{sec:s_blocks}

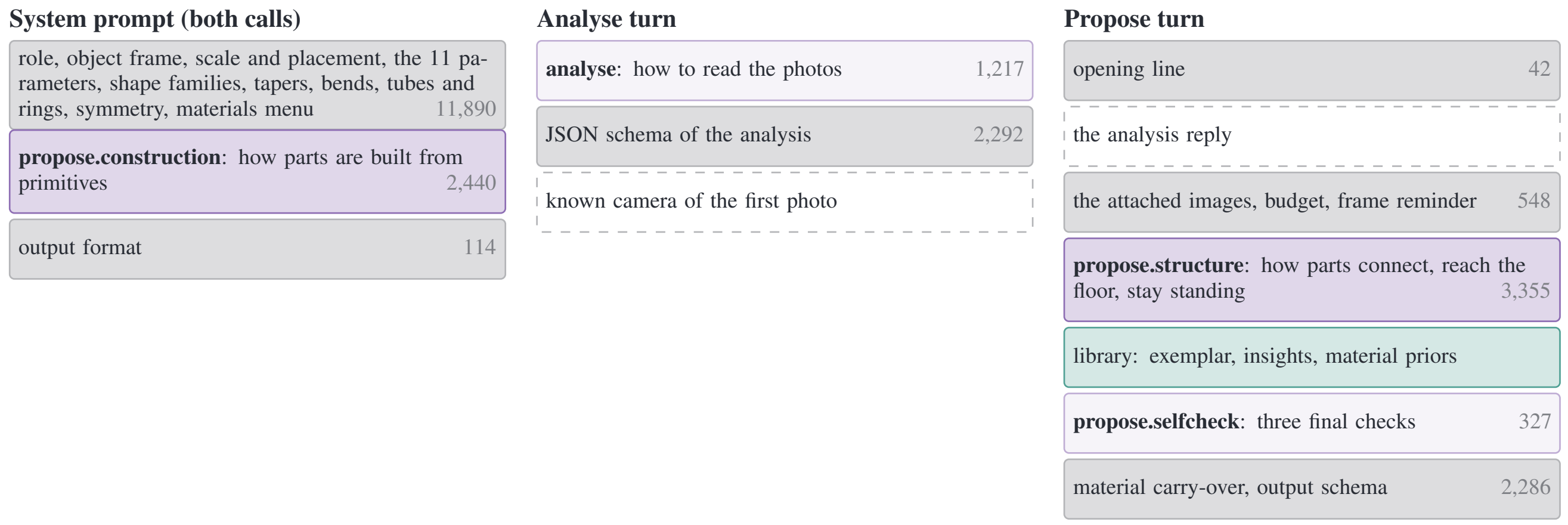
\begin{figure}[H]
  \centering
\begin{tikzpicture}[x=1cm,y=1cm,font=\scriptsize]
\node[anchor=south west, inner sep=0, font=\footnotesize\bfseries, text=inkc] at (0.00,0.08) {System prompt (both calls)};
\node[anchor=north west, rectangle, rounded corners=1.5pt, line width=0.5pt, fill=inkc!16, draw=inkc!35, minimum width=5.5cm, minimum height=0.92cm, text width=5.3cm, align=left, inner sep=0.1cm, text=inkc] at (0.00,0.00) {role, object frame, scale and placement, the 11 parameters, shape families, tapers, bends, tubes and rings, symmetry, materials menu\hfill\textcolor{inkc!60}{11,890}};
\node[anchor=north west, rectangle, rounded corners=1.5pt, line width=0.5pt, fill=evoc!22, draw=evoc!80, minimum width=5.5cm, minimum height=0.92cm, text width=5.3cm, align=left, inner sep=0.1cm, text=inkc] at (0.00,-0.99) {\textbf{propose.construction}: how parts are built from primitives\hfill\textcolor{inkc!60}{2,440}};
\node[anchor=north west, rectangle, rounded corners=1.5pt, line width=0.5pt, fill=inkc!16, draw=inkc!35, minimum width=5.5cm, minimum height=0.66cm, text width=5.3cm, align=left, inner sep=0.1cm, text=inkc] at (0.00,-1.98) {output format\hfill\textcolor{inkc!60}{114}};
\node[anchor=south west, inner sep=0, font=\footnotesize\bfseries, text=inkc] at (5.85,0.08) {Analyse turn};
\node[anchor=north west, rectangle, rounded corners=1.5pt, line width=0.5pt, fill=evoc!6, draw=evoc!45, minimum width=5.5cm, minimum height=0.66cm, text width=5.3cm, align=left, inner sep=0.1cm, text=inkc] at (5.85,0.00) {\textbf{analyse}: how to read the photos\hfill\textcolor{inkc!60}{1,217}};
\node[anchor=north west, rectangle, rounded corners=1.5pt, line width=0.5pt, fill=inkc!16, draw=inkc!35, minimum width=5.5cm, minimum height=0.66cm, text width=5.3cm, align=left, inner sep=0.1cm, text=inkc] at (5.85,-0.73) {JSON schema of the analysis\hfill\textcolor{inkc!60}{2,292}};
\node[anchor=north west, rectangle, rounded corners=1.5pt, line width=0.5pt, fill=white, draw=inkc!35, dashed, minimum width=5.5cm, minimum height=0.66cm, text width=5.3cm, align=left, inner sep=0.1cm, text=inkc] at (5.85,-1.46) {known camera of the first photo};
\node[anchor=south west, inner sep=0, font=\footnotesize\bfseries, text=inkc] at (11.70,0.08) {Propose turn};
\node[anchor=north west, rectangle, rounded corners=1.5pt, line width=0.5pt, fill=inkc!16, draw=inkc!35, minimum width=5.5cm, minimum height=0.66cm, text width=5.3cm, align=left, inner sep=0.1cm, text=inkc] at (11.70,0.00) {opening line\hfill\textcolor{inkc!60}{42}};
\node[anchor=north west, rectangle, rounded corners=1.5pt, line width=0.5pt, fill=white, draw=inkc!35, dashed, minimum width=5.5cm, minimum height=0.66cm, text width=5.3cm, align=left, inner sep=0.1cm, text=inkc] at (11.70,-0.73) {the analysis reply};
\node[anchor=north west, rectangle, rounded corners=1.5pt, line width=0.5pt, fill=inkc!16, draw=inkc!35, minimum width=5.5cm, minimum height=0.66cm, text width=5.3cm, align=left, inner sep=0.1cm, text=inkc] at (11.70,-1.46) {the attached images, budget, frame reminder\hfill\textcolor{inkc!60}{548}};
\node[anchor=north west, rectangle, rounded corners=1.5pt, line width=0.5pt, fill=evoc!22, draw=evoc!80, minimum width=5.5cm, minimum height=0.92cm, text width=5.3cm, align=left, inner sep=0.1cm, text=inkc] at (11.70,-2.19) {\textbf{propose.structure}: how parts connect, reach the floor, stay standing\hfill\textcolor{inkc!60}{3,355}};
\node[anchor=north west, rectangle, rounded corners=1.5pt, line width=0.5pt, fill=libc!18, draw=libc!75, minimum width=5.5cm, minimum height=0.66cm, text width=5.3cm, align=left, inner sep=0.1cm, text=inkc] at (11.70,-3.18) {library: exemplar, insights, material priors};
\node[anchor=north west, rectangle, rounded corners=1.5pt, line width=0.5pt, fill=evoc!6, draw=evoc!45, minimum width=5.5cm, minimum height=0.66cm, text width=5.3cm, align=left, inner sep=0.1cm, text=inkc] at (11.70,-3.91) {\textbf{propose.selfcheck}: three final checks\hfill\textcolor{inkc!60}{327}};
\node[anchor=north west, rectangle, rounded corners=1.5pt, line width=0.5pt, fill=inkc!16, draw=inkc!35, minimum width=5.5cm, minimum height=0.66cm, text width=5.3cm, align=left, inner sep=0.1cm, text=inkc] at (11.70,-4.64) {material carry-over, output schema\hfill\textcolor{inkc!60}{2,286}};
\end{tikzpicture}
  \caption{\textbf{One PhysDec-Evolved prediction, block by block,} in order within each column. Numbers are characters in the evolved prompt. \textcolor{evoc}{\textbf{Dark purple}}: rewritten by GEPA. \textcolor{evoc}{Light purple}: evolvable, kept as in the seed. Grey: frozen. \textcolor{libc}{\textbf{Teal}}: retrieved per object (\cref{sec:s_library}). Dashed: filled in per object.}
  \label{fig:s_blocks}
\end{figure}

Self-evolution (Sec.~\mainevolve{}) never rewrites a prompt as a whole.
The prompts of \cref{sec:s_correction} are cut at fixed anchors into frozen blocks and four evolvable components (\cref{fig:s_blocks,tab:s_blocks}), and only two components are rewritten.
The frozen blocks hold the definitions that everything else relies on: the frame, the parameters, the materials menu and the output schemas.
Every candidate is checked to contain them verbatim.

\begin{table}[H]
  \centering\footnotesize
  \caption{\textbf{The four evolvable components.}}
  \label{tab:s_blocks}
  \begin{tabular}{@{}L{3.0cm}L{3.2cm}L{6.9cm}L{2.7cm}@{}}
    \toprule
    Component & Where it sits & What it says & During GEPA \\
    \midrule
    \texttt{propose.construction} & system prompt, \texttt{\# Objective} & how to build parts out of primitives & rewritten \\
    \texttt{propose.structure} & Propose turn, after the frame reminder & how parts connect, reach the floor and stay standing & rewritten \\
    \texttt{analyse} & Analyse turn, before the schema & how to read the photos and list the parts & kept from the seed \\
    \texttt{propose.selfcheck} & Propose turn, after the library & three checks before replying & kept from the seed \\
    \bottomrule
  \end{tabular}
\end{table}

\textbf{Serving text.}
A PhysDec-Evolved prediction has no later rounds, so three passages that describe them are removed from the frozen blocks:
the system prompt's \texttt{\# View conventions} section, which describes renders the model is never shown;
the analysis schema's sentence ``Only 'free\_standing' objects are checked for whether they stand up.'';
and the Propose sentence ``You will be shown a render of this and get to correct it, so commit to your best reading now rather than hedging.''
The seed's \texttt{analyse} component also ends its sentence on facing with ``nothing downstream can correct it.'' in place of a clause about renders.

\textbf{Rewriting.}
GEPA~\cite{agrawal2026gepa} draws a minibatch of six training objects from six categories, predicts each object with a candidate prompt and collects the verifier feedback as text.
Feedback on connectivity, floor contact, the drop test and the review's \emph{structure} criterion goes to \texttt{propose.structure}; feedback on geometry, the silhouette and the other review criteria goes to \texttt{propose.construction}.
Opus 5.5 then rewrites the block under \cref{lst:reflection}, with the system prompt ``You rewrite instructions for an AI assistant.''
A rewrite is thrown out if it restates a frozen definition (12 text patterns), grows past twice the seed, names or points at one particular object, or leaves a frozen block damaged in the assembled prompt.
The model gets one retry that names the reason; after that the iteration is skipped.

\begin{lstlisting}[caption={\textbf{Rewrite prompt.} GEPA's own instruction-proposal template, with our note on the block inserted before its last line.},label=lst:reflection]
I provided an assistant with the following instructions to perform a task for me:
```
@@<the block's current text>@@
```

The following are examples of different task inputs provided to the assistant along with the assistant's response for each of them, and some feedback on how the assistant's response could be better:
```
@@<for each object of the minibatch: category, reference support, the analysis, the primitives returned, the verifier feedback for this block, the score>@@
```

Your task is to write a new instruction for the assistant.

Read the inputs carefully and identify the input format and infer detailed task description about the task I wish to solve with the assistant.

Read all the assistant responses and the corresponding feedback. Identify all niche and domain specific factual information about the task and include it in the instruction, as a lot of it may not be available to the assistant in the future. The assistant may have utilized a generalizable strategy to solve the task, if so, include that in the instruction as well.

Context: the instruction above is ONE block of a larger prompt. The rest of that prompt already defines the coordinate frame and axes, what every primitive parameter means (sizes, exponents, rotations), the shape families, the materials menu and the JSON output format, and those definitions are fixed. Do not restate, paraphrase or redefine any of them. Text containing any of these phrases is discarded, even in passing: "half-extent", "semi-axis", "full length", "full extent", "Z-up", "+X is" (or any "+X/+Y/+Z is/means/points/faces"), "e1 controls" (or shapes/governs/defines), "density", "kg/m", "radius 0.5", "hole =", "waist =", "screen-right". Say "size" for a primitive's extent instead. Write only this block: @@<role, see below>@@. Keep it under @@<limit>@@ characters; longer text is discarded. Do not mention particular objects from the examples; write rules that carry over to objects you have not seen.@@<rejected>@@

Provide the new instructions within ``` blocks.

@@[filled in as follows, not part of the template]@@
@@<role>@@ for propose.structure: rules for how the parts connect, rest on the floor and stay standing
@@<role>@@ for propose.construction: guidance on choosing, sizing and placing primitives to match the object's shape
@@<limit>@@ in characters: propose.structure 3303, propose.construction 2307
@@<rejected>@@ is empty on the first attempt; on the second: Your previous rewrite (<n> characters) was discarded because: <reasons>. Write it again without these problems.
\end{lstlisting}

\textbf{Scoring a candidate.}
GEPA spends a budget of 500 evaluations on the training objects and compares candidates on 26 validation objects.
Each prediction is made without online correction and evaluated with the scoring rule of Eq.~(\mainscore{}) and the verifiers of Sec.~\mainverifiers{}.
During evolution the review score $v$ is one grade per evaluation.
At the end, the three best candidates each ran three times on the validation objects with two grades per evaluation, and the highest mean won.

\section{Prompts before and after evolution}
\label{sec:s_before}

\Cref{tab:s_stages} separates the prompts involved, because ``before'' can mean two things.
\emph{Base} in Table~\mainevolution{} uses the original prompts: one call analyzes the images and a second proposes a decomposition, without retrieval or online correction.
The \emph{seed} is where GEPA starts, and it adds a block of structure rules.
\Cref{tab:structure,tab:construction} print the seed and the evolved text in full.

\begin{table}[H]
  \centering\footnotesize
  \caption{\textbf{What each prompt contains.}}
  \label{tab:s_stages}
  \begin{tabular}{@{}L{3.6cm}L{3.9cm}L{4.4cm}L{4.1cm}@{}}
    \toprule
    Block & Base (Table~\mainevolution{}) & Seed & PhysDec-Evolved (Table~\mainevolution{}) \\
    \midrule
    \texttt{propose.construction} & original (\cref{lst:system}) & original & evolved (\cref{tab:construction}) \\
    \texttt{propose.structure} & empty & 8 rules (\cref{tab:structure}, left) & evolved, 10 rules (\cref{tab:structure}, right) \\
    \texttt{analyse} & original (\cref{lst:analyse}) & one clause changed (\cref{sec:s_blocks}) & as the seed \\
    \texttt{propose.selfcheck} & original (\cref{lst:propose}) & original & original \\
    Frozen blocks & as in \cref{lst:system,lst:analyse,lst:propose} & three passages removed (\cref{sec:s_blocks}) & as the seed \\
    Library & none & retrieved & retrieved \\
    \bottomrule
  \end{tabular}
\end{table}

\begingroup
\footnotesize\ttfamily\raggedright
\setlength{\LTpre}{4pt}\setlength{\LTpost}{6pt}
\begin{longtable}{@{}p{0.487\linewidth}@{\hspace{0.026\linewidth}}p{0.487\linewidth}@{}}
  \caption{\normalfont\textbf{\texttt{propose.structure}, before and after.} \oldw{Orange} words appear only in the seed, \neww{blue} words only in the evolved text. A row with no seed text is a new rule.}\label{tab:structure}\\
  \toprule
  \normalfont\textbf{Seed} & \normalfont\textbf{Evolved} \\
  \midrule
  \endfirsthead
  \toprule
  \normalfont\textbf{Seed} & \normalfont\textbf{Evolved} \\
  \midrule
  \endhead
  \bottomrule
  \endlastfoot
Structure rules (apply before writing any primitive): & Structure rules (apply before writing any primitive): \\ \addlinespace[3pt]
1. Every part must touch the main \oldw{body. Connected parts should touch} or \oldw{slightly overlap at joints;} a part that floats even \oldw{slightly} counts as a separate piece and is penalized. Faces that only meet and edges that only kiss do not \oldw{count as connected:} a connector \oldw{must overlap every part it joins, even when you slim it} to match the silhouette. & 1. \neww{One piece.} Every part must touch the main \neww{body, directly} or \neww{through} a \neww{chain of touching parts. A} part that floats even \neww{a hair away} counts as a separate piece and is penalized. Faces that only meet and edges that only kiss do not \neww{count. Give every joint a real overlap of roughly 10-30\% of the thinner part's size, and keep it after slimming} a connector to match the silhouette. \neww{A mounted part (lid, knob, cap, shade, sight, ring on a rod) must sink a little into its host.} \\ \addlinespace[3pt]
2. \oldw{Free-standing objects rest on one} ground plane. \oldw{All parts} meant to touch the floor (legs, base, \oldw{wheels)} must reach the same lowest z. Sibling parts with the same cross-section \oldw{(e.g. four legs) must} share \oldw{the same} bottom height. & 2. \neww{One} ground plane. \neww{For free-standing objects, every part} meant to touch the floor (legs, base, \neww{plinth, wheels, side boards, runners)} must reach the same lowest z. Sibling parts with the same cross-section \neww{(a set of legs, a pair of panels)} share \neww{one} bottom height. \neww{The ground check also flags any part whose footprint matches a floor-reaching part but stops short, such as a bottom board between floor-length sides, or a back panel spanning the width above a base. Either extend that part to the shared lowest z or clearly change its footprint. For leaning, splayed or rotated parts, check the lowest corner, not the centre.} \\ \addlinespace[3pt]
3. Weight \oldw{must sit} over the footprint. \oldw{Choose} the real material and \oldw{form first (a lampshade} is usually a thin shell, a \oldw{lamp} pole a \oldw{tube); only then} adjust \oldw{geometry.} Never pick a material just to make the object stand. & 3. Weight over the footprint. \neww{First choose} the real material and \neww{form: a cover or housing} is usually a thin shell, \neww{and} a pole \neww{or rod is} a \neww{tube. Then} adjust \neww{the geometry so the centre of mass sits well inside the support area, with a clear margin and not near its edge. For a long cantilever, widen the base, move the base under the heavy end, or keep a solid mass behind the pivot.} Never pick a material just to make the object stand. \\ \addlinespace[3pt]
4. Build \oldw{thin,} long, or branching structures (arms, frames, \oldw{rails)} as thin parts, never as one thick \oldw{block: a} curved slender part \oldw{as} ONE \oldw{tube, a} straight one \oldw{as one} thin \oldw{primitive, each overlapping} the part it \oldw{joins.} & 4. \neww{Thin structures stay thin.} Build long, \neww{slender} or branching structures (arms, frames, \neww{rails, rods, struts)} as thin parts, never as one thick \neww{block. A} curved slender part \neww{is} ONE \neww{tube along its path. A} straight one \neww{is a single} thin \neww{primitive. Each must overlap what it joins at both ends. Put a tube's first spine point inside} the part it \neww{grows from and its last inside the part it reaches.} \\ \addlinespace[3pt]
5. Parts repeated evenly around a central stem \oldw{(the arms and candles of a chandelier, the legs of a swivel base)} are ONE written part with a repeat, in the real number of copies. Start each copy inside the stem so \oldw{that} every copy overlaps \oldw{it; never} leave the axis empty where \oldw{the stem meets them.} & 5. \neww{Radial repeats.} Parts repeated evenly around a central stem are ONE written part with a repeat, in the real number of copies. Start each copy inside the stem so every copy overlaps \neww{it. Never} leave the axis empty where \neww{they meet.} \\ \addlinespace[3pt]
6. For a rotated primitive, work out which object axis each \oldw{of its} local \oldw{axes ends up on, and give} each \oldw{extent (a} panel's thickness above \oldw{all)} to the local axis that lands on that direction. & 6. \neww{Rotations.} For a rotated primitive, work out which object axis each local \neww{axis lands on. Give} each \neww{size, a} panel's thickness above \neww{all,} to the local axis that lands on that direction. \neww{After rotating, recheck the overlaps with neighbours and, where needed, floor contact.} \\ \addlinespace[3pt]
7. Leave out any primitive that would sit almost entirely inside \oldw{another; every} part \oldw{should cover} some \oldw{visible} outer surface of its own. & 7. \neww{No hidden parts.} Leave out any primitive that would sit almost entirely inside \neww{another. Every} part \neww{must show} some outer surface of its own. \\ \addlinespace[3pt]
8. Model only the object. A \oldw{person} or other reference \oldw{object} shown for scale gets no primitive. & 8. Model only the object. A \neww{person, hand, wall, water} or other reference shown for scale \neww{or context} gets no primitive. \\ \addlinespace[3pt]
--- & \neww{9. Other supports. For wall-mounted, handheld, hanging, lying-flat or afloat objects, the floor and drop tests are skipped, but rule 1 still applies in full. Keep every part connected and the object in its natural pose.} \\ \addlinespace[3pt]
--- & \neww{10. Final pass. List each part's lowest z and its contact partner with the overlap amount. Estimate the centre of mass against the footprint. Then fix any gap, any floor part short of the shared lowest z, any part not linked to the main body, and any balance that sits near tipping.} \\ \addlinespace[3pt]

\end{longtable}

\begin{longtable}{@{}p{0.487\linewidth}@{\hspace{0.026\linewidth}}p{0.487\linewidth}@{}}
  \caption{\normalfont\textbf{\texttt{propose.construction}, before and after,} aligned by topic; the evolved text is in its own order. The seed's two bullets on one primitive per part and on stacking merge into one. A row with no seed text is a new bullet.}\label{tab:construction}\\
  \toprule
  \normalfont\textbf{Seed} & \normalfont\textbf{Evolved} \\
  \midrule
  \endfirsthead
  \toprule
  \normalfont\textbf{Seed} & \normalfont\textbf{Evolved} \\
  \midrule
  \endhead
  \bottomrule
  \endlastfoot
Reproduce the object's \oldw{geometry} with \oldw{AS FEW} primitives as \oldw{possible;} SuperDec \oldw{optimises exactly this accuracy/compactness trade-off. Concretely:} & Reproduce the object's \neww{shape} with \neww{as few} primitives as \neww{possible.} SuperDec \neww{scores the trade-off between accuracy and compactness.} \\ \addlinespace[3pt]
- \oldw{One} primitive per functional part. \oldw{Do not} tile a flat surface with several \oldw{boxes when one flatter box will do. - Do not} stack primitives \oldw{redundantly} on the same volume. \oldw{Interpenetration AT JOINTS is expected and not penalised:} a \oldw{wing root belongs inside the fuselage, a leg's} top \oldw{belongs} inside the seat. \oldw{Overlap two primitives only where the real parts actually meet.} & - \neww{Use one} primitive per functional part. \neww{Never} tile a flat surface with several \neww{boxes, and never} stack primitives on the same volume. \neww{Let parts overlap only at real joints, such as} a \neww{leg} top inside the seat. \\ \addlinespace[3pt]
- \oldw{DO represent parts that are hidden or occluded} in the \oldw{photo. You are reconstructing the object, not the visible pixels: a chair photographed from the front still has four legs. But do not add parts} the object \oldw{does not have -- an airborne aircraft with the gear up has no landing gear.} & - \neww{Build every part listed} in the \neww{analysis, with its count, including hidden parts. Add nothing} the object \neww{lacks.} \\ \addlinespace[3pt]
--- & \neww{- Scale is the biggest error source, and outputs tend to come out too small. Set the longest axis first. Then size every other axis so that the union of all parts spans the stated extents ratio exactly. Before output, add up the min and max positions plus sizes along each axis and fix any shortfall.} \\ \addlinespace[3pt]
--- & \neww{- Do not let vertical stacks come out squat. The supports, base or cage must run from the floor all the way to the part they carry.} \\ \addlinespace[3pt]
--- & \neww{- Follow the fractions given in the descriptions, such as "0.41 wide" or "rear 40\%". Keep part-to-part scale right: a rim stated as wide must be wider than a cap above it, and a cab stated as long must not be shortened. Never shrink small details. Place each part at the stated height. Elements described as "slightly below mid-height" must not drift lower.} \\ \addlinespace[3pt]
--- & \neww{- Keep feet and supports inside the footprint of the body they carry, slightly inset from its corners.} \\ \addlinespace[3pt]
--- & \neww{- For sharp boxes, use the lowest exponents, 0.1 or less. Values of 0.15 to 0.3 already read as rounded. Round edges only when the description says soft or rounded.} \\ \addlinespace[3pt]
- \oldw{Do} not \oldw{fight} a \oldw{taper or a curve with exponents -- raising e1 rounds off both ends and narrows the waist. Give a tapered part a "taper" and a curved one a "bend\_deg", or make a curved slender part a tube.} & - \neww{Use "taper" for tapers and "bend\_deg" for curves, matching the direction and depth described. Keep a shallow cone shallow and broad: rim at full width, low height, near-linear sides,} not a \neww{steep rounded bowl.} \\ \addlinespace[3pt]
--- & \neww{- Give handles and loops that bow away from a surface real curvature and clearance.} \\ \addlinespace[3pt]
--- & \neww{- Model wheels as their own rubber discs or cylinders, never as fender blocks.} \\ \addlinespace[3pt]
- \oldw{Where} a moulded shell \oldw{flows continuously from one part into another (a seat into its back, a fuselage into its tail cone)} split \oldw{it} at the point of maximum \oldw{curvature, the visual "corner",} and let the two primitives \oldw{interpenetrate} through the fillet. & - \neww{At} a moulded shell \neww{transition,} split at the point of maximum \neww{curvature} and let the two primitives \neww{overlap} through the fillet. \\ \addlinespace[3pt]
--- & \neww{- Identity features justify extra primitives. These include windows and windshields, as dark thin glass panels on the cab faces. They also include separate cushions or slats with gaps, paired tabs, flared feet and button grids. For a button grid, a few row strips read better than one flat slab.} \\ \addlinespace[3pt]
--- & \neww{- When a part has a face of a different colour or material, add a thin cap primitive for that face.} \\ \addlinespace[3pt]
--- & \neww{- Mirror symmetric pairs exactly.} \\ \addlinespace[3pt]

\end{longtable}
\endgroup

\textbf{What changed.}
The structure block grows from 1,835 to 3,355 characters and the construction block from 1,282 to 2,440.
The structure rules become more exact (a joint overlap of 10--30\% of the thinner part, the lowest corner of a rotated part) and gain a rule for other support types and a final pass.
The new construction bullets name specific errors, such as results that come out too small, stacks that come out squat and exponents that round off sharp boxes.
Some rules also describe the verifiers themselves (rules 2 and 9), since the verifiers' feedback is what the rewrite reads.
On the 26 validation objects the evolved prompt raises the mean VLM score over the seed from 70.8 to 72.7.
Table~\mainevolution{} reports the test results.

\section{Insights and the retrieval library}
\label{sec:s_library}

The library comes from PhysDec-Loop runs on the training objects and never reads test objects.
It holds 144 insights, 81 exemplars and material priors for all 13 categories.

\textbf{Insights.}
When a revision improves $S_r$ of Eq.~(\mainscore{}) without adding a piece or a fall, its two rounds form a before-and-after correction pair; the library rests on 411 pairs.
A pair whose gain comes from the review, with Chamfer distance about unchanged, has the type \emph{semantic}.
Pairs are grouped by category and type, and Opus 5.5 reads up to eight at a time under \cref{lst:distil}, returning operations on the current list: add, edit, upvote or downvote.
A rule is refused if it names one object, describes objects instead of saying what to do, runs too long or has an unknown scope.
Upvotes set an insight's weight, which orders retrieval.
\Cref{tab:s_insights} shows the highest-weight insight of every category.

\begin{lstlisting}[caption={\textbf{Distilling insights.}},label=lst:distil]
@@[system prompt]@@
You curate a short list of lessons for a model that rebuilds objects from photos as superquadric primitives in one shot, with no feedback. Each lesson must help on objects it has never seen. Reply with a single JSON object and nothing else.

@@[user prompt]@@
Below are corrections a reconstructor made to its own @@<category>@@ reconstructions after seeing feedback (true-surface Chamfer distance, connectivity, floor contact, a drop test). Each one made the result better without breaking anything. @@[for type "semantic" this sentence is instead: Each one left Chamfer about where it was but made the result clearly more like the photographed object to the reviewer -- a part added or reshaped, say -- without breaking anything.]@@ They are of type "@@<type>@@".

The model that will read the lessons gets no feedback: it sees the photos once and must get it right the first time. Turn what these corrections reveal into lessons it can apply up front.

The model's prompt already contains these rules; do not add lessons that repeat them:
@@<the structure rules of the seed prompt, Section C, left column>@@

Current lessons:
@@<one per line: id [category/role, weight w] when ... | do ... | why ...; the list starts empty>@@

@@[up to 8 corrections of one category and type, largest gain first]@@
Correction 1
- before: Chamfer @@<cd>@@, @@<n>@@ piece(s), parts just off the floor: @@<parts>@@, fell: @@<fell>@@, @@<n>@@ primitives
- after: Chamfer @@<cd>@@, @@<n>@@ piece(s), parts just off the floor: @@<parts>@@, fell: @@<fell>@@, @@<n>@@ primitives
- a reviewer comparing the photos with renders of the result scored it @@<a>@@ -> @@<b>@@/100: @@<the criteria that changed, with evidence>@@
- what changed: @@<parts added, removed, resized, moved or given another material>@@
- the reconstructor's own reasoning: @@<the critique that made this revision>@@

Return operations on the list:
- {"op": "ADD", "scope": {"category": "@@<category>@@" or "global", "part_role": one of ['shade', 'wheel', 'screen', 'armrest', 'back', 'seat', 'base', 'leg', 'pole', 'arm', 'top', 'door', 'drawer', 'shelf', 'frame', 'hull', 'mast', 'speaker', 'wing', 'fuselage', 'tail', 'engine', 'barrel', 'stock', 'scope', 'magazine', 'driver', 'keypad', 'body'] or "any"}, "when": "<visible situation in the photo>", "do": "<imperative instruction about the primitives>", "why": "<what goes wrong otherwise>"}
- {"op": "EDIT", "id": "...", ...fields to replace}
- {"op": "UPVOTE", "id": "..."} when a correction confirms an existing lesson
- {"op": "DOWNVOTE", "id": "..."} when a correction contradicts one

Rules: a lesson says what to do and how, never what such objects look like. Never refer to a particular object, photo, run or round. Use "global" only when the lesson holds for every kind of object. Prefer UPVOTE or EDIT over adding a near-duplicate. Keep "when" under 150 characters, "do" under 250 and "why" under 150 -- one sentence each; longer lessons are discarded. Return at most 3 operations, or none if nothing generalises.

Reply with JSON: {"operations": [...]}
\end{lstlisting}

\begingroup
\scriptsize\setlength{\LTpre}{4pt}\setlength{\LTpost}{6pt}
\begin{longtable}{@{}L{1.75cm}c@{\hspace{5pt}}L{4.2cm}L{5.9cm}L{4.3cm}@{}}
  \caption{\normalfont\small\textbf{Insights from the library:} the highest-weight insight of each category and the top three global ones, as stored. Scope is category/part role; $w$ is the weight.}\label{tab:s_insights}\\
  \toprule
  Scope & $w$ & When & Do & Why \\
  \midrule
  \endfirsthead
  \toprule
  Scope & $w$ & When & Do & Why \\
  \midrule
  \endhead
  \bottomrule
  \endlastfoot
global/any & 10 & A part that does not stand on the floor has its lowest point near the ground, such as a brace, a skirt or a low bracket. & Compute that part's lowest z, including any rotation, and either set it exactly to the ground z of the floor-touching parts or raise it clearly above the floor. & A bottom that ends slightly above the floor is flagged as just off the ground and breaks clean floor contact. \\ \addlinespace[2pt]
global/any & 5 & A box-like part such as a carcass, top slab, panel or leg shows crisp, square corners and edges in the photos. & Give that primitive near-zero shape exponents (about 0.1 or less) so its corners come out square, and round corners only as much as the photos show. & Default rounded exponents around 0.2 or more shave off the corners, leaving the true corner surface uncovered along every edge. \\ \addlinespace[2pt]
global/any & 3 & A panel or part is tilted by a rotation, such as a reclined backrest, sloped leg-rest or angled brace that must join another part at one end. & After choosing the rotation sign, compute the heights of both ends and check that the raised end matches the photo; then place the low end inside the part it joins. & A wrong rotation sign raises the wrong end, so the part floats above its neighbour and splits off, and the model may fall in the drop test. \\ \addlinespace[2pt]
airplane/engine & 3 & Engine nacelles or pods are visible on the wings or the rear fuselage, sometimes with gear or wheels hanging beneath them. & Count the pods per side in the photos and model only that many. Measure each pod's spanwise offset, hang height, length and depth against fuselage length in the front, side and top views. Place pods as far outboard and as low as shown. & Pods guessed from memory come out too many, too small, too far inboard or too high, adding false surface and missing the real one. \\ \addlinespace[2pt]
bench/frame & 4 & Intermediate supports, uprights or under-seat cross braces are visible along a bench's length between the two end frames. & Count every such support in the photos, including one at the middle if shown, and place each at its measured fraction of seat length rather than at habitual inset positions. & Supports placed by habit cover no real surface, and the true supports at other positions or at the centre stay uncovered. \\ \addlinespace[2pt]
cabinet/shelf & 2 & The interior of a cabinet is visible through an opening, open top or glass door. & Add an interior shelf wherever a photo shows its edge, including through glass, and extend its sides and back slightly into the side and back panels so it overlaps them. & Missing shelves leave real interior surface uncovered, guessed shelves add false surface, and shelves that only graze the inner faces float loose. \\ \addlinespace[2pt]
car/body & 2 & A car's cabin or greenhouse sits on the lower body, visible in side, front and top views. & In the side view, measure the cabin's front and rear ends as fractions of body length and its roof height above the beltline. In the front view, measure roof width against base width, and taper only by that measured ratio. & Guessed cabins stop short of the tail or overhang it, and strong tapers collapse the roof or pull the sides inside the true glass. \\ \addlinespace[2pt]
chair/seat & 4 & A chair seat slab or cushion is visible edge-on in a side or front view, with legs, a base or arms around it. & Measure the seat top's height and the seat edge thickness as fractions of total chair height in the side view, then set the seat's center z and half-height so both match, without defaulting to a thin slab. & Default seats come out too thin or at the wrong height, which leaves a band of real seat rim and seat-top surface uncovered. \\ \addlinespace[2pt]
display/screen & 5 & A display is built as a front screen slab plus a rear housing, wedge or bezel block behind it. & Measure the bezel and housing depth as fractions of screen width in side or top views. Size the slabs to those depths, then set the screen's back face slightly behind the housing's front face so the two overlap. & A too-thin screen slab leaves edge surface uncovered, and a screen that only meets the housing face counts as a separate piece and drops off. \\ \addlinespace[2pt]
lamp/base & 3 & A lamp or post stands on a foot, pedestal or plinth visible in a front or side view below its pole or body. & Measure the base's width, height and flared foot as fractions of total lamp height, and where the pole rises on it (centre or rear), then size the base and offset its centre from the pole to match. & Default bases come out too small, thin or centred under the pole, leaving the real foot rim and its forward part uncovered. \\ \addlinespace[2pt]
loudspeaker/body & 2 & A speaker enclosure's front, side and top are visible, whether it is built as one box or as several side-by-side strips or blocks. & Measure the enclosure's width and depth as fractions of its height in the front and side views, then size the box, or the combined outer faces of all strips, to exactly those ratios rather than default depth or footprint. & Guessed enclosures come out too deep, too shallow or too small in footprint, which adds false surface or leaves bands of real flank uncovered. \\ \addlinespace[2pt]
rifle/magazine & 4 & A magazine, pistol grip or foregrip hangs below a rifle body and is the lowest part near the floor in a side view. & Measure each hanging part's fore-aft width, its drop below the bore and its position along the body as fractions of overall length; size it no narrower or shorter than measured, then set its lowest z to the ground or clearly above it. & Hanging parts sized by habit come out too slim, too short and misplaced, leaving real surface uncovered and their bottoms hovering just off the floor. \\ \addlinespace[2pt]
sofa/seat & 3 & Sofa armrests or arm blocks rest on or beside a seat base or cushion that is visible edge-on in a side or front view. & Measure the seat base thickness from the photographed edge and set its half-height from it, then check that each arm's bottom sinks below the seat top and its inner face reaches inside the seat's side edge. & A thin default seat leaves the arms hovering just above or beside it, so they split off as separate pieces and fall in the drop test. \\ \addlinespace[2pt]
table/leg & 3 & A table's full height and top length are both visible in a side or front view. & Measure total height as a fraction of top length and leg and top thickness against that height, then set leg length and slab thicknesses from those ratios, not from defaults. & Default heights make the table too squat or too tall and default thicknesses bulk it up, leaving the true top surface uncovered. \\ \addlinespace[2pt]
telephone/body & 4 & A handheld or desk telephone body is visible face-on in a front view and edge-on in a side view, possibly as a wedge thicker at one end. & Measure the body's width and its front-to-back thickness at top and bottom as fractions of its height, then set the width half-size, depth half-size and any taper to match, never padding the depth past the photographed edge. & Padded slabs add false flank surface without covering recessed detail, and too-thin or too-wide ones leave real casing uncovered or tip over. \\ \addlinespace[2pt]
watercraft/body & 4 & A raised cockpit, engine housing or deck block sits on the aft or fore deck of a hull, seen in side and top views. & Measure the block's fore and aft ends as fractions of hull length and its height above the deck as a fraction of hull depth, run it as far toward the transom as shown, and move rails or struts on it to match. & Blocks sized by habit stop short of the stern and sit too low, leaving the real aft and upper housing surface uncovered. \\ \addlinespace[2pt]

\end{longtable}
\endgroup

\begin{figure}[H]
  \centering
  \input{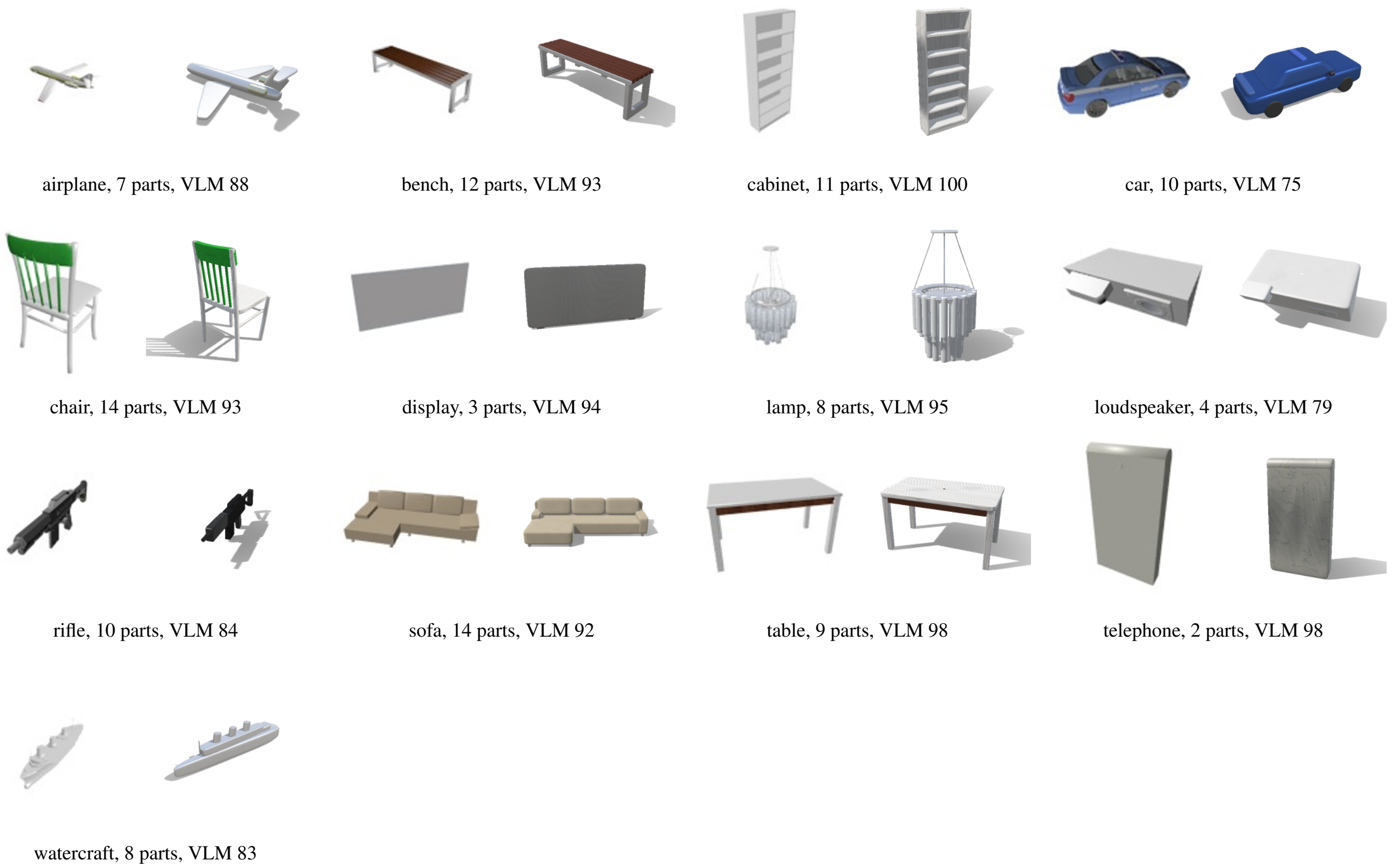}
  \caption{\textbf{Library exemplars:} the highest-graded exemplar of each category. Each thumbnail is what the model is shown, the training photo on the left and the decomposition in its materials on the right.}
  \label{fig:s_exemplars}
\end{figure}

\textbf{Exemplars.}
An exemplar is a delivered PhysDec-Loop result that is physically clean, with Chamfer distance at most its category's median and a VLM score at least its category's median.
The model sees the thumbnail (\cref{fig:s_exemplars}) and the decomposition as JSON, under an instruction to take the part breakdown and joints from it and every proportion from its own photos.

\textbf{Material priors.}
Priors count the material and form chosen for each part role in clean results: 17 human-checked results, plus those whose materials scored full marks in every rubric grade.
\Cref{tab:s_priors} prints the two most frequent choices per role, which is what a prompt receives.

\begin{table}[H]
  \centering\scriptsize
  \caption{\small\textbf{Material priors,} as material/form, the more frequent first.}
  \label{tab:s_priors}
  \begin{tabular}{@{}L{1.5cm}L{15.5cm}@{}}
    \toprule
    Category & Part role and its two most frequent choices \\
    \midrule
    
airplane & \textit{engine} metal/shell, metal/solid; \textit{fuselage} metal/shell; \textit{hull} metal/shell; \textit{pole} metal/solid; \textit{tail} metal/shell, metal/solid; \textit{wing} metal/shell, metal/solid \\ \addlinespace[2pt]
bench & \textit{arm} metal/tube, metal/solid; \textit{armrest} metal/tube, wood/solid; \textit{back} wood/solid, metal/tube; \textit{base} metal/solid; \textit{frame} metal/tube, wood/solid; \textit{leg} metal/tube, wood/solid; \textit{pole} metal/tube, wood/solid; \textit{seat} wood/solid, metal/shell \\ \addlinespace[2pt]
cabinet & \textit{back} wood/solid, rubber/solid; \textit{base} wood/solid, metal/solid; \textit{body} metal/shell, wood/shell; \textit{door} wood/solid, metal/shell; \textit{drawer} wood/solid, wood/shell; \textit{frame} metal/shell; \textit{pole} wood/solid; \textit{screen} glass/solid; \textit{shelf} wood/solid; \textit{top} wood/solid; \textit{wheel} rubber/solid \\ \addlinespace[2pt]
car & \textit{body} metal/shell; \textit{engine} metal/shell; \textit{hull} metal/shell; \textit{tail} metal/shell; \textit{wheel} rubber/shell; \textit{wing} metal/solid \\ \addlinespace[2pt]
chair & \textit{arm} upholstery/solid, plastic/solid; \textit{armrest} plastic/solid, upholstery/solid; \textit{back} metal/tube, upholstery/solid; \textit{base} upholstery/solid, metal/solid; \textit{frame} wood/solid, metal/tube; \textit{leg} metal/tube, wood/solid; \textit{pole} metal/tube, wood/solid; \textit{seat} upholstery/solid, wood/solid; \textit{wheel} rubber/solid, plastic/solid; \textit{wing} upholstery/solid \\ \addlinespace[2pt]
display & \textit{arm} metal/solid, metal/tube; \textit{back} plastic/shell, plastic/solid; \textit{base} metal/solid, plastic/solid; \textit{body} plastic/shell; \textit{frame} metal/tube; \textit{leg} metal/tube; \textit{pole} plastic/solid, metal/tube; \textit{screen} plastic/shell; \textit{shelf} plastic/solid \\ \addlinespace[2pt]
lamp & \textit{arm} metal/tube, plastic/solid; \textit{base} metal/solid, plastic/shell; \textit{body} ceramic/shell; \textit{frame} metal/tube; \textit{leg} ceramic/solid, metal/solid; \textit{pole} metal/tube, ceramic/solid; \textit{shade} fabric/shell, metal/shell; \textit{top} metal/solid \\ \addlinespace[2pt]
loudspeaker & \textit{back} plastic/shell, plastic/solid; \textit{base} metal/solid; \textit{body} plastic/shell; \textit{driver} plastic/solid, plastic/shell \\ \addlinespace[2pt]
rifle & \textit{arm} plastic/solid; \textit{barrel} metal/tube, metal/solid; \textit{body} metal/shell; \textit{frame} metal/solid, plastic/solid; \textit{leg} metal/tube, plastic/solid; \textit{magazine} metal/shell, metal/solid; \textit{pole} plastic/solid, metal/tube; \textit{scope} metal/solid, metal/tube; \textit{stock} metal/tube, wood/solid \\ \addlinespace[2pt]
sofa & \textit{arm} upholstery/solid, leather/solid; \textit{armrest} leather/solid, upholstery/solid; \textit{back} upholstery/solid, leather/solid; \textit{base} upholstery/solid, leather/solid; \textit{leg} metal/tube, wood/solid; \textit{seat} upholstery/solid, leather/solid \\ \addlinespace[2pt]
table & \textit{back} metal/tube, wood/solid; \textit{base} metal/shell, wood/solid; \textit{drawer} metal/shell, plastic/solid; \textit{frame} metal/tube, wood/solid; \textit{leg} wood/solid, metal/tube; \textit{pole} wood/solid, metal/tube; \textit{seat} wood/solid; \textit{shelf} glass/solid, wood/solid; \textit{top} wood/solid, glass/solid \\ \addlinespace[2pt]
telephone & \textit{back} plastic/solid; \textit{body} plastic/shell, metal/shell; \textit{keypad} plastic/solid; \textit{screen} glass/solid; \textit{top} plastic/shell \\ \addlinespace[2pt]
watercraft & \textit{arm} plastic/solid; \textit{body} plastic/shell, metal/shell; \textit{frame} wood/solid; \textit{hull} plastic/shell, metal/shell; \textit{mast} metal/tube, fabric/shell; \textit{pole} metal/tube; \textit{tail} metal/solid, wood/solid; \textit{top} plastic/shell \\ \addlinespace[2pt]

    \bottomrule
  \end{tabular}
\end{table}

\begin{figure}[H]
  \centering\scriptsize
  \begin{tabular}{@{}c@{\hspace{4pt}}c@{\hspace{4pt}}l@{\hspace{4pt}}c@{\hspace{4pt}}c@{}}
    \toprule
    Input & Retrieved exemplar & Retrieved insights and priors & Base & PhysDec-Evolved \\
    \midrule
    
\begin{minipage}[t]{1.9cm}\centering\vspace{0pt}\includegraphics[width=\linewidth,height=2.25cm,keepaspectratio]{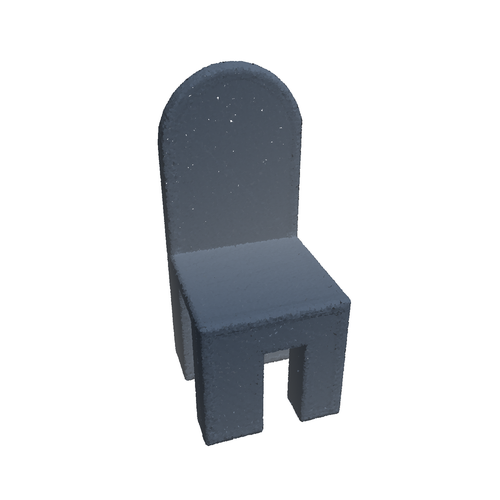}\end{minipage} & \begin{minipage}[t]{3.3cm}\centering\vspace{0pt}\includegraphics[width=\linewidth]{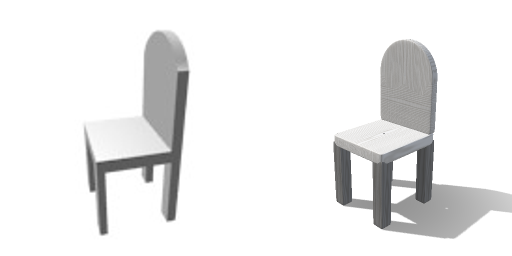}\end{minipage} & \begin{minipage}[t]{7.1cm}\vspace{0pt}\raggedright\fontsize{6}{7}\selectfont \textbf{When} a chair seat slab or cushion is visible edge-on in a side or front view, with legs, a base or arms around it: Measure the seat top's height and the seat edge thickness as fractions of total chair height in the side view, then set the seat's center z and half-height so both match, without defaulting to a thin slab. \textbf{When} chair legs are visible below the seat in front, side and top views: Measure leg thickness and the spread of the feet as fractions of seat width in the front and top views. Put leg tops flush with the seat corners and tilt splayed legs outward so the feet land at the measured spread. \textcolor{inkc!60}{(+6 more)}\\[1pt]\textbf{Priors} \textit{back} metal/tube, upholstery/solid; \textit{leg} metal/tube, wood/solid; \textit{seat} upholstery/solid, wood/solid\end{minipage} & \begin{minipage}[t]{2.0cm}\centering\vspace{0pt}\includegraphics[width=\linewidth,height=2.25cm,keepaspectratio]{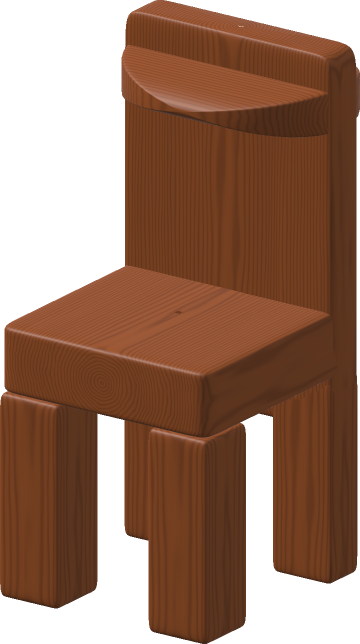}\\[1pt]{\tiny VLM 71.2, 7 parts}\end{minipage} & \begin{minipage}[t]{2.0cm}\centering\vspace{0pt}\includegraphics[width=\linewidth,height=2.25cm,keepaspectratio]{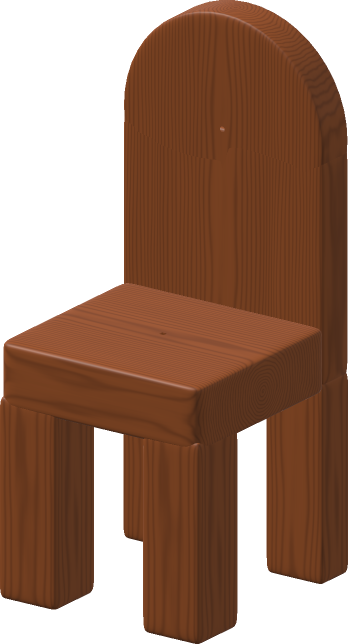}\\[1pt]{\tiny VLM 97.1, 7 parts}\end{minipage} \\ \addlinespace[5pt]
\begin{minipage}[t]{1.9cm}\centering\vspace{0pt}\includegraphics[width=\linewidth,height=2.25cm,keepaspectratio]{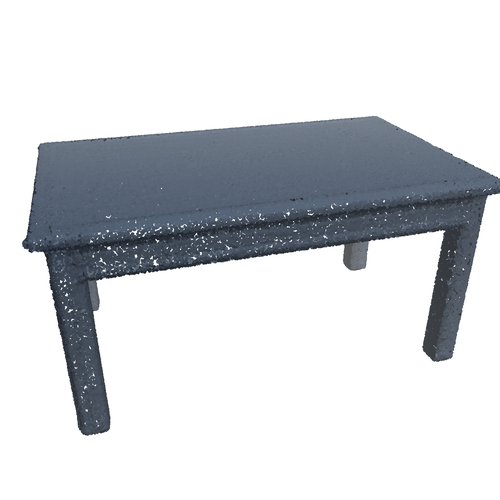}\end{minipage} & \begin{minipage}[t]{3.3cm}\centering\vspace{0pt}\includegraphics[width=\linewidth]{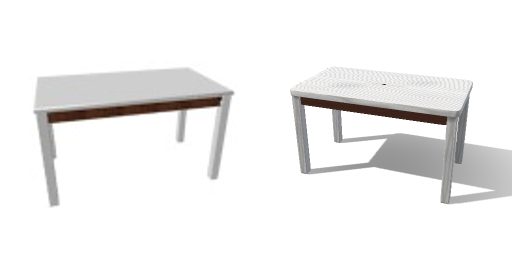}\end{minipage} & \begin{minipage}[t]{7.1cm}\vspace{0pt}\raggedright\fontsize{6}{7}\selectfont \textbf{When} a table's full height and top length are both visible in a side or front view: Measure total height as a fraction of top length and leg and top thickness against that height, then set leg length and slab thicknesses from those ratios, not from defaults. \textbf{When} table legs are visible under a tabletop in top, front or side views, whether the top is rectangular, round or polygonal: Count the legs in the photos and put one under each photographed corner or vertex. Measure each leg's inset from the top's edge along both axes as fractions of top length and width, and place legs from those fractions. \textcolor{inkc!60}{(+5 more)}\\[1pt]\textbf{Priors} \textit{top} wood/solid, glass/solid; \textit{frame} metal/tube, wood/solid; \textit{leg} wood/solid, metal/tube\end{minipage} & \begin{minipage}[t]{2.0cm}\centering\vspace{0pt}\includegraphics[width=\linewidth,height=2.25cm,keepaspectratio]{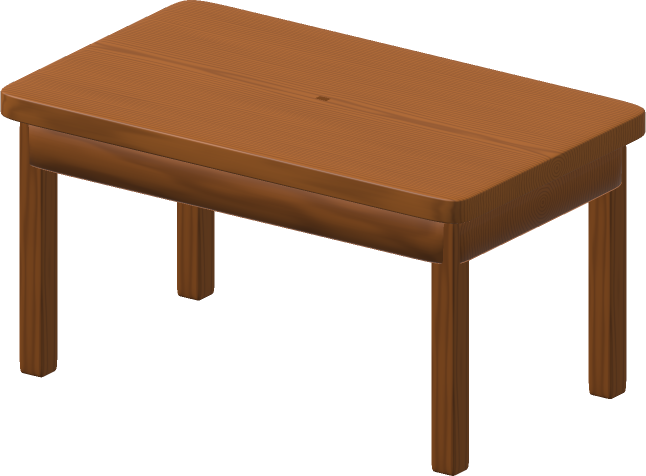}\\[1pt]{\tiny VLM 86.5, 6 parts}\end{minipage} & \begin{minipage}[t]{2.0cm}\centering\vspace{0pt}\includegraphics[width=\linewidth,height=2.25cm,keepaspectratio]{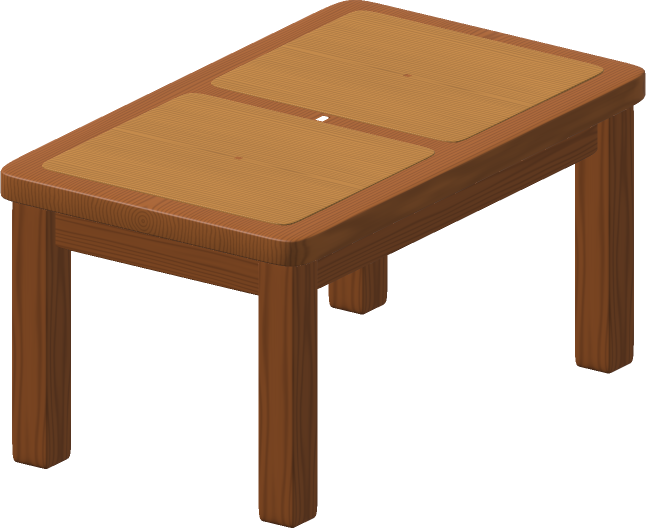}\\[1pt]{\tiny VLM 100.0, 11 parts}\end{minipage} \\ \addlinespace[5pt]
\begin{minipage}[t]{1.9cm}\centering\vspace{0pt}\includegraphics[width=\linewidth,height=2.25cm,keepaspectratio]{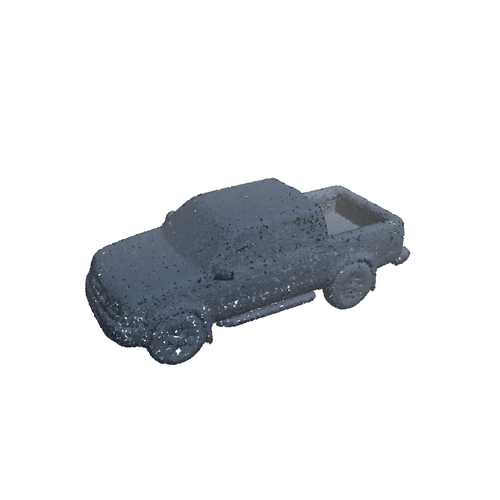}\end{minipage} & \begin{minipage}[t]{3.3cm}\centering\vspace{0pt}\includegraphics[width=\linewidth]{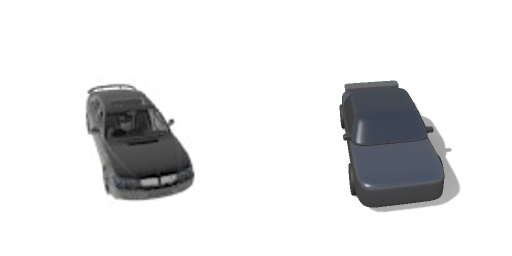}\end{minipage} & \begin{minipage}[t]{7.1cm}\vspace{0pt}\raggedright\fontsize{6}{7}\selectfont \textbf{When} a car's cabin or greenhouse sits on the lower body, visible in side, front and top views: In the side view, measure the cabin's front and rear ends as fractions of body length and its roof height above the beltline. In the front view, measure roof width against base width, and taper only by that measured ratio. \textbf{When} wheels stand outboard of a narrow body or nose, as on open-wheel cars, so the wheel's inner face does not reach the bodywork: Add one thin axle or suspension member per wheel pair that runs from inside each wheel hub through the body, overlapping both. Then check each wheel's inner face against the body edge at that axle position. \textcolor{inkc!60}{(+6 more)}\\[1pt]\textbf{Priors} \textit{body} metal/shell; \textit{wheel} rubber/shell\end{minipage} & \begin{minipage}[t]{2.0cm}\centering\vspace{0pt}\includegraphics[width=\linewidth,height=2.25cm,keepaspectratio]{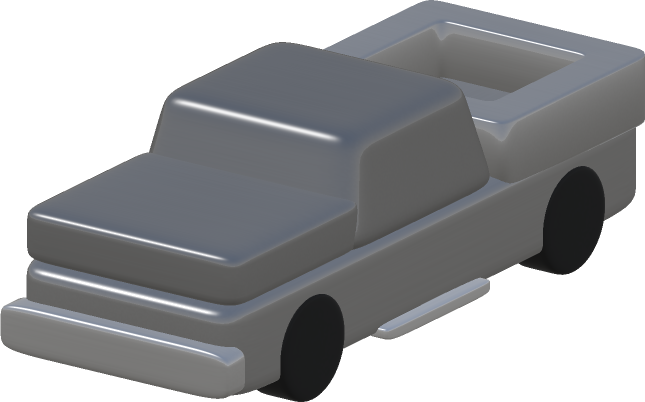}\\[1pt]{\tiny VLM 59.6, 11 parts}\end{minipage} & \begin{minipage}[t]{2.0cm}\centering\vspace{0pt}\includegraphics[width=\linewidth,height=2.25cm,keepaspectratio]{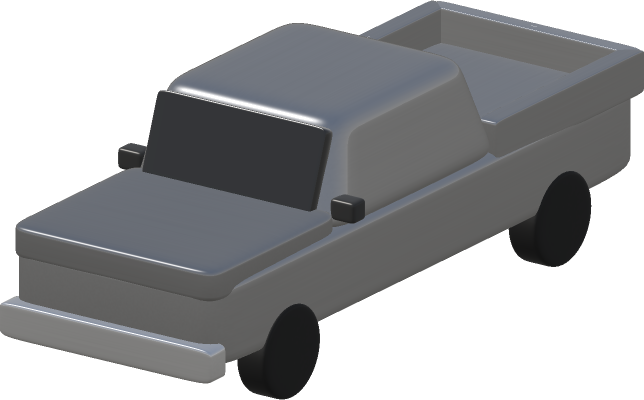}\\[1pt]{\tiny VLM 76.9, 14 parts}\end{minipage} \\ \addlinespace[5pt]
\begin{minipage}[t]{1.9cm}\centering\vspace{0pt}\includegraphics[width=\linewidth,height=2.25cm,keepaspectratio]{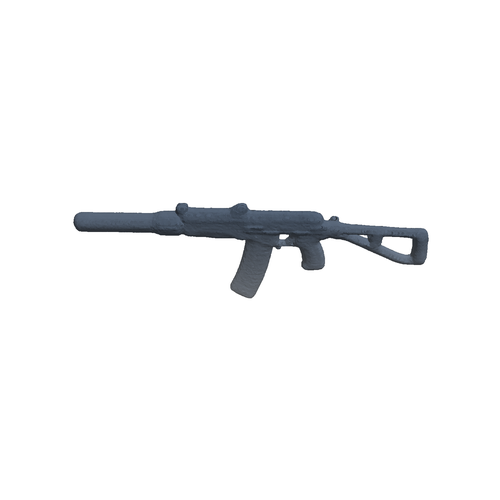}\end{minipage} & \begin{minipage}[t]{3.3cm}\centering\vspace{0pt}\includegraphics[width=\linewidth]{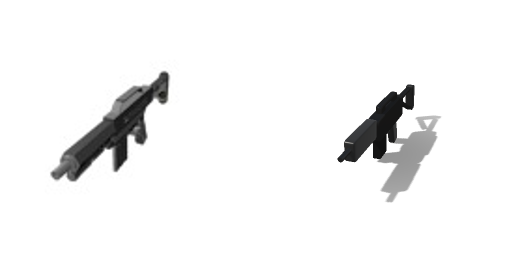}\end{minipage} & \begin{minipage}[t]{7.1cm}\vspace{0pt}\raggedright\fontsize{6}{7}\selectfont \textbf{When} a magazine, pistol grip or foregrip hangs below a rifle body and is the lowest part near the floor in a side view: Measure each hanging part's fore-aft width, its drop below the bore and its position along the body as fractions of overall length; size it no narrower or shorter than measured, then set its lowest z to the ground or clearly above it. \textbf{When} a rifle's buttstock, action, fore-end and exposed barrel are visible end to end in a side view: Measure where each segment starts and ends along the long axis as fractions of overall length, then size and place every segment and its attached scope, bolt and trigger guard from those fractions. \textcolor{inkc!60}{(+6 more)}\\[1pt]\textbf{Priors} \textit{scope} metal/solid, metal/tube; \textit{magazine} metal/shell, metal/solid; \textit{stock} metal/tube, wood/solid\end{minipage} & \begin{minipage}[t]{2.0cm}\centering\vspace{0pt}\includegraphics[width=\linewidth,height=2.25cm,keepaspectratio]{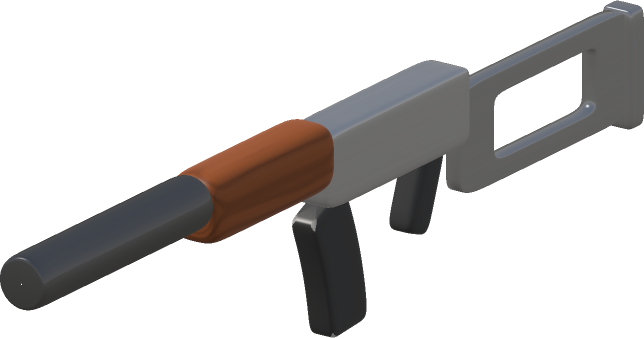}\\[1pt]{\tiny VLM 74.0, 7 parts}\end{minipage} & \begin{minipage}[t]{2.0cm}\centering\vspace{0pt}\includegraphics[width=\linewidth,height=2.25cm,keepaspectratio]{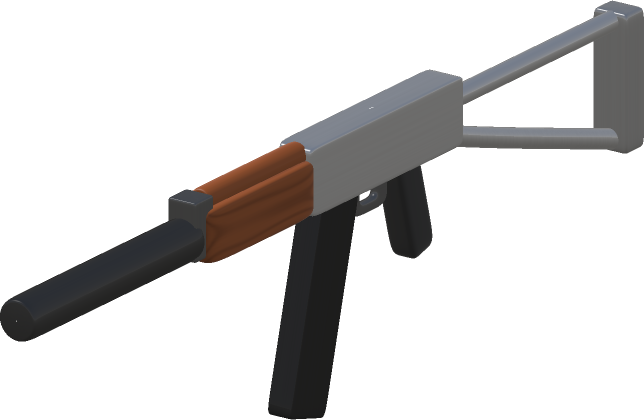}\\[1pt]{\tiny VLM 82.7, 11 parts}\end{minipage} \\ \addlinespace[5pt]

    \bottomrule
  \end{tabular}
  \caption{\textbf{Retrieval on four test objects,} chosen to be legible and among the higher scores, so they do not show the average gain; Table~\mainevolution{} does. Two of the retrieved insights are shown, those of the object's category first. \emph{Base} uses the original prompts; \emph{PhysDec-Evolved} uses the evolved prompts and the library. Both are two-call predictions. Both are drawn as returned by the model, before refinement, from one fixed view with the input photo nearest it, and carry the VLM score of Table~\mainevolution{}. The figure shows what was retrieved and returned, not that one insight caused one change.}
  \label{fig:s_retrieval}
\end{figure}

\textbf{Retrieval.}
The analysis reply names a category and parts.
The category filters the exemplars, and cosine similarity of DINOv2~\cite{oquab2024dinov2} embeddings averaged over the four photos picks the nearest one; an object never retrieves itself.
Up to eight insights follow, global or of this category and for a part role in the analysis, by weight.
Then come the priors for those roles.
If the block outgrows the Propose turn itself, the lowest-weight insights go first.
\Cref{fig:s_retrieval} follows four test objects through this, and \cref{lst:library} prints one block as injected.

\begin{lstlisting}[breakatwhitespace=false,caption={\textbf{Library text as injected} into the Propose turn for the chair of \cref{fig:s_retrieval}, after \texttt{propose.structure}.},label=lst:library]
Reference decomposition of a similar object. The image "library exemplar" shows its photo (left) and that decomposition drawn in its materials (right). Use its part breakdown and how the parts connect; take every proportion from your own photos.
{"primitives":[{"name":"seat","size":[0.1933,0.1804,0.03],"shape":[0.1,0.1],"rotation_deg":[0.0,0.0,0.0],"position":[0.0,0.0,-0.0601],"material":"wood","form":"solid"},{"name":"front_left_leg","size":[0.032,0.032,0.1803],"shape":[0.1,0.1],"rotation_deg":[0.0,0.0,0.0],"position":[0.1612,0.1482,-0.249],"material":"wood","form":"solid"},{"name":"front_right_leg","size":[0.032,0.032,0.1803],"shape":[0.1,0.1],"rotation_deg":[0.0,0.0,0.0],"position":[0.1612,-0.1482,-0.249],"material":"wood","form":"solid"},{"name":"rear_left_leg","size":[0.032,0.032,0.1803],"shape":[0.1,0.1],"rotation_deg":[0.0,0.0,0.0],"position":[-0.1612,0.1482,-0.249],"material":"wood","form":"solid"},{"name":"rear_right_leg","size":[0.032,0.032,0.1803],"shape":[0.1,0.1],"rotation_deg":[0.0,0.0,0.0],"position":[-0.1612,-0.1482,-0.249],"material":"wood","form":"solid"},{"name":"backrest_panel","size":[0.0214,0.1804,0.1417],"shape":[0.1,0.1],"rotation_deg":[0.0,0.0,0.0],"position":[-0.1716,0.0,0.1075],"material":"wood","form":"solid"},{"name":"backrest_arch","size":[0.1804,0.1804,0.0214],"shape":[0.1,1.0],"rotation_deg":[0.0,90.0,0.0],"position":[-0.1716,0.0,0.2492],"material":"wood","form":"solid"}]}

Lessons from earlier reconstructions (apply the ones that match your object):
- When A part that does not stand on the floor has its lowest point near the ground, such as a brace, a skirt or a low bracket: Compute that part's lowest z, including any rotation, and either set it exactly to the ground z of the floor-touching parts or raise it clearly above the floor. (A bottom that ends slightly above the floor is flagged as just off the ground and breaks clean floor contact)
- When A box-like part such as a carcass, top slab, panel or leg shows crisp, square corners and edges in the photos: Give that primitive near-zero shape exponents (about 0.1 or less) so its corners come out square, and round corners only as much as the photos show. (Default rounded exponents around 0.2 or more shave off the corners, leaving the true corner surface uncovered along every edge)
- When A chair seat slab or cushion is visible edge-on in a side or front view, with legs, a base or arms around it: Measure the seat top's height and the seat edge thickness as fractions of total chair height in the side view, then set the seat's center z and half-height so both match, without defaulting to a thin slab. (Default seats come out too thin or at the wrong height, which leaves a band of real seat rim and seat-top surface uncovered)
- When A panel or part is tilted by a rotation, such as a reclined backrest, sloped leg-rest or angled brace that must join another part at one end: After choosing the rotation sign, compute the heights of both ends and check that the raised end matches the photo; then place the low end inside the part it joins. (A wrong rotation sign raises the wrong end, so the part floats above its neighbour and splits off, and the model may fall in the drop test)
- When Chair legs are visible below the seat in front, side and top views: Measure leg thickness and the spread of the feet as fractions of seat width in the front and top views. Put leg tops flush with the seat corners and tilt splayed legs outward so the feet land at the measured spread. (Legs set by habit come out too thin, too far inboard or too upright, which leaves the real outer leg surface uncovered)
- When A part narrows toward one end in the photos, either in width or in height, and you plan to taper it: Set the deform axis to the direction along which it narrows, put the nonzero taper on the local axis that actually shrinks (width or height), and pick the sign so the end shown wide stays wide. (A taper on the wrong local axis or with the wrong sign wedges the part the wrong way, leaving true surface uncovered and making the object tip)
- When A whole layout must be reversed end-to-end along one axis after its front and back were read the wrong way round: Negate every part's position along that axis, negate rotations that tilt parts toward or away from it, and flip the sign of any taper or bend along it; then check that each sloped end still matches the photo. (Moving the positions alone leaves parts sloped and tapered the old way, so tilted panels and tapered ends point backward)
- When A part and its mirrored twin taper, bevel or bend toward one end, such as left and right wings, fins or arms: Give the mirrored copy the opposite taper or bend sign along the mirrored local axis, then confirm that both copies narrow toward their outer ends. (Reusing the same taper sign makes the mirrored copy widen toward its tip, so one side comes out badly misshapen)

Material priors for parts like yours. When the photo shows the material, follow the photo.
- back: metal/tube, upholstery/solid
- leg: metal/tube, wood/solid
- seat: upholstery/solid, wood/solid
\end{lstlisting}

\section{VLM score rubric}
\label{sec:s_rubric}

The VLM scores of Tables~\mainbaselines{}--\mainablation{}, and every review inside the loop, use the same rubric.
\Cref{tab:s_rubric,lst:rubric} give all seven criteria, including material plausibility.
A judge sees only images: the four input photos, and the result drawn through each photo's camera (\cref{fig:s_images}e).
It never sees the primitives, the reasoning or the method's name.
Results with materials are drawn in them; methods that predict none are drawn in a plain material.
For our results only the turn around the object is fitted, to the photos' outlines; baselines already sit in the photos' frame.

The judge grades seven criteria from 0 to 4, writing its evidence before each score.
The overall score is
\begin{equation*}
  v = 100\cdot\frac{\sum_k w_k s_k}{4\sum_k w_k},
\end{equation*}
with weights summing to 13, then capped at 30 if \emph{identity} is at most 1 and at 50 if \emph{inventory} is at most 1.
Each grade is rounded to one decimal place after applying these caps.
Each result is graded twice, independently, and the two scores are averaged.
An unreadable reply is asked again once; a result with no usable grade is left out of the mean.
The judge is Claude Sonnet 5.5; GPT-6.1 Sol also grades the same pictures in Table~\mainbaselines{}, with its lowest reasoning effort.

\begin{table}[H]
  \centering\footnotesize
  \caption{\textbf{Rubric criteria and weights.} The weights sum to 13.}
  \label{tab:s_rubric}
  \begin{tabular}{@{}lcL{10.8cm}@{}}
    \toprule
    Criterion & Weight & Asks about \\
    \midrule
    proportions & 3 & height, width and depth, and the outline in every view \\
    placement & 3 & each major part's relative size, position, attachment and symmetry \\
    inventory & 2 & the major parts against the judge's own list: missing or extra \\
    part\_shape & 2 & each major part's own shape, including an open part drawn solid \\
    identity & 1 & whether someone would recognise this exact object \\
    structure & 1 & whether parts touch what holds them and it would stand or hang like the real object \\
    materials & 1 & whether each part's material agrees with the input images \\
    \bottomrule
  \end{tabular}
\end{table}

\begin{lstlisting}[caption={\textbf{Rubric, system prompt.}},label=lst:rubric_system]
You grade 3D reconstructions of real objects. Each reconstruction is built from a small number of simple primitive shapes -- rounded boxes, cylinders, ellipsoids, tapered or bent blocks, tubes -- so it is blocky by design. Grade how well the primitives capture the object in the photos, not how realistic the rendering looks. You only see images. Reply with a single JSON object and nothing else.
\end{lstlisting}

\lstinputlisting[caption={\textbf{Rubric, user prompt.} Repeat grades add a last line, ``(Independent grading pass 2.)'', so that each is a new sample.},label=lst:rubric]{supp/prompts/rubric.txt}

\begin{lstlisting}[caption={\textbf{A rubric reply.} A real example: the first of the two grades of the chair's delivered result in \cref{fig:s_images}e. The two grades give 90.4 and 84.6, so this result scores 87.5. White space is condensed here.},label=lst:rubric_reply]
{
  "major_parts": ["arched back panel", "seat block", "four legs"],
  "proportions": {"score": 4, "evidence": "In all four views the overall height-to-width-to-depth and the outline closely match the photos. The tall arched back, boxy seat and short legs have about the right ratios."},
  "placement": {"score": 3, "evidence": "The back panel sits at the rear edge of the seat and the four legs are at the corners, as in the photos. In the photos the back panel runs down to the floor as the rear legs. In the reconstruction the back stops at the seat and separate rear legs sit below it. The result looks nearly the same from the viewpoints shown."},
  "inventory": {"score": 4, "evidence": "missing: nothing; extra: nothing. Seat, back and four legs are all present."},
  "part_shape": {"score": 3, "evidence": "The arched back is a thick flat panel, the seat is a square slab and the legs are square blocks, all matching the photos. The back panel is shaped as a separate piece rather than continuing into the rear legs, which is a minor difference."},
  "identity": {"score": 4, "evidence": "It is clearly the same chunky wooden chair with an arched back and thick block legs."},
  "structure": {"score": 4, "evidence": "The legs reach the floor, the seat rests on them and the back is attached to the seat. It would stand like the real chair."},
  "materials": {"score": 4, "evidence": "Brown wood in the reconstruction matches the brown wooden look of the photos. The grain texture is a harmless addition."},
  "orientation": "yes"
}
\end{lstlisting}

\section{Chamfer distance and decomposition quality}
\label{sec:s_chamfer}
\label{sec:s_part_errors}

\paragraph{Limitations of Chamfer distance.}
Chamfer distance (CD) measures surface proximity, but does not fully capture the quality of a primitive decomposition.
Small parts can be important for recognizing or supporting an object while contributing little to its average surface error; CD also does not directly assess connectivity, stability or material-dependent mass distributions.
Prior work shows that CD can overlook local point-density mismatches~\cite{wu2021dcd}, ignores surface normals~\cite{li2025ncd}, and can miss fine geometric details that matter to human perception~\cite{luan2024saucd,feng2026locase}.
When used as a mesh reconstruction loss, CD can also lead to clustered vertices and twisted faces~\cite{zeng2023cd2}.
These observations motivate evaluating CD alongside visual quality and physical plausibility.

\begin{figure}[!t]
  \centering
\setlength{\tabcolsep}{0pt}\renewcommand{\arraystretch}{0}
\def\gw{\dimexpr(\linewidth-0.4em-0.4em-1.6em-0.4em-0.4em)/6\relax}
\begin{tabular}{@{}c@{\hspace{0.4em}}c@{\hspace{0.4em}}c@{\hspace{1.6em}}c@{\hspace{0.4em}}c@{\hspace{0.4em}}c@{}}
\includegraphics[width=\gw]{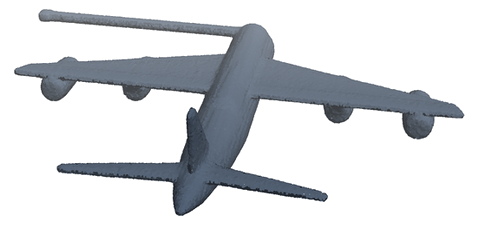} & \includegraphics[width=\gw]{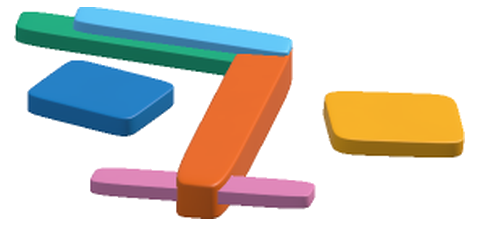} & \includegraphics[width=\gw]{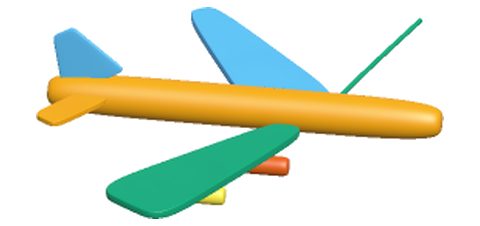} & \includegraphics[width=\gw]{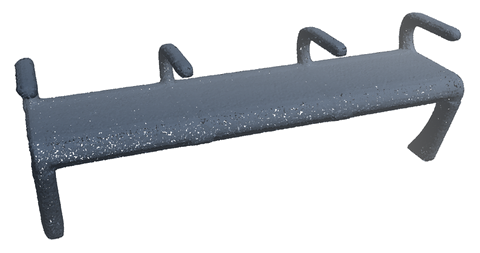} & \includegraphics[width=\gw]{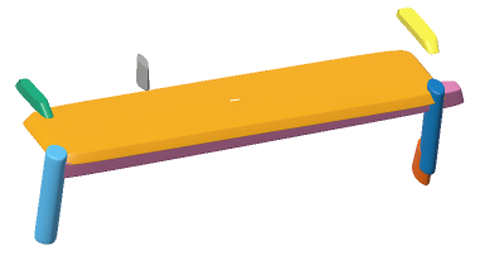} & \includegraphics[width=\gw]{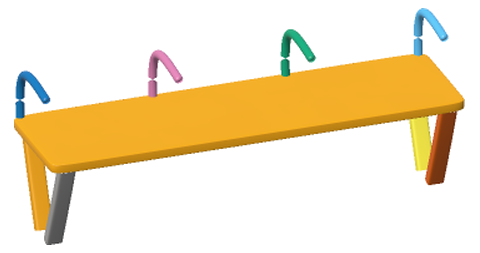} \\[0.4ex]
\parbox[t]{\gw}{\centering\scriptsize\renewcommand{\arraystretch}{1}airplane} & \parbox[t]{\gw}{\centering\scriptsize\renewcommand{\arraystretch}{1}SuperDec\\CD \textbf{1.80}, VLM 29} & \parbox[t]{\gw}{\centering\scriptsize\renewcommand{\arraystretch}{1}PhysDec-Loop\\CD 2.27, VLM \textbf{63}} & \parbox[t]{\gw}{\centering\scriptsize\renewcommand{\arraystretch}{1}bench} & \parbox[t]{\gw}{\centering\scriptsize\renewcommand{\arraystretch}{1}SuperFlex\\CD \textbf{1.15}, VLM 50} & \parbox[t]{\gw}{\centering\scriptsize\renewcommand{\arraystretch}{1}PhysDec-Loop\\CD 1.28, VLM \textbf{86}} \\[1.6ex]
\includegraphics[width=\gw]{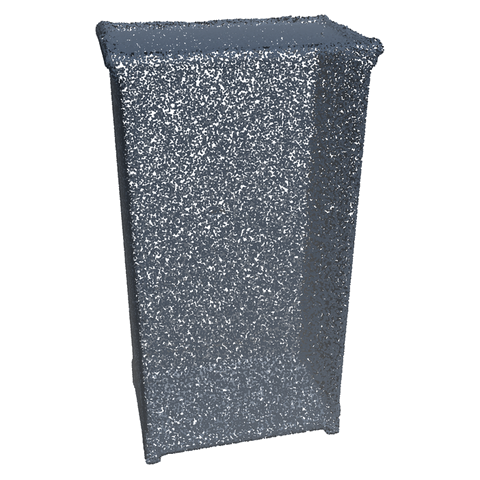} & \includegraphics[width=\gw]{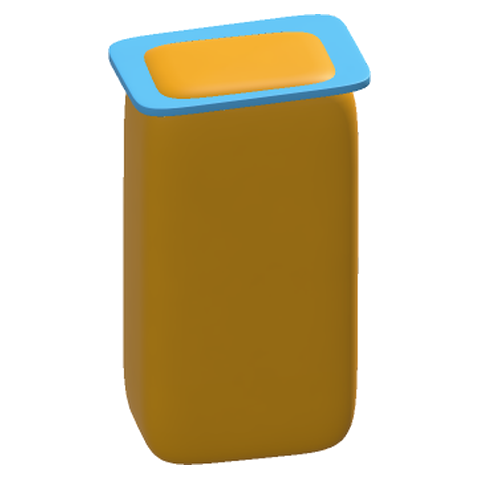} & \includegraphics[width=\gw]{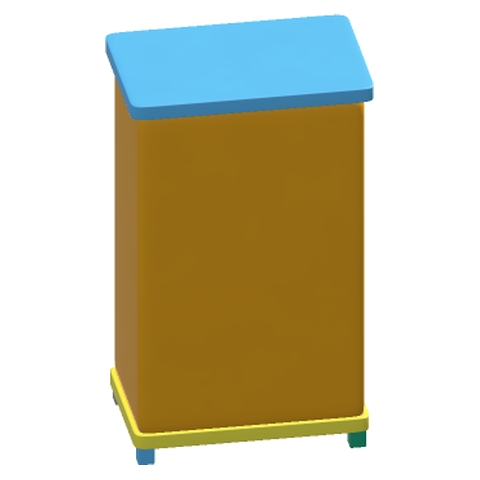} & \includegraphics[width=\gw]{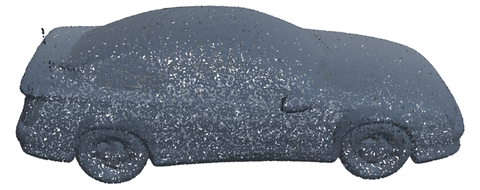} & \includegraphics[width=\gw]{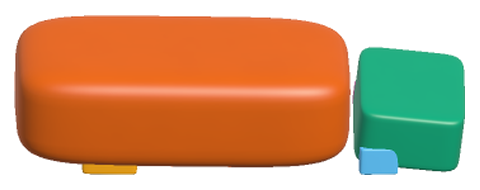} & \includegraphics[width=\gw]{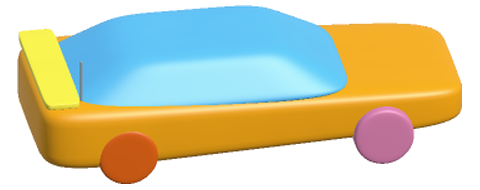} \\[0.4ex]
\parbox[t]{\gw}{\centering\scriptsize\renewcommand{\arraystretch}{1}cabinet} & \parbox[t]{\gw}{\centering\scriptsize\renewcommand{\arraystretch}{1}SuperDec\\CD \textbf{1.47}, VLM 30} & \parbox[t]{\gw}{\centering\scriptsize\renewcommand{\arraystretch}{1}PhysDec-Loop\\CD 1.94, VLM \textbf{85}} & \parbox[t]{\gw}{\centering\scriptsize\renewcommand{\arraystretch}{1}car} & \parbox[t]{\gw}{\centering\scriptsize\renewcommand{\arraystretch}{1}SuperDec\\CD \textbf{1.62}, VLM 24} & \parbox[t]{\gw}{\centering\scriptsize\renewcommand{\arraystretch}{1}PhysDec-Loop\\CD 1.92, VLM \textbf{79}} \\[1.6ex]
\includegraphics[width=\gw]{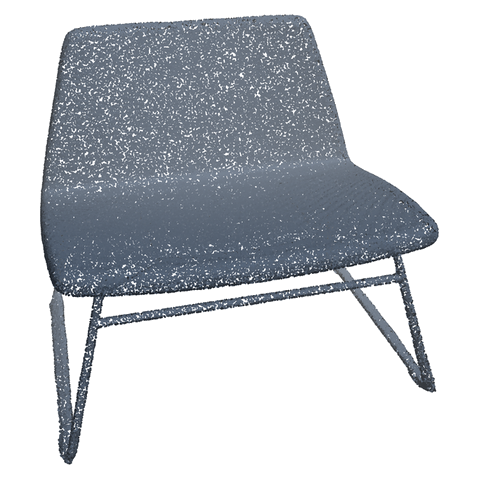} & \includegraphics[width=\gw]{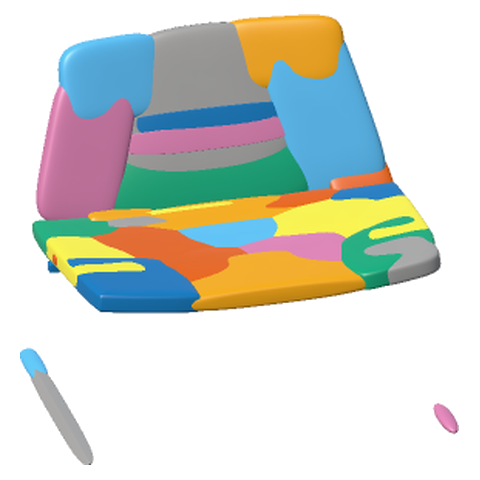} & \includegraphics[width=\gw]{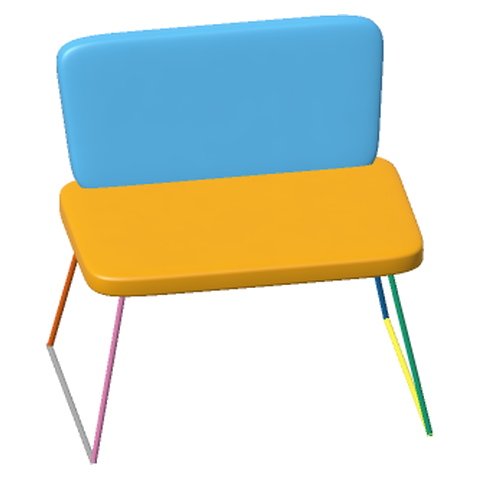} & \includegraphics[width=\gw]{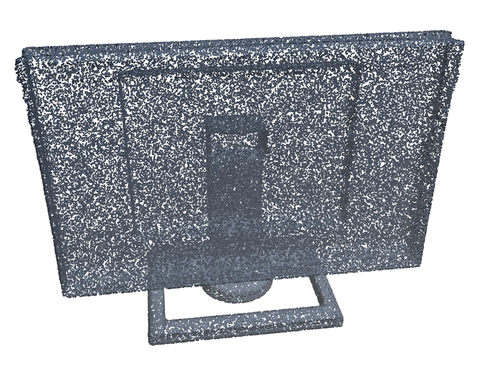} & \includegraphics[width=\gw]{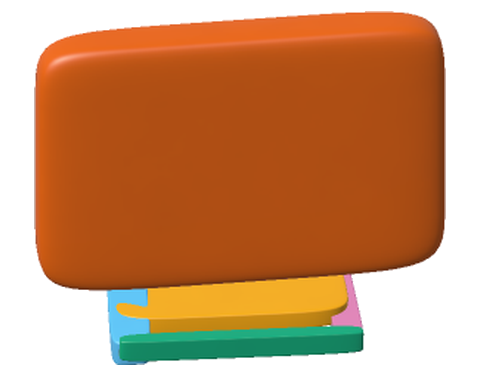} & \includegraphics[width=\gw]{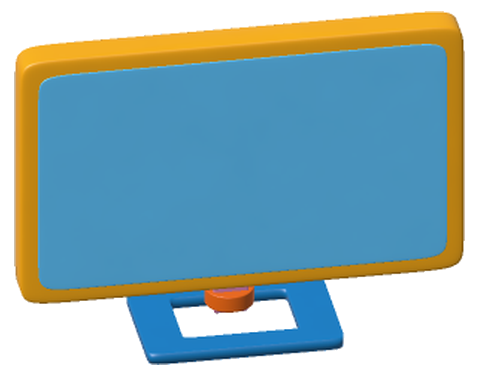} \\[0.4ex]
\parbox[t]{\gw}{\centering\scriptsize\renewcommand{\arraystretch}{1}chair} & \parbox[t]{\gw}{\centering\scriptsize\renewcommand{\arraystretch}{1}Marching-Prim.\\CD \textbf{1.83}, VLM 38} & \parbox[t]{\gw}{\centering\scriptsize\renewcommand{\arraystretch}{1}PhysDec-Loop\\CD 2.21, VLM \textbf{83}} & \parbox[t]{\gw}{\centering\scriptsize\renewcommand{\arraystretch}{1}display} & \parbox[t]{\gw}{\centering\scriptsize\renewcommand{\arraystretch}{1}SuperDec\\CD \textbf{2.01}, VLM 62} & \parbox[t]{\gw}{\centering\scriptsize\renewcommand{\arraystretch}{1}PhysDec-Loop\\CD 2.32, VLM \textbf{85}} \\[1.6ex]
\includegraphics[width=\gw]{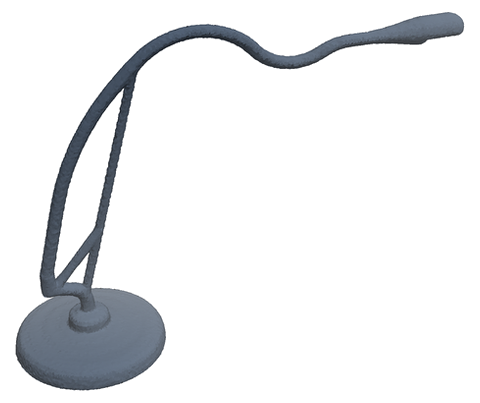} & \includegraphics[width=\gw]{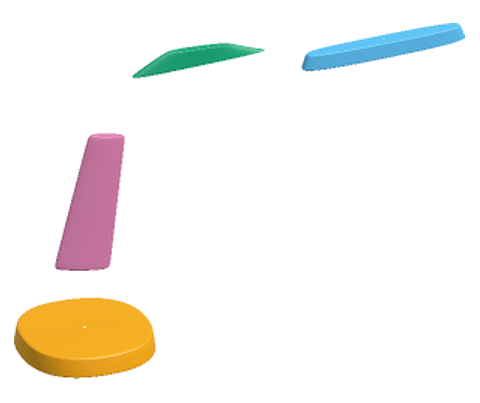} & \includegraphics[width=\gw]{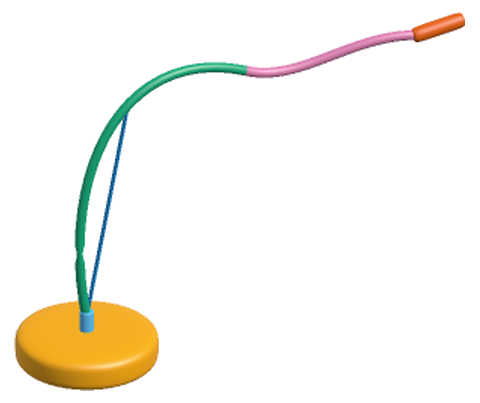} & \includegraphics[width=\gw]{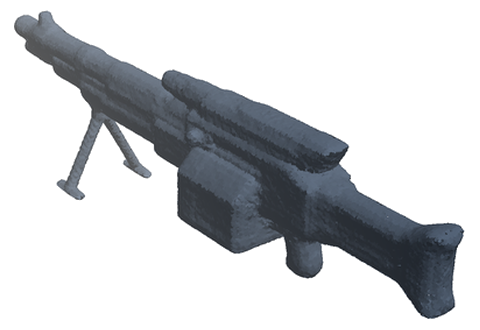} & \includegraphics[width=\gw]{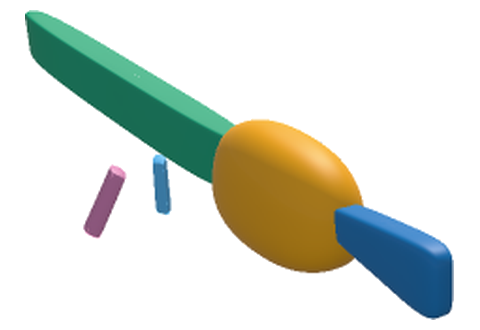} & \includegraphics[width=\gw]{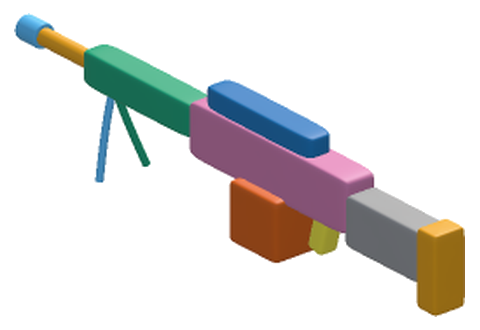} \\[0.4ex]
\parbox[t]{\gw}{\centering\scriptsize\renewcommand{\arraystretch}{1}lamp} & \parbox[t]{\gw}{\centering\scriptsize\renewcommand{\arraystretch}{1}SuperFlex\\CD \textbf{1.16}, VLM 28} & \parbox[t]{\gw}{\centering\scriptsize\renewcommand{\arraystretch}{1}PhysDec-Loop\\CD 1.45, VLM \textbf{83}} & \parbox[t]{\gw}{\centering\scriptsize\renewcommand{\arraystretch}{1}rifle} & \parbox[t]{\gw}{\centering\scriptsize\renewcommand{\arraystretch}{1}SuperFlex\\CD \textbf{1.07}, VLM 30} & \parbox[t]{\gw}{\centering\scriptsize\renewcommand{\arraystretch}{1}PhysDec-Loop\\CD 1.26, VLM \textbf{79}} \\[1.6ex]
\includegraphics[width=\gw]{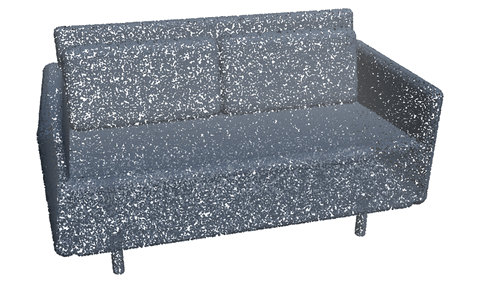} & \includegraphics[width=\gw]{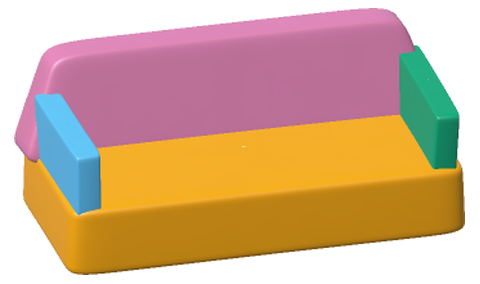} & \includegraphics[width=\gw]{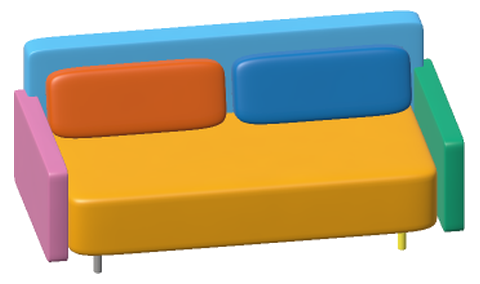} & \includegraphics[width=\gw]{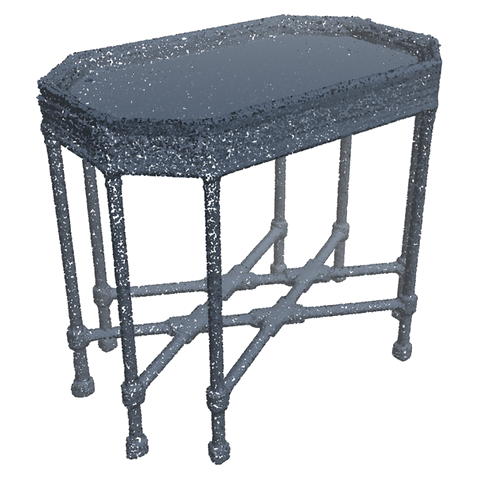} & \includegraphics[width=\gw]{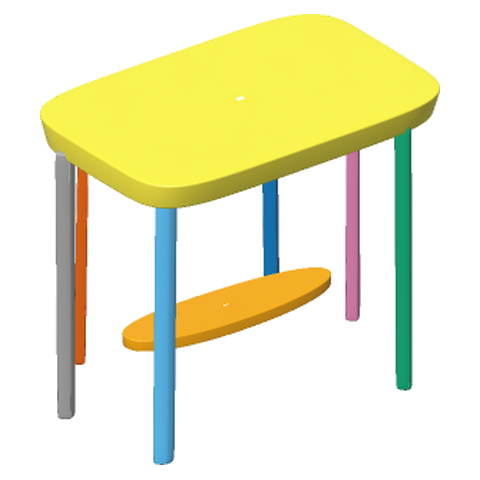} & \includegraphics[width=\gw]{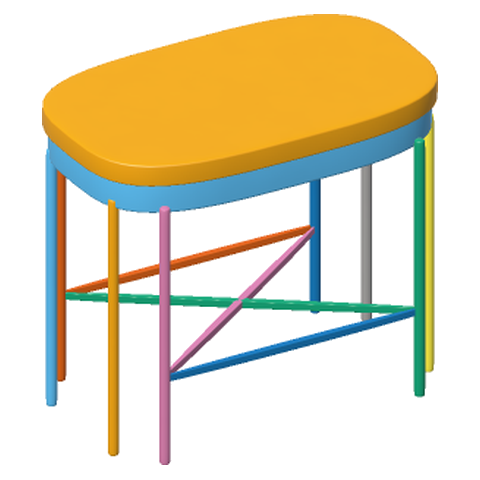} \\[0.4ex]
\parbox[t]{\gw}{\centering\scriptsize\renewcommand{\arraystretch}{1}sofa} & \parbox[t]{\gw}{\centering\scriptsize\renewcommand{\arraystretch}{1}SuperFlex\\CD \textbf{1.67}, VLM 54} & \parbox[t]{\gw}{\centering\scriptsize\renewcommand{\arraystretch}{1}PhysDec-Loop\\CD 2.05, VLM \textbf{100}} & \parbox[t]{\gw}{\centering\scriptsize\renewcommand{\arraystretch}{1}table} & \parbox[t]{\gw}{\centering\scriptsize\renewcommand{\arraystretch}{1}SuperFlex\\CD \textbf{2.52}, VLM 38} & \parbox[t]{\gw}{\centering\scriptsize\renewcommand{\arraystretch}{1}PhysDec-Loop\\CD 2.87, VLM \textbf{74}} \\[1.6ex]
\includegraphics[width=\gw]{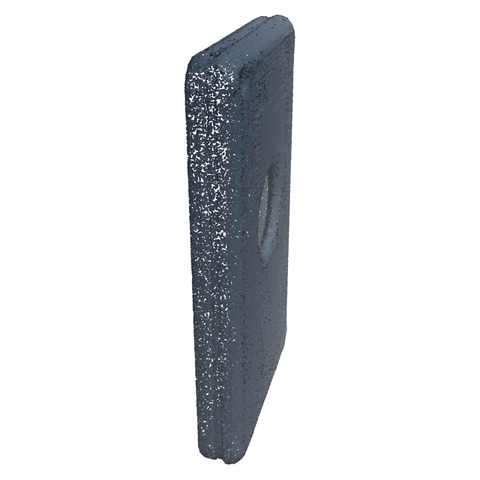} & \includegraphics[width=\gw]{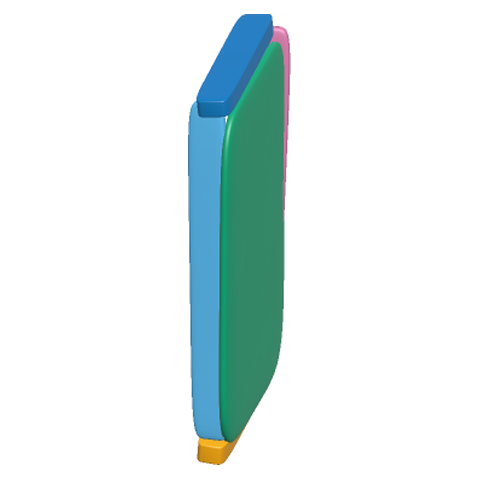} & \includegraphics[width=\gw]{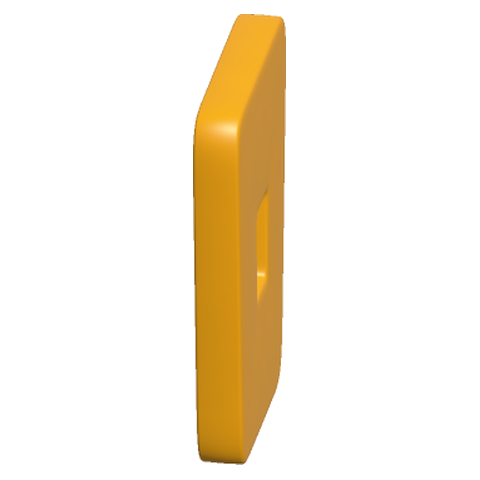} & \includegraphics[width=\gw]{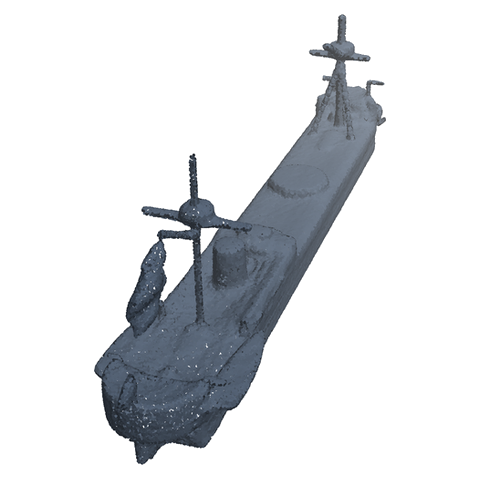} & \includegraphics[width=\gw]{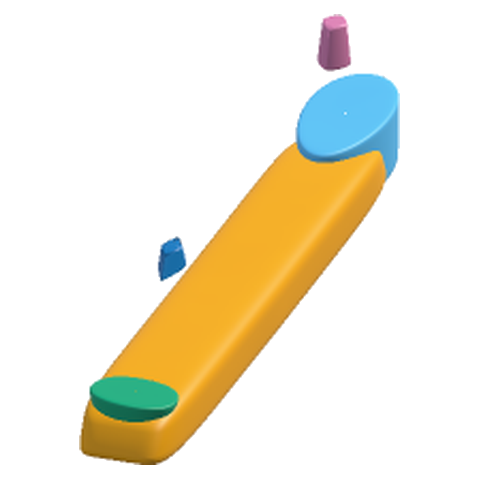} & \includegraphics[width=\gw]{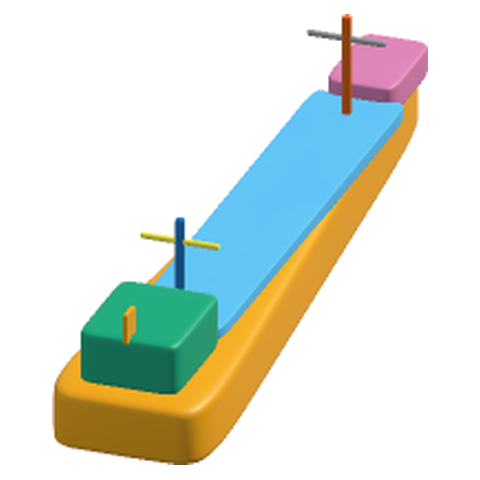} \\[0.4ex]
\parbox[t]{\gw}{\centering\scriptsize\renewcommand{\arraystretch}{1}telephone} & \parbox[t]{\gw}{\centering\scriptsize\renewcommand{\arraystretch}{1}SuperDec\\CD \textbf{1.34}, VLM 30} & \parbox[t]{\gw}{\centering\scriptsize\renewcommand{\arraystretch}{1}PhysDec-Loop\\CD 1.78, VLM \textbf{77}} & \parbox[t]{\gw}{\centering\scriptsize\renewcommand{\arraystretch}{1}watercraft} & \parbox[t]{\gw}{\centering\scriptsize\renewcommand{\arraystretch}{1}SuperFlex\\CD \textbf{1.01}, VLM 43} & \parbox[t]{\gw}{\centering\scriptsize\renewcommand{\arraystretch}{1}PhysDec-Loop\\CD 1.31, VLM \textbf{76}} \\[1.6ex]
\end{tabular}

  \caption{\textbf{Lower Chamfer distance, but a lower VLM score.} Objects on which a baseline's CD is at least 10\% lower than PhysDec-Loop's, but both VLM judges score our decomposition higher; one per shown category, with the largest gap in Sonnet score. Each result is drawn from the input photo's viewpoint, one flat colour per part. CD ($\times100$) and the VLM score (Claude Sonnet~5.5) are those of Table~\mainbaselines{}; the lower CD and the higher score are in bold. Our CD is measured after the evaluation-time alignment and part refinement; the decomposition drawn is the generated one, which the judges graded.}
  \label{fig:s_cd_flip_visual}
\end{figure}

We compare PhysDec-Loop with SQ, EMS, Marching-Primitives, SuperDec and SuperFlex on the 585 objects in Table~\mainbaselines{}.
Across these five baselines, 573 of the 2{,}925 object--method comparisons have a lower baseline CD, covering 249 objects.
Despite the lower CD, Claude Sonnet~5.5 prefers our decomposition in 524 of these pairs (91\%), with GPT-6.1 Sol agreeing in 432.
\Cref{fig:s_cd_flip_visual} illustrates these disagreements using the reported CDs: reference-assisted values for PhysDec-Loop and the baselines' original evaluation values.
We select examples where SuperDec, SuperFlex or Marching-Primitives has at least 10\% lower CD.

\paragraph{Optimization as an optional agent tool.}
When lower CD is a priority, we can provide reference-based optimization as an additional tool for PhysDec-Loop.
The tool aligns a copy of the current decomposition to the reference point cloud and optimizes its part scales and poses.
It returns the suggested changes, the six parts with the largest residual errors and up to three reference regions that remain uncovered.
\textbf{Critique} uses this feedback to guide the next \textbf{Propose} step, allowing the agent to revise the decomposition under the existing geometric and physical checks.
The fitted copy is used for diagnosis; the agent produces the revised primitives.
The initial proposal, other verifiers, round budget and selection rule remain unchanged.
Optimization adds 7.6--20.8\,s per round on an Apple M2 chip using eight CPU cores, so we keep this tool optional rather than include it in the main pipeline.

\paragraph{Chamfer evaluation protocol.}
The ShapeNet baselines take ground-truth point clouds as input (converted to a reference SDF for Marching-Primitives), whereas our VLM proposes primitive parameters from images.
Even a geometrically exact decomposition can receive an elevated CD from a small mismatch in global scale, rotation or translation.
For our predictions, we estimate a global similarity alignment and fit each part's scale, rotation and translation, keeping its primitive family, shape parameters and material assignment fixed.
The fit minimizes symmetric Chamfer-$L_1$ between 8192 sampled surface points and 4096 reference points using Adam for 300 steps, with an initial learning rate of 0.01 and cosine decay.
Final evaluation samples 4096 surface points and reports the mean of the two directed nearest-neighbor distances, multiplied by 100.
\Cref{tab:s_cd_steps} reports CD after alignment, per-axis scaling and part refinement; refinements that change the physical-check outcome are rolled back.
These values measure reference-assisted geometric fit.

\begin{table}[H]
  \centering\footnotesize
  \caption{\textbf{Our mean Chamfer distance after each evaluation step} (CD, $\times100$), on the predictions reported in the main tables. From left to right, each column adds one step: no fitting (only the fixed change of axis convention), global similarity alignment, per-axis scaling and per-part refinement. Refinements that would change the physical-check outcome are rolled back.}
  \label{tab:s_cd_steps}
  \begin{tabular}{@{}lcccc@{}}
    \toprule
    & & Similarity & + Per-axis & + Part refinement \\
    Method & Unaligned & alignment & scaling & (reported) \\
    \midrule
    \multicolumn{5}{@{}l}{\emph{ShapeNet, 585 objects (Tables~\mainbaselines{} and~\mainevolution{})}} \\
    Base & 4.031 & 2.496 & 1.882 & 1.526 \\
    PhysDec-Evolved & 3.961 & 2.470 & 1.883 & 1.528 \\
    PhysDec-Loop & 3.773 & 2.249 & 1.756 & 1.448 \\
    \midrule
    \multicolumn{5}{@{}l}{\emph{Google Scanned Objects, 43 objects (Table~\mainunseen{})}} \\
    Base & 6.121 & 3.280 & 2.665 & 2.127 \\
    PhysDec-Evolved & 6.348 & 3.353 & 2.736 & 2.123 \\
    PhysDec-Loop & 5.125 & 2.826 & 2.375 & 1.894 \\
    \bottomrule
  \end{tabular}
\end{table}

\paragraph{Results with the optimization tool.}
On 585 ShapeNet objects, the agent with this tool achieves lower CD at each evaluation stage than the main pipeline (\cref{tab:s_cd_steps,tab:s_part_error_steps}).
Its generated decompositions have a 98.5\% physical-check pass rate and a VLM score of 78.7.
We also assess the effect of applying evaluation-time transformations to these decompositions.
Similarity alignment retains their original physical and visual scores.
For per-axis scaling, we stretch each decomposition about the center of its footprint, approximating each part by its nearest rotation and local-axis scales.
Part refinement rolls back changes that alter the physical-check outcome before stretching, then applies the retained decomposition's per-axis fit.
We rerun physical checks and visual grading on the scaled and refined geometries.

Per-axis scaling reduces CD from 1.878 to 1.517, with a pass rate of 98.1\% and a VLM score of 78.0.
Optimization therefore offers a way to reduce CD, while these results reinforce that surface fit alone is insufficient to assess decomposition quality.

\begin{table}[!htbp]
  \centering\footnotesize
  \caption{\textbf{Effect of evaluation-time alignment and refinement} on PhysDec-Loop with per-part error feedback, evaluated on 585 ShapeNet objects. Each row adds one step and reports CD ($\times100$), physical-check pass rate (\%) and VLM score (Claude Sonnet~5.5). The final row rolls back refinements that change the physical-check outcome before per-axis scaling. Physical checks and visual grading assess the geometry at each step, with similarity alignment retaining the generated decomposition's scores.}
  \label{tab:s_part_error_steps}
  \setlength{\tabcolsep}{5pt}
  \begin{tabular}{@{}lccc@{}}
    \toprule
    Evaluation step & CD$\downarrow$ & Phys.$\uparrow$ & VLM$\uparrow$ \\
    \midrule
    Unaligned (generated) & 3.446 & 98.5 & 78.7 \\
    + Similarity alignment & 1.878 & 98.5 & 78.7 \\
    + Per-axis scaling & 1.517 & 98.1 & 78.0 \\
    + Part refinement & 1.344 & 97.3 & 72.8 \\
    \bottomrule
  \end{tabular}
\end{table}

\section{Additional evaluation details}
\label{sec:s_evaluation}
\paragraph{Physical model.}
We use the same connectivity and support checks for all methods, with one shared support label per test object.
Two primitives are connected when their sampled surfaces are within $2\%$ of the object's bounding-box diagonal or their volumes overlap.
Each connected component is represented as a single rigid body; disconnected components remain separate bodies.
Convex collision pieces approximate each primitive, with multiple pieces for non-convex shapes.
Thus, the checks assess connectivity and stability assuming rigid connections between touching parts; they do not simulate joint strength or attachment failure.

MuJoCo uses gravity of $9.81\,\mathrm{m/s^2}$, a 2\,ms time step, and sliding, torsional and rolling friction coefficients of 1, 0.02 and 0.01.
Standing and resting tests run for up to 4\,s, with early termination after settling.
The resting test tries four orientations, placing the object's front, back or either side on the support plane, and accepts an object that remains stable in at least one orientation.
Wall contact is evaluated geometrically: surface points within $1\%$ of the bounding-box diagonal of the rearmost surface must cover at least $3\%$ of the projected outline.
The suspension test runs for 5\,s and checks lateral balance and disconnected parts; its allowable center-of-mass offset is $2\%$ of the span.
For airborne and floating objects, this is a balance proxy rather than a simulation of flight or buoyancy.

\paragraph{Materials and mass.}
PhysDec uses the effective density associated with each predicted material and construction form; methods without material predictions use a uniform density of $1000\,\mathrm{kg/m^3}$.
The material choices are inferred without ground-truth material or mass annotations.
The resulting mass distributions are modeling assumptions for the physical checks, rather than measurements of the objects' true masses.
The lamp example in the main paper holds the geometry fixed and changes the density assignment to illustrate its effect on stability.

\paragraph{Primitive count.}
For PhysDec, \#Prim. counts entries in the primitive program: a complete radial array counts once, although its copies are expanded for rendering and simulation.
This count measures program compactness, rather than the number of collision pieces or equal geometric complexity across different primitive vocabularies.

\paragraph{Self-evolution.}
In Table~\mainevolution{} of the main paper, Base and PhysDec-Evolved predictions use no correction loop.
Base and PhysDec-Evolved each require two model calls per prediction; PhysDec-Loop call counts also include VLM review.
For the ShapeNet and unseen-category comparisons, calls count only the run producing the reported result.
The comparison evaluates the combined effect of evolved instructions and retrieved experience.
The seed used to initialize prompt evolution already includes structure rules and retrieval, as detailed in \cref{tab:s_stages}; it differs from the original-prompt Base predictor.
Consequently, the Base-to-Evolved improvement does not isolate automatic prompt rewriting from these other changes.
Call counts describe prediction-time model invocations, excluding offline construction of the library and prompt evolution; they are not measurements of wall-clock time or token cost.

\paragraph{Reference component counts.}
The connectivity penalty in Eq.~(\mainscore{}) assumes a single connected object.
During offline self-evolution, the evaluator instead penalizes $\max(0,c_r-c_{\mathrm{ref}})$, where $c_{\mathrm{ref}}$ is stored in the training and validation reference labels as \texttt{pieces\_ref}.
These counts are initialized from the ground-truth surface point clouds: we voxelize at $1\%$ of the bounding-box diagonal, dilate by one voxel to bridge small gaps, and count connected regions containing at least $1\%$ of the points, with a minimum count of one.
This accommodates samples containing multiple separate objects, such as a monitor with detached speakers.
Missing reference counts default to one, and these labels are used only for offline evaluation, not as inputs to PhysDec-Evolved at deployment.
PhysDec-Loop also supports an explicit object count in scene context when a crop contains multiple objects; ordinary single-object runs use one.

\paragraph{Ablations.}
Table~\mainablation{} uses 195 ShapeNet test objects, with one run per variant and no resampling.
Ablated verifiers are removed from feedback, scoring and selection.
The second view is about $90^\circ$ from the single view.
Model configurations use GPT-6.1 Sol with Sonnet 5.5 review, or Qwen3.8-Flash-Next (FP8) with Qwen3.8-27B review.
Sonnet 5.5 grades every row.
The ``w/o point cloud'' variant removes point-cloud feedback from the correction loop; reference points are still used by the common Chamfer evaluation protocol in \cref{sec:s_chamfer}.


\section{Agreement with human judgments}
\label{sec:s_userstudy}

We assess how well VLM scores reflect human judgments of reconstruction quality.
Table~\mainuserstudy{} of the main paper summarizes the findings.

\paragraph{Setup.}
We randomly sample 65 objects, five per category, from the 390 test objects with available baseline predictions at the time of the study.
Each object is reconstructed by SuperDec, SuperFlex and PhysDec-Loop, yielding 195 results with the VLM scores used in Table~\mainbaselines{}.
Five participants rate all results on a five-point scale from 1 (bad) to 5 (excellent).

Participants compare each reconstruction with the four input images using the same rendered views as the VLM judges.
Methods are anonymized and randomly ordered for each object, with identical rendering styles and no VLM scores shown.
Instructions cover the six geometric and structural criteria in \cref{tab:s_rubric}; material assignments are not shown to participants but account for $1/13$ of the VLM rubric's total weight.

\paragraph{Evaluation scope and participant information.}
The questionnaire collects ratings of 3D reconstruction quality to compare methods and assess agreement with VLM scores.
Its purpose is to evaluate the reconstructions, rather than participants' personal characteristics or experiences.
Participants were informed that their ratings would be used in this CVPR submission.

\begin{figure}[H]
  \centering
  \includegraphics[width=0.85\linewidth]{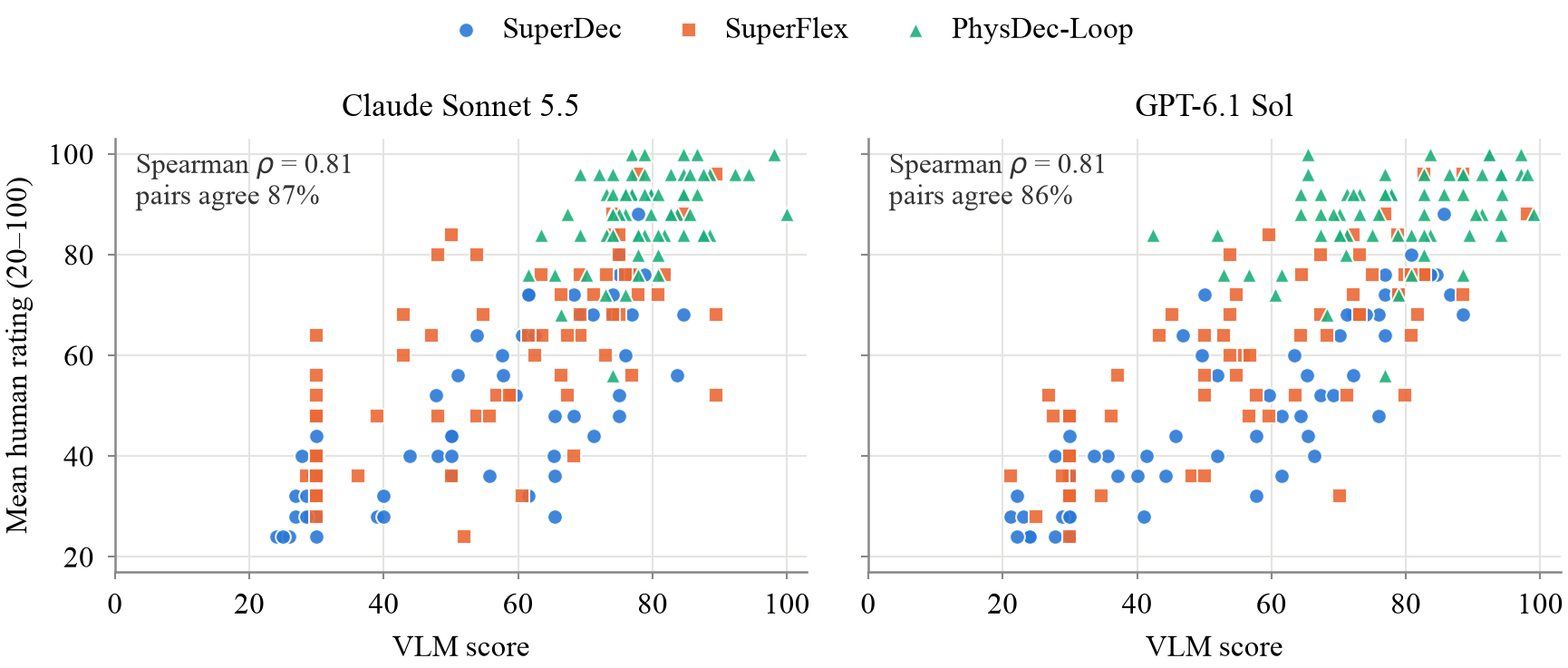}
  \caption{\textbf{VLM scores versus mean human ratings}, rescaled to 20--100. Vertical bands at 30 and 50 reflect the rubric's score caps.}
  \label{fig:s_userstudy}
\end{figure}

\paragraph{Results.}
Both VLM judges correlate strongly with mean human ratings (Spearman $\rho=0.81$) and match human preferences in 86--87\% of within-object comparisons, excluding ties (\cref{tab:s_userstudy,fig:s_userstudy}).
They also reproduce the human ranking of methods, with PhysDec-Loop rated highest (\cref{tab:s_userstudy_methods}).

Inter-rater reliability is Krippendorff's $\alpha=0.74$ (interval).
Comparing each participant with the mean of the other four gives an average $\rho=0.86$ and 94.9\% pairwise agreement.
Thus, the VLM scores broadly reflect human judgments, although agreement remains below that between participants.

\begin{table}[H]
  \centering\footnotesize
  \caption{\textbf{Agreement with human ratings.} Pairwise agreement excludes tied comparisons within each object. Brackets give 95\% confidence intervals for Spearman $\rho$, bootstrapped over objects. The lower block averages leave-one-participant-out comparisons.}
  \label{tab:s_userstudy}
  \begin{tabular}{@{}lcccc@{}}
    \toprule
    & Spearman $\rho$ & Kendall $\tau$ & Pearson $r$ & Pairwise agreement (\%) \\
    \midrule
    \multicolumn{5}{@{}l}{\emph{Against the mean of all five participants}} \\
    \quad Claude Sonnet 5.5 & 0.81 {\scriptsize[0.75, 0.86]} & 0.64 & 0.82 & 86.7 \\
    \quad GPT-6.1 Sol & 0.81 {\scriptsize[0.76, 0.85]} & 0.63 & 0.83 & 86.4 \\
    \multicolumn{5}{@{}l}{\emph{Against the mean of the other four participants}} \\
    \quad A participant & 0.86 & 0.74 & 0.86 & 94.9 \\
    \quad Claude Sonnet 5.5 & 0.81 & 0.64 & 0.81 & 86.9 \\
    \quad GPT-6.1 Sol & 0.80 & 0.63 & 0.83 & 86.5 \\
    \bottomrule
  \end{tabular}
\end{table}

\begin{table}[H]
  \centering\footnotesize
  \caption{\textbf{Mean scores} on 65 objects. Human ratings are linearly rescaled from 1--5 to 20--100.}
  \label{tab:s_userstudy_methods}
  \begin{tabular}{@{}lccc@{}}
    \toprule
    Method & User & Sonnet 5.5 & GPT-6.1 Sol \\
    \midrule
    SuperDec & 47.1 & 51.4 & 50.8 \\
    SuperFlex & 59.3 & 57.2 & 56.6 \\
    PhysDec-Loop & 87.9 & 79.2 & 78.6 \\
    \bottomrule
  \end{tabular}
\end{table}

\section{Random object comparisons}
\label{sec:s_comparisons}

\begin{figure}[p]
  \centering
\setlength{\tabcolsep}{0pt}\renewcommand{\arraystretch}{0}
\def\gw{\dimexpr(\textwidth-7\dimexpr0.25em\relax)/8\relax}
\begin{tabular}{@{}c*{7}{@{\hspace{0.25em}}c}@{}}
\parbox[b]{\gw}{\centering\scriptsize\renewcommand{\arraystretch}{1}Input} & \parbox[b]{\gw}{\centering\scriptsize\renewcommand{\arraystretch}{1}SQ} & \parbox[b]{\gw}{\centering\scriptsize\renewcommand{\arraystretch}{1}EMS} & \parbox[b]{\gw}{\centering\scriptsize\renewcommand{\arraystretch}{1}Marching-Prim.} & \parbox[b]{\gw}{\centering\scriptsize\renewcommand{\arraystretch}{1}SuperDec} & \parbox[b]{\gw}{\centering\scriptsize\renewcommand{\arraystretch}{1}SuperFlex} & \parbox[b]{\gw}{\centering\scriptsize\renewcommand{\arraystretch}{1}PhysDec\\(Evolved)} & \parbox[b]{\gw}{\centering\scriptsize\renewcommand{\arraystretch}{1}PhysDec\\(Loop)} \\[0.5ex]
\includegraphics[width=\gw]{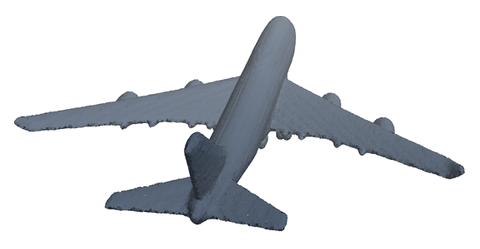} & \includegraphics[width=\gw]{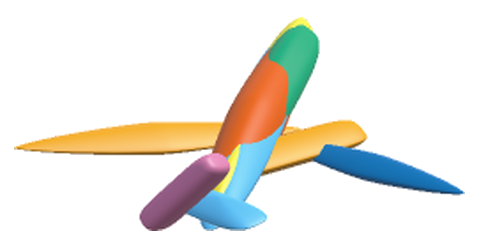} & \includegraphics[width=\gw]{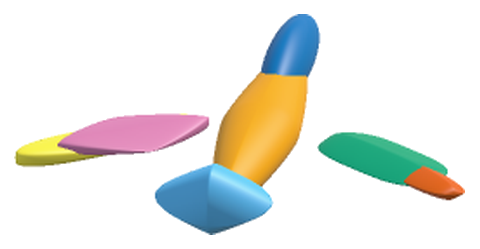} & \includegraphics[width=\gw]{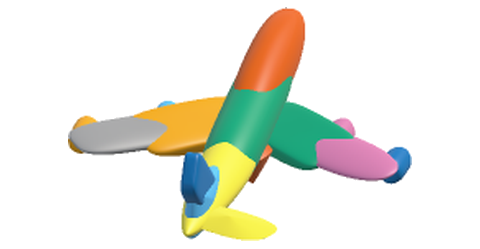} & \includegraphics[width=\gw]{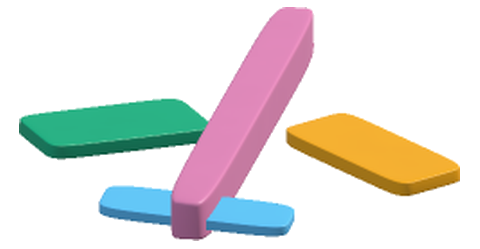} & \includegraphics[width=\gw]{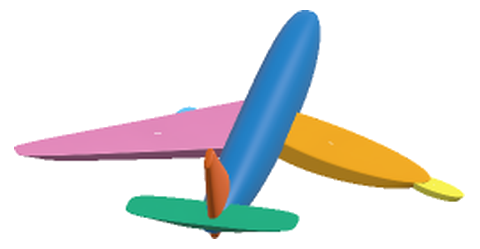} & \includegraphics[width=\gw]{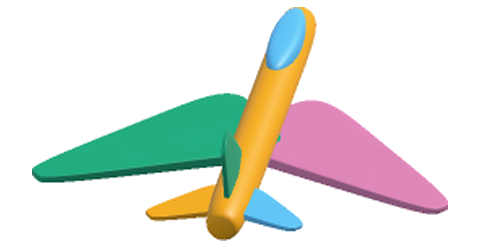} & \includegraphics[width=\gw]{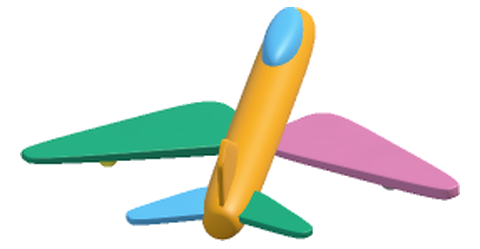} \\[0.4ex]
\scriptsize airplane & \scriptsize 53 & \scriptsize 28 & \scriptsize \textbf{81} & \scriptsize 39 & \scriptsize 50 & \scriptsize 69 & \scriptsize 64 \\[1.4ex]
\includegraphics[width=\gw]{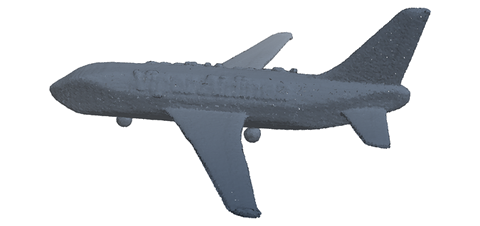} & \includegraphics[width=\gw]{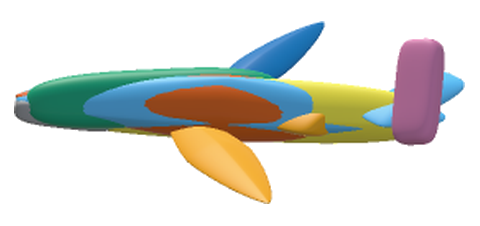} & \includegraphics[width=\gw]{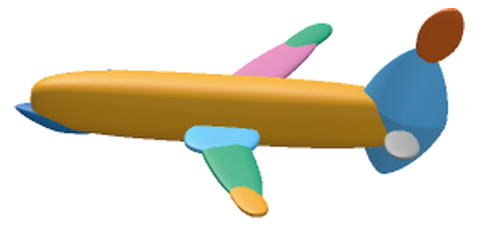} & \includegraphics[width=\gw]{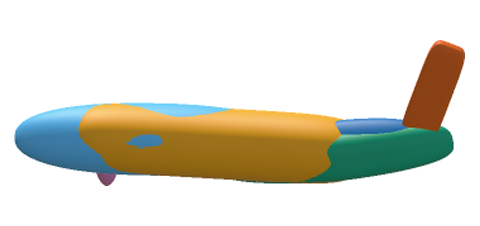} & \includegraphics[width=\gw]{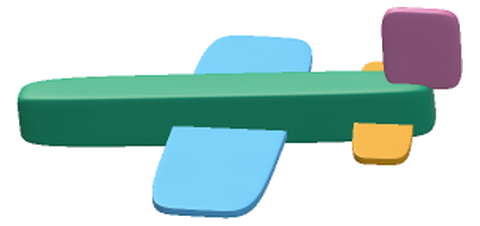} & \includegraphics[width=\gw]{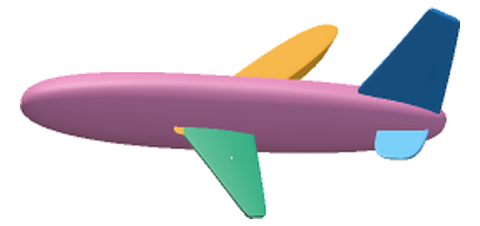} & \includegraphics[width=\gw]{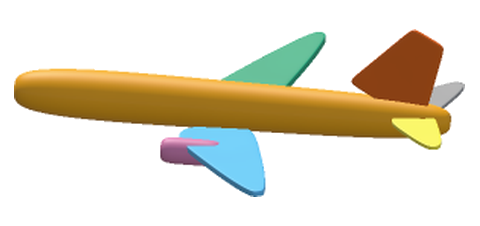} & \includegraphics[width=\gw]{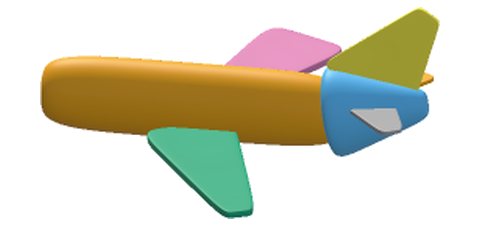} \\[0.4ex]
\scriptsize airplane & \scriptsize 56 & \scriptsize 66 & \scriptsize 30 & \scriptsize 57 & \scriptsize 75 & \scriptsize \textbf{79} & \scriptsize 77 \\[1.4ex]
\includegraphics[width=\gw]{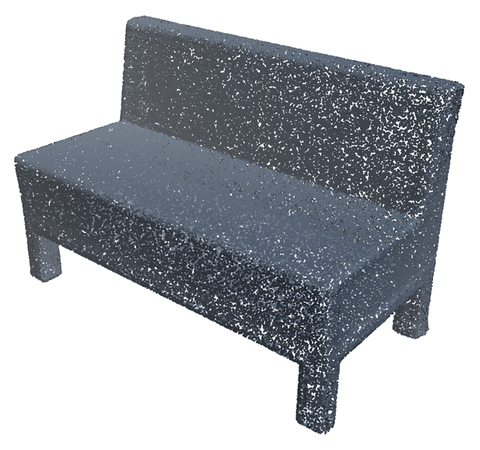} & \includegraphics[width=\gw]{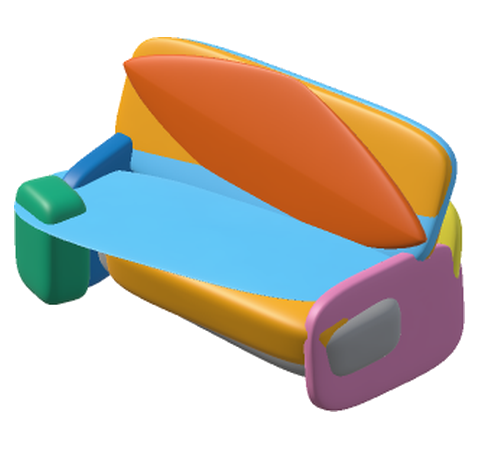} & \includegraphics[width=\gw]{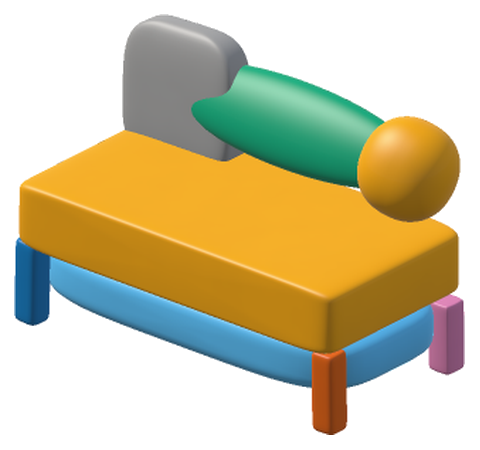} & \includegraphics[width=\gw]{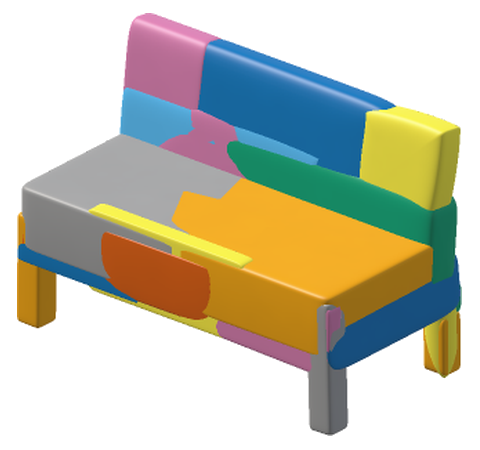} & \includegraphics[width=\gw]{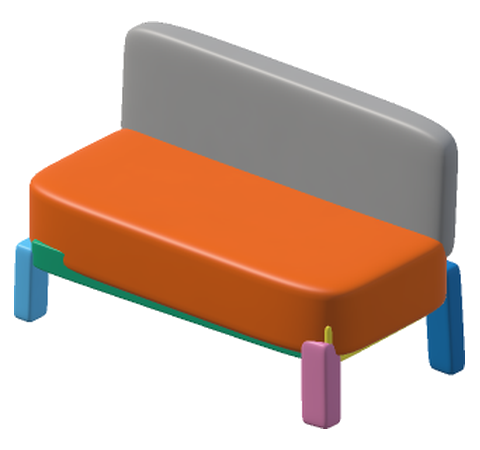} & \includegraphics[width=\gw]{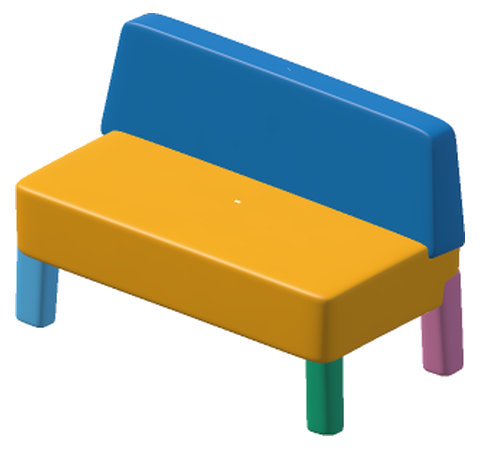} & \includegraphics[width=\gw]{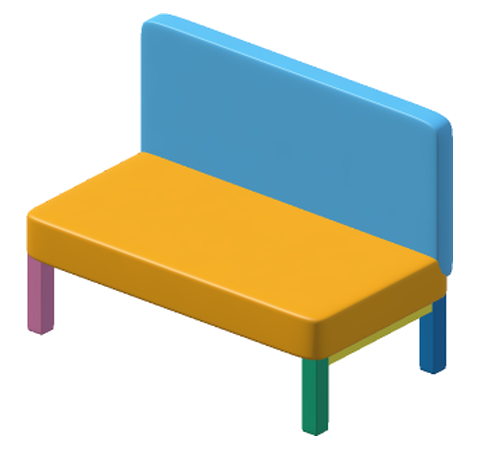} & \includegraphics[width=\gw]{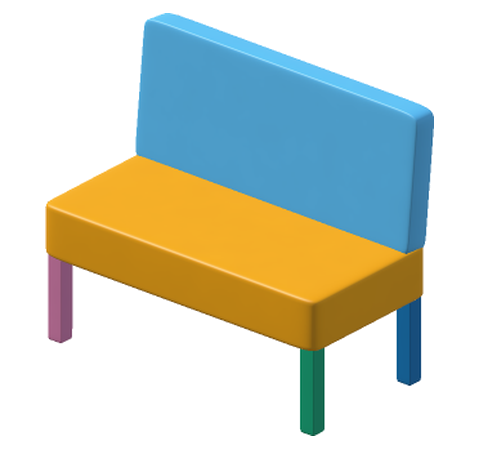} \\[0.4ex]
\scriptsize bench & \scriptsize 38 & \scriptsize 34 & \scriptsize 90 & \scriptsize 77 & \scriptsize 85 & \scriptsize 91 & \scriptsize \textbf{92} \\[1.4ex]
\includegraphics[width=\gw]{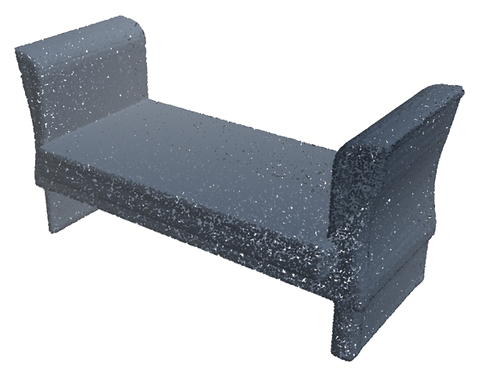} & \includegraphics[width=\gw]{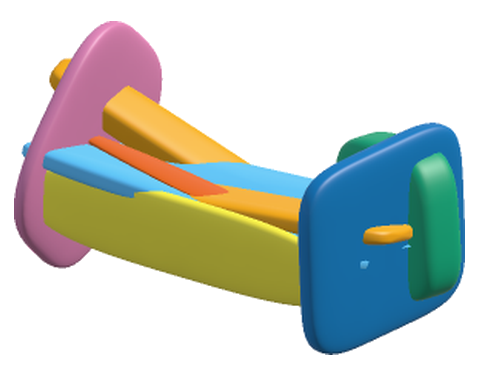} & \includegraphics[width=\gw]{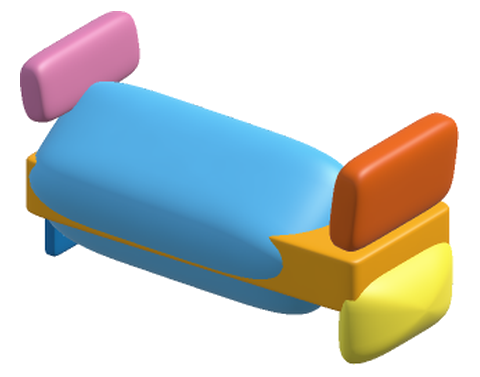} & \includegraphics[width=\gw]{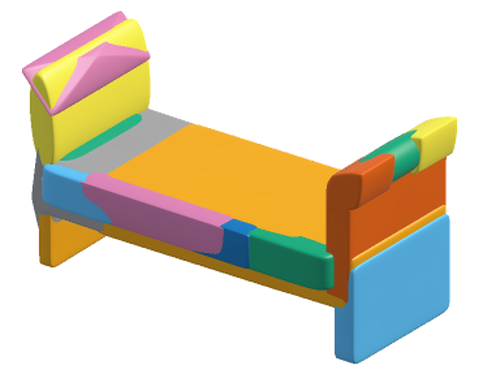} & \includegraphics[width=\gw]{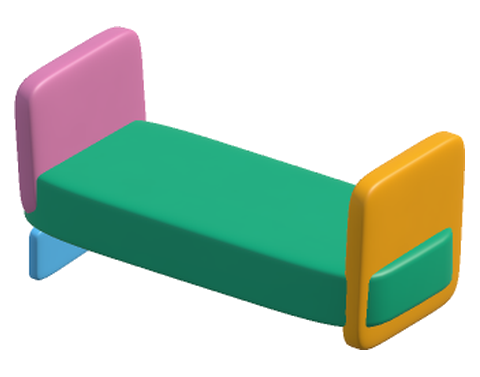} & \includegraphics[width=\gw]{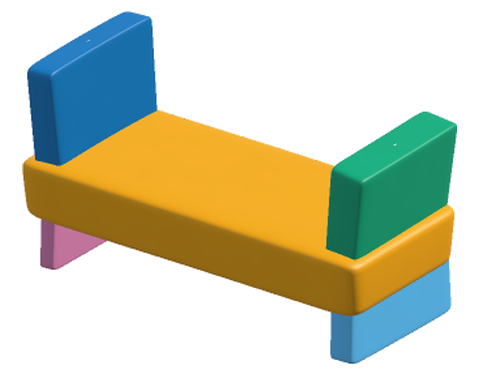} & \includegraphics[width=\gw]{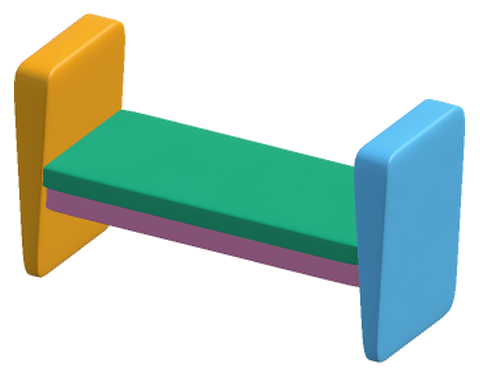} & \includegraphics[width=\gw]{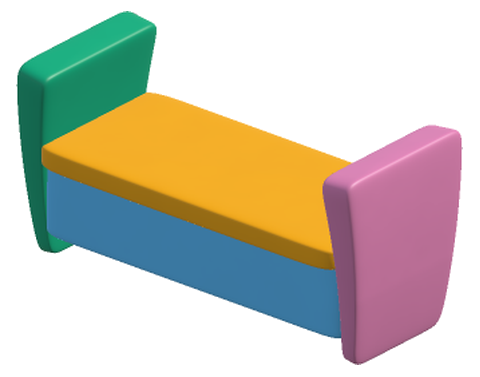} \\[0.4ex]
\scriptsize bench & \scriptsize 48 & \scriptsize 72 & \scriptsize 68 & \scriptsize 68 & \scriptsize 74 & \scriptsize 63 & \scriptsize \textbf{76} \\[1.4ex]
\includegraphics[width=\gw]{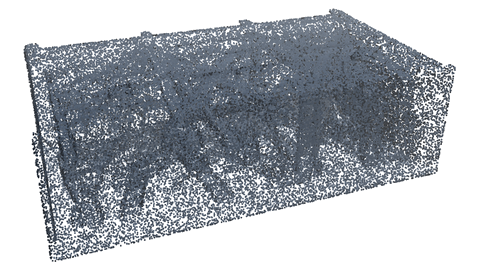} & \includegraphics[width=\gw]{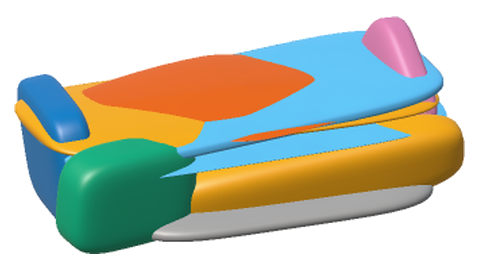} & \includegraphics[width=\gw]{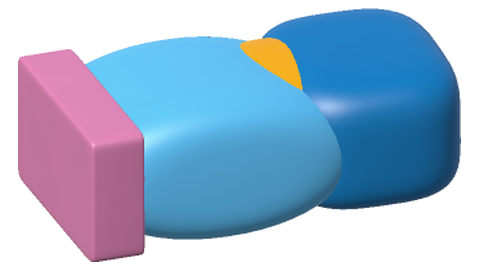} & \includegraphics[width=\gw]{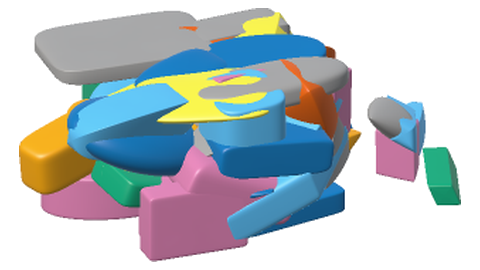} & \includegraphics[width=\gw]{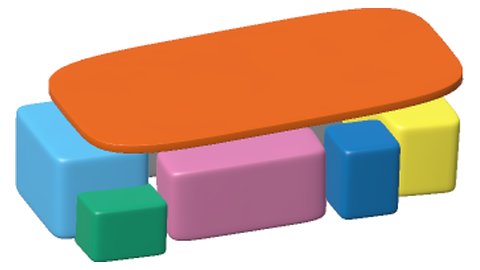} & \includegraphics[width=\gw]{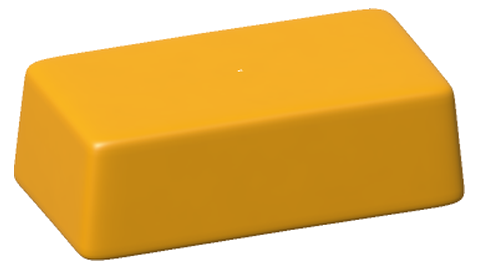} & \includegraphics[width=\gw]{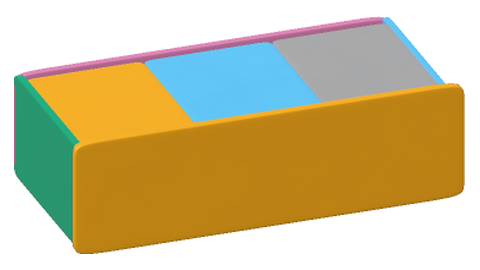} & \includegraphics[width=\gw]{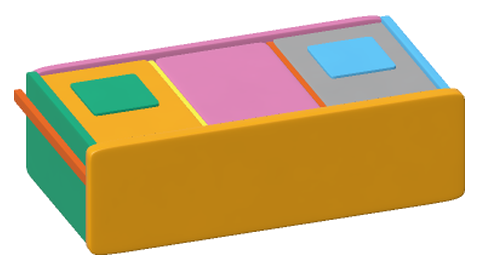} \\[0.4ex]
\scriptsize cabinet & \scriptsize 30 & \scriptsize 28 & \scriptsize 28 & \scriptsize 30 & \scriptsize 30 & \scriptsize 67 & \scriptsize \textbf{77} \\[1.4ex]
\includegraphics[width=\gw]{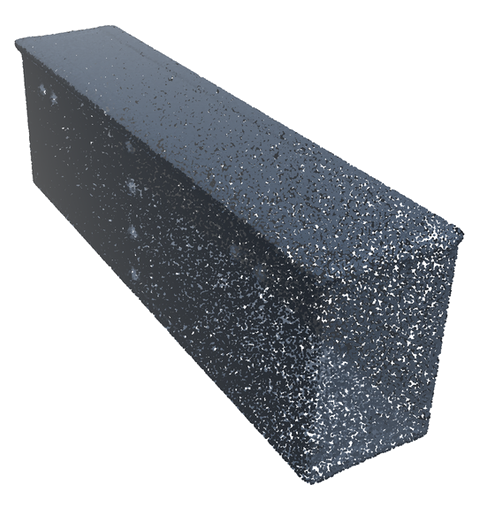} & \includegraphics[width=\gw]{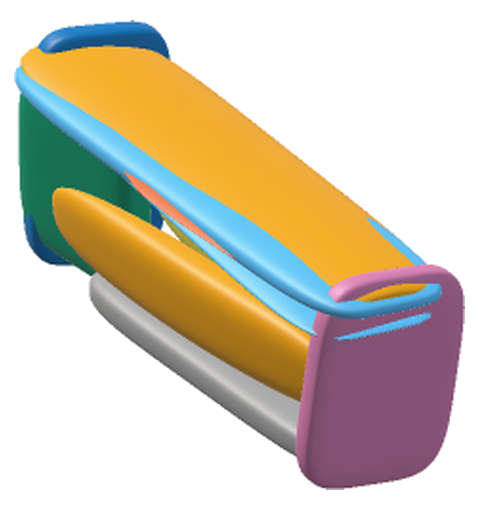} & \includegraphics[width=\gw]{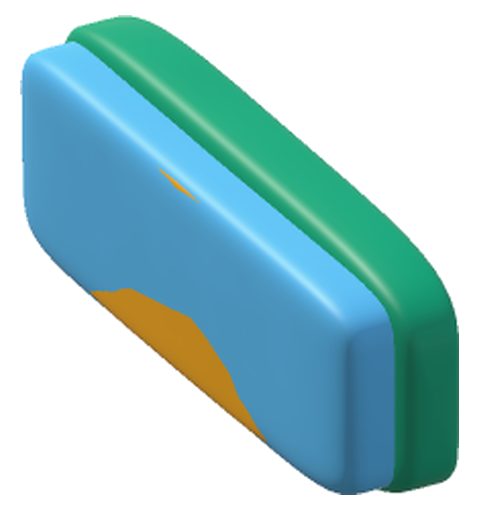} & \includegraphics[width=\gw]{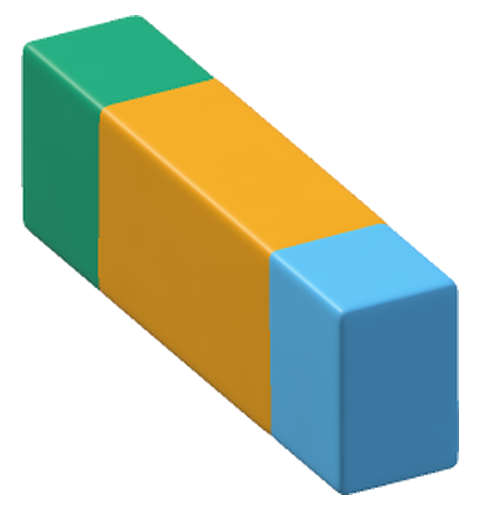} & \includegraphics[width=\gw]{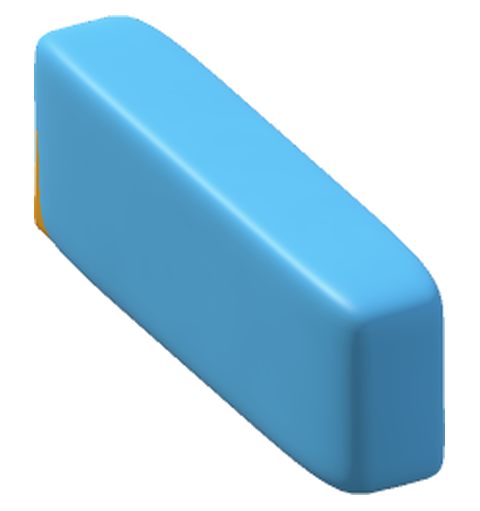} & \includegraphics[width=\gw]{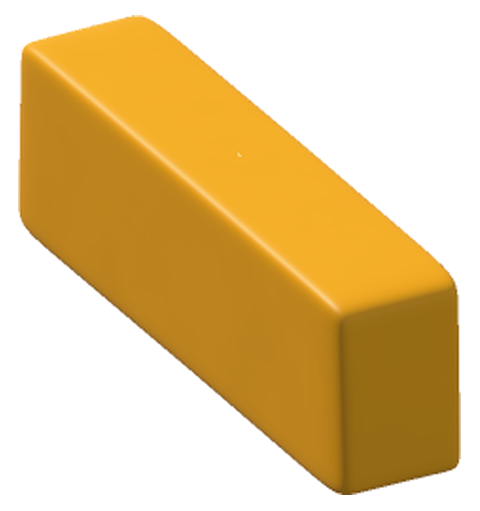} & \includegraphics[width=\gw]{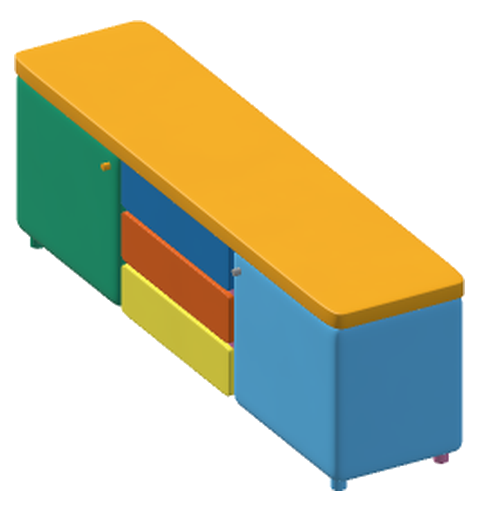} & \includegraphics[width=\gw]{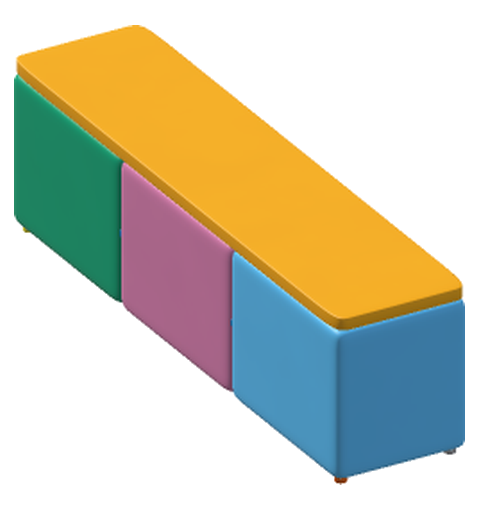} \\[0.4ex]
\scriptsize cabinet & \scriptsize 41 & \scriptsize 30 & \scriptsize 56 & \scriptsize 30 & \scriptsize 48 & \scriptsize \textbf{90} & \scriptsize 85 \\[1.4ex]
\includegraphics[width=\gw]{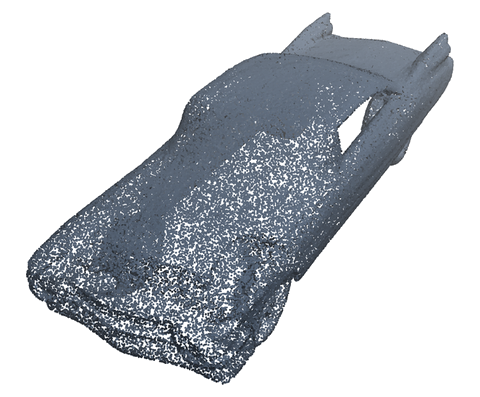} & \includegraphics[width=\gw]{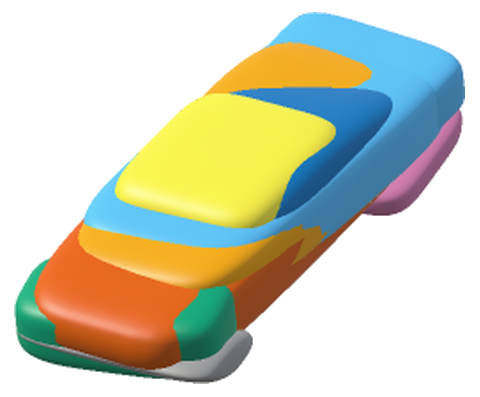} & \includegraphics[width=\gw]{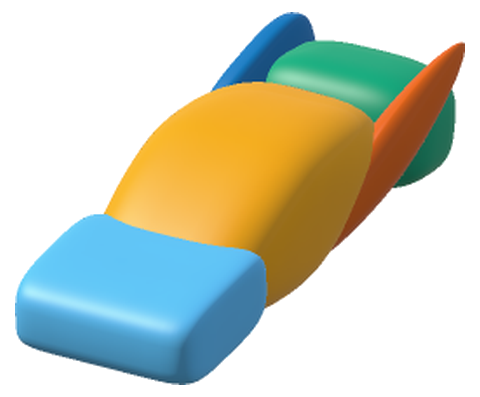} & \includegraphics[width=\gw]{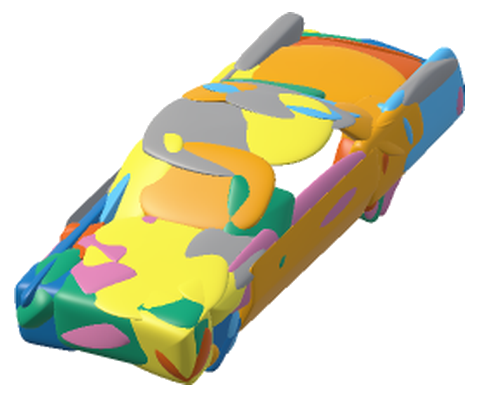} & \includegraphics[width=\gw]{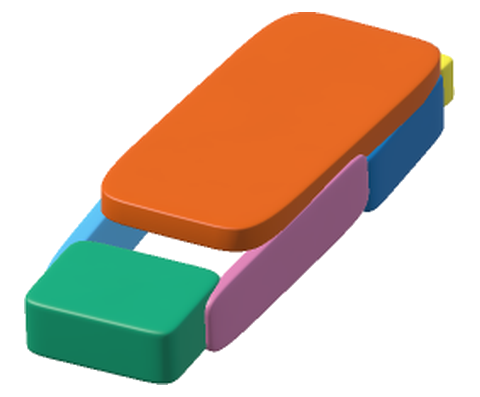} & \includegraphics[width=\gw]{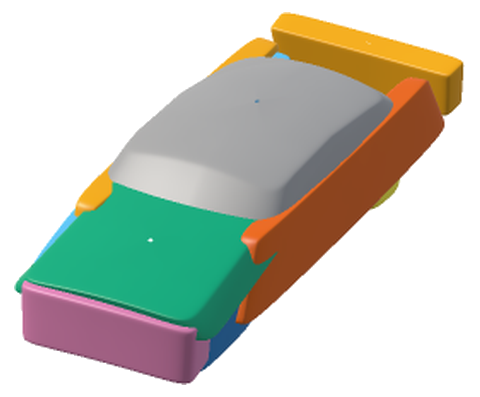} & \includegraphics[width=\gw]{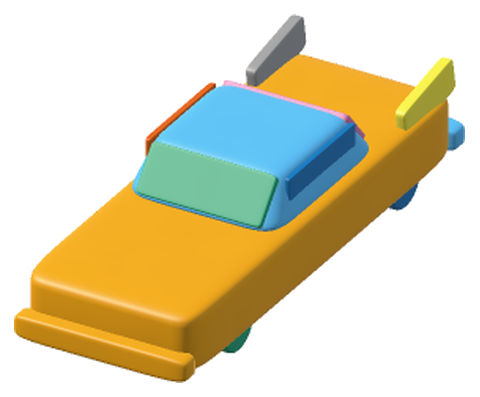} & \includegraphics[width=\gw]{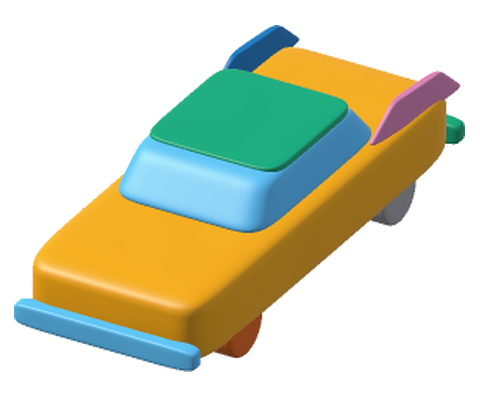} \\[0.4ex]
\scriptsize car & \scriptsize 55 & \scriptsize 51 & \scriptsize 69 & \scriptsize 30 & \scriptsize 57 & \scriptsize 73 & \scriptsize \textbf{78} \\[1.4ex]
\includegraphics[width=\gw]{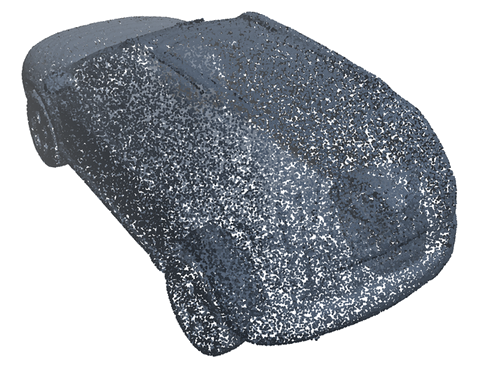} & \includegraphics[width=\gw]{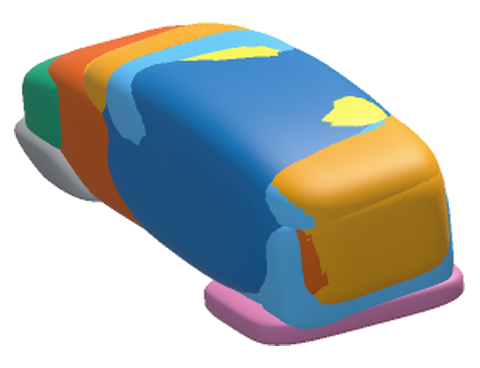} & \includegraphics[width=\gw]{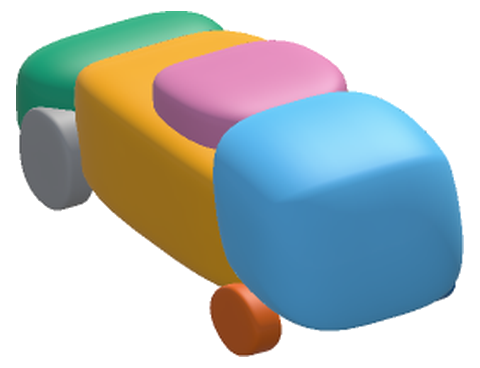} & \includegraphics[width=\gw]{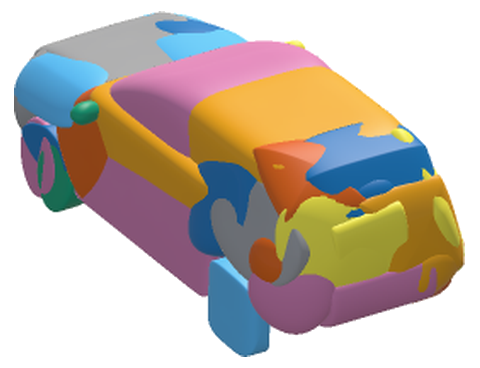} & \includegraphics[width=\gw]{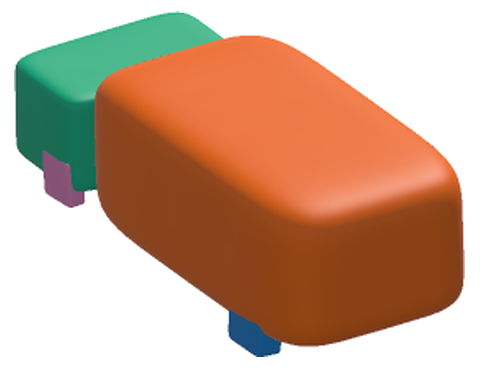} & \includegraphics[width=\gw]{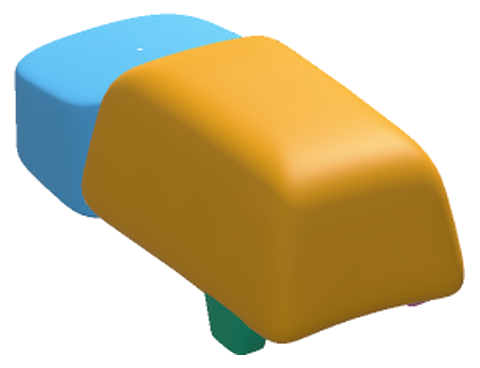} & \includegraphics[width=\gw]{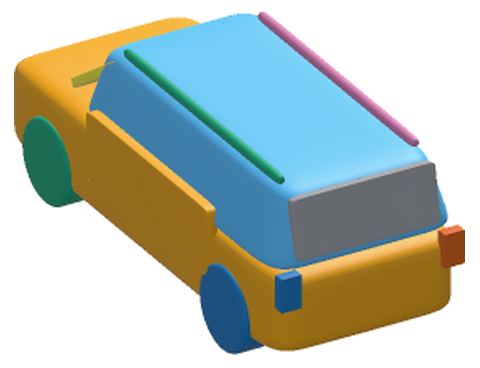} & \includegraphics[width=\gw]{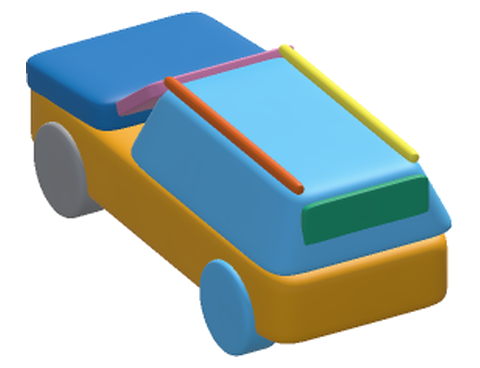} \\[0.4ex]
\scriptsize car & \scriptsize 48 & \scriptsize 52 & \scriptsize 67 & \scriptsize 30 & \scriptsize 30 & \scriptsize 67 & \scriptsize \textbf{72} \\[1.4ex]
\end{tabular}

  \caption{\textbf{Random objects, as predicted} (rows 1--8 of 26). Two objects per category, drawn at random and not selected. Each result is drawn from the input photo's viewpoint, one flat colour per part. Under each result is its VLM score (Claude Sonnet~5.5); the highest in each row is in bold.}
  \label{fig:s_more_visual_1}
\end{figure}

\begin{figure}[p]
  \centering
\setlength{\tabcolsep}{0pt}\renewcommand{\arraystretch}{0}
\def\gw{\dimexpr(\textwidth-7\dimexpr0.25em\relax)/8\relax}
\begin{tabular}{@{}c*{7}{@{\hspace{0.25em}}c}@{}}
\parbox[b]{\gw}{\centering\scriptsize\renewcommand{\arraystretch}{1}Input} & \parbox[b]{\gw}{\centering\scriptsize\renewcommand{\arraystretch}{1}SQ} & \parbox[b]{\gw}{\centering\scriptsize\renewcommand{\arraystretch}{1}EMS} & \parbox[b]{\gw}{\centering\scriptsize\renewcommand{\arraystretch}{1}Marching-Prim.} & \parbox[b]{\gw}{\centering\scriptsize\renewcommand{\arraystretch}{1}SuperDec} & \parbox[b]{\gw}{\centering\scriptsize\renewcommand{\arraystretch}{1}SuperFlex} & \parbox[b]{\gw}{\centering\scriptsize\renewcommand{\arraystretch}{1}PhysDec\\(Evolved)} & \parbox[b]{\gw}{\centering\scriptsize\renewcommand{\arraystretch}{1}PhysDec\\(Loop)} \\[0.5ex]
\includegraphics[width=\gw]{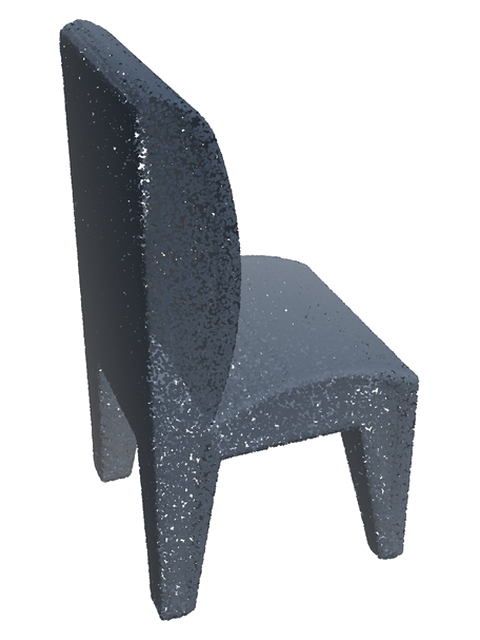} & \includegraphics[width=\gw]{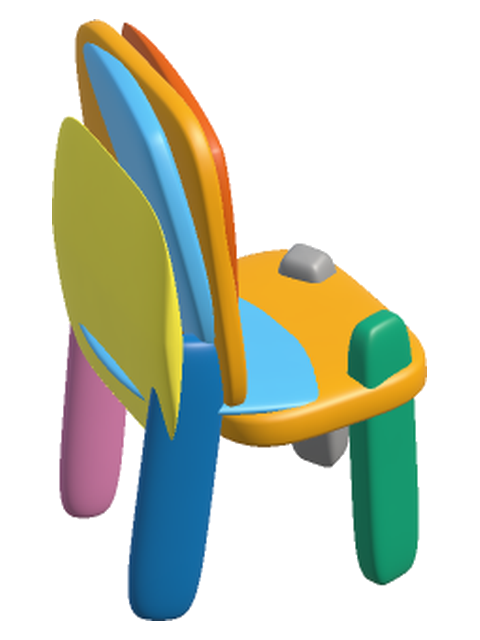} & \includegraphics[width=\gw]{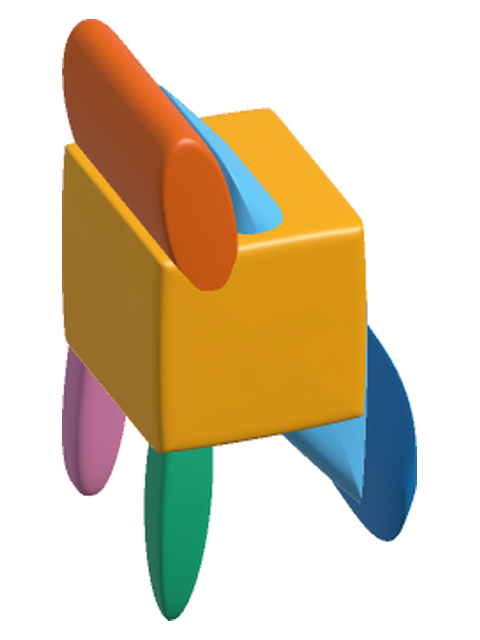} & \includegraphics[width=\gw]{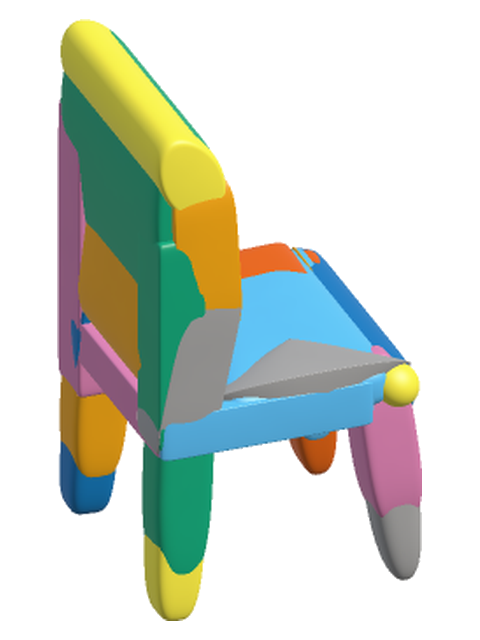} & \includegraphics[width=\gw]{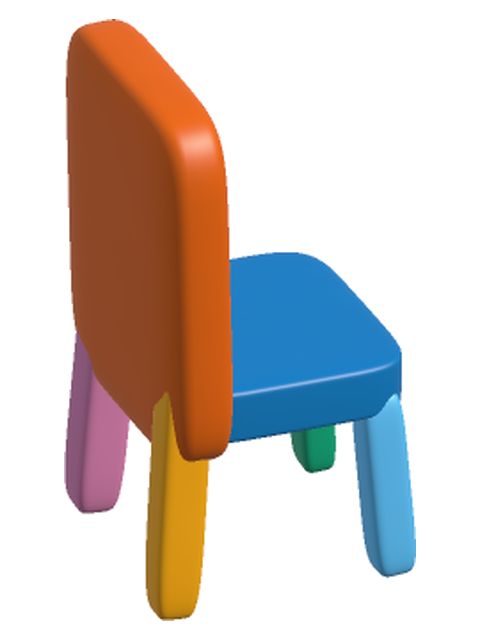} & \includegraphics[width=\gw]{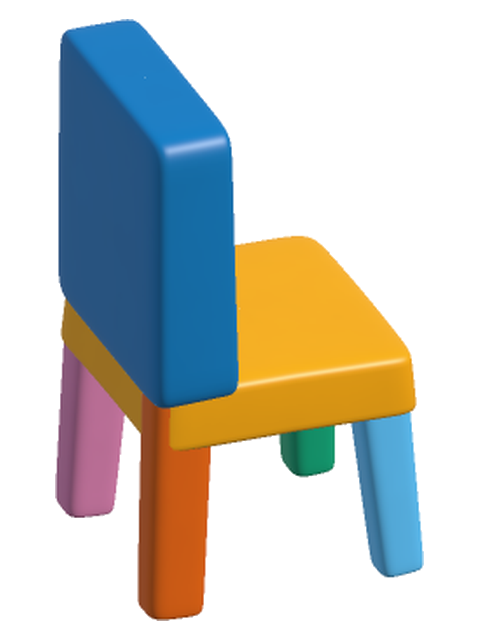} & \includegraphics[width=\gw]{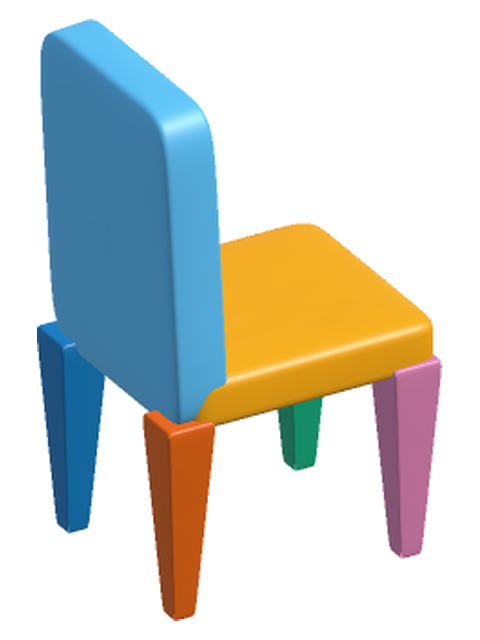} & \includegraphics[width=\gw]{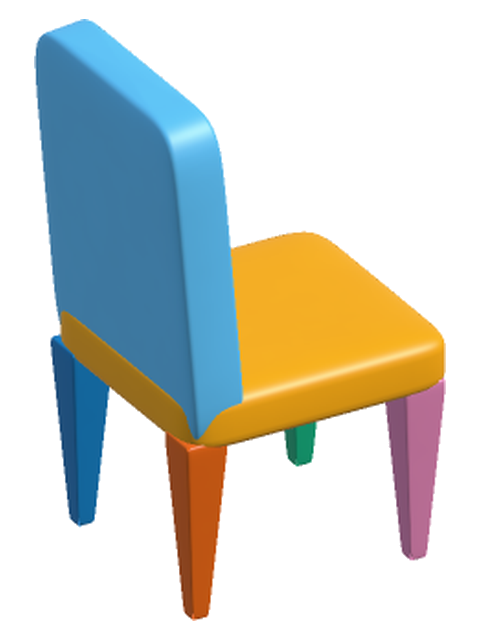} \\[0.4ex]
\scriptsize chair & \scriptsize 64 & \scriptsize 56 & \scriptsize 75 & \scriptsize 79 & \scriptsize 75 & \scriptsize 75 & \scriptsize \textbf{81} \\[1.4ex]
\includegraphics[width=\gw]{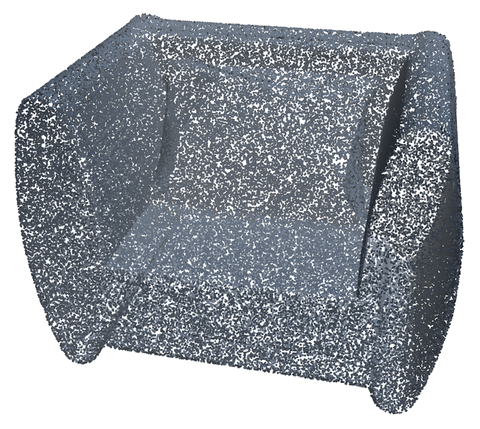} & \includegraphics[width=\gw]{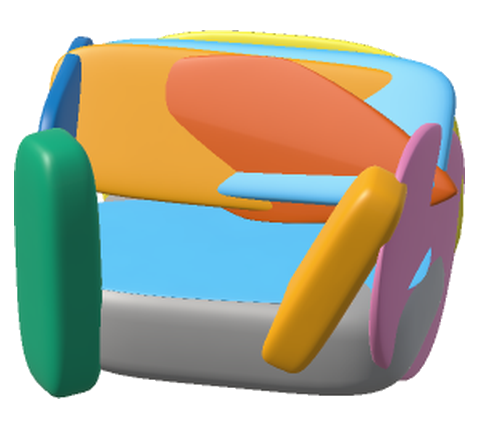} & \includegraphics[width=\gw]{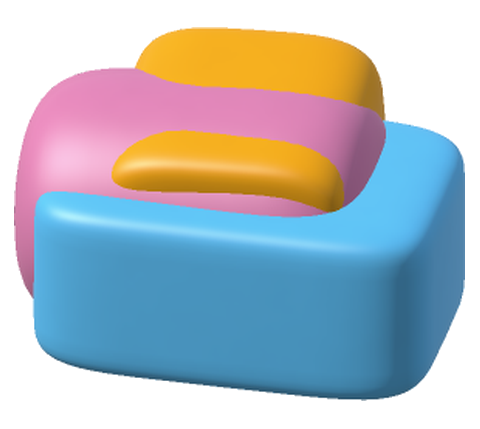} & \includegraphics[width=\gw]{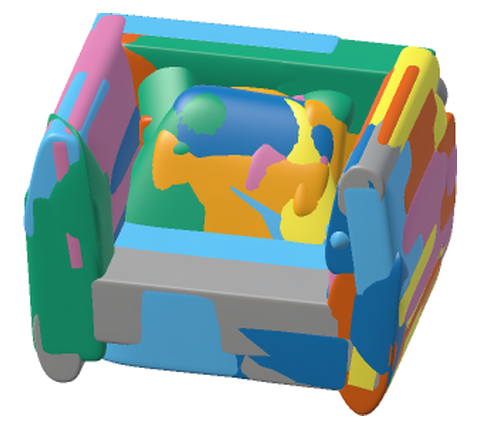} & \includegraphics[width=\gw]{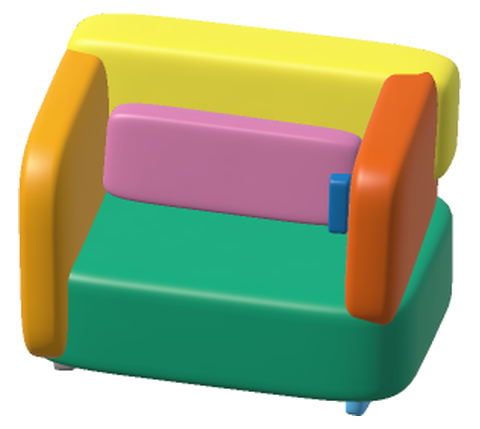} & \includegraphics[width=\gw]{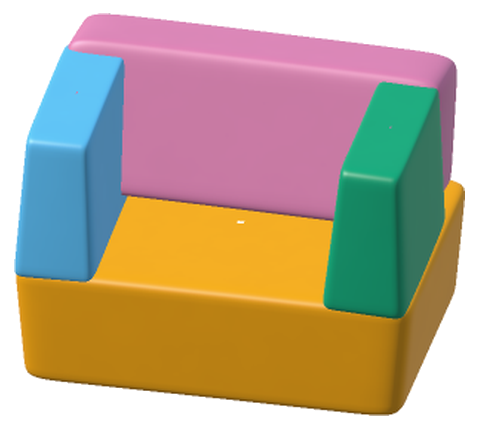} & \includegraphics[width=\gw]{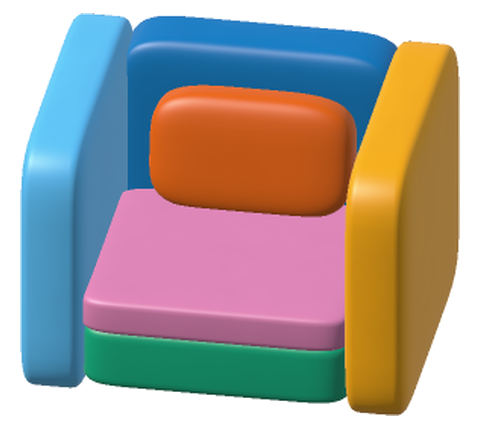} & \includegraphics[width=\gw]{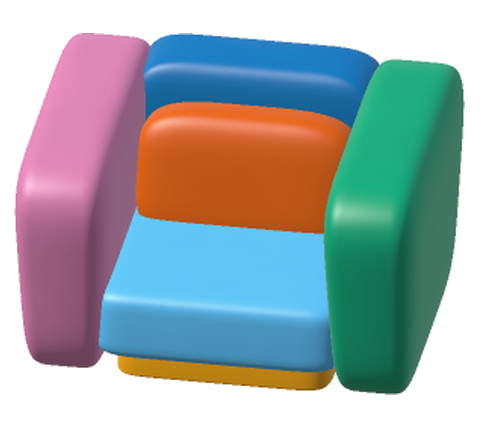} \\[0.4ex]
\scriptsize chair & \scriptsize 49 & \scriptsize 58 & \scriptsize \textbf{82} & \scriptsize 70 & \scriptsize 71 & \scriptsize 79 & \scriptsize 60 \\[1.4ex]
\includegraphics[width=\gw]{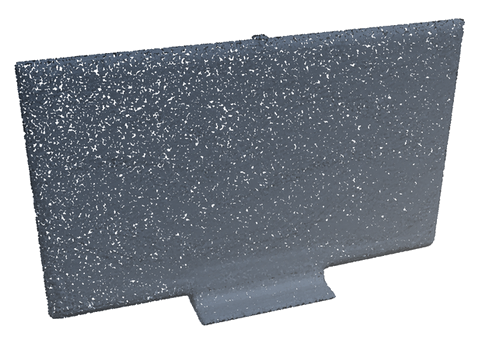} & \includegraphics[width=\gw]{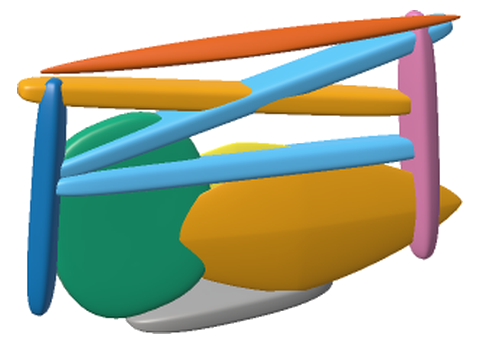} & \includegraphics[width=\gw]{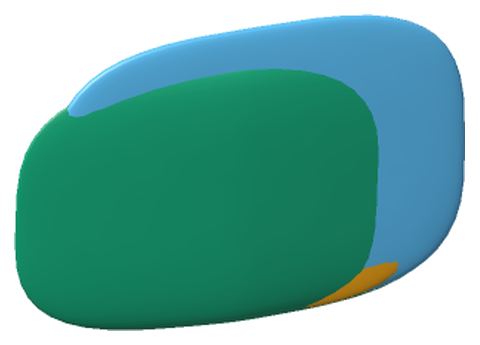} & \includegraphics[width=\gw]{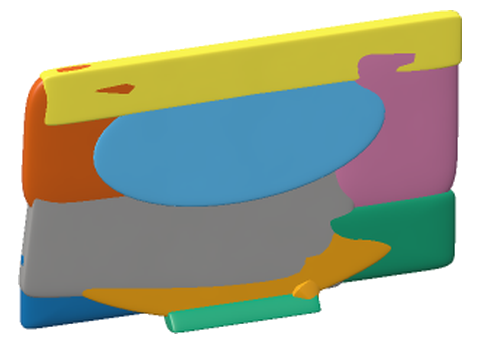} & \includegraphics[width=\gw]{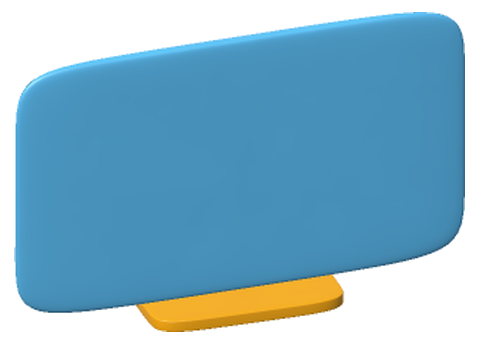} & \includegraphics[width=\gw]{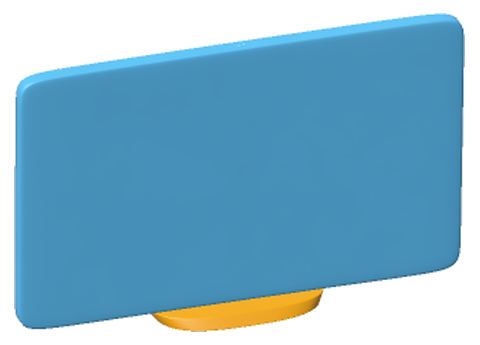} & \includegraphics[width=\gw]{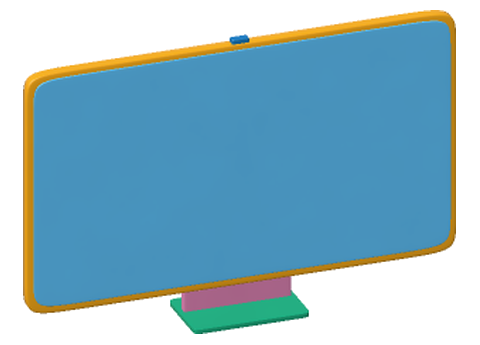} & \includegraphics[width=\gw]{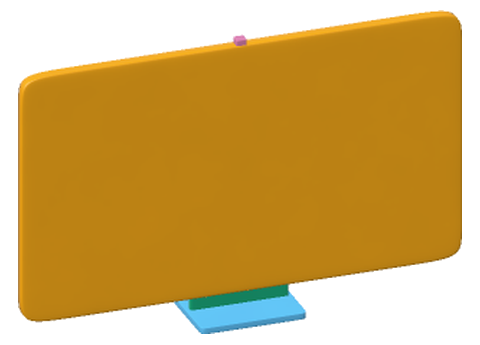} \\[0.4ex]
\scriptsize display & \scriptsize 23 & \scriptsize 26 & \scriptsize 66 & \scriptsize 71 & \scriptsize 76 & \scriptsize \textbf{83} & \scriptsize \textbf{83} \\[1.4ex]
\includegraphics[width=\gw]{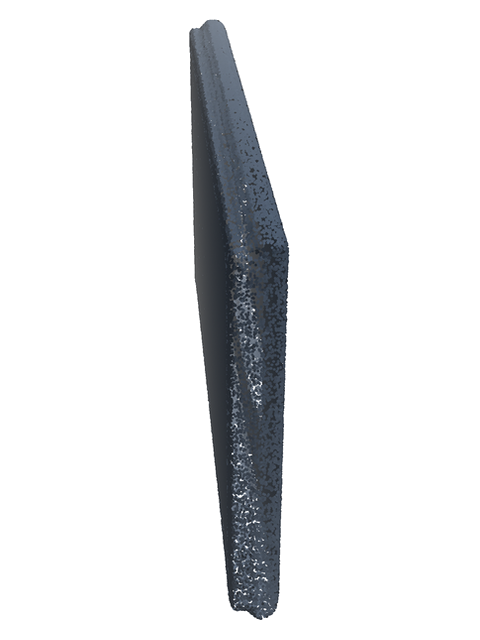} & \includegraphics[width=\gw]{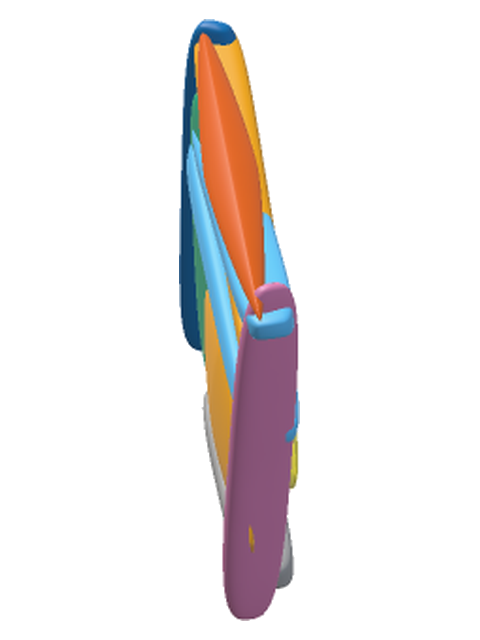} & \includegraphics[width=\gw]{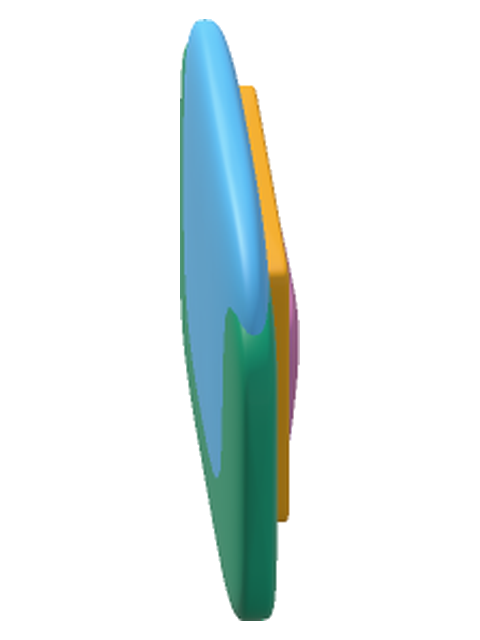} & \includegraphics[width=\gw]{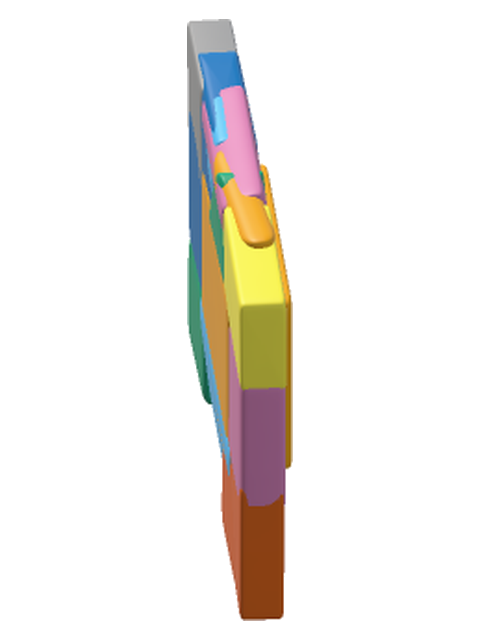} & \includegraphics[width=\gw]{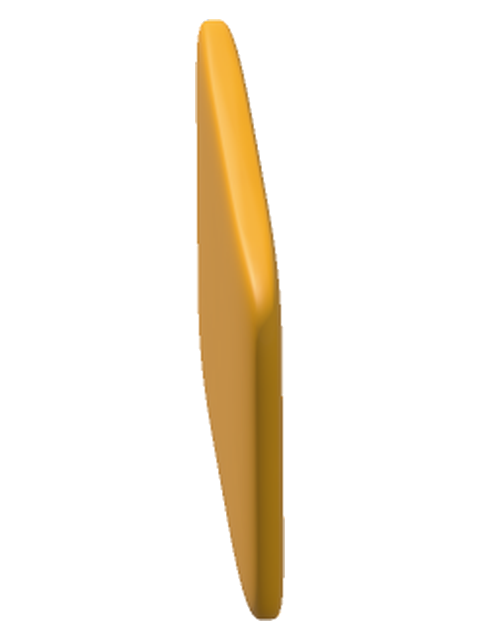} & \includegraphics[width=\gw]{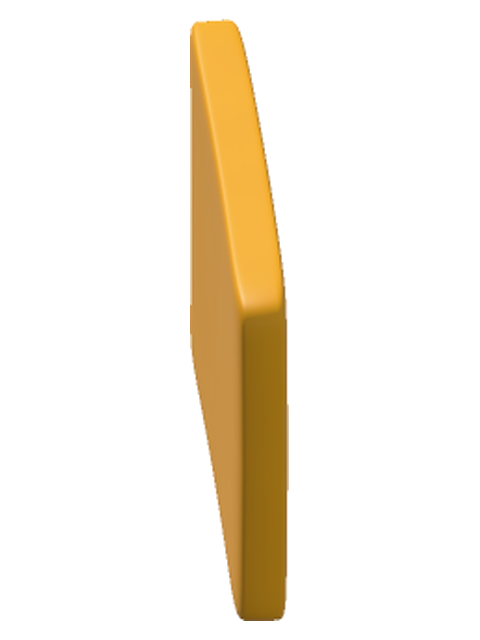} & \includegraphics[width=\gw]{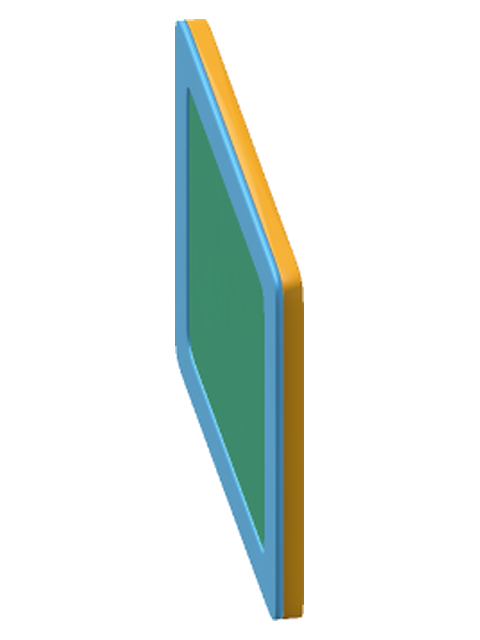} & \includegraphics[width=\gw]{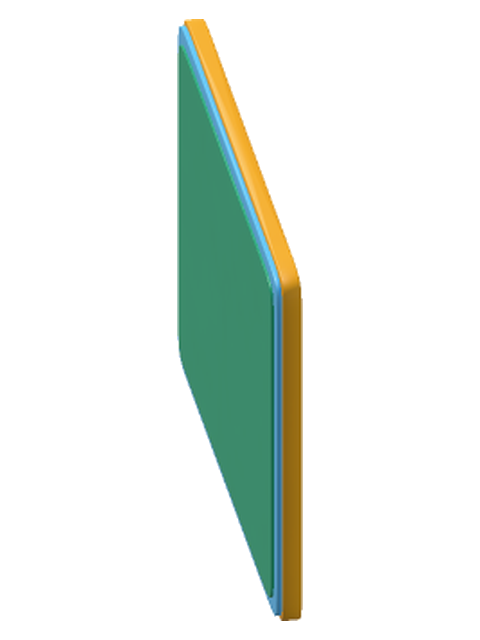} \\[0.4ex]
\scriptsize display & \scriptsize 29 & \scriptsize 58 & \scriptsize 62 & \scriptsize 59 & \scriptsize 62 & \scriptsize \textbf{88} & \scriptsize 80 \\[1.4ex]
\includegraphics[width=\gw]{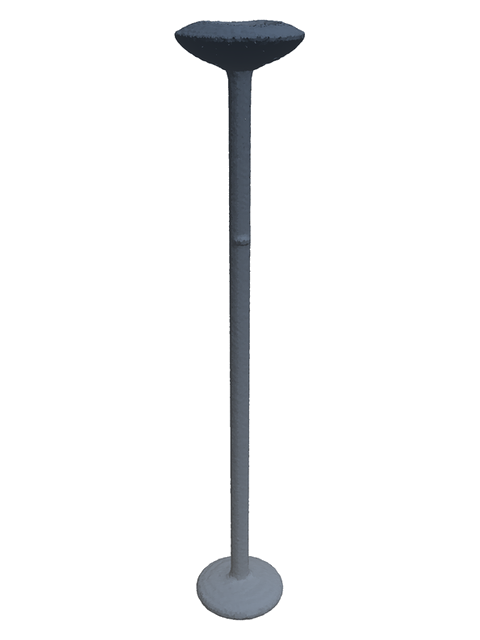} & \includegraphics[width=\gw]{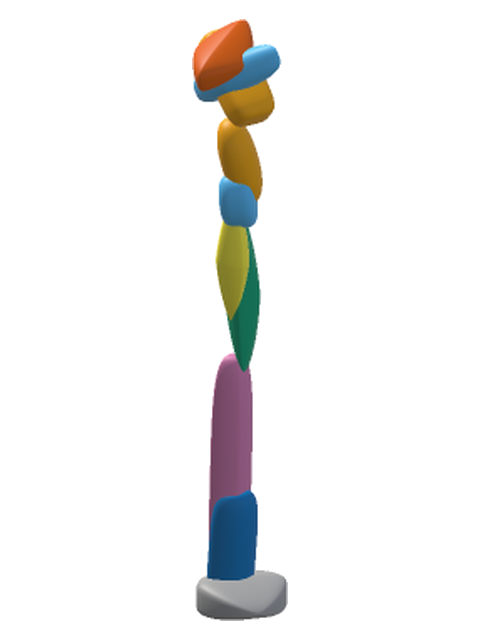} & \includegraphics[width=\gw]{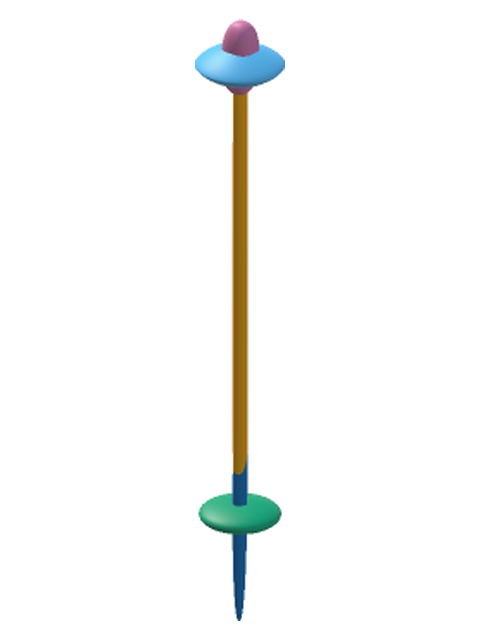} & \includegraphics[width=\gw]{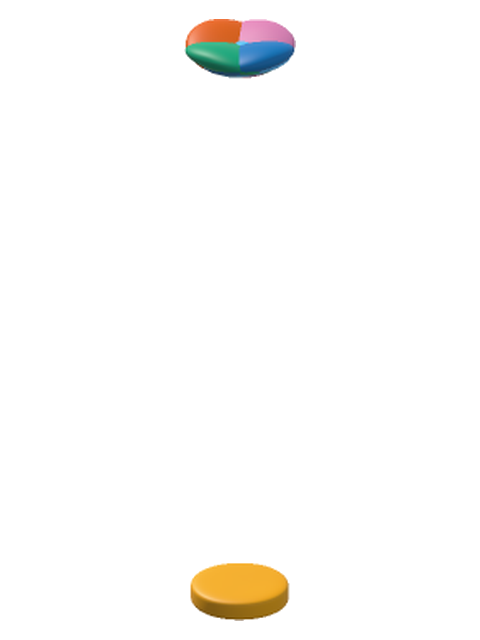} & \includegraphics[width=\gw]{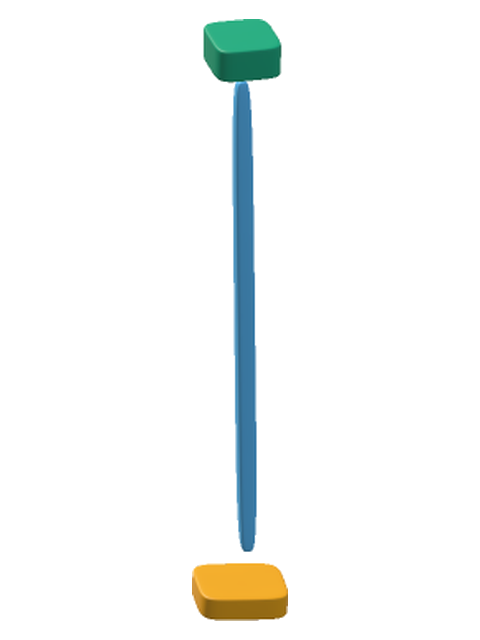} & \includegraphics[width=\gw]{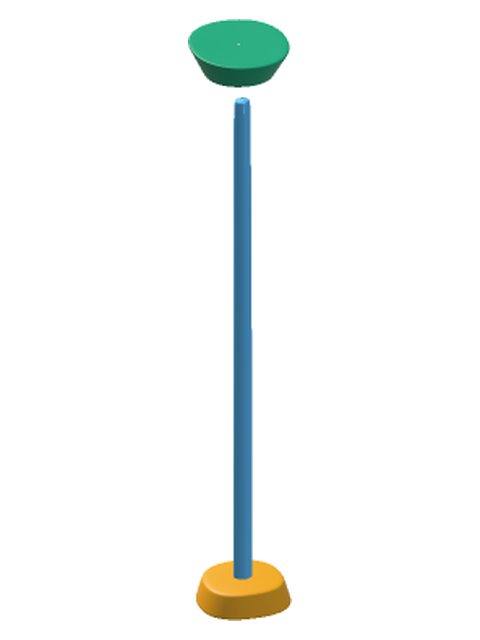} & \includegraphics[width=\gw]{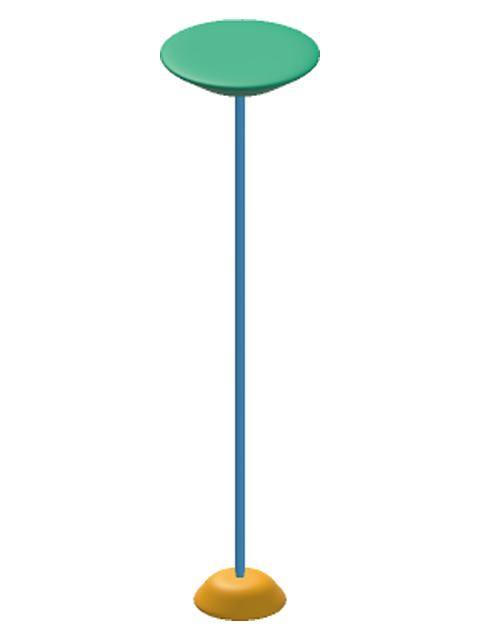} & \includegraphics[width=\gw]{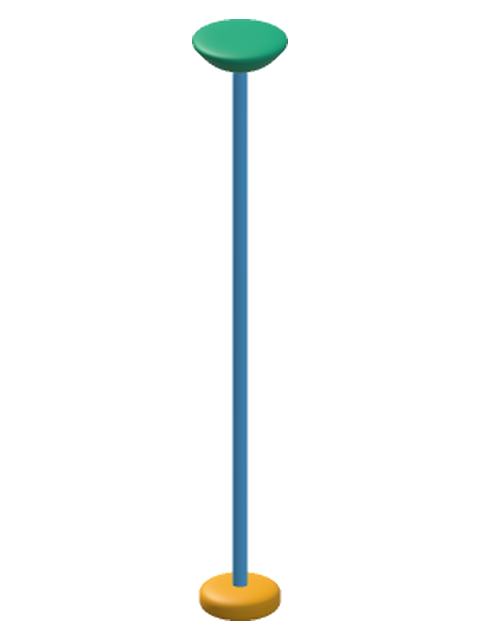} \\[0.4ex]
\scriptsize lamp & \scriptsize 63 & \scriptsize 71 & \scriptsize 30 & \scriptsize 69 & \scriptsize 79 & \scriptsize 88 & \scriptsize \textbf{96} \\[1.4ex]
\includegraphics[width=\gw]{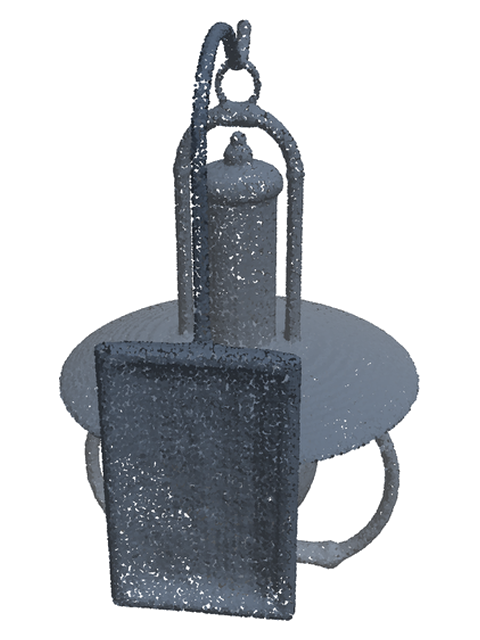} & \includegraphics[width=\gw]{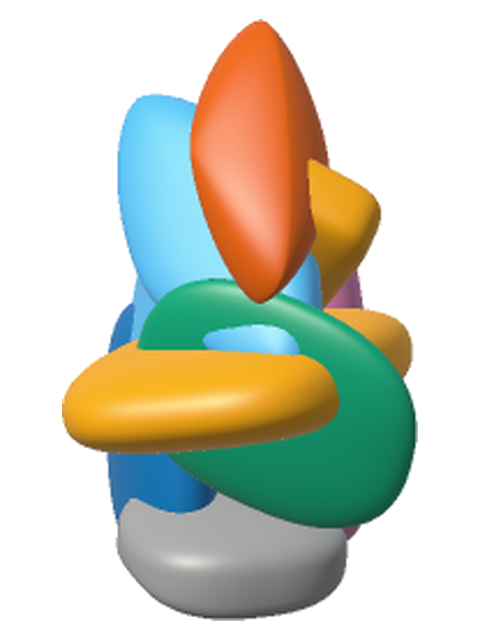} & \includegraphics[width=\gw]{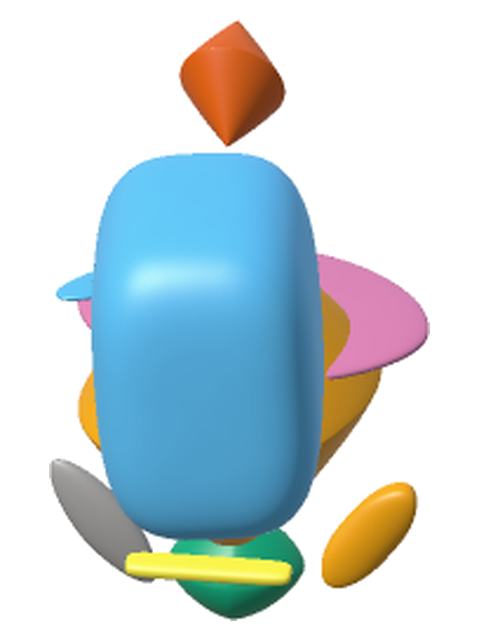} & \includegraphics[width=\gw]{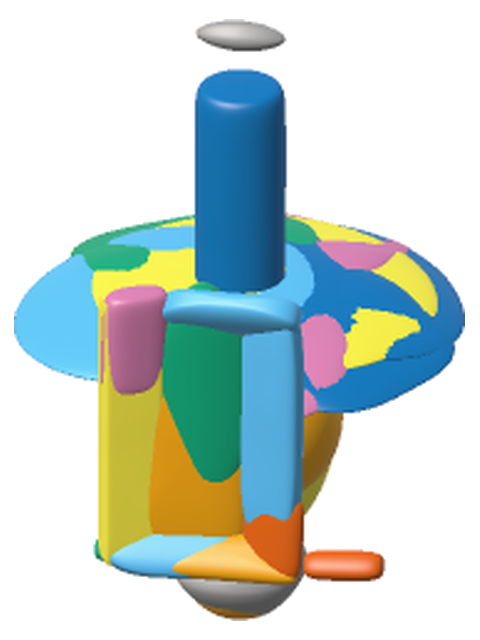} & \includegraphics[width=\gw]{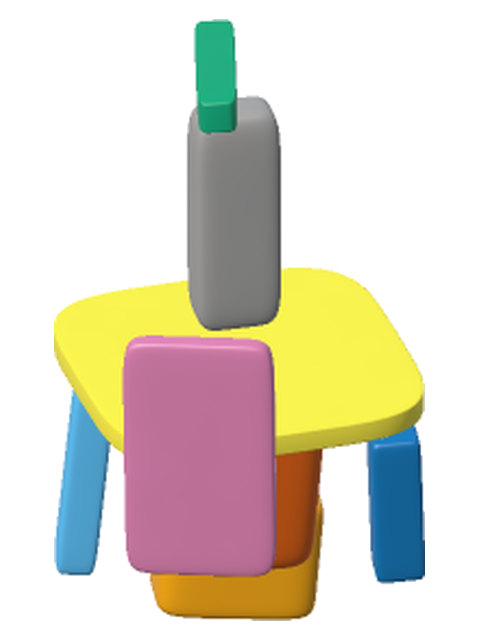} & \includegraphics[width=\gw]{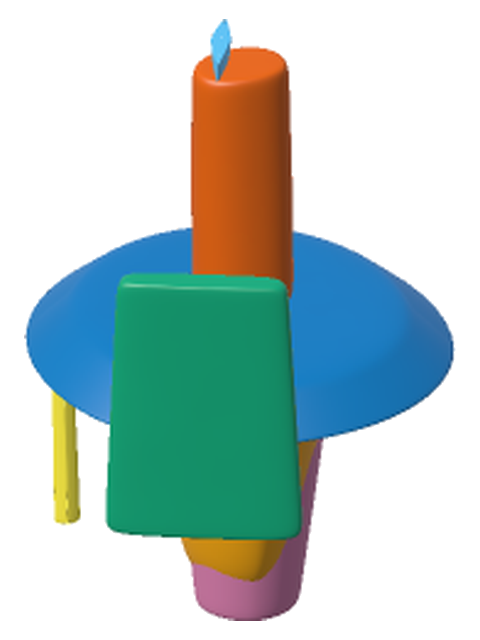} & \includegraphics[width=\gw]{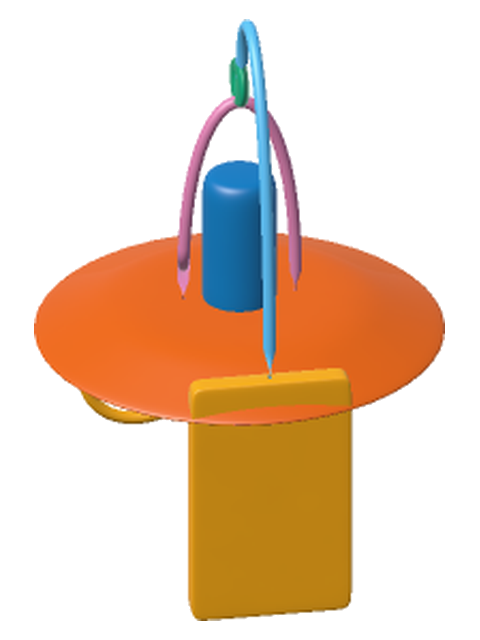} & \includegraphics[width=\gw]{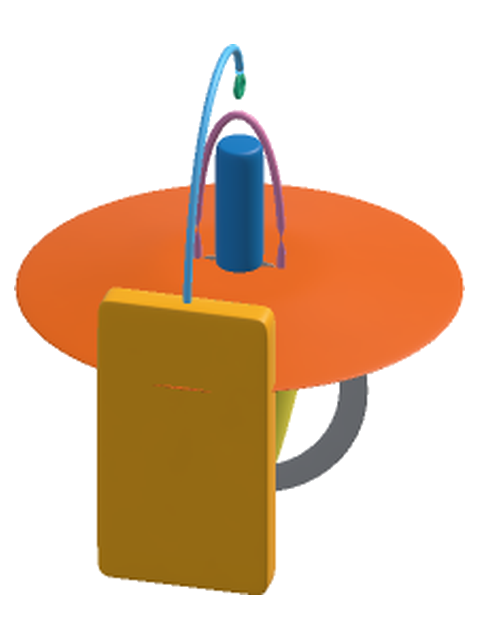} \\[0.4ex]
\scriptsize lamp & \scriptsize 29 & \scriptsize 29 & \scriptsize 44 & \scriptsize 30 & \scriptsize 48 & \scriptsize \textbf{75} & \scriptsize 61 \\[1.4ex]
\end{tabular}

  \caption{\textbf{Random objects, as predicted} (rows 9--14 of 26). Two objects per category, drawn at random and not selected. Each result is drawn from the input photo's viewpoint, one flat colour per part. Under each result is its VLM score (Claude Sonnet~5.5); the highest in each row is in bold.}
  \label{fig:s_more_visual_2}
\end{figure}

\begin{figure}[p]
  \centering
\setlength{\tabcolsep}{0pt}\renewcommand{\arraystretch}{0}
\def\gw{\dimexpr(\textwidth-7\dimexpr0.25em\relax)/8\relax}
\begin{tabular}{@{}c*{7}{@{\hspace{0.25em}}c}@{}}
\parbox[b]{\gw}{\centering\scriptsize\renewcommand{\arraystretch}{1}Input} & \parbox[b]{\gw}{\centering\scriptsize\renewcommand{\arraystretch}{1}SQ} & \parbox[b]{\gw}{\centering\scriptsize\renewcommand{\arraystretch}{1}EMS} & \parbox[b]{\gw}{\centering\scriptsize\renewcommand{\arraystretch}{1}Marching-Prim.} & \parbox[b]{\gw}{\centering\scriptsize\renewcommand{\arraystretch}{1}SuperDec} & \parbox[b]{\gw}{\centering\scriptsize\renewcommand{\arraystretch}{1}SuperFlex} & \parbox[b]{\gw}{\centering\scriptsize\renewcommand{\arraystretch}{1}PhysDec\\(Evolved)} & \parbox[b]{\gw}{\centering\scriptsize\renewcommand{\arraystretch}{1}PhysDec\\(Loop)} \\[0.5ex]
\includegraphics[width=\gw]{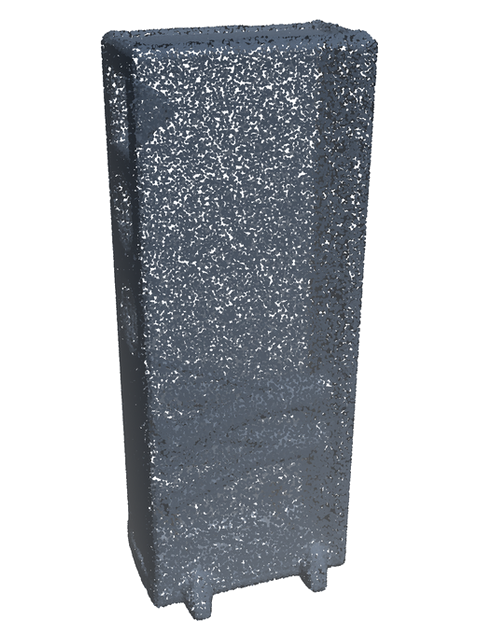} & \includegraphics[width=\gw]{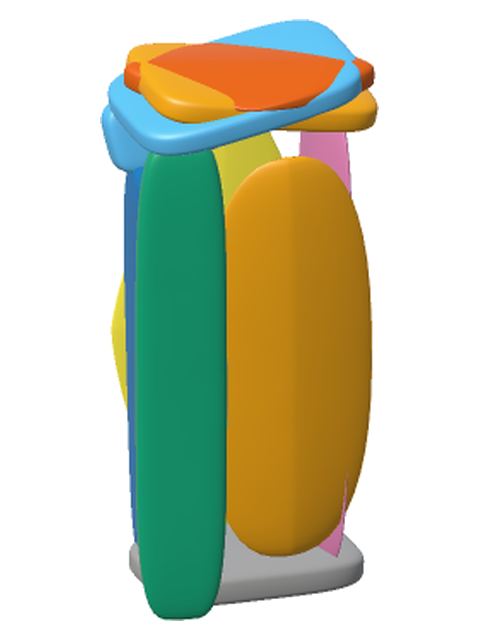} & \includegraphics[width=\gw]{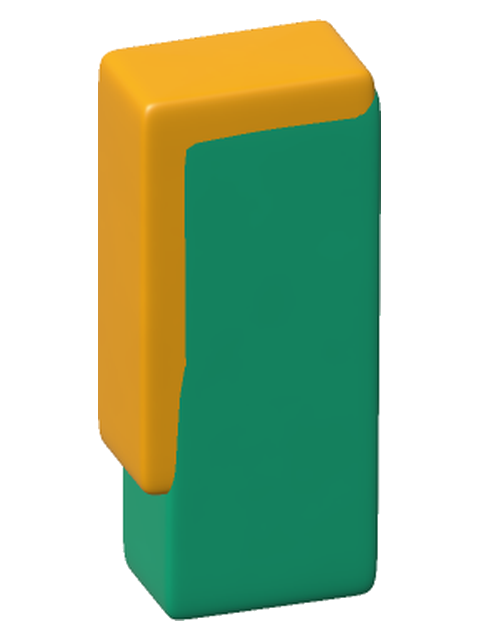} & \includegraphics[width=\gw]{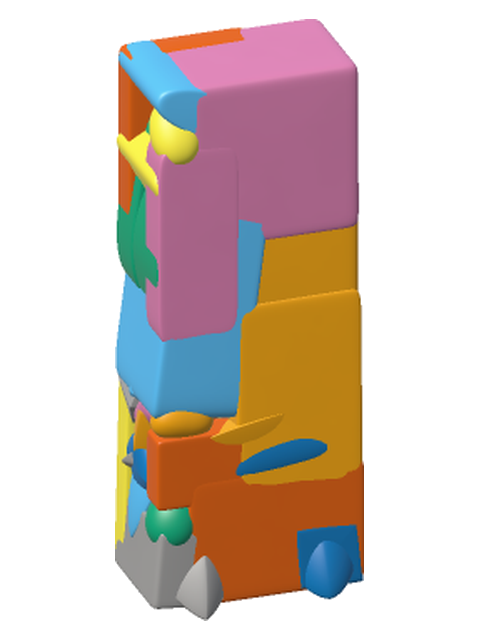} & \includegraphics[width=\gw]{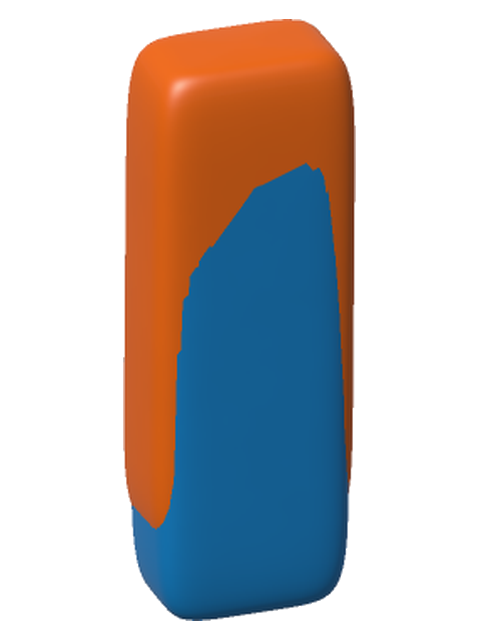} & \includegraphics[width=\gw]{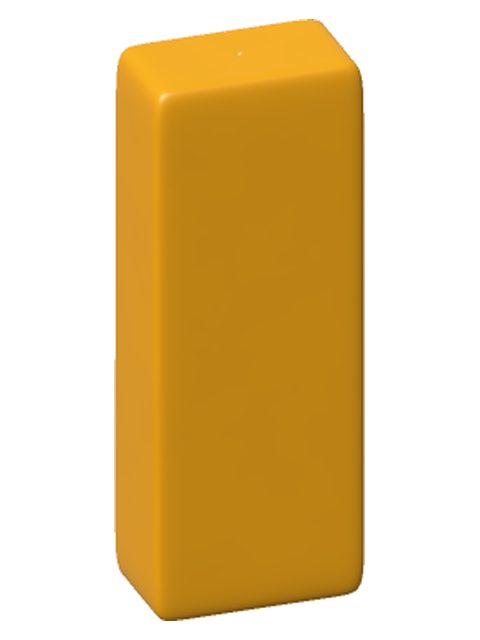} & \includegraphics[width=\gw]{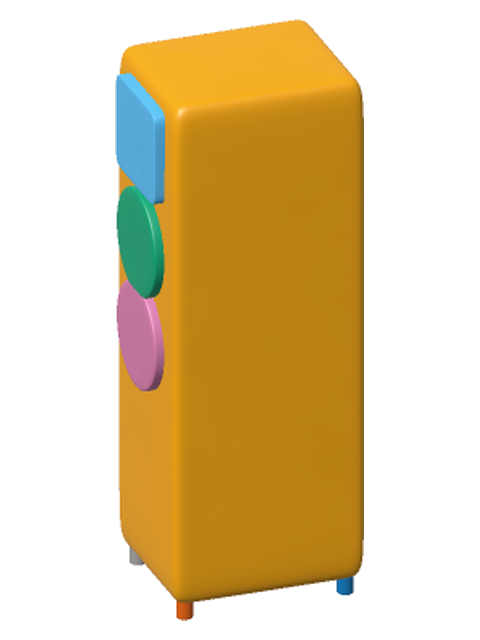} & \includegraphics[width=\gw]{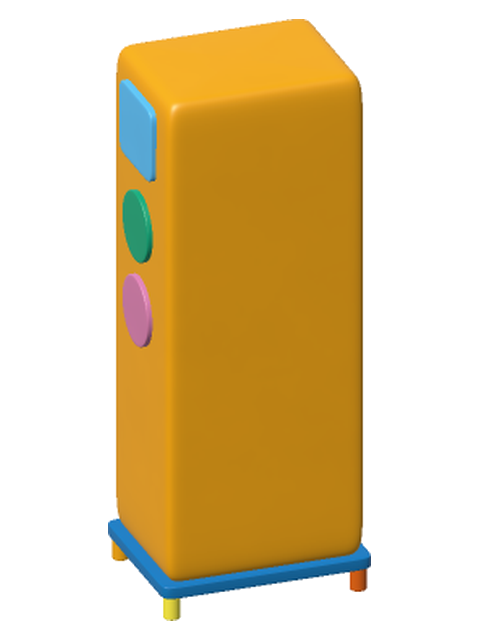} \\[0.4ex]
\scriptsize loudspeaker & \scriptsize 30 & \scriptsize 30 & \scriptsize 57 & \scriptsize 30 & \scriptsize 30 & \scriptsize \textbf{70} & \scriptsize 69 \\[1.4ex]
\includegraphics[width=\gw]{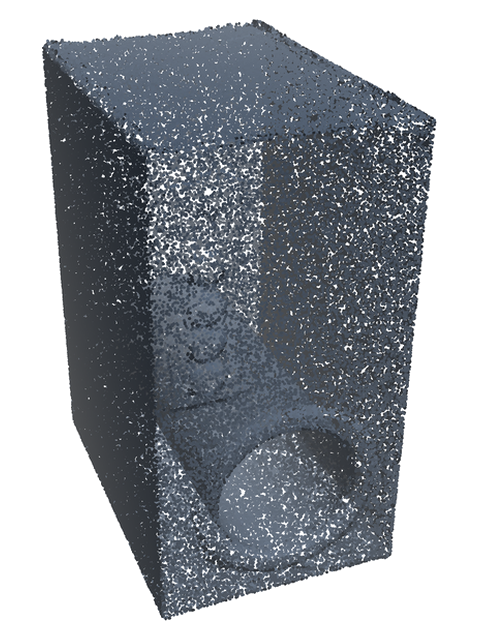} & \includegraphics[width=\gw]{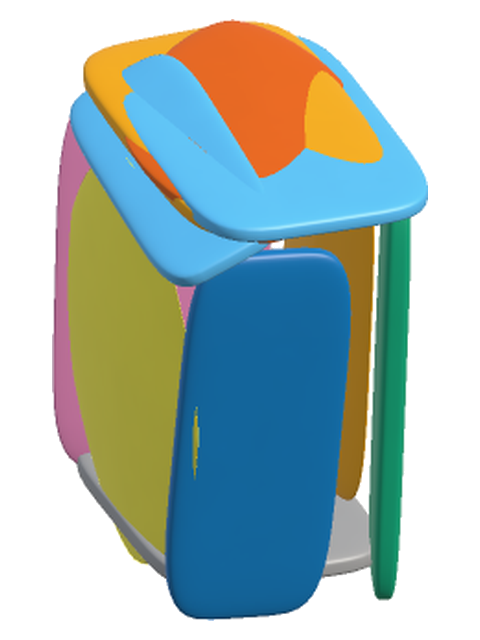} & \includegraphics[width=\gw]{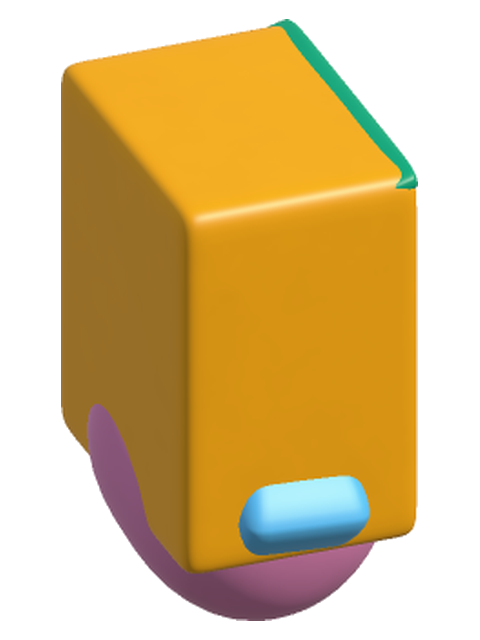} & \includegraphics[width=\gw]{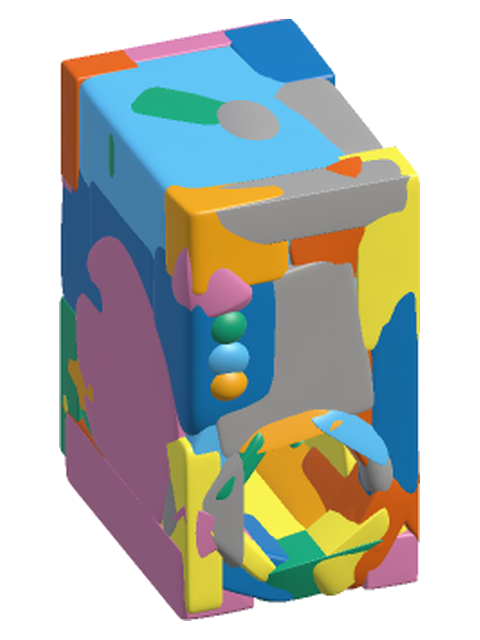} & \includegraphics[width=\gw]{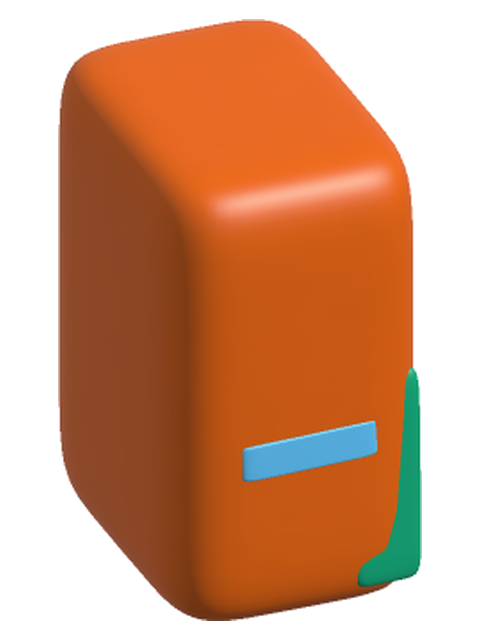} & \includegraphics[width=\gw]{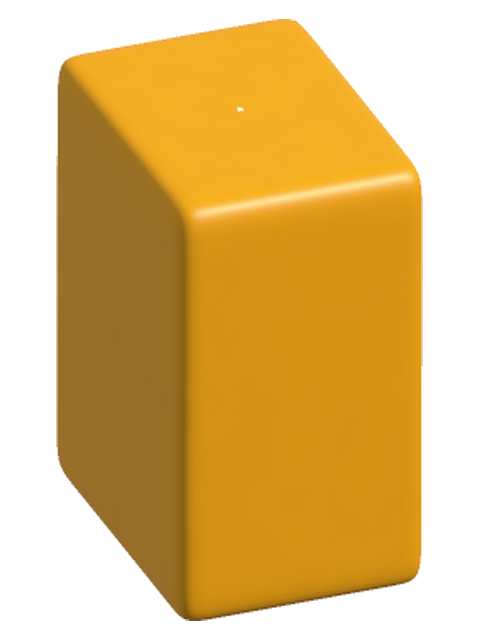} & \includegraphics[width=\gw]{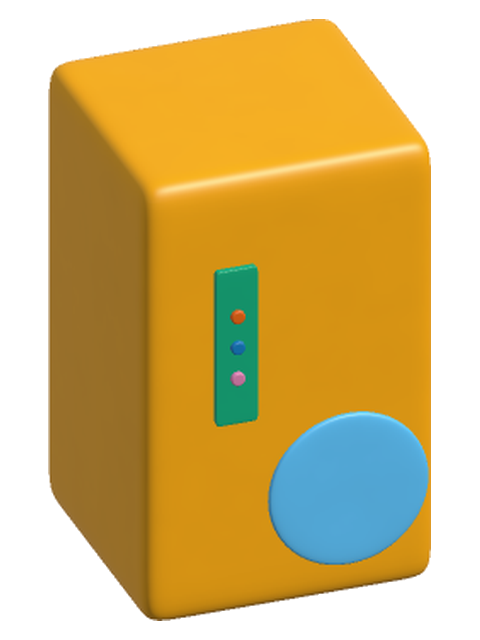} & \includegraphics[width=\gw]{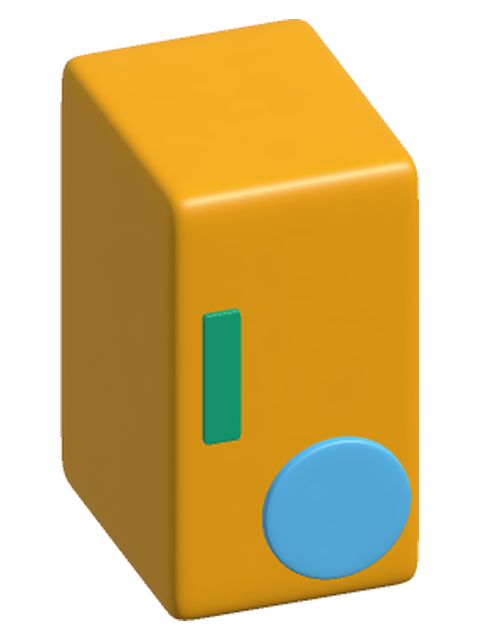} \\[0.4ex]
\scriptsize loudspeaker & \scriptsize 30 & \scriptsize 30 & \scriptsize 69 & \scriptsize 44 & \scriptsize 30 & \scriptsize 79 & \scriptsize \textbf{81} \\[1.4ex]
\includegraphics[width=\gw]{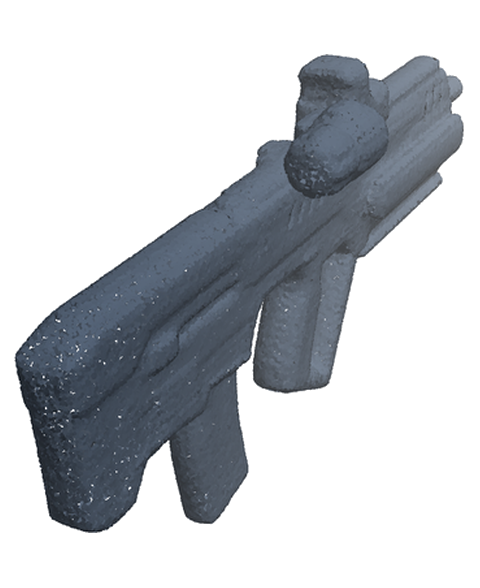} & \includegraphics[width=\gw]{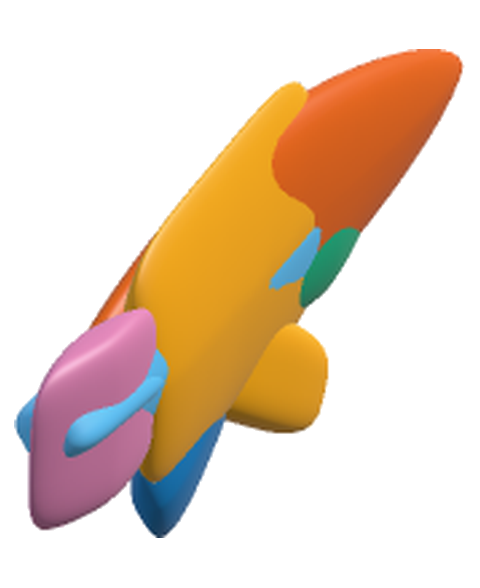} & \includegraphics[width=\gw]{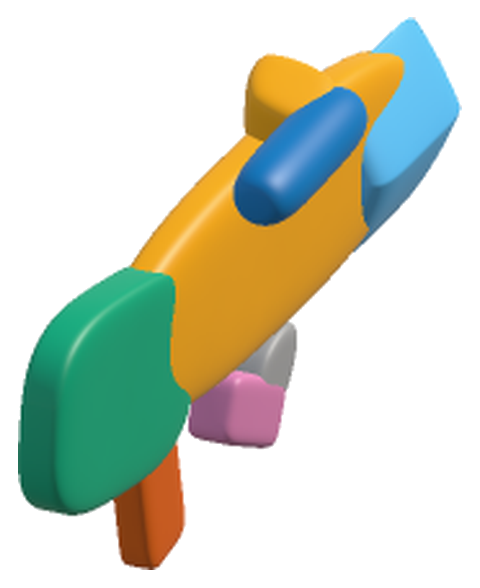} & \includegraphics[width=\gw]{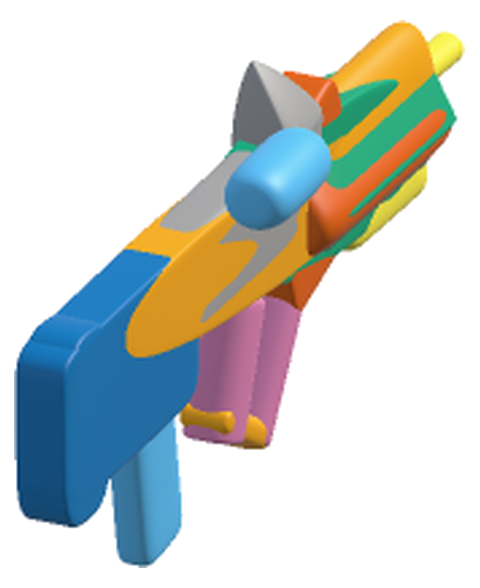} & \includegraphics[width=\gw]{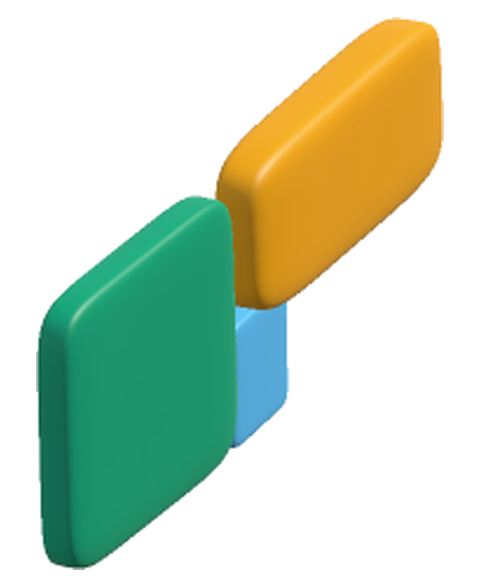} & \includegraphics[width=\gw]{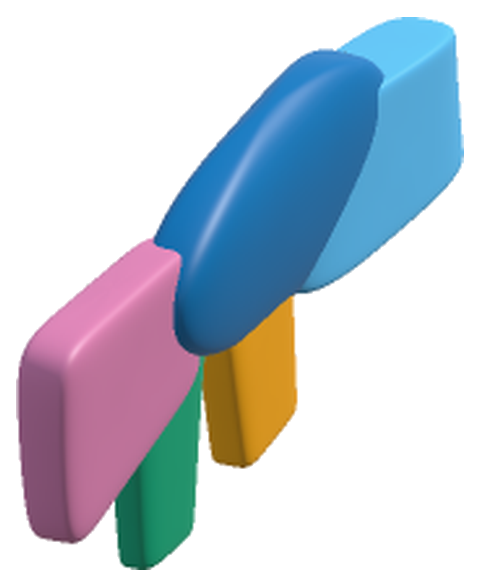} & \includegraphics[width=\gw]{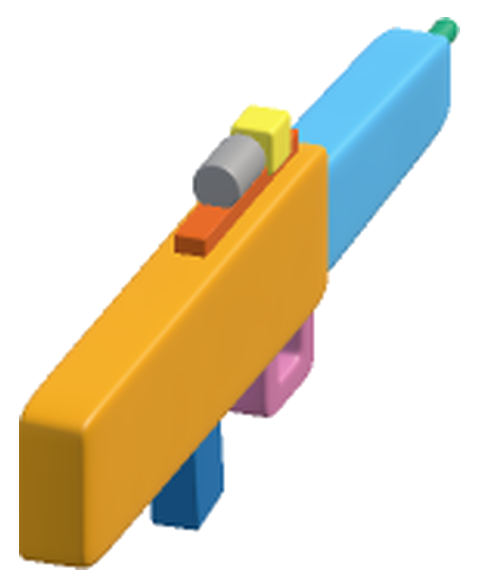} & \includegraphics[width=\gw]{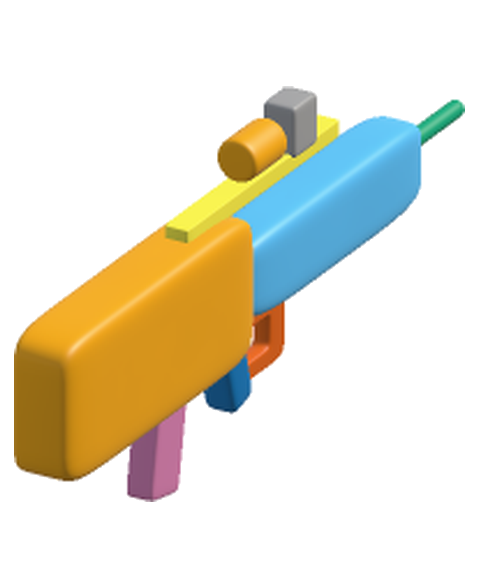} \\[0.4ex]
\scriptsize rifle & \scriptsize 29 & \scriptsize 42 & \scriptsize 59 & \scriptsize 26 & \scriptsize 30 & \scriptsize 62 & \scriptsize \textbf{65} \\[1.4ex]
\includegraphics[width=\gw]{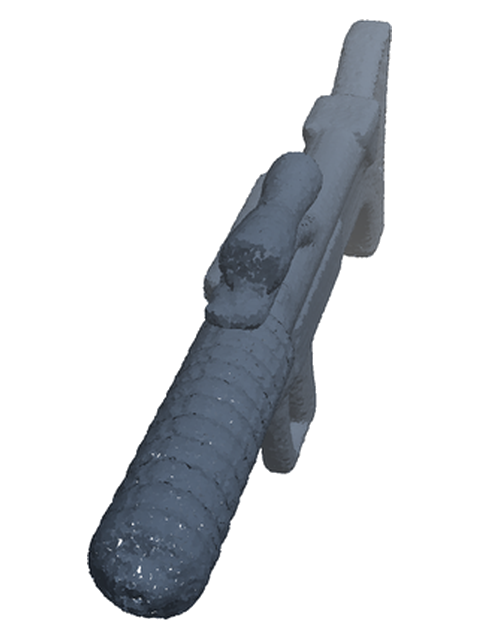} & \includegraphics[width=\gw]{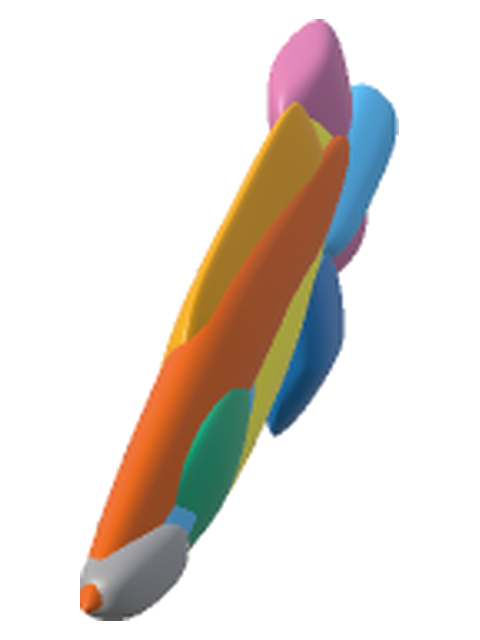} & \includegraphics[width=\gw]{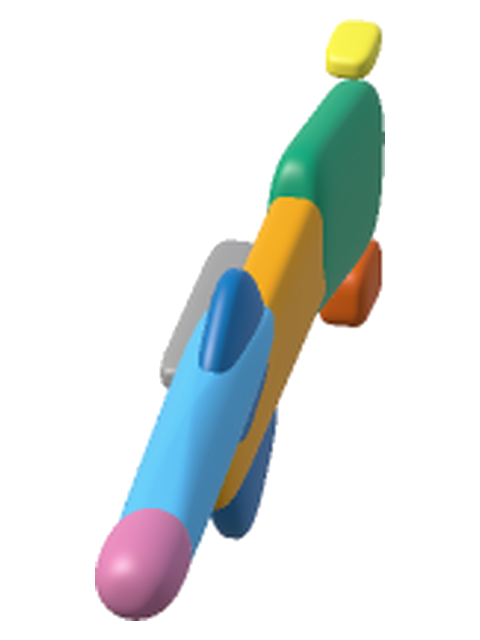} & \includegraphics[width=\gw]{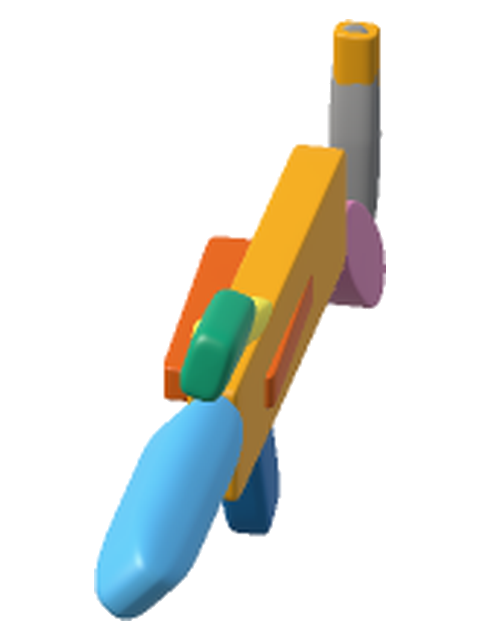} & \includegraphics[width=\gw]{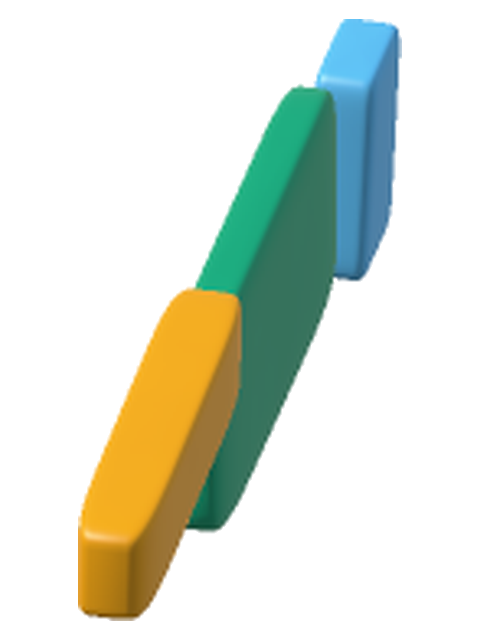} & \includegraphics[width=\gw]{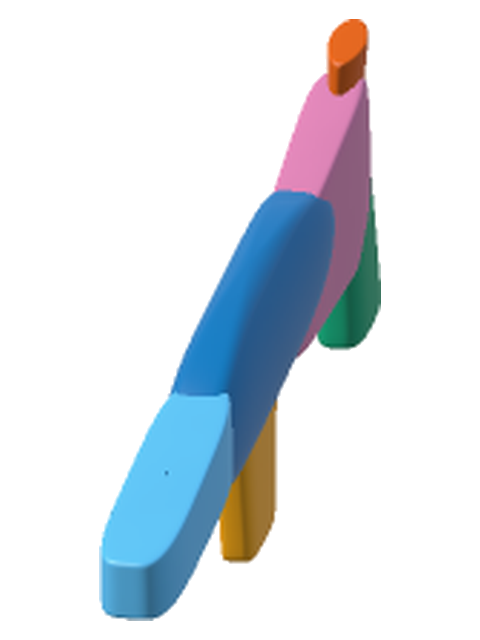} & \includegraphics[width=\gw]{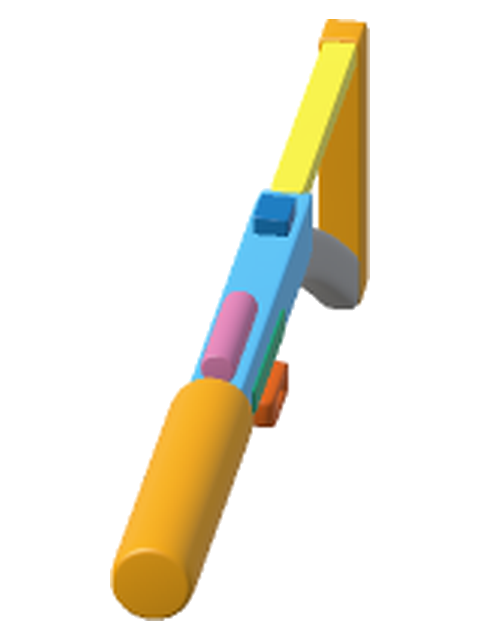} & \includegraphics[width=\gw]{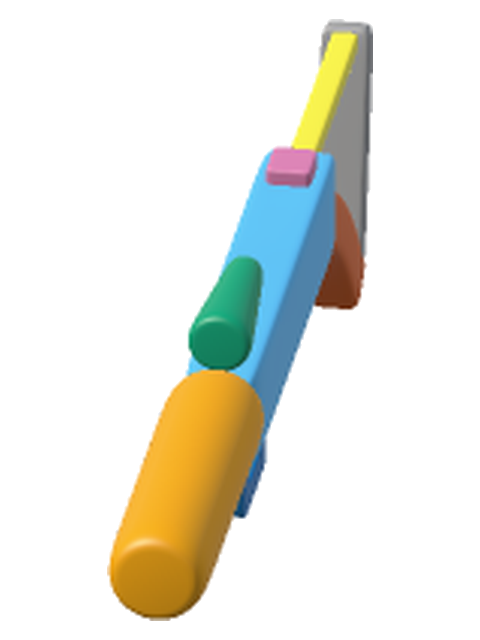} \\[0.4ex]
\scriptsize rifle & \scriptsize 30 & \scriptsize 55 & \scriptsize 54 & \scriptsize 30 & \scriptsize 52 & \scriptsize 75 & \scriptsize \textbf{76} \\[1.4ex]
\includegraphics[width=\gw]{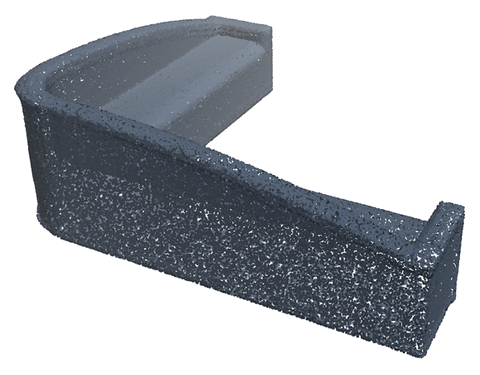} & \includegraphics[width=\gw]{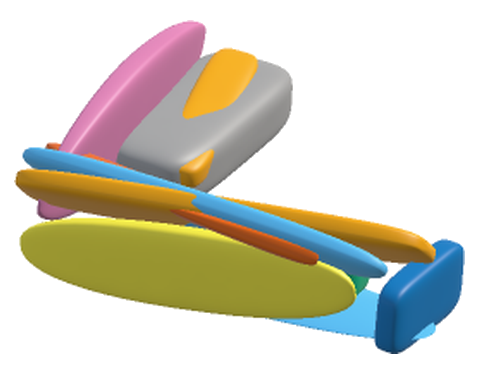} & \includegraphics[width=\gw]{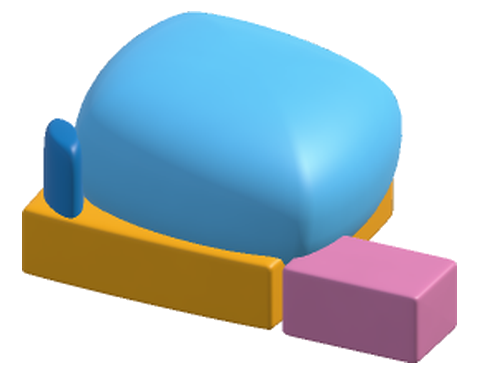} & \includegraphics[width=\gw]{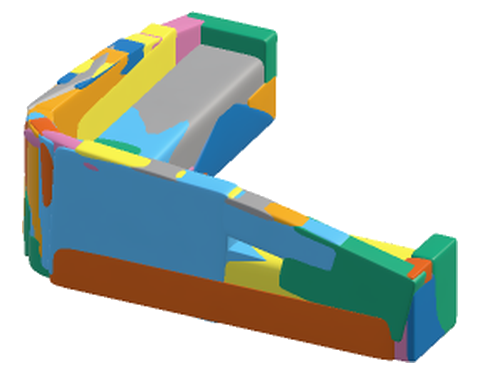} & \includegraphics[width=\gw]{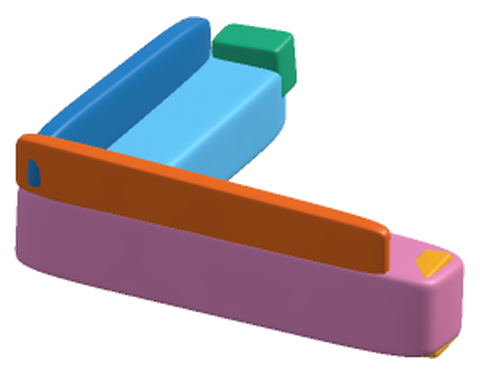} & \includegraphics[width=\gw]{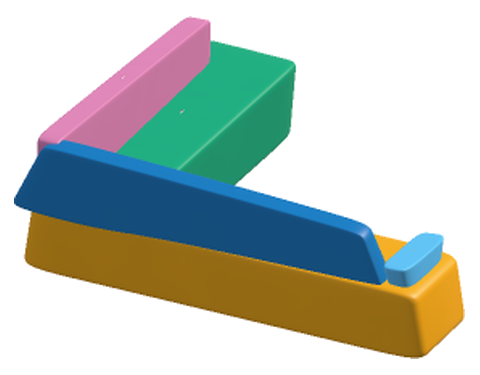} & \includegraphics[width=\gw]{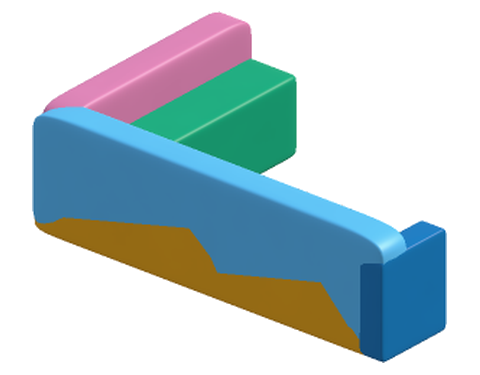} & \includegraphics[width=\gw]{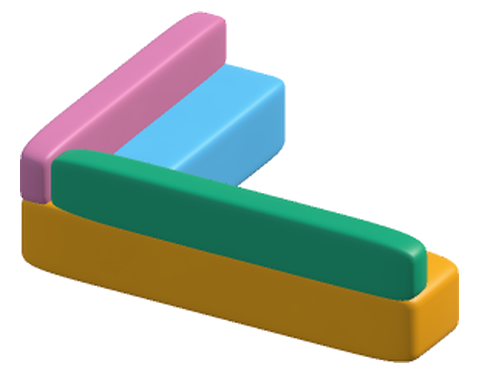} \\[0.4ex]
\scriptsize sofa & \scriptsize 24 & \scriptsize 29 & \scriptsize 75 & \scriptsize 59 & \scriptsize \textbf{77} & \scriptsize 73 & \scriptsize 73 \\[1.4ex]
\includegraphics[width=\gw]{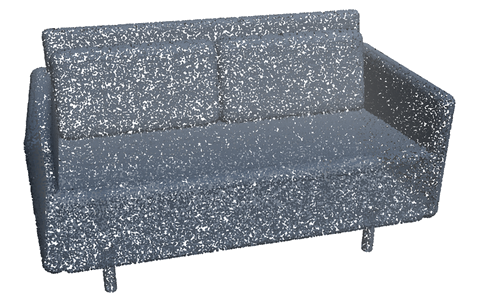} & \includegraphics[width=\gw]{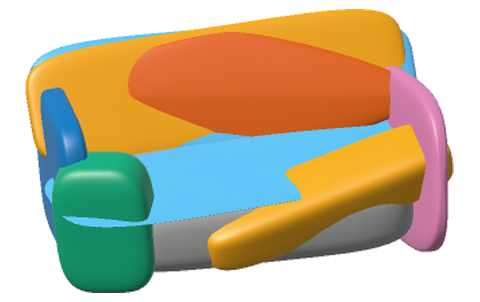} & \includegraphics[width=\gw]{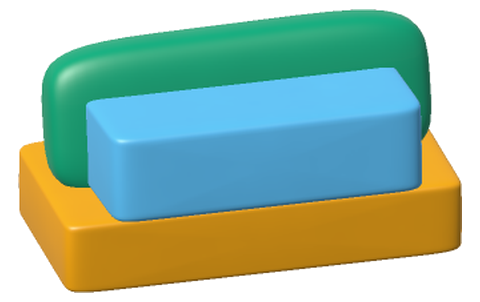} & \includegraphics[width=\gw]{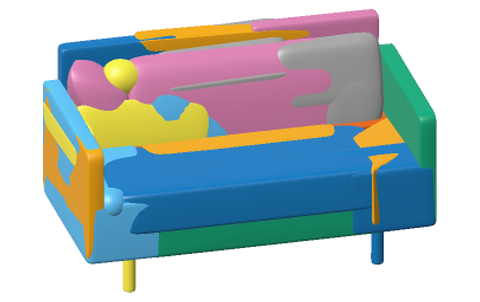} & \includegraphics[width=\gw]{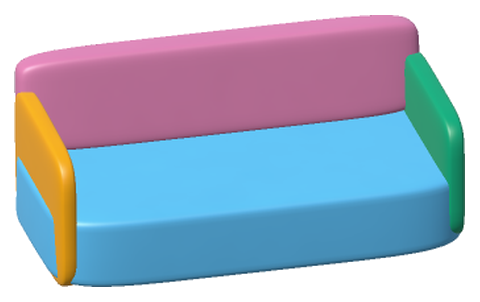} & \includegraphics[width=\gw]{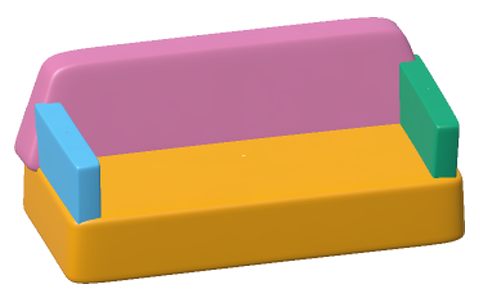} & \includegraphics[width=\gw]{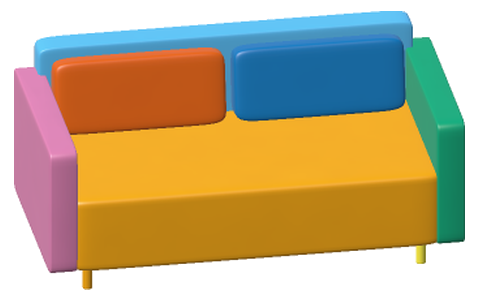} & \includegraphics[width=\gw]{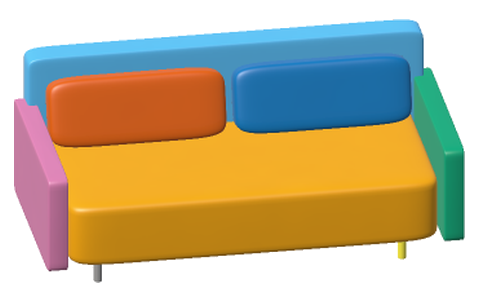} \\[0.4ex]
\scriptsize sofa & \scriptsize 38 & \scriptsize 30 & \scriptsize 65 & \scriptsize 56 & \scriptsize 54 & \scriptsize \textbf{100} & \scriptsize \textbf{100} \\[1.4ex]
\end{tabular}

  \caption{\textbf{Random objects, as predicted} (rows 15--20 of 26). Two objects per category, drawn at random and not selected. Each result is drawn from the input photo's viewpoint, one flat colour per part. Under each result is its VLM score (Claude Sonnet~5.5); the highest in each row is in bold.}
  \label{fig:s_more_visual_3}
\end{figure}

\begin{figure}[p]
  \centering
\setlength{\tabcolsep}{0pt}\renewcommand{\arraystretch}{0}
\def\gw{\dimexpr(\textwidth-7\dimexpr0.25em\relax)/8\relax}
\begin{tabular}{@{}c*{7}{@{\hspace{0.25em}}c}@{}}
\parbox[b]{\gw}{\centering\scriptsize\renewcommand{\arraystretch}{1}Input} & \parbox[b]{\gw}{\centering\scriptsize\renewcommand{\arraystretch}{1}SQ} & \parbox[b]{\gw}{\centering\scriptsize\renewcommand{\arraystretch}{1}EMS} & \parbox[b]{\gw}{\centering\scriptsize\renewcommand{\arraystretch}{1}Marching-Prim.} & \parbox[b]{\gw}{\centering\scriptsize\renewcommand{\arraystretch}{1}SuperDec} & \parbox[b]{\gw}{\centering\scriptsize\renewcommand{\arraystretch}{1}SuperFlex} & \parbox[b]{\gw}{\centering\scriptsize\renewcommand{\arraystretch}{1}PhysDec\\(Evolved)} & \parbox[b]{\gw}{\centering\scriptsize\renewcommand{\arraystretch}{1}PhysDec\\(Loop)} \\[0.5ex]
\includegraphics[width=\gw]{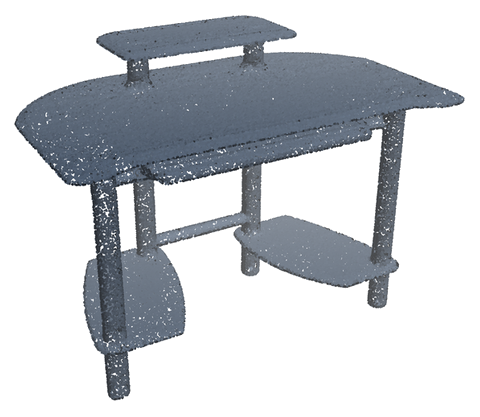} & \includegraphics[width=\gw]{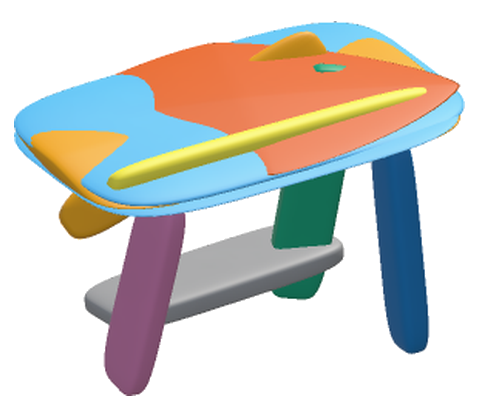} & \includegraphics[width=\gw]{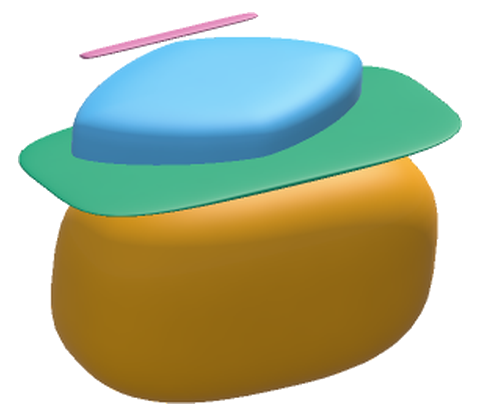} & \includegraphics[width=\gw]{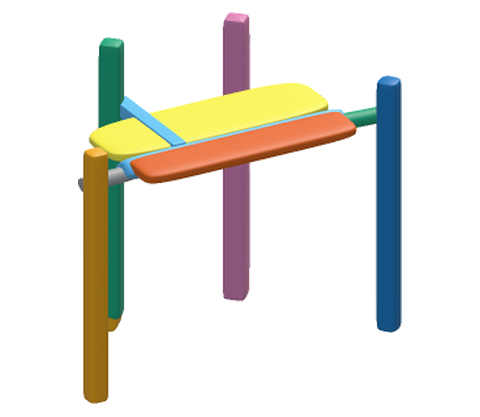} & \includegraphics[width=\gw]{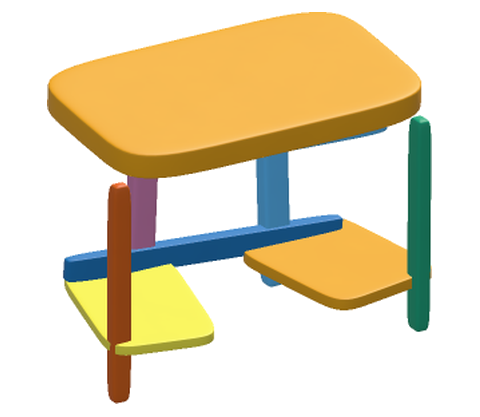} & \includegraphics[width=\gw]{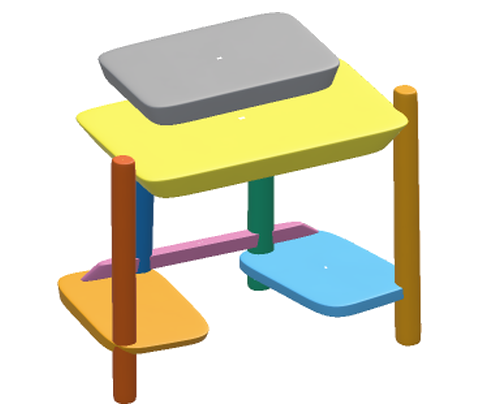} & \includegraphics[width=\gw]{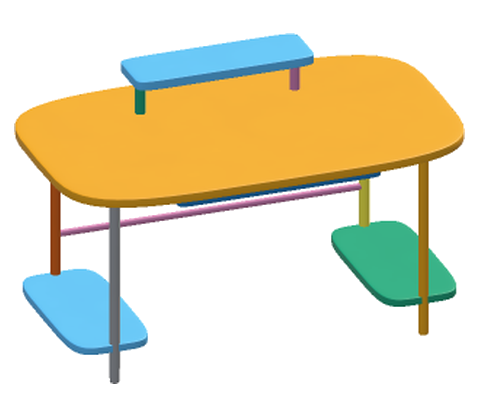} & \includegraphics[width=\gw]{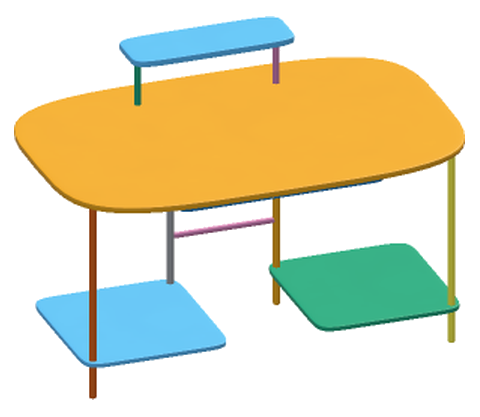} \\[0.4ex]
\scriptsize table & \scriptsize 37 & \scriptsize 26 & \scriptsize 24 & \scriptsize 51 & \scriptsize 50 & \scriptsize \textbf{74} & \scriptsize 73 \\[1.4ex]
\includegraphics[width=\gw]{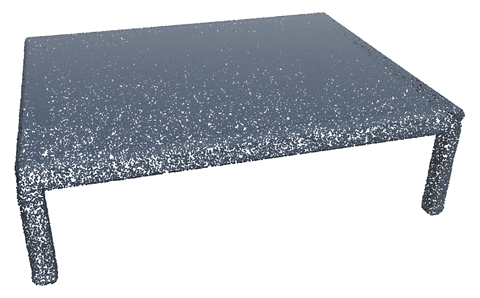} & \includegraphics[width=\gw]{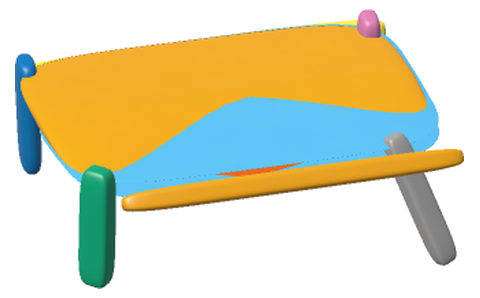} & \includegraphics[width=\gw]{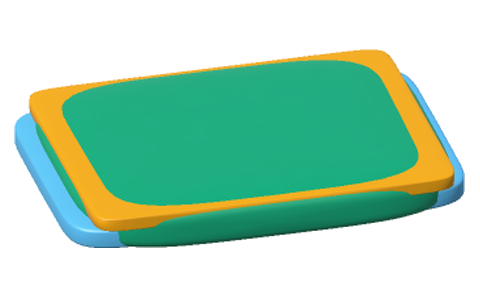} & \includegraphics[width=\gw]{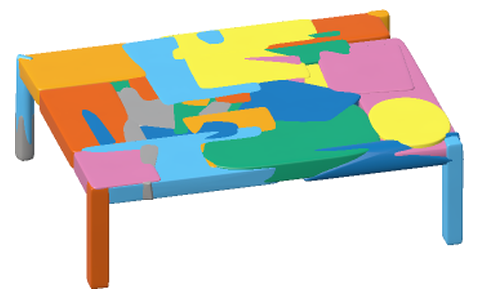} & \includegraphics[width=\gw]{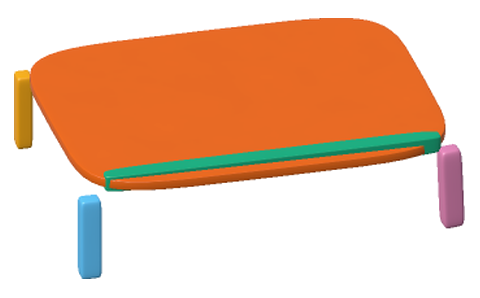} & \includegraphics[width=\gw]{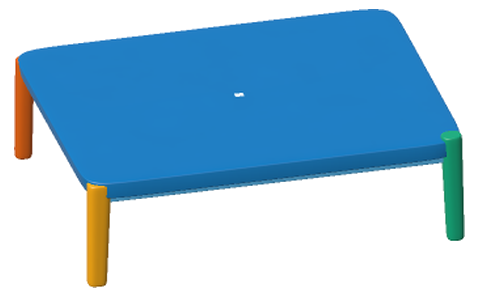} & \includegraphics[width=\gw]{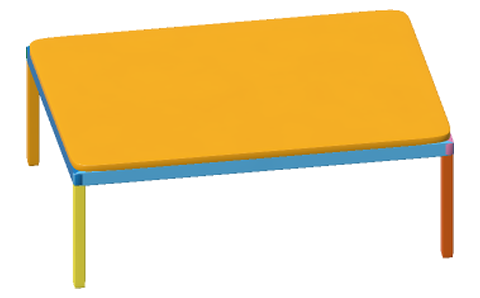} & \includegraphics[width=\gw]{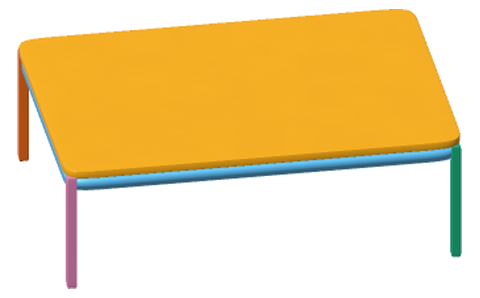} \\[0.4ex]
\scriptsize table & \scriptsize 38 & \scriptsize 27 & \scriptsize 73 & \scriptsize 54 & \scriptsize 78 & \scriptsize 83 & \scriptsize \textbf{84} \\[1.4ex]
\includegraphics[width=\gw]{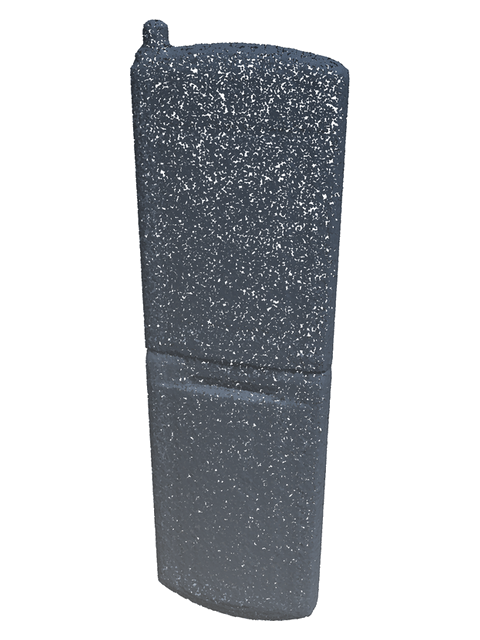} & \includegraphics[width=\gw]{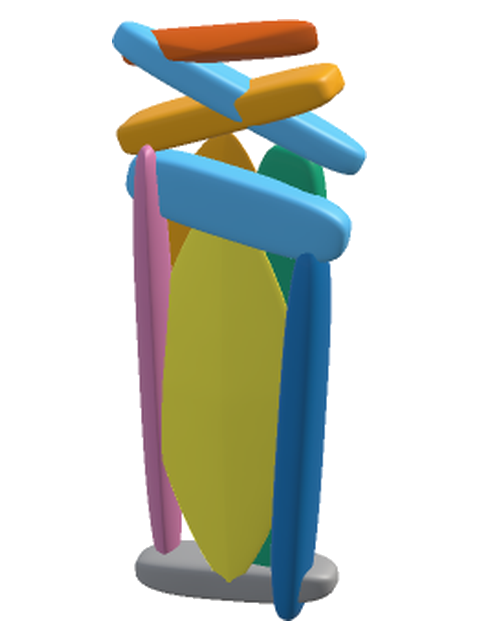} & \includegraphics[width=\gw]{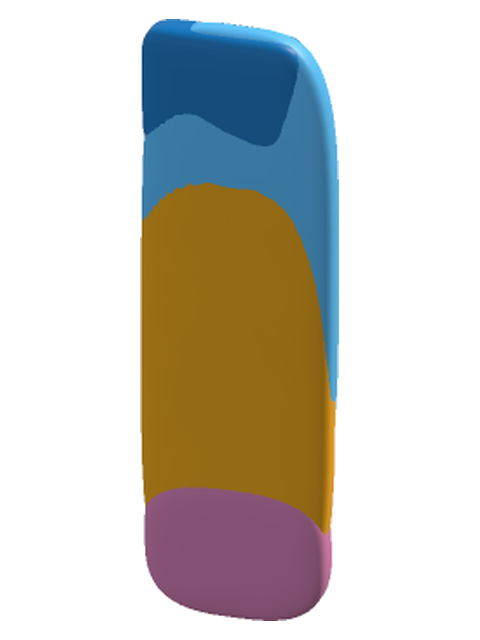} & \includegraphics[width=\gw]{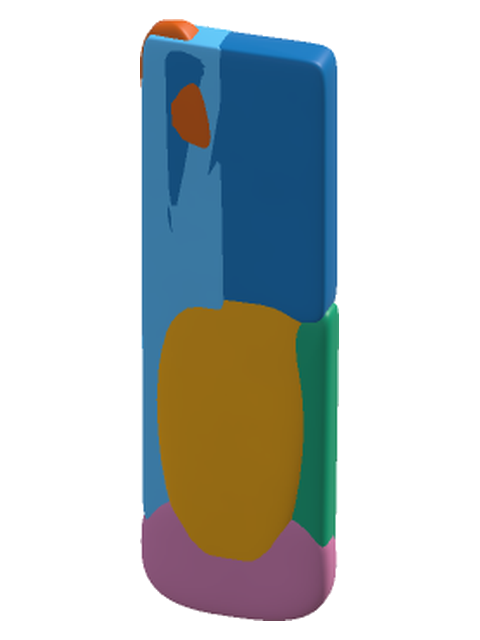} & \includegraphics[width=\gw]{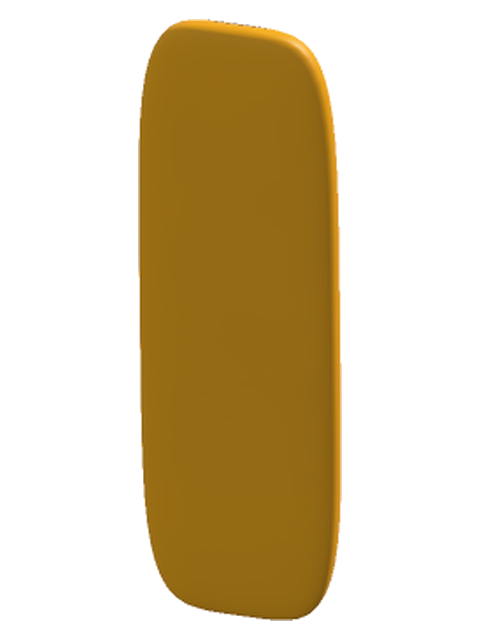} & \includegraphics[width=\gw]{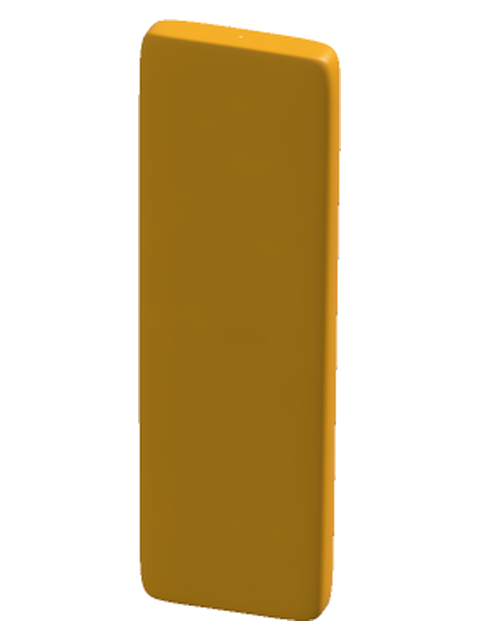} & \includegraphics[width=\gw]{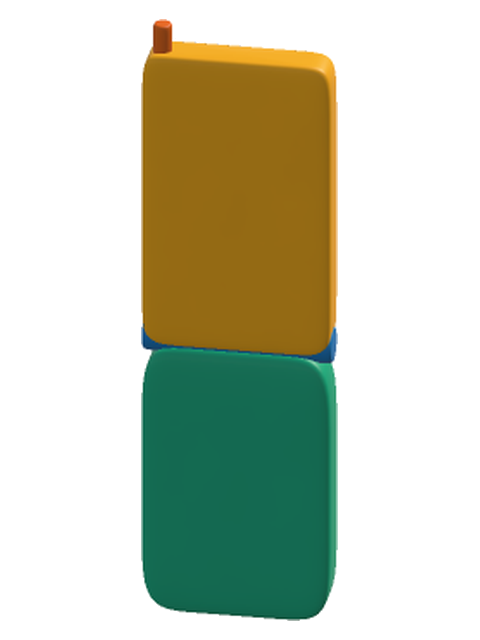} & \includegraphics[width=\gw]{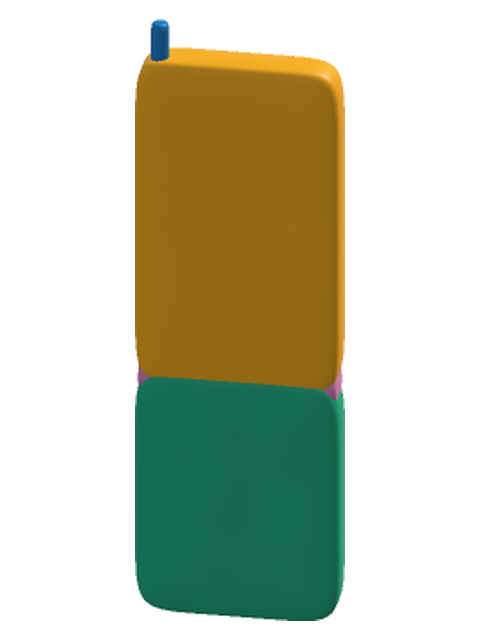} \\[0.4ex]
\scriptsize telephone & \scriptsize 29 & \scriptsize 58 & \scriptsize 64 & \scriptsize 50 & \scriptsize 40 & \scriptsize \textbf{82} & \scriptsize 78 \\[1.4ex]
\includegraphics[width=\gw]{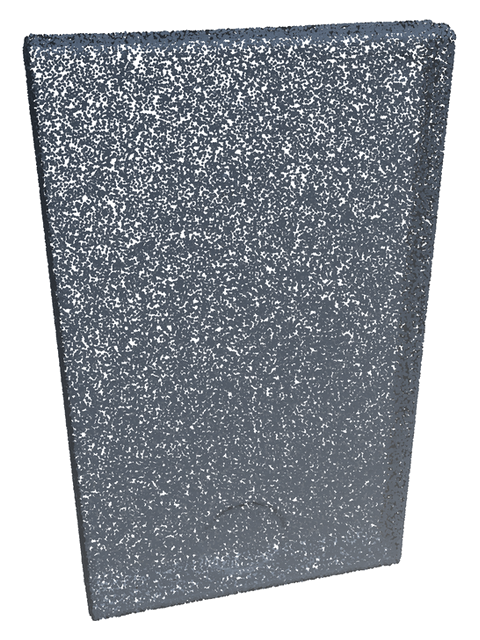} & \includegraphics[width=\gw]{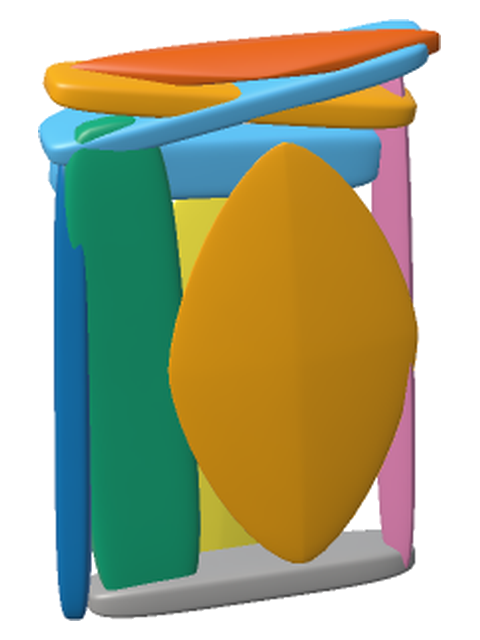} & \includegraphics[width=\gw]{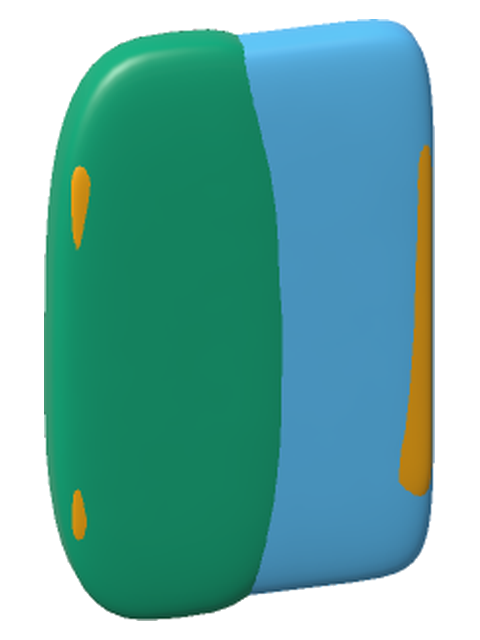} & \includegraphics[width=\gw]{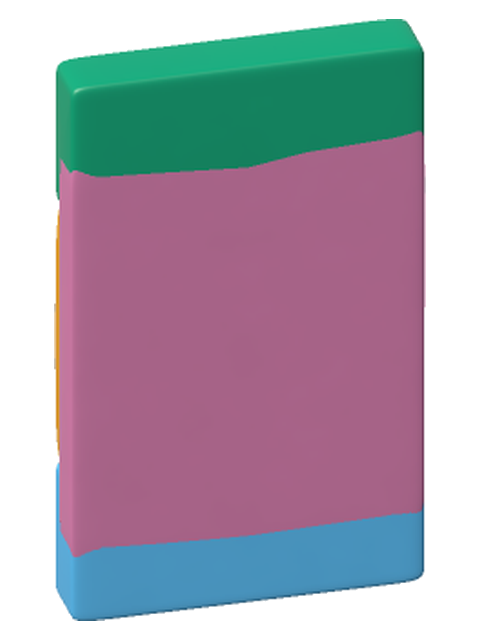} & \includegraphics[width=\gw]{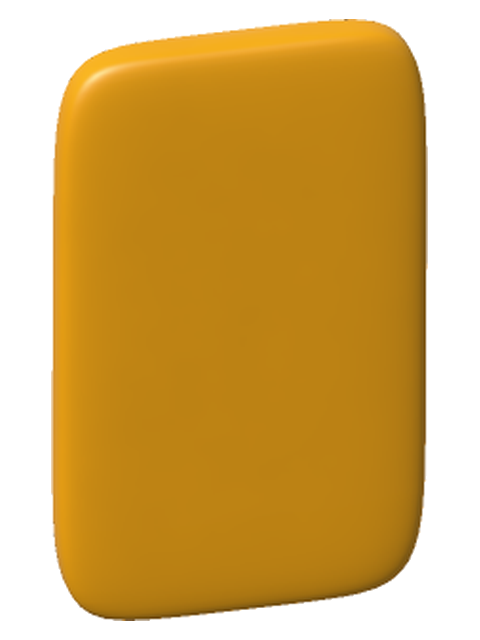} & \includegraphics[width=\gw]{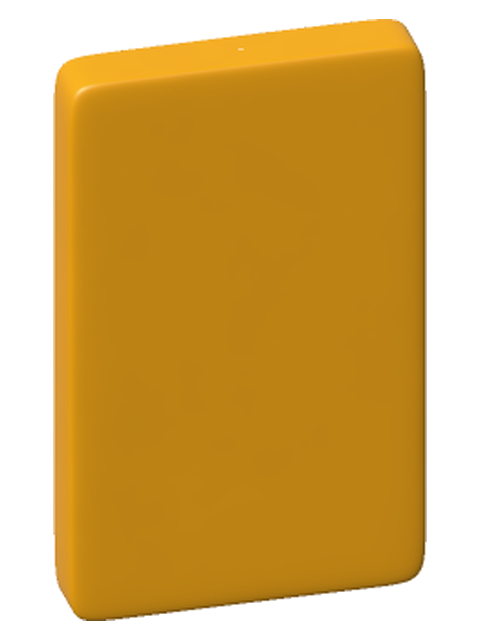} & \includegraphics[width=\gw]{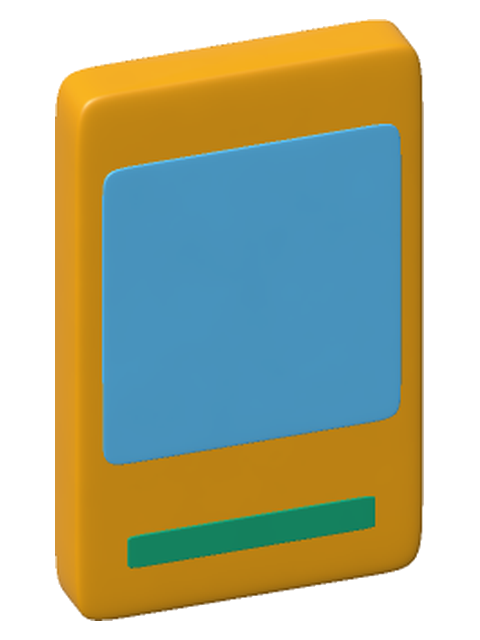} & \includegraphics[width=\gw]{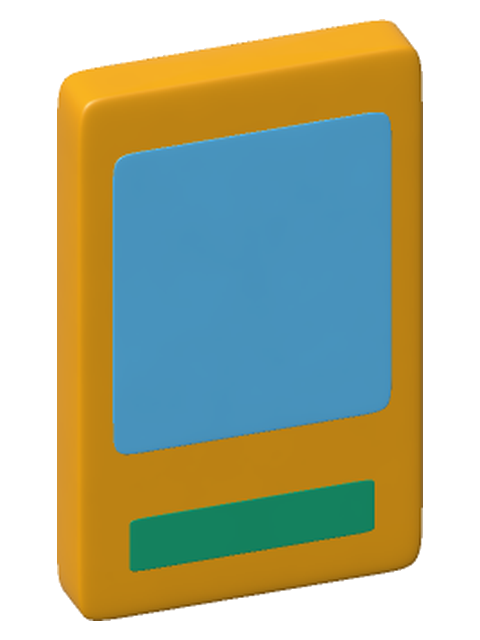} \\[0.4ex]
\scriptsize telephone & \scriptsize 27 & \scriptsize 30 & \scriptsize 61 & \scriptsize 30 & \scriptsize 44 & \scriptsize 79 & \scriptsize \textbf{83} \\[1.4ex]
\includegraphics[width=\gw]{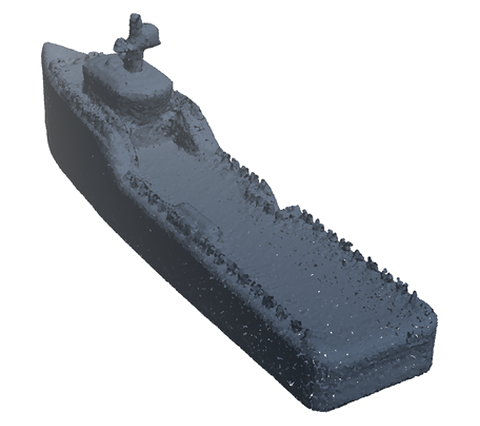} & \includegraphics[width=\gw]{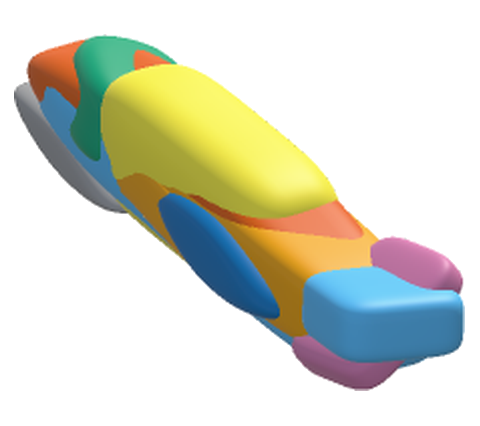} & \includegraphics[width=\gw]{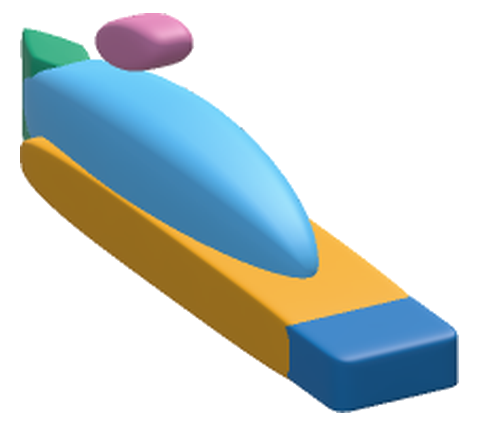} & \includegraphics[width=\gw]{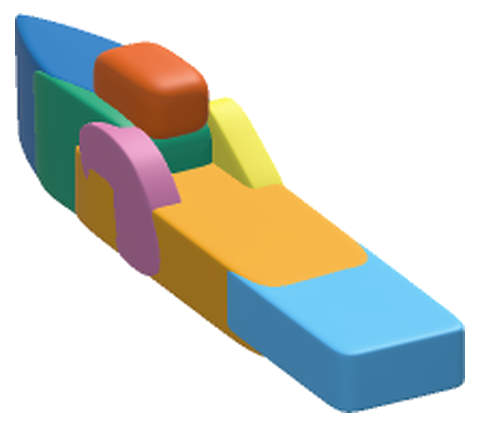} & \includegraphics[width=\gw]{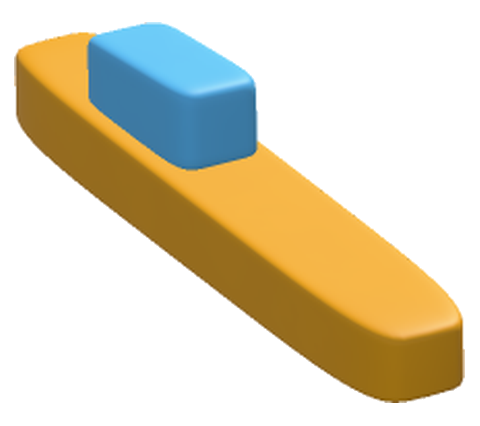} & \includegraphics[width=\gw]{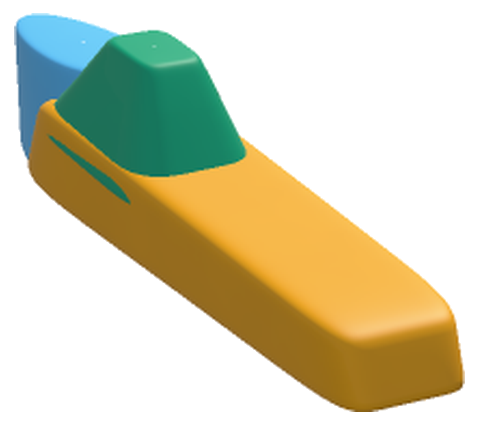} & \includegraphics[width=\gw]{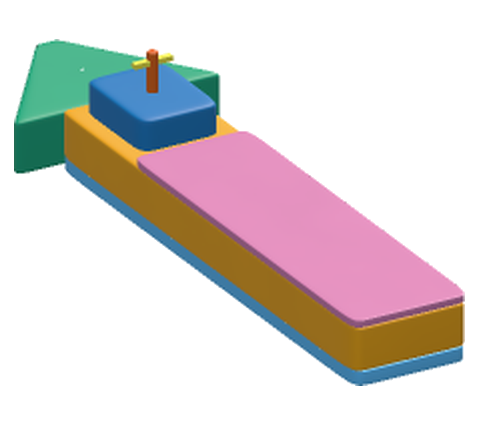} & \includegraphics[width=\gw]{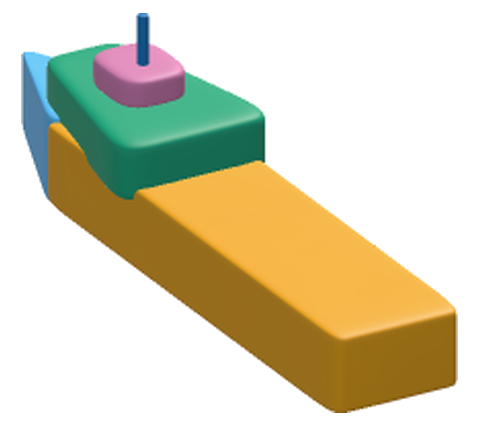} \\[0.4ex]
\scriptsize watercraft & \scriptsize 54 & \scriptsize 62 & \scriptsize 77 & \scriptsize 65 & \scriptsize 77 & \scriptsize 63 & \scriptsize \textbf{78} \\[1.4ex]
\includegraphics[width=\gw]{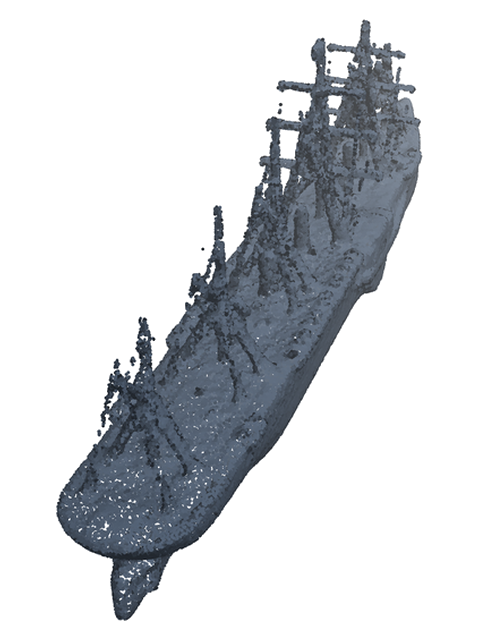} & \includegraphics[width=\gw]{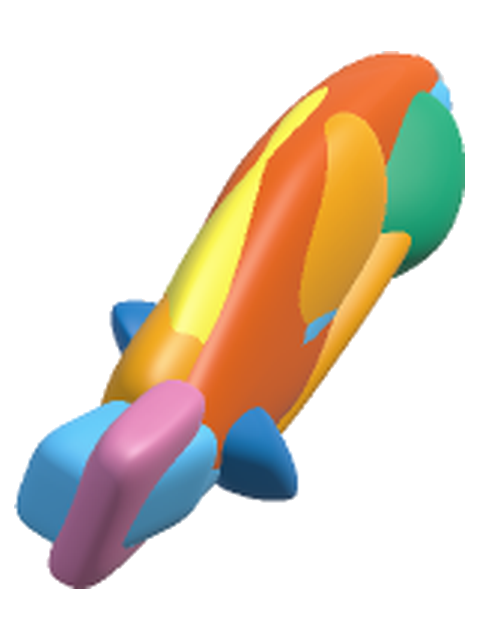} & \includegraphics[width=\gw]{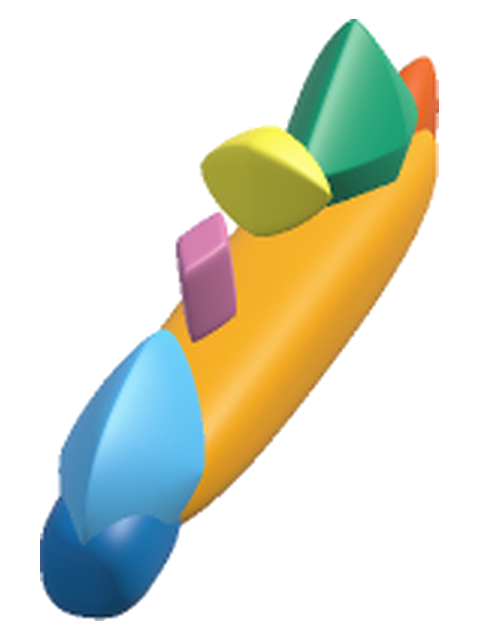} & \includegraphics[width=\gw]{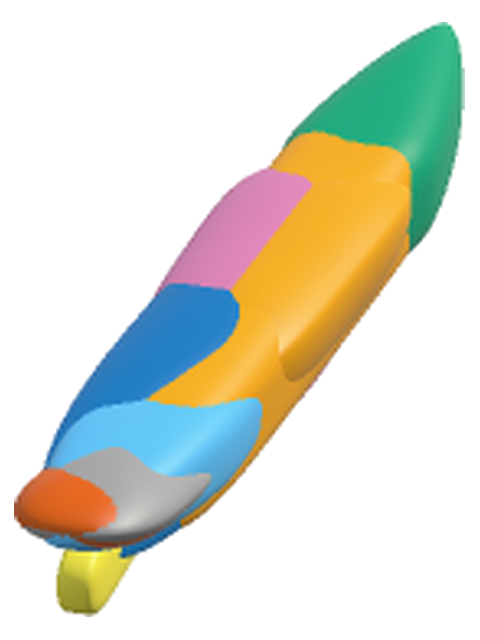} & \includegraphics[width=\gw]{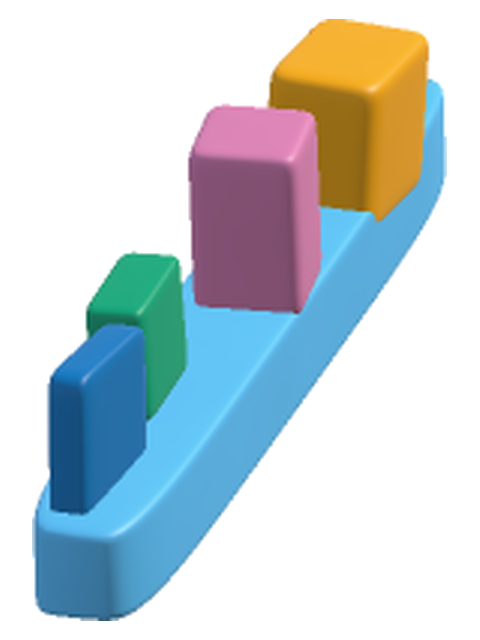} & \includegraphics[width=\gw]{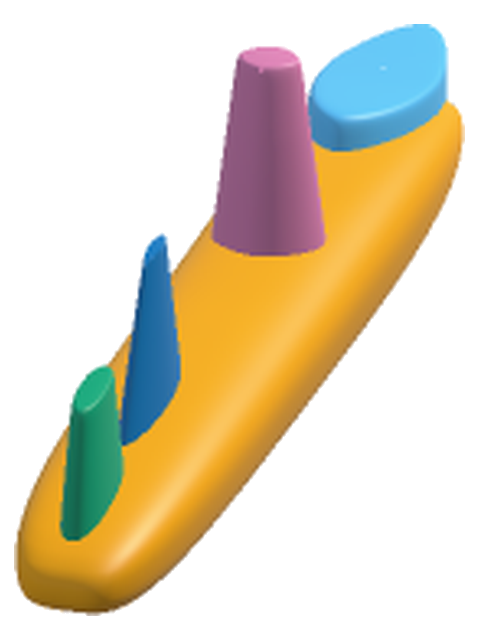} & \includegraphics[width=\gw]{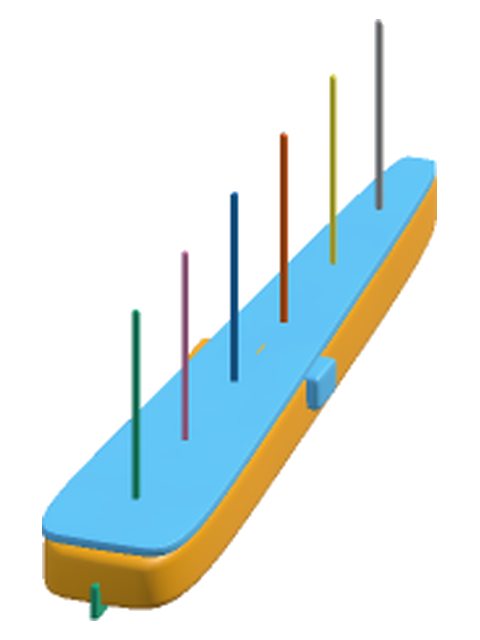} & \includegraphics[width=\gw]{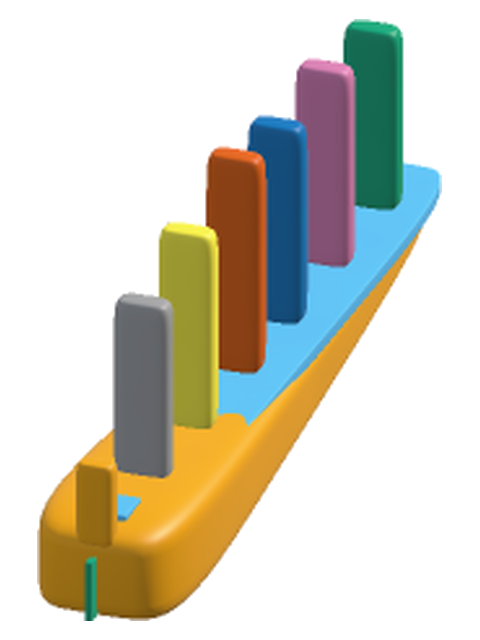} \\[0.4ex]
\scriptsize watercraft & \scriptsize 30 & \scriptsize 43 & \scriptsize 50 & \scriptsize 55 & \scriptsize 53 & \scriptsize \textbf{77} & \scriptsize 73 \\[1.4ex]
\end{tabular}

  \caption{\textbf{Random objects, as predicted} (rows 21--26 of 26). Two objects per category, drawn at random and not selected. Each result is drawn from the input photo's viewpoint, one flat colour per part. Under each result is its VLM score (Claude Sonnet~5.5); the highest in each row is in bold.}
  \label{fig:s_more_visual_4}
\end{figure}

\Crefrange{fig:s_more_visual_1}{fig:s_more_visual_4} show 26 objects, two per category.
They are drawn at random from the 585 test objects, leaving out every object shown elsewhere in the paper and this supplement, and are not selected.
A PhysDec result has the highest VLM score on 23 of them.
Marching-Primitives has it on an airplane and a chair, and SuperFlex on a sofa.
Over the 26, the mean VLM score is 77.4 for PhysDec-Loop and 77.1 for PhysDec-Evolved, and from 39.4 to 60.7 for the baselines.

\begin{figure}[p]
  \centering
\setlength{\tabcolsep}{0pt}\renewcommand{\arraystretch}{0}
\def\gw{\dimexpr(\textwidth-7\dimexpr0.25em\relax)/8\relax}
\begin{tabular}{@{}c*{7}{@{\hspace{0.25em}}c}@{}}
\parbox[b]{\gw}{\centering\scriptsize\renewcommand{\arraystretch}{1}Input} & \parbox[b]{\gw}{\centering\scriptsize\renewcommand{\arraystretch}{1}SQ} & \parbox[b]{\gw}{\centering\scriptsize\renewcommand{\arraystretch}{1}EMS} & \parbox[b]{\gw}{\centering\scriptsize\renewcommand{\arraystretch}{1}Marching-Prim.} & \parbox[b]{\gw}{\centering\scriptsize\renewcommand{\arraystretch}{1}SuperDec} & \parbox[b]{\gw}{\centering\scriptsize\renewcommand{\arraystretch}{1}SuperFlex} & \parbox[b]{\gw}{\centering\scriptsize\renewcommand{\arraystretch}{1}PhysDec\\(Evolved)} & \parbox[b]{\gw}{\centering\scriptsize\renewcommand{\arraystretch}{1}PhysDec\\(Loop)} \\[0.5ex]
\includegraphics[width=\gw]{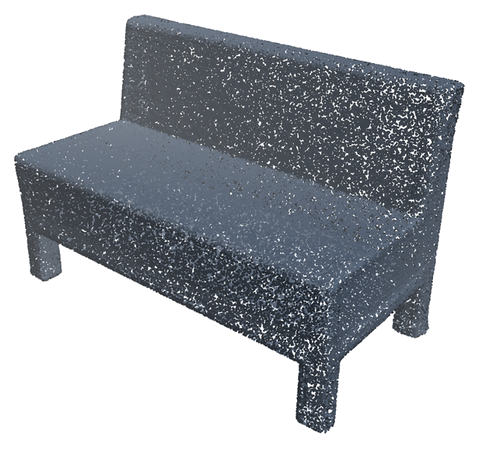} & \includegraphics[width=\gw]{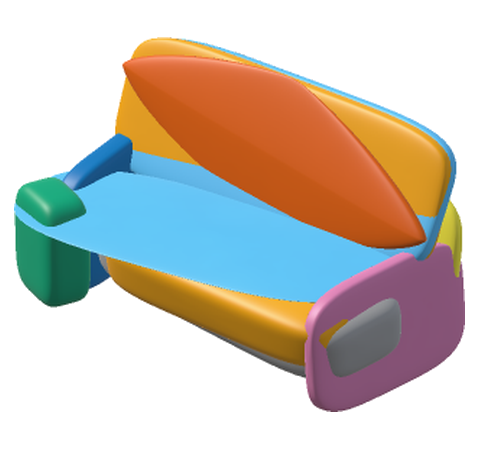} & \includegraphics[width=\gw]{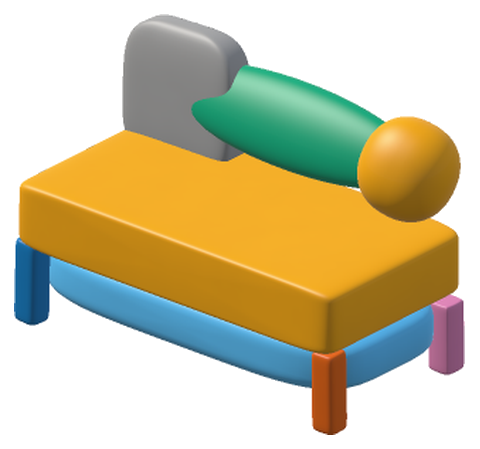} & \includegraphics[width=\gw]{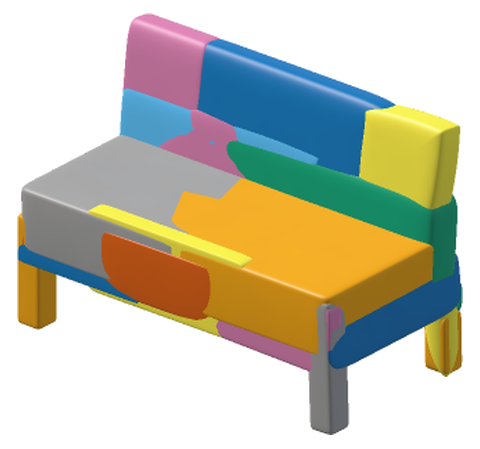} & \includegraphics[width=\gw]{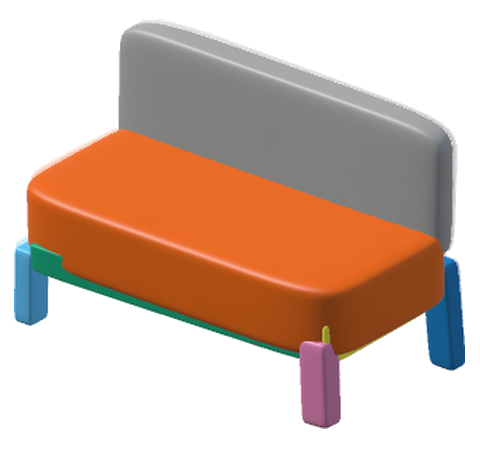} & \includegraphics[width=\gw]{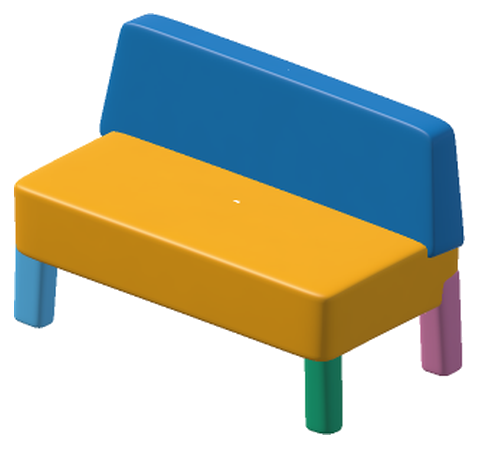} & \includegraphics[width=\gw]{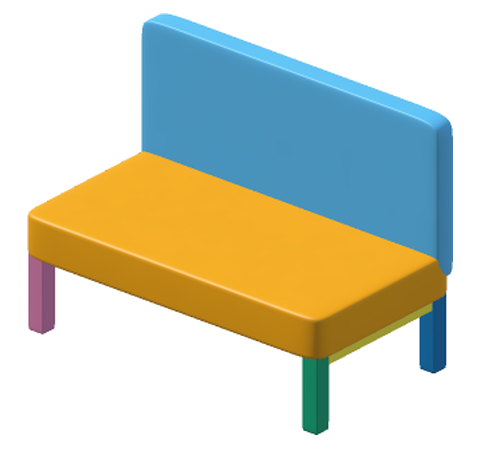} & \includegraphics[width=\gw]{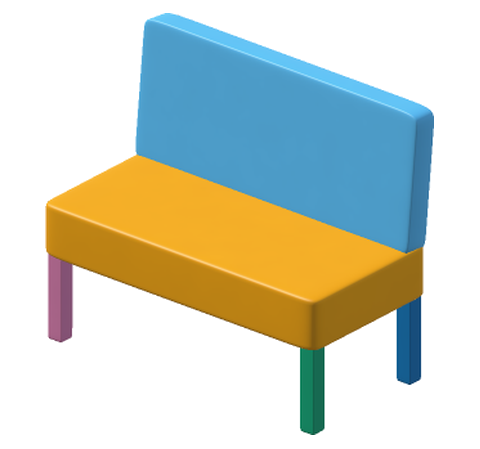} \\[0.4ex]
\scriptsize bench & \scriptsize \textcolor{passc}{stands} & \scriptsize \textcolor{passc}{stands} & \scriptsize \textcolor{passc}{stands} & \scriptsize \textcolor{passc}{stands} & \scriptsize \textcolor{passc}{stands} & \scriptsize \textcolor{passc}{stands} & \scriptsize \textcolor{passc}{stands} \\[1.4ex]
\includegraphics[width=\gw]{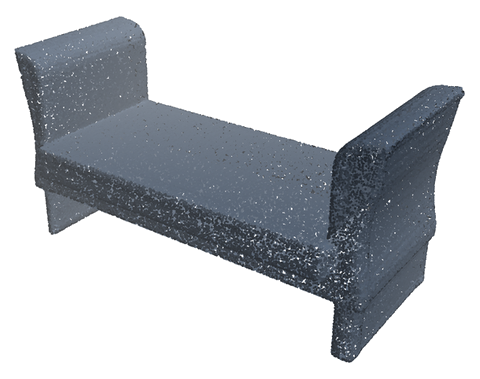} & \includegraphics[width=\gw]{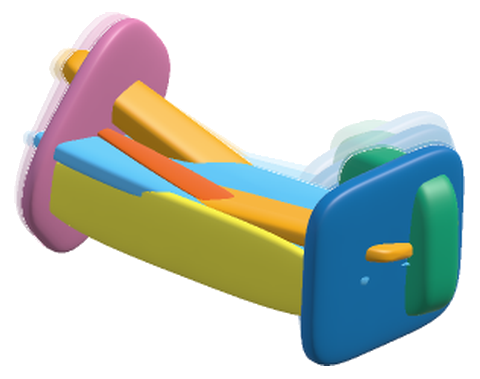} & \includegraphics[width=\gw]{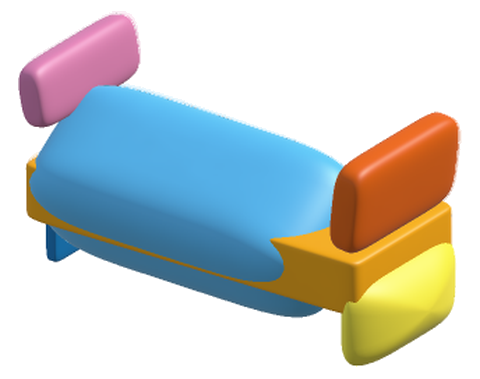} & \includegraphics[width=\gw]{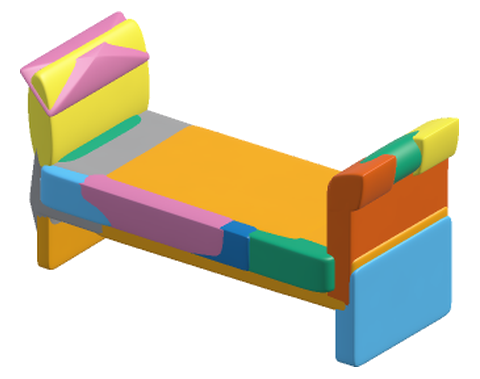} & \includegraphics[width=\gw]{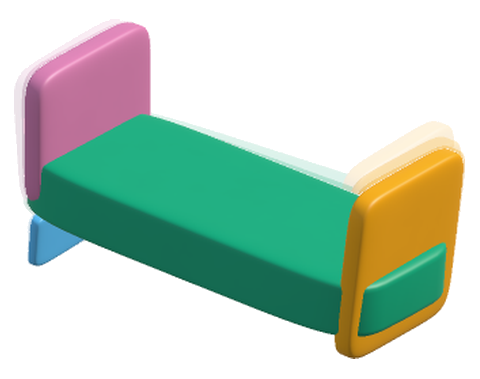} & \includegraphics[width=\gw]{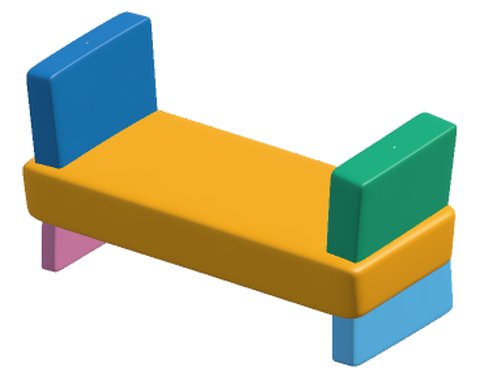} & \includegraphics[width=\gw]{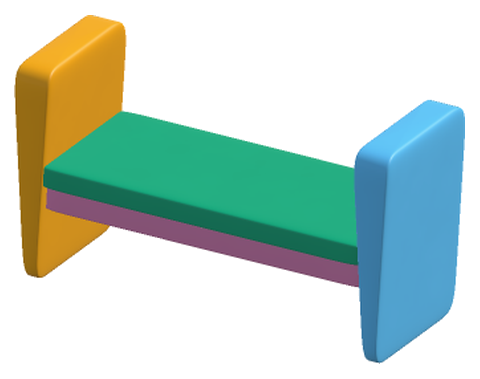} & \includegraphics[width=\gw]{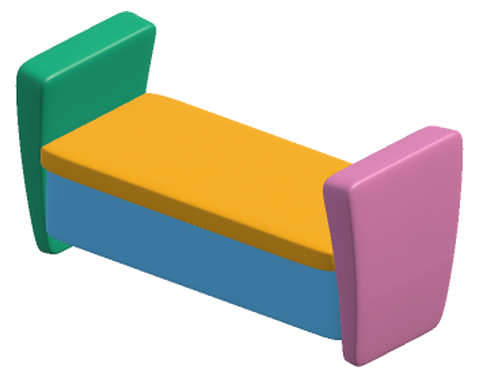} \\[0.4ex]
\scriptsize bench & \scriptsize \textcolor{passc}{stands} & \scriptsize \textcolor{passc}{stands} & \scriptsize \textcolor{passc}{stands} & \scriptsize \textcolor{passc}{stands} & \scriptsize \textcolor{passc}{stands} & \scriptsize \textcolor{passc}{stands} & \scriptsize \textcolor{passc}{stands} \\[1.4ex]
\includegraphics[width=\gw]{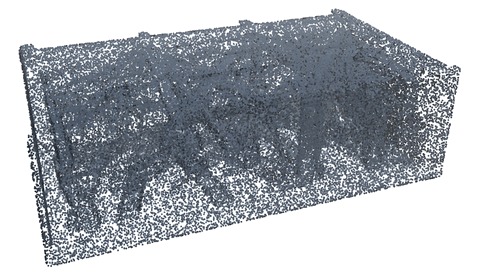} & \includegraphics[width=\gw]{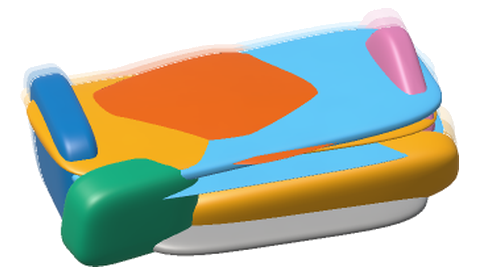} & \includegraphics[width=\gw]{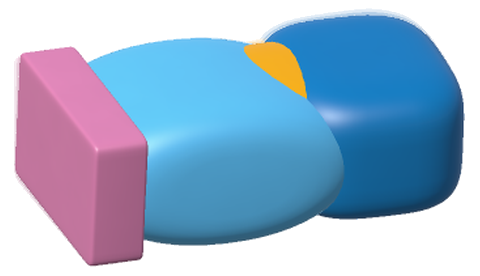} & \includegraphics[width=\gw]{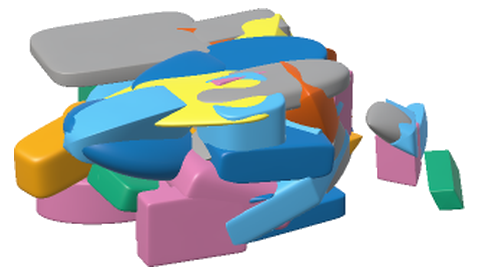} & \includegraphics[width=\gw]{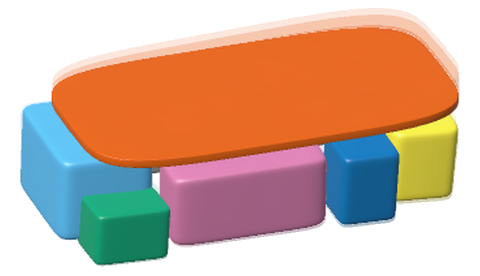} & \includegraphics[width=\gw]{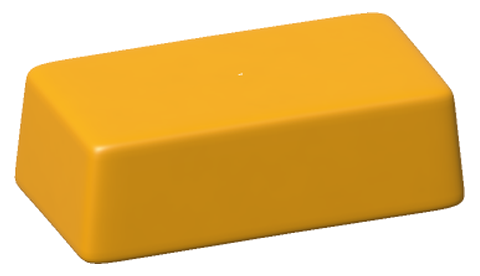} & \includegraphics[width=\gw]{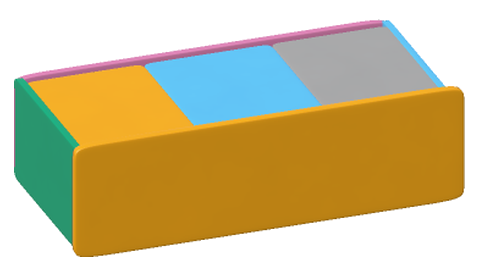} & \includegraphics[width=\gw]{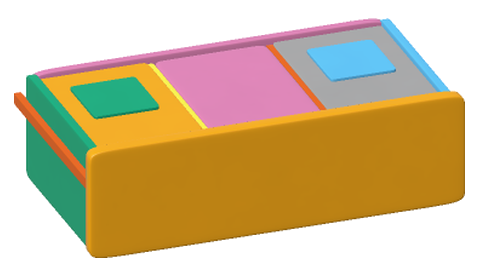} \\[0.4ex]
\scriptsize cabinet & \scriptsize \textcolor{failc}{falls} & \scriptsize \textcolor{passc}{stands} & \scriptsize \textcolor{failc}{2 pieces} & \scriptsize \textcolor{failc}{2 pieces} & \scriptsize \textcolor{passc}{stands} & \scriptsize \textcolor{passc}{stands} & \scriptsize \textcolor{passc}{stands} \\[1.4ex]
\includegraphics[width=\gw]{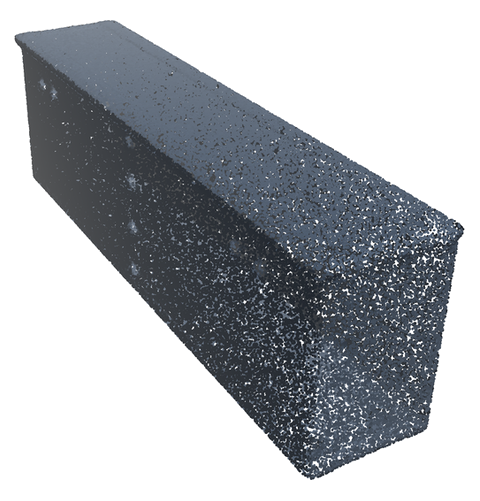} & \includegraphics[width=\gw]{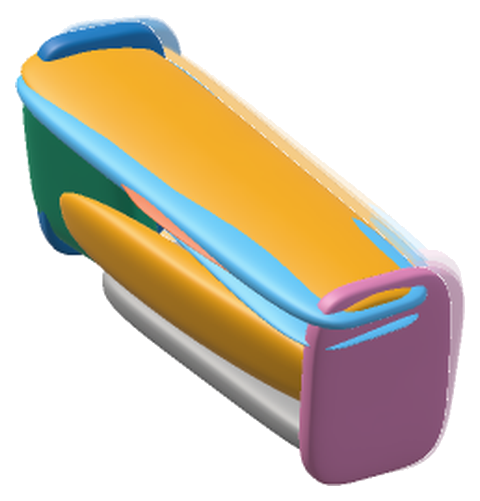} & \includegraphics[width=\gw]{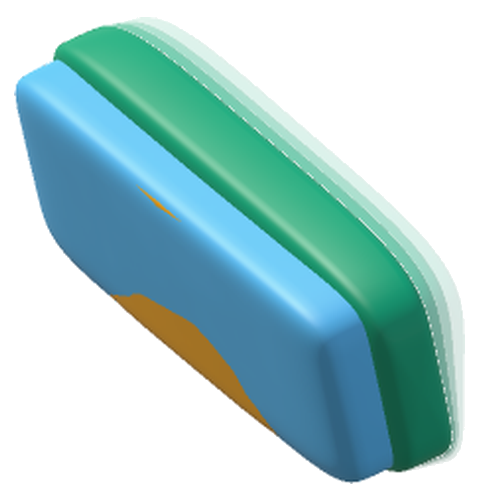} & \includegraphics[width=\gw]{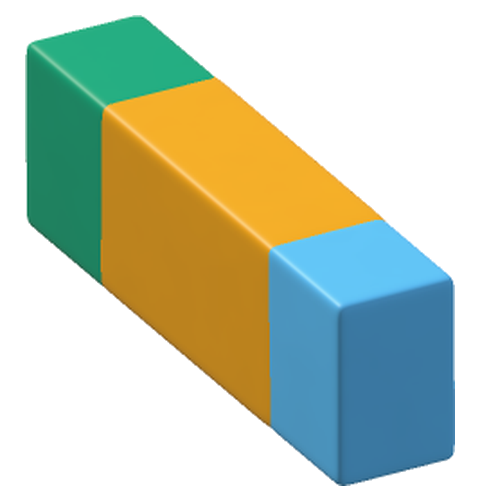} & \includegraphics[width=\gw]{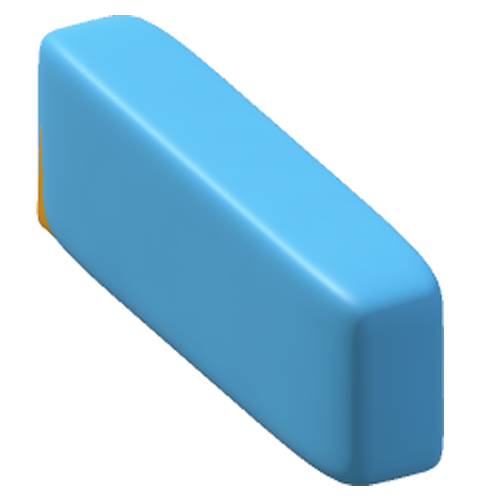} & \includegraphics[width=\gw]{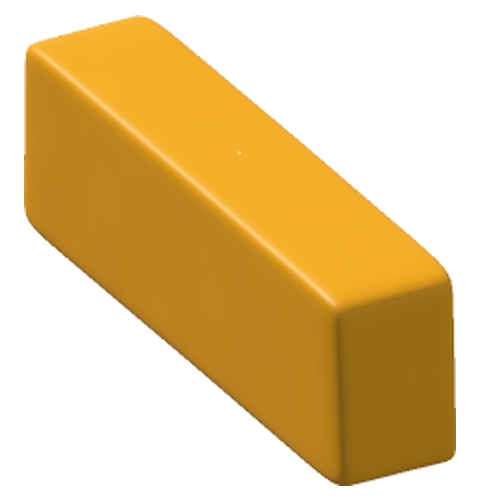} & \includegraphics[width=\gw]{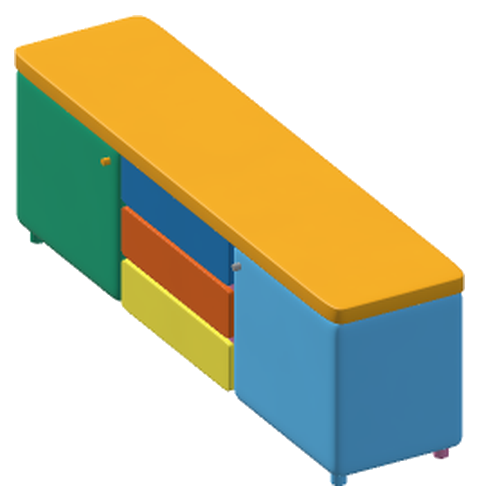} & \includegraphics[width=\gw]{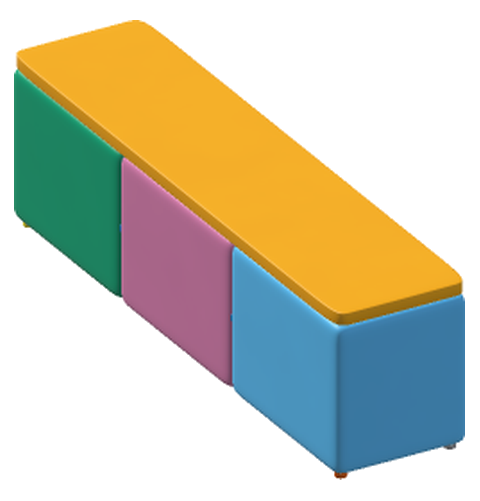} \\[0.4ex]
\scriptsize cabinet & \scriptsize \textcolor{passc}{stands} & \scriptsize \textcolor{failc}{falls} & \scriptsize \textcolor{passc}{stands} & \scriptsize \textcolor{passc}{stands} & \scriptsize \textcolor{passc}{stands} & \scriptsize \textcolor{passc}{stands} & \scriptsize \textcolor{passc}{stands} \\[1.4ex]
\includegraphics[width=\gw]{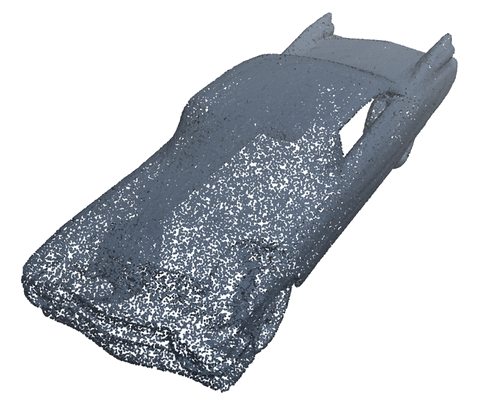} & \includegraphics[width=\gw]{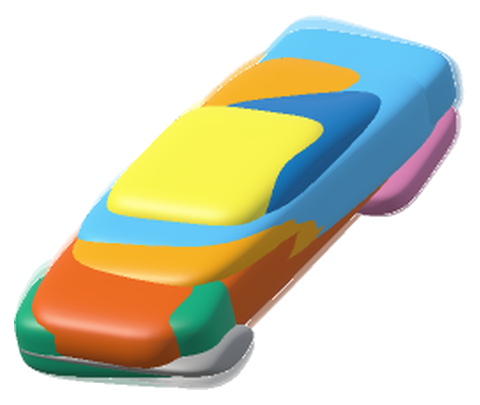} & \includegraphics[width=\gw]{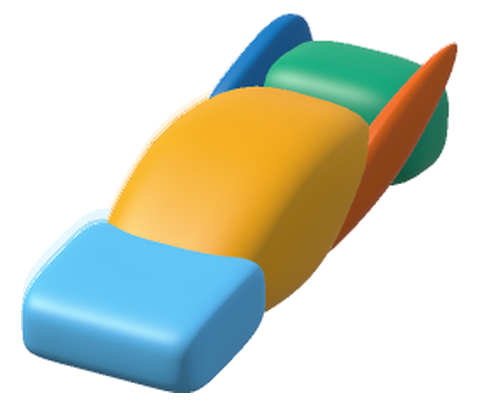} & \includegraphics[width=\gw]{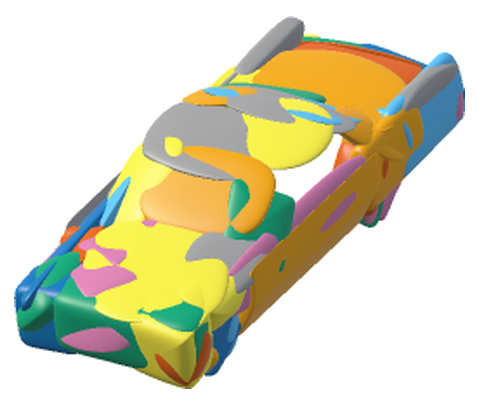} & \includegraphics[width=\gw]{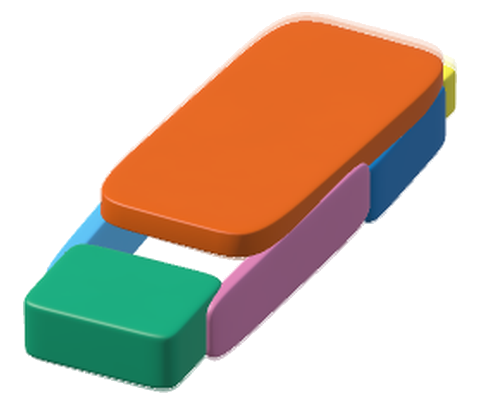} & \includegraphics[width=\gw]{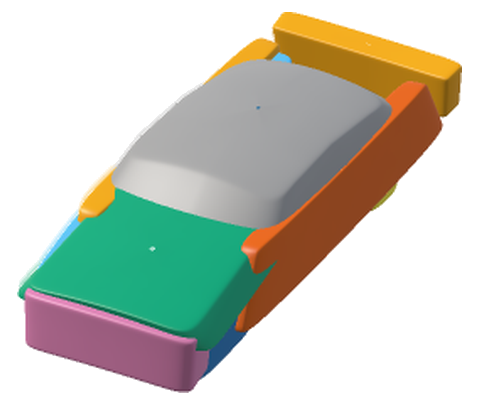} & \includegraphics[width=\gw]{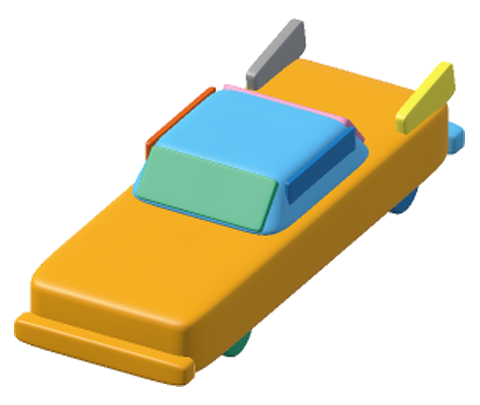} & \includegraphics[width=\gw]{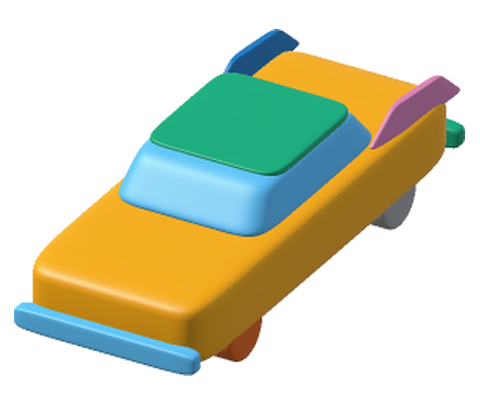} \\[0.4ex]
\scriptsize car & \scriptsize \textcolor{passc}{stands} & \scriptsize \textcolor{passc}{stands} & \scriptsize \textcolor{passc}{stands} & \scriptsize \textcolor{passc}{stands} & \scriptsize \textcolor{passc}{stands} & \scriptsize \textcolor{passc}{stands} & \scriptsize \textcolor{passc}{stands} \\[1.4ex]
\includegraphics[width=\gw]{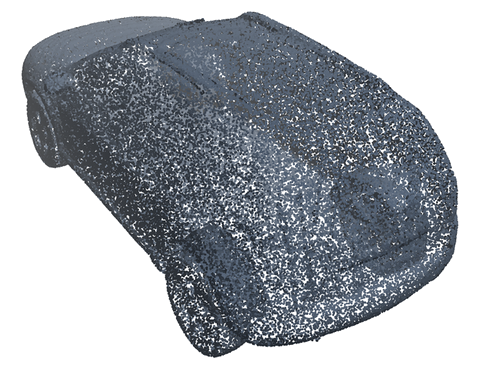} & \includegraphics[width=\gw]{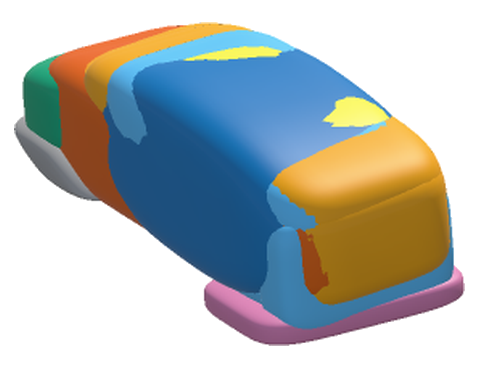} & \includegraphics[width=\gw]{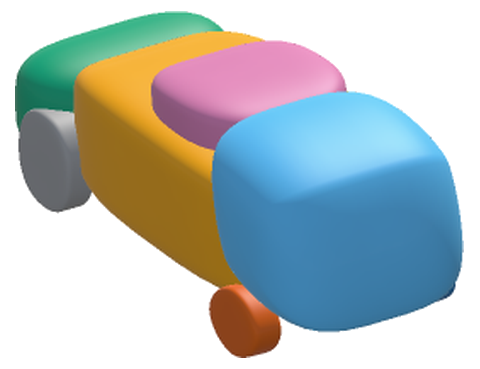} & \includegraphics[width=\gw]{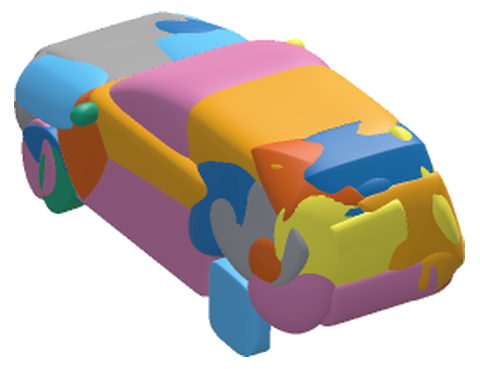} & \includegraphics[width=\gw]{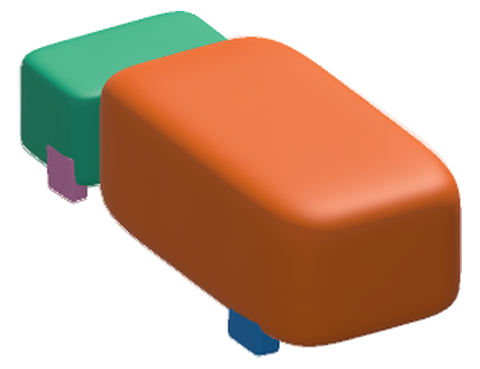} & \includegraphics[width=\gw]{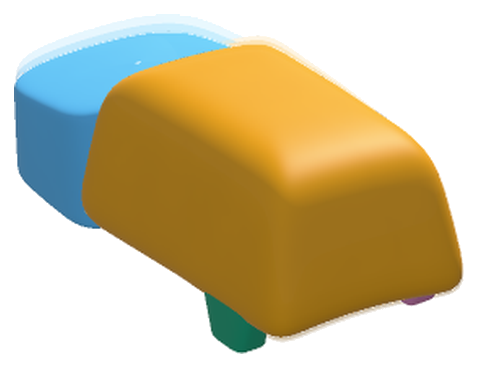} & \includegraphics[width=\gw]{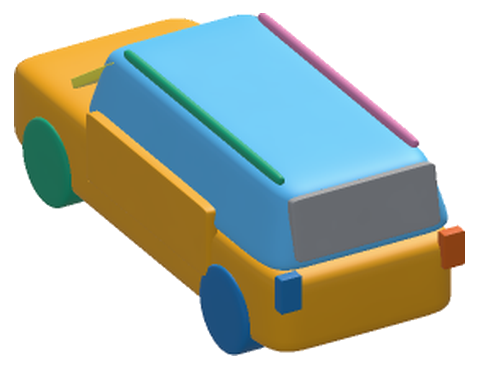} & \includegraphics[width=\gw]{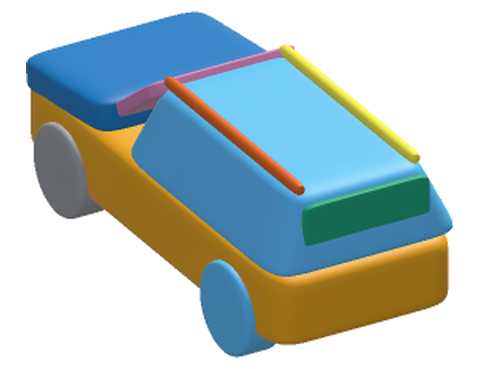} \\[0.4ex]
\scriptsize car & \scriptsize \textcolor{passc}{stands} & \scriptsize \textcolor{passc}{stands} & \scriptsize \textcolor{passc}{stands} & \scriptsize \textcolor{passc}{stands} & \scriptsize \textcolor{passc}{stands} & \scriptsize \textcolor{passc}{stands} & \scriptsize \textcolor{passc}{stands} \\[1.4ex]
\includegraphics[width=\gw]{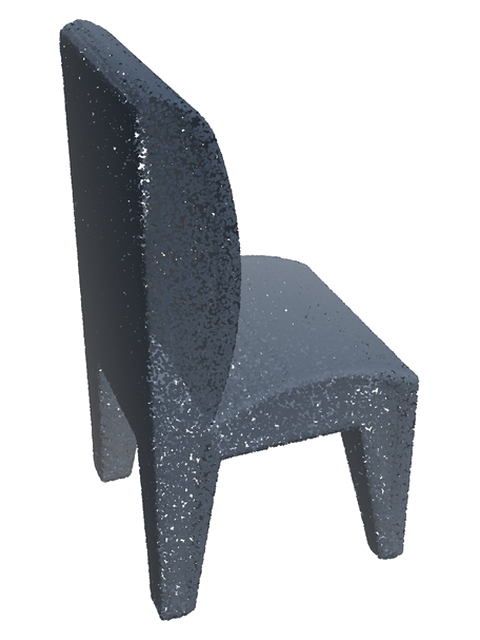} & \includegraphics[width=\gw]{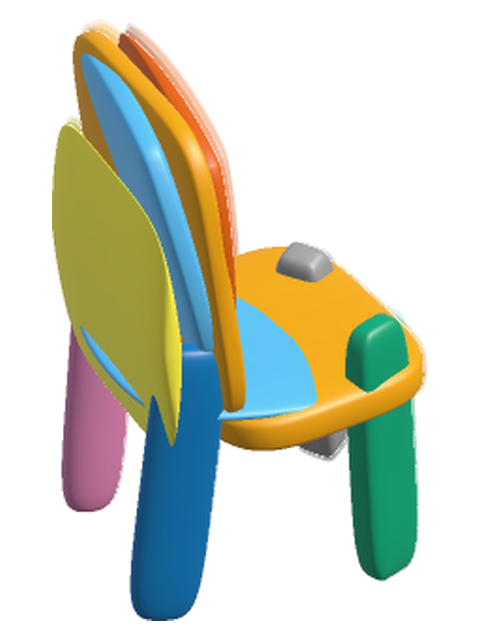} & \includegraphics[width=\gw]{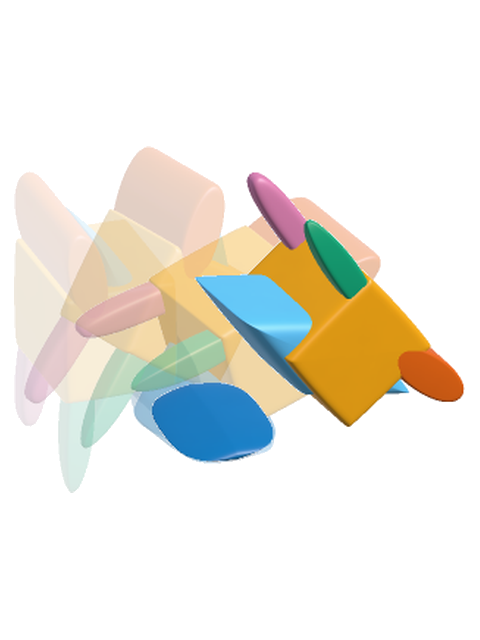} & \includegraphics[width=\gw]{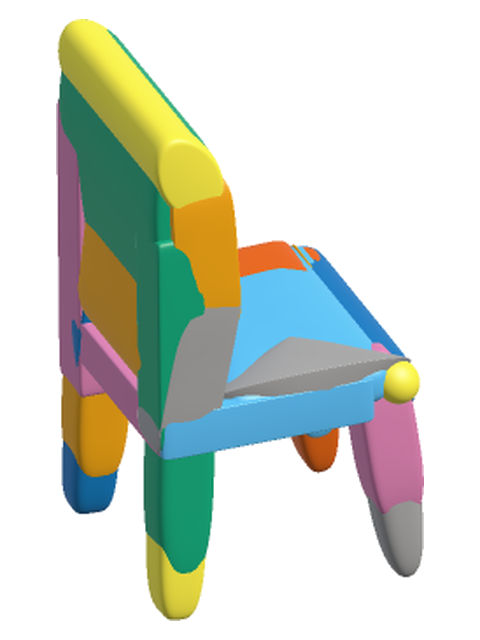} & \includegraphics[width=\gw]{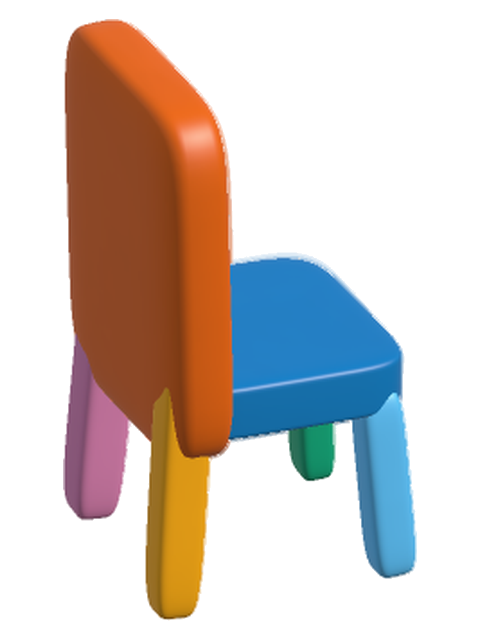} & \includegraphics[width=\gw]{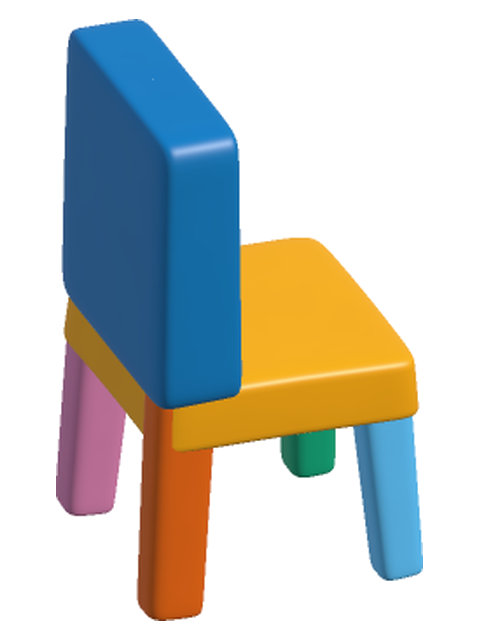} & \includegraphics[width=\gw]{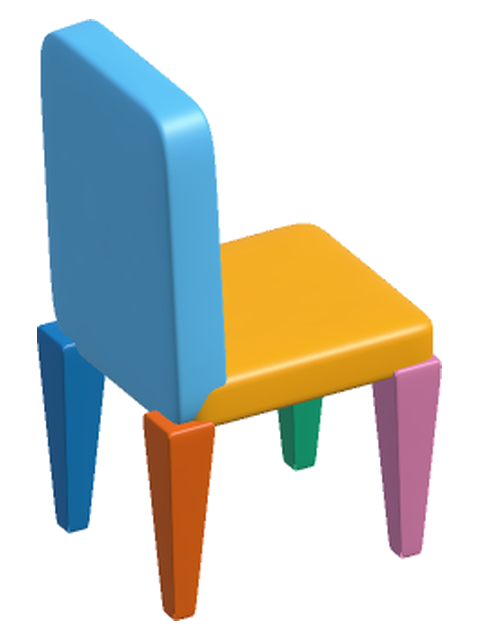} & \includegraphics[width=\gw]{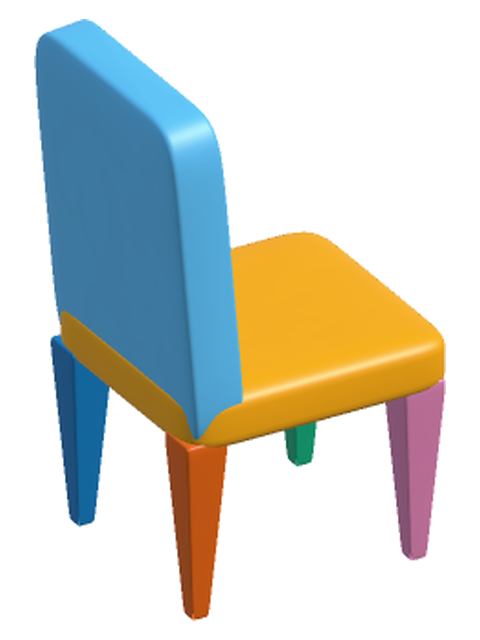} \\[0.4ex]
\scriptsize chair & \scriptsize \textcolor{passc}{stands} & \scriptsize \textcolor{failc}{2 pieces, falls} & \scriptsize \textcolor{passc}{stands} & \scriptsize \textcolor{passc}{stands} & \scriptsize \textcolor{passc}{stands} & \scriptsize \textcolor{passc}{stands} & \scriptsize \textcolor{passc}{stands} \\[1.4ex]
\includegraphics[width=\gw]{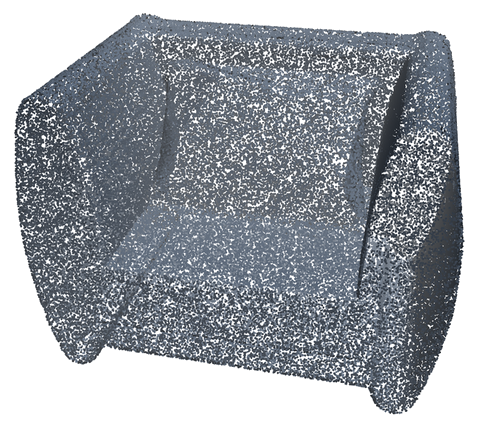} & \includegraphics[width=\gw]{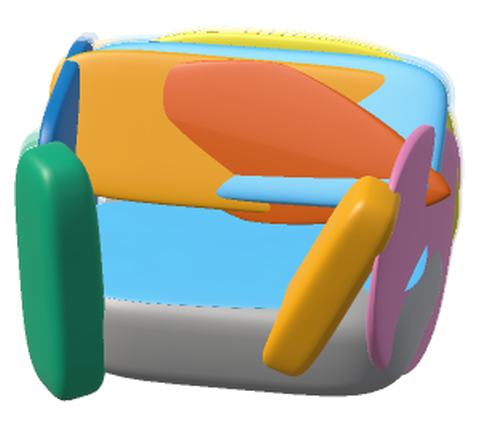} & \includegraphics[width=\gw]{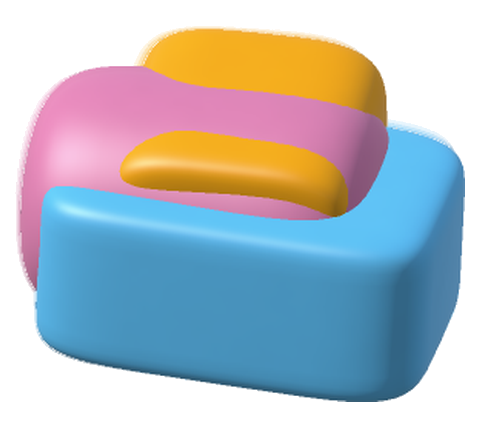} & \includegraphics[width=\gw]{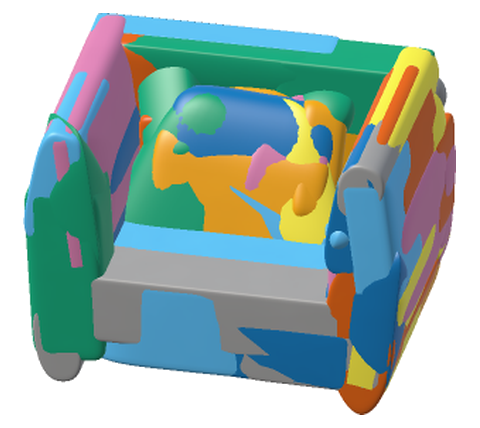} & \includegraphics[width=\gw]{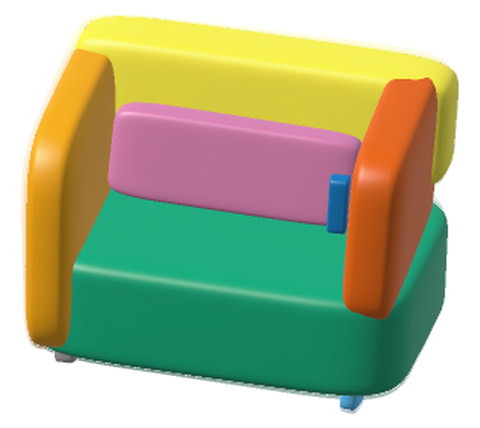} & \includegraphics[width=\gw]{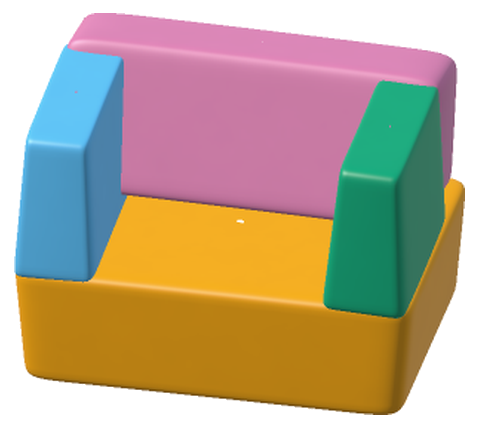} & \includegraphics[width=\gw]{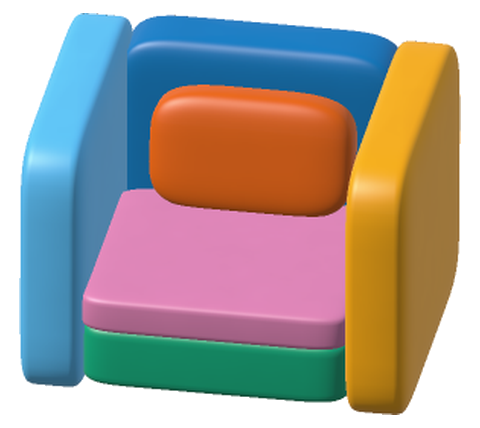} & \includegraphics[width=\gw]{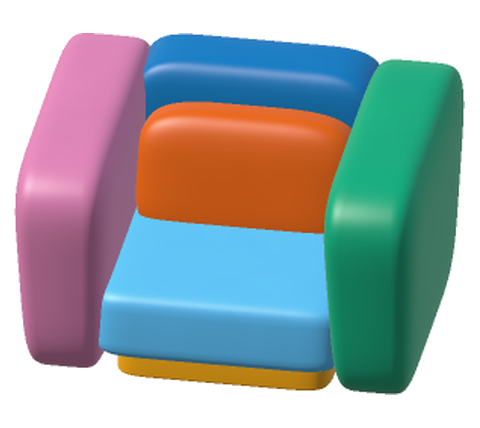} \\[0.4ex]
\scriptsize chair & \scriptsize \textcolor{passc}{stands} & \scriptsize \textcolor{passc}{stands} & \scriptsize \textcolor{passc}{stands} & \scriptsize \textcolor{passc}{stands} & \scriptsize \textcolor{passc}{stands} & \scriptsize \textcolor{passc}{stands} & \scriptsize \textcolor{passc}{stands} \\[1.4ex]
\end{tabular}

  \caption{\textbf{Random objects in simulation} (rows 1--8 of 16). The free-standing objects of \crefrange{fig:s_more_visual_1}{fig:s_more_visual_4} after $4$\,s of the drop test, each with its own materials (the baselines, which have none, with one uniform density). Faint overlays trace motion. Under each result is the physical-check verdict, in green if it passes and in red if it fails.}
  \label{fig:s_more_physics_1}
\end{figure}

\begin{figure}[p]
  \centering
\setlength{\tabcolsep}{0pt}\renewcommand{\arraystretch}{0}
\def\gw{\dimexpr(\textwidth-7\dimexpr0.25em\relax)/8\relax}
\begin{tabular}{@{}c*{7}{@{\hspace{0.25em}}c}@{}}
\parbox[b]{\gw}{\centering\scriptsize\renewcommand{\arraystretch}{1}Input} & \parbox[b]{\gw}{\centering\scriptsize\renewcommand{\arraystretch}{1}SQ} & \parbox[b]{\gw}{\centering\scriptsize\renewcommand{\arraystretch}{1}EMS} & \parbox[b]{\gw}{\centering\scriptsize\renewcommand{\arraystretch}{1}Marching-Prim.} & \parbox[b]{\gw}{\centering\scriptsize\renewcommand{\arraystretch}{1}SuperDec} & \parbox[b]{\gw}{\centering\scriptsize\renewcommand{\arraystretch}{1}SuperFlex} & \parbox[b]{\gw}{\centering\scriptsize\renewcommand{\arraystretch}{1}PhysDec\\(Evolved)} & \parbox[b]{\gw}{\centering\scriptsize\renewcommand{\arraystretch}{1}PhysDec\\(Loop)} \\[0.5ex]
\includegraphics[width=\gw]{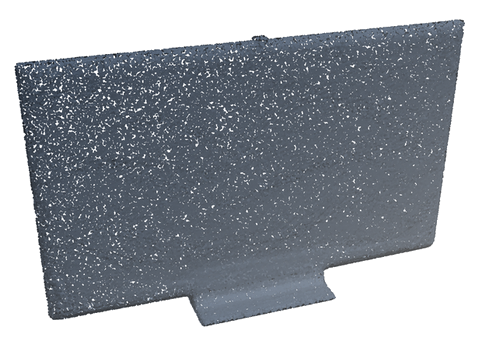} & \includegraphics[width=\gw]{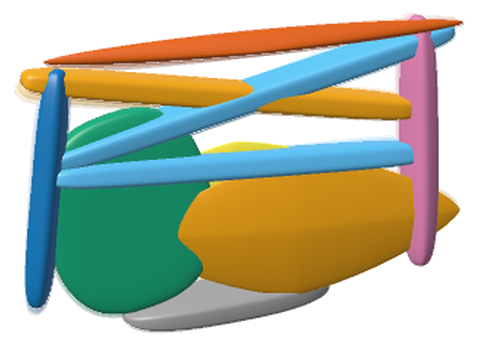} & \includegraphics[width=\gw]{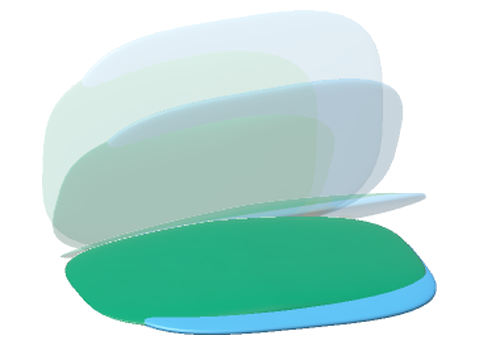} & \includegraphics[width=\gw]{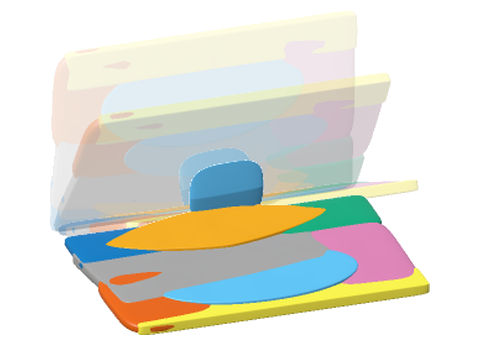} & \includegraphics[width=\gw]{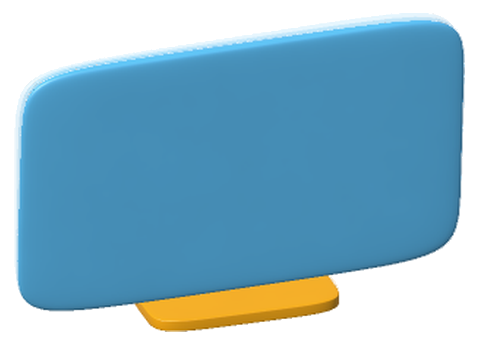} & \includegraphics[width=\gw]{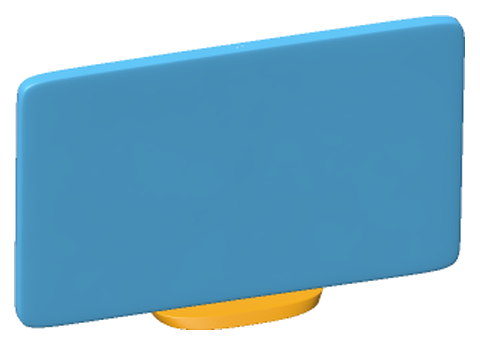} & \includegraphics[width=\gw]{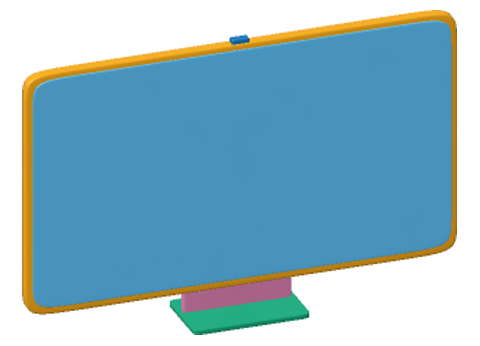} & \includegraphics[width=\gw]{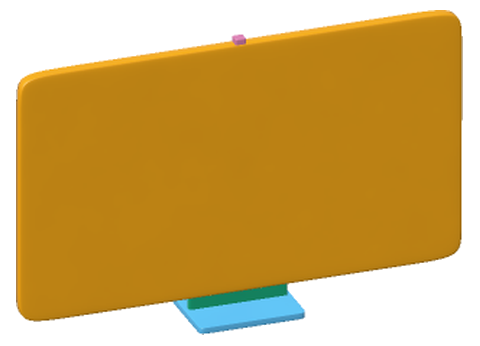} \\[0.4ex]
\scriptsize display & \scriptsize \textcolor{passc}{stands} & \scriptsize \textcolor{failc}{falls} & \scriptsize \textcolor{failc}{falls} & \scriptsize \textcolor{passc}{stands} & \scriptsize \textcolor{passc}{stands} & \scriptsize \textcolor{passc}{stands} & \scriptsize \textcolor{passc}{stands} \\[1.4ex]
\includegraphics[width=\gw]{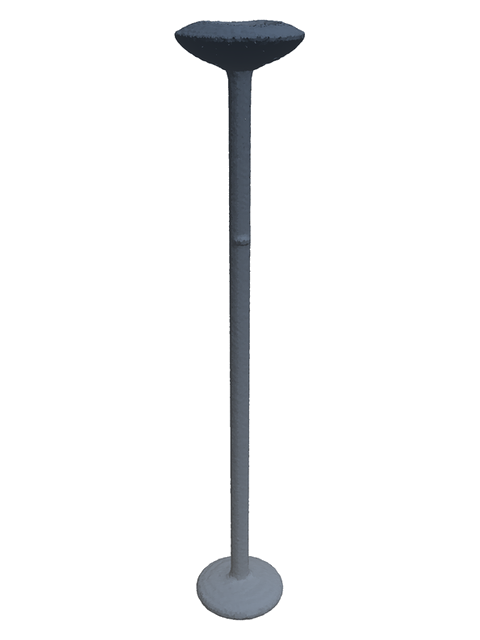} & \includegraphics[width=\gw]{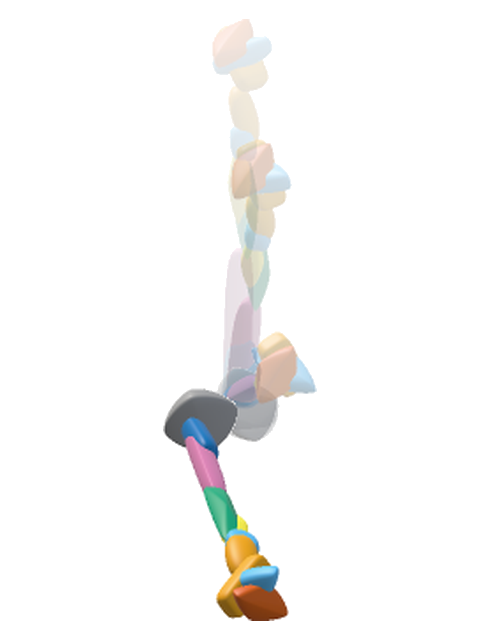} & \includegraphics[width=\gw]{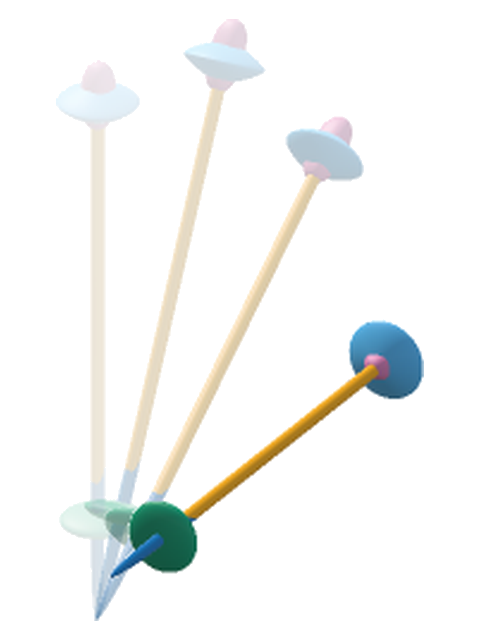} & \includegraphics[width=\gw]{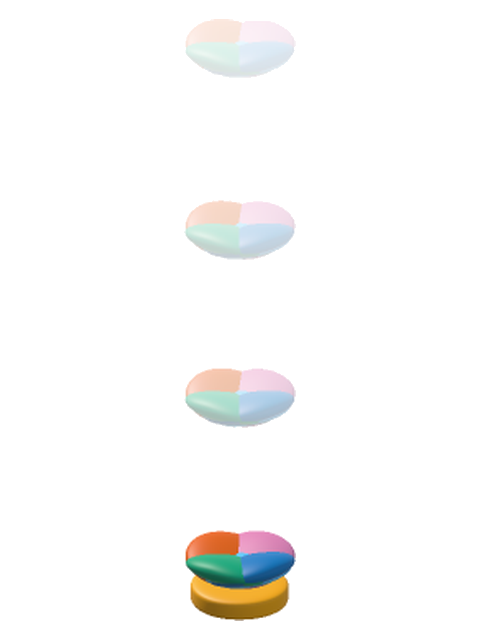} & \includegraphics[width=\gw]{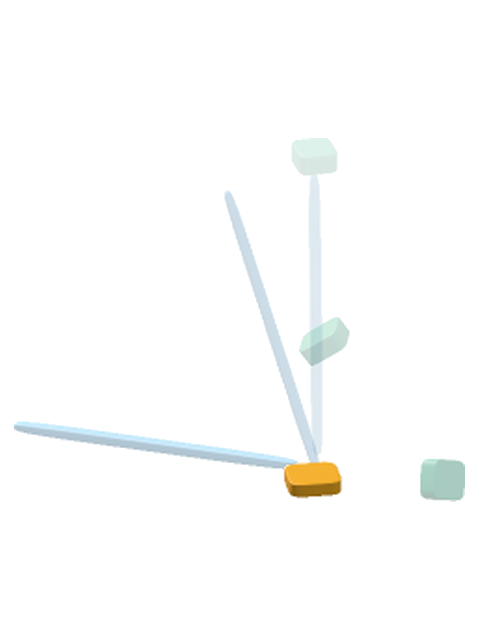} & \includegraphics[width=\gw]{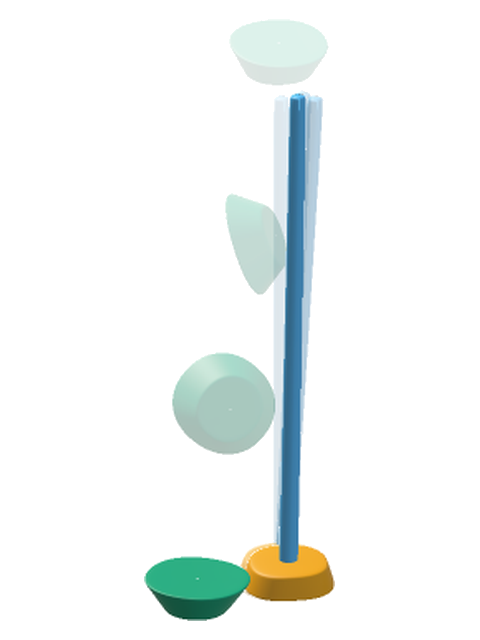} & \includegraphics[width=\gw]{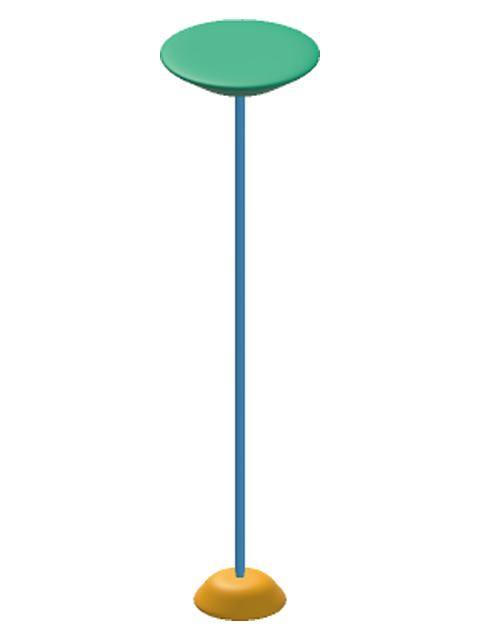} & \includegraphics[width=\gw]{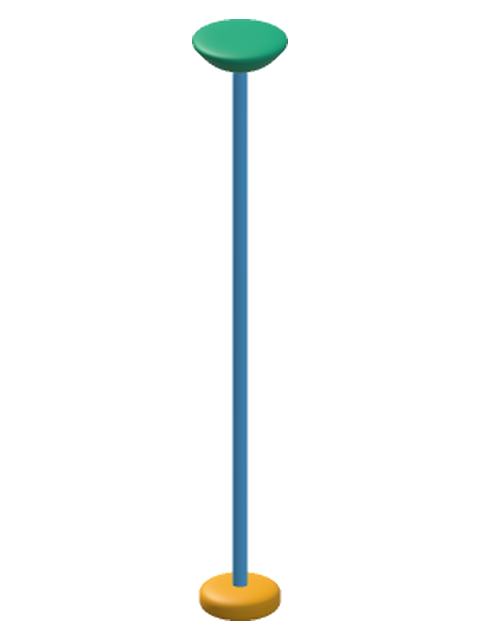} \\[0.4ex]
\scriptsize lamp & \scriptsize \textcolor{failc}{falls} & \scriptsize \textcolor{failc}{falls} & \scriptsize \textcolor{failc}{2 pieces} & \scriptsize \textcolor{failc}{3 pieces, falls} & \scriptsize \textcolor{failc}{2 pieces, falls} & \scriptsize \textcolor{passc}{stands} & \scriptsize \textcolor{passc}{stands} \\[1.4ex]
\includegraphics[width=\gw]{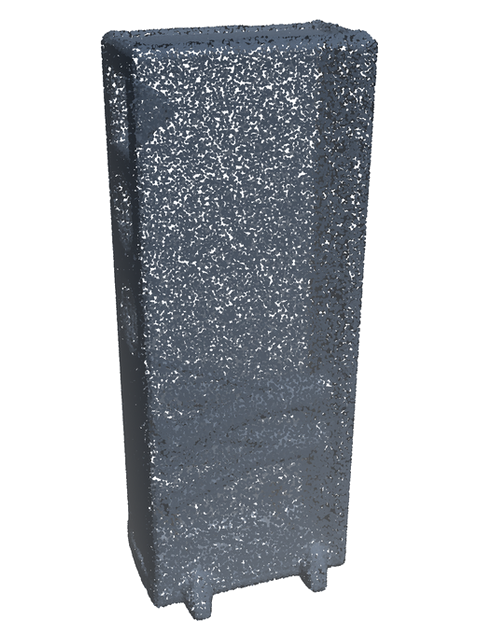} & \includegraphics[width=\gw]{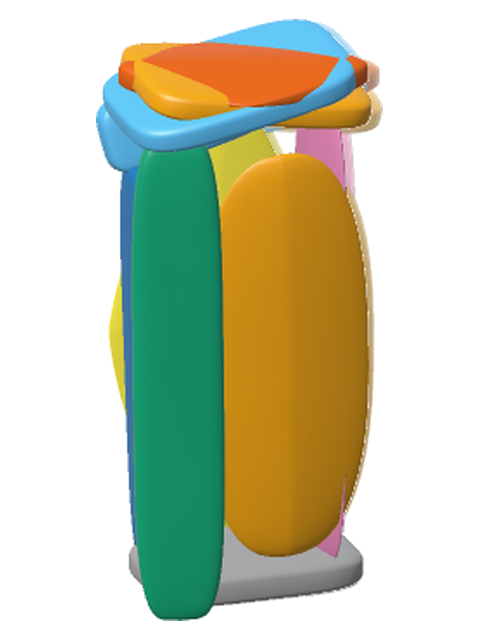} & \includegraphics[width=\gw]{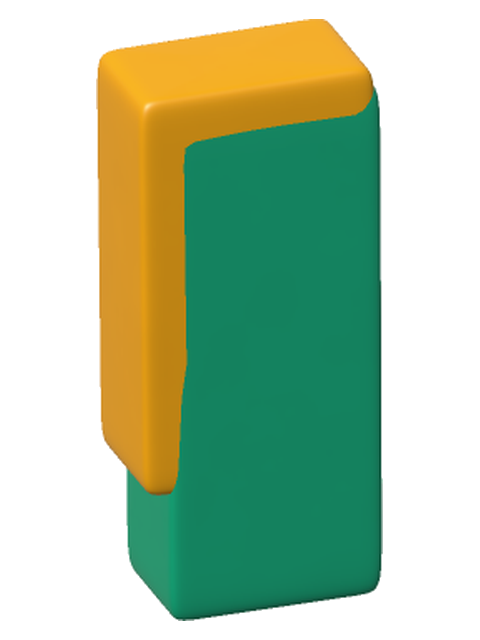} & \includegraphics[width=\gw]{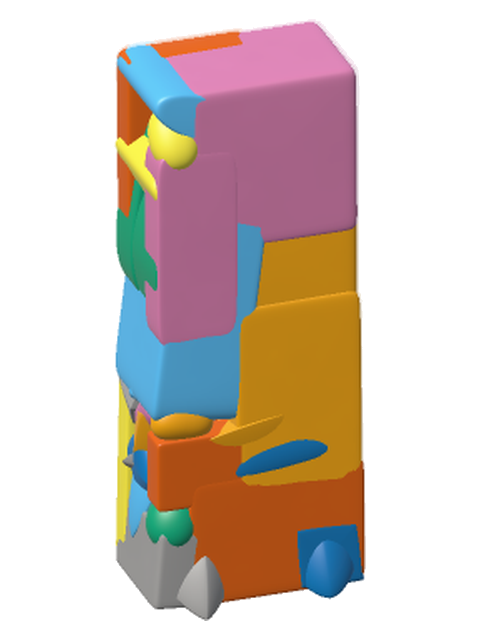} & \includegraphics[width=\gw]{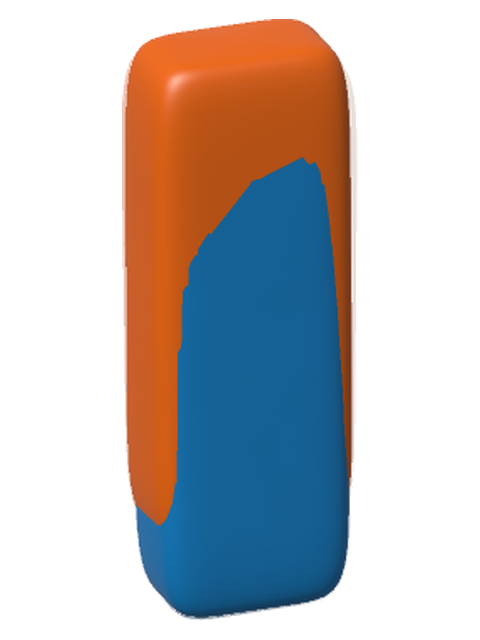} & \includegraphics[width=\gw]{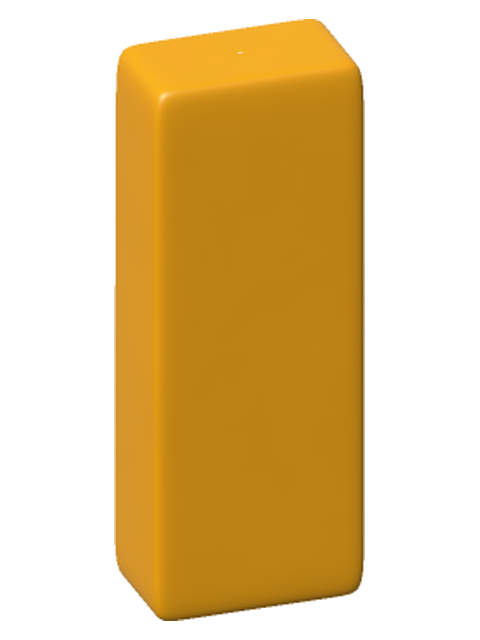} & \includegraphics[width=\gw]{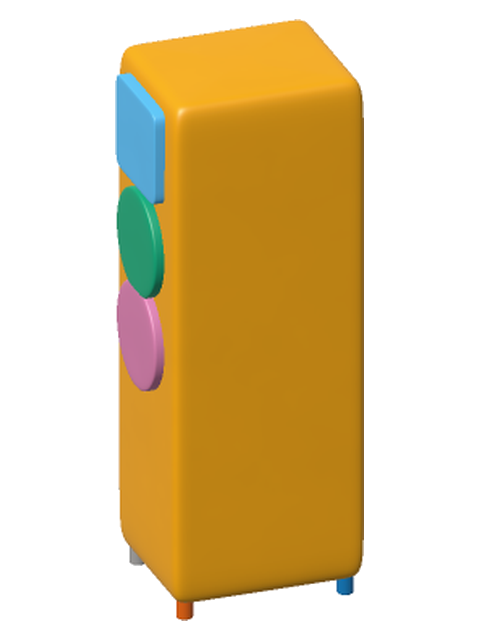} & \includegraphics[width=\gw]{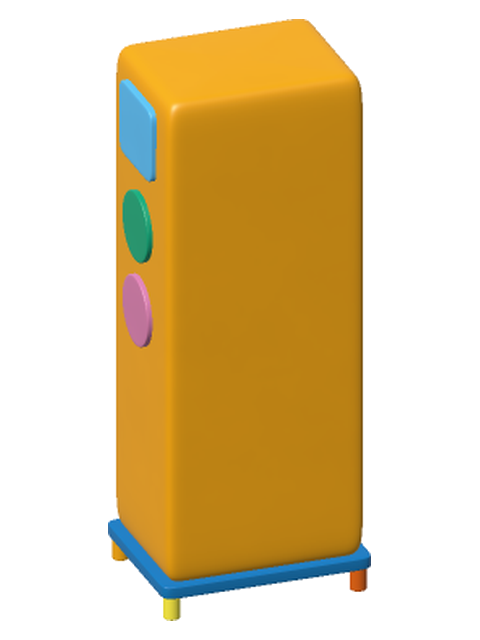} \\[0.4ex]
\scriptsize loudspeaker & \scriptsize \textcolor{passc}{stands} & \scriptsize \textcolor{passc}{stands} & \scriptsize \textcolor{passc}{stands} & \scriptsize \textcolor{passc}{stands} & \scriptsize \textcolor{passc}{stands} & \scriptsize \textcolor{passc}{stands} & \scriptsize \textcolor{passc}{stands} \\[1.4ex]
\includegraphics[width=\gw]{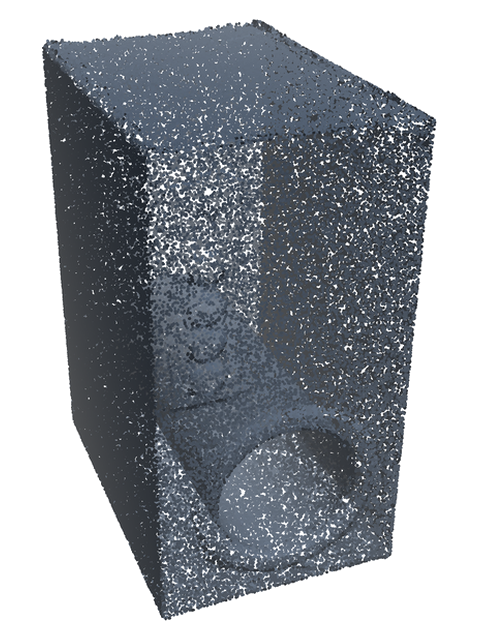} & \includegraphics[width=\gw]{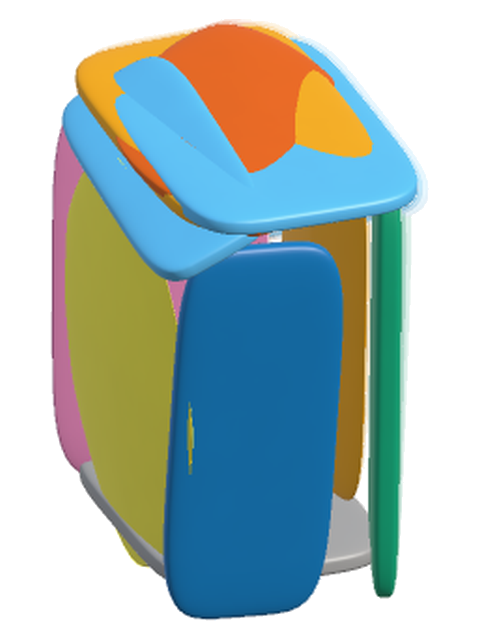} & \includegraphics[width=\gw]{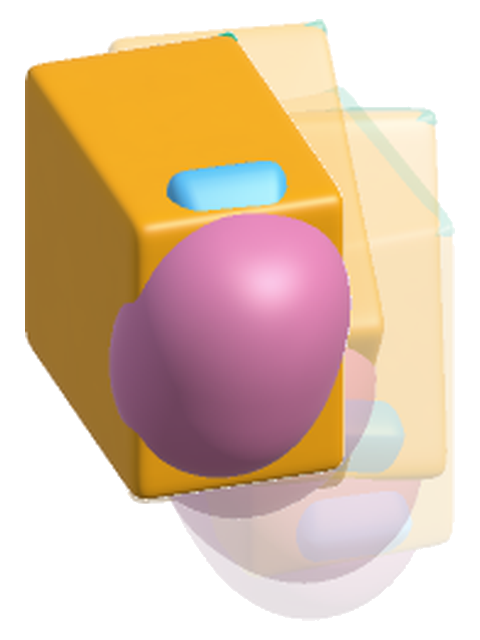} & \includegraphics[width=\gw]{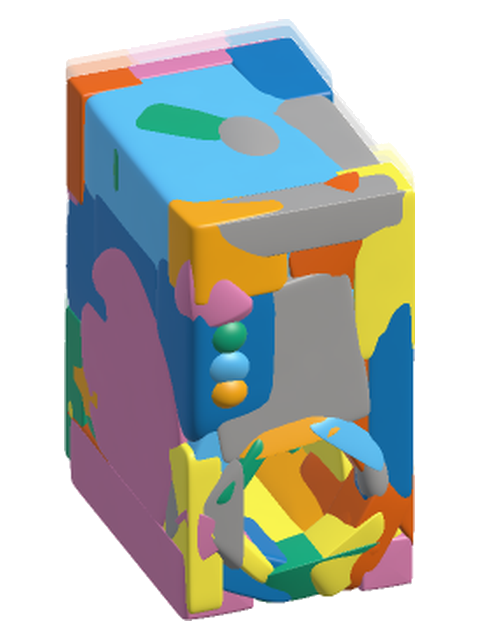} & \includegraphics[width=\gw]{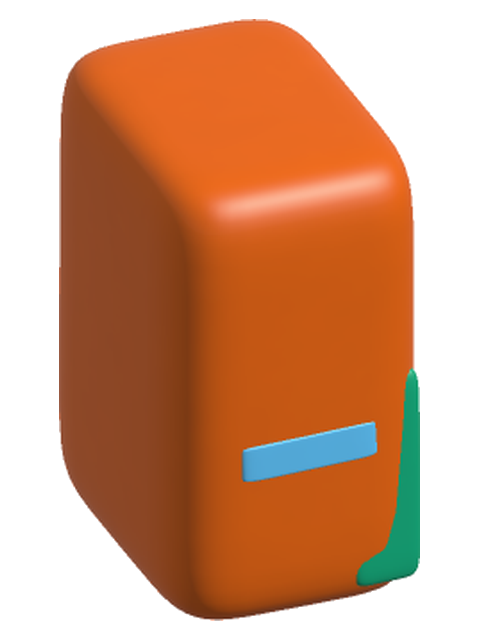} & \includegraphics[width=\gw]{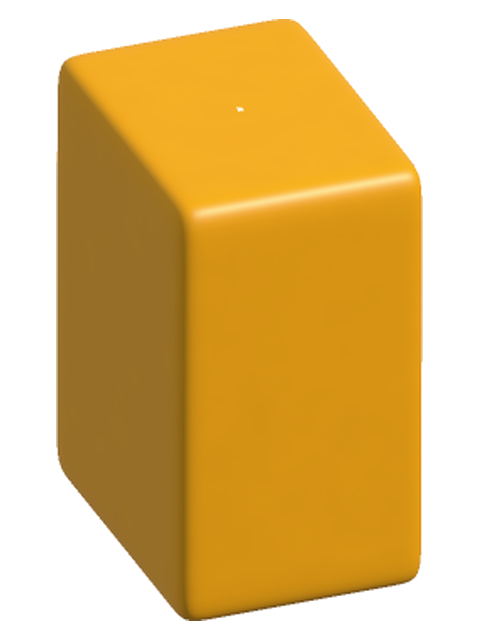} & \includegraphics[width=\gw]{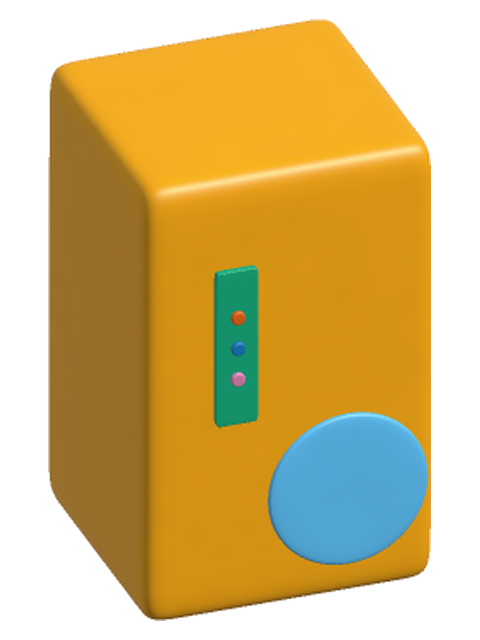} & \includegraphics[width=\gw]{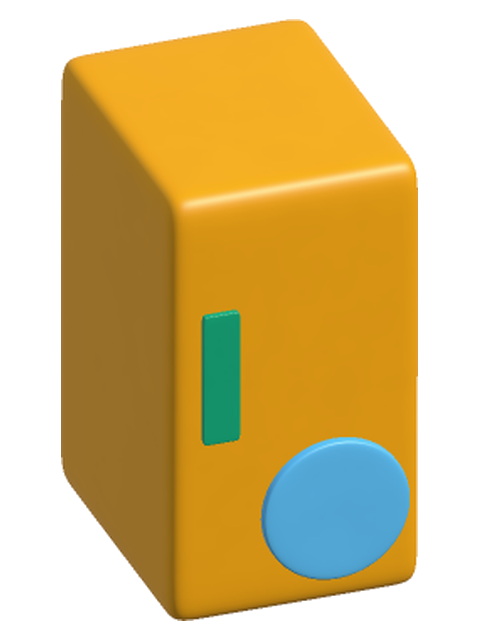} \\[0.4ex]
\scriptsize loudspeaker & \scriptsize \textcolor{passc}{stands} & \scriptsize \textcolor{failc}{falls} & \scriptsize \textcolor{failc}{2 pieces} & \scriptsize \textcolor{passc}{stands} & \scriptsize \textcolor{passc}{stands} & \scriptsize \textcolor{passc}{stands} & \scriptsize \textcolor{passc}{stands} \\[1.4ex]
\includegraphics[width=\gw]{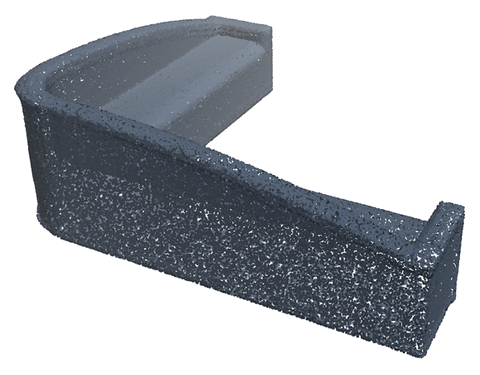} & \includegraphics[width=\gw]{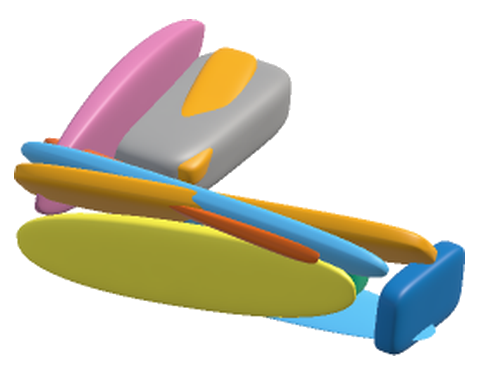} & \includegraphics[width=\gw]{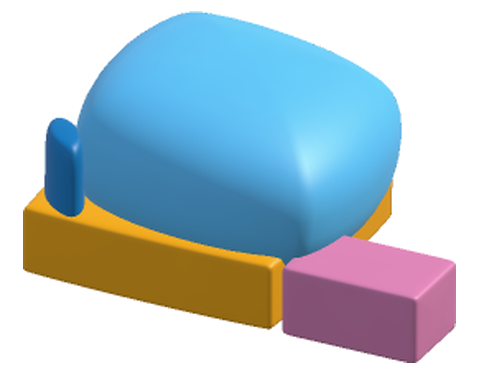} & \includegraphics[width=\gw]{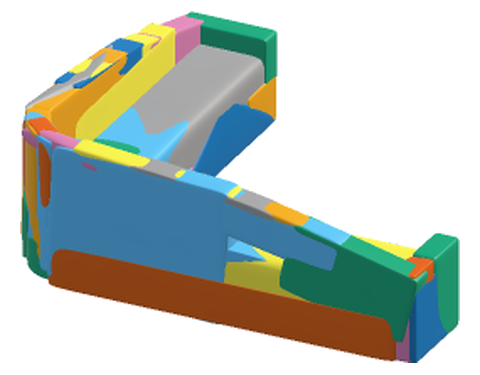} & \includegraphics[width=\gw]{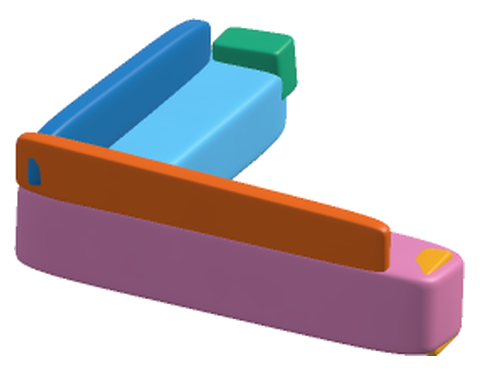} & \includegraphics[width=\gw]{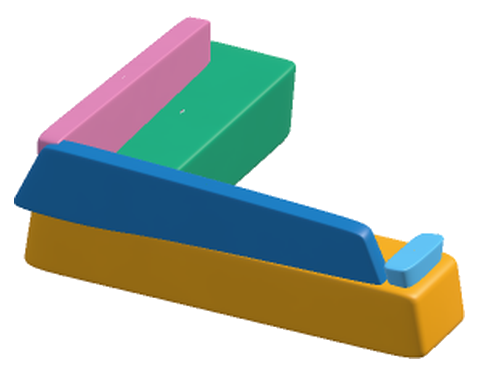} & \includegraphics[width=\gw]{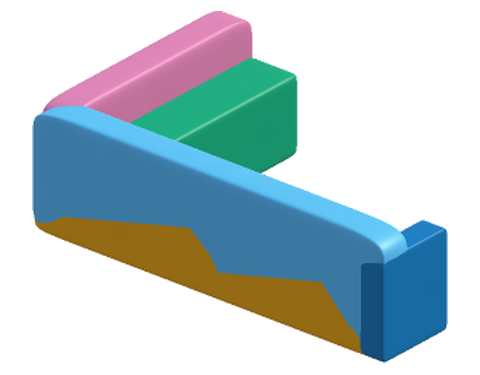} & \includegraphics[width=\gw]{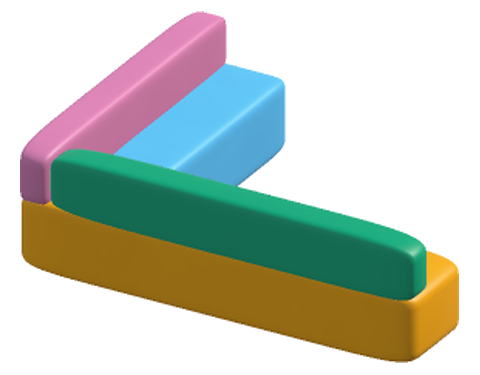} \\[0.4ex]
\scriptsize sofa & \scriptsize \textcolor{passc}{stands} & \scriptsize \textcolor{passc}{stands} & \scriptsize \textcolor{passc}{stands} & \scriptsize \textcolor{passc}{stands} & \scriptsize \textcolor{passc}{stands} & \scriptsize \textcolor{passc}{stands} & \scriptsize \textcolor{passc}{stands} \\[1.4ex]
\includegraphics[width=\gw]{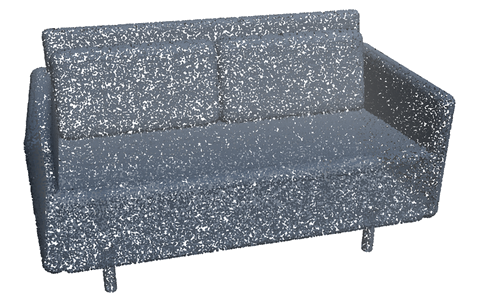} & \includegraphics[width=\gw]{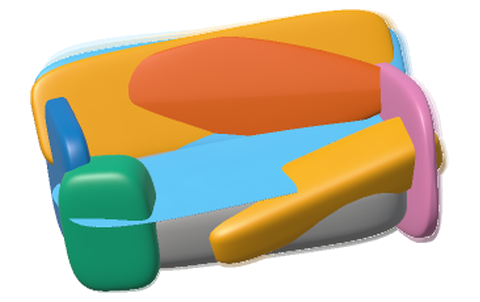} & \includegraphics[width=\gw]{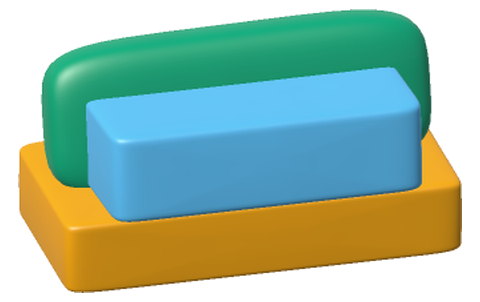} & \includegraphics[width=\gw]{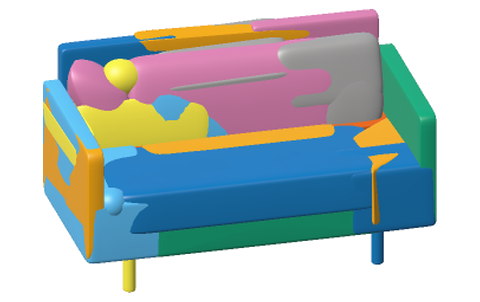} & \includegraphics[width=\gw]{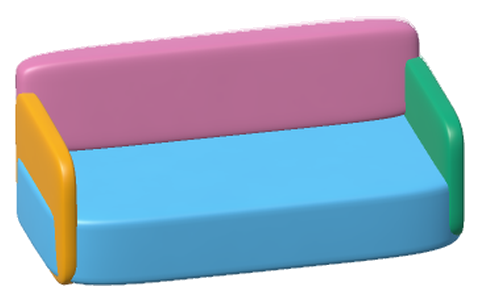} & \includegraphics[width=\gw]{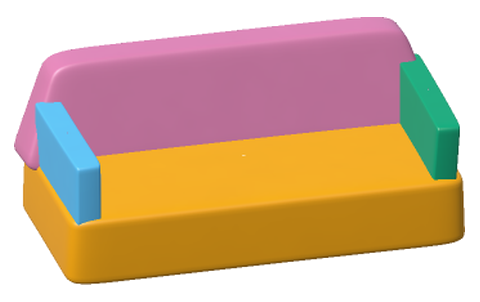} & \includegraphics[width=\gw]{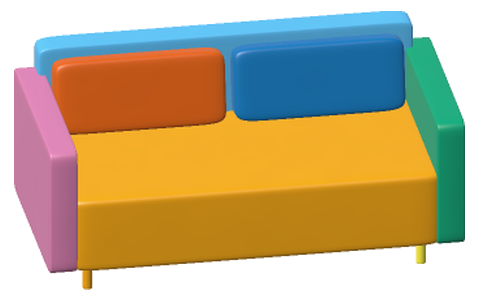} & \includegraphics[width=\gw]{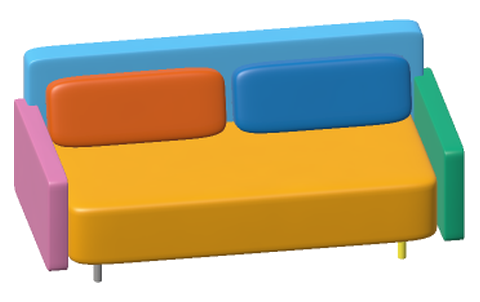} \\[0.4ex]
\scriptsize sofa & \scriptsize \textcolor{passc}{stands} & \scriptsize \textcolor{passc}{stands} & \scriptsize \textcolor{passc}{stands} & \scriptsize \textcolor{passc}{stands} & \scriptsize \textcolor{passc}{stands} & \scriptsize \textcolor{passc}{stands} & \scriptsize \textcolor{passc}{stands} \\[1.4ex]
\includegraphics[width=\gw]{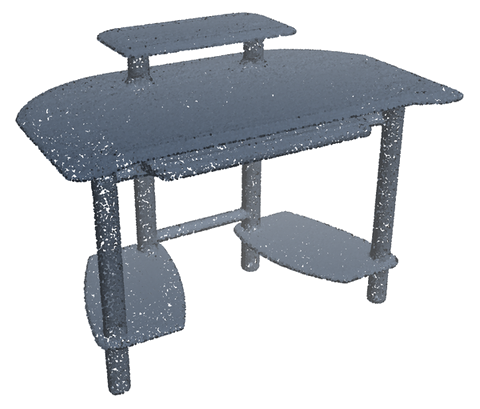} & \includegraphics[width=\gw]{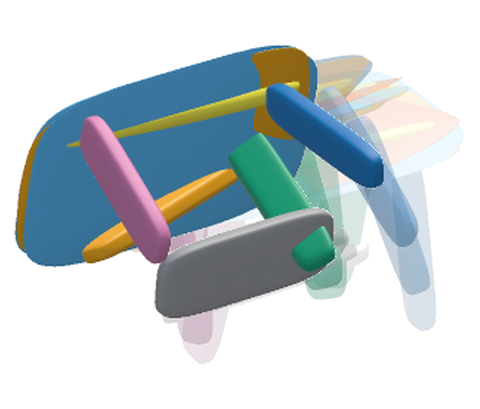} & \includegraphics[width=\gw]{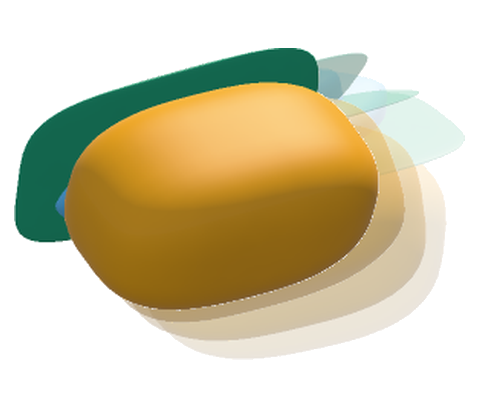} & \includegraphics[width=\gw]{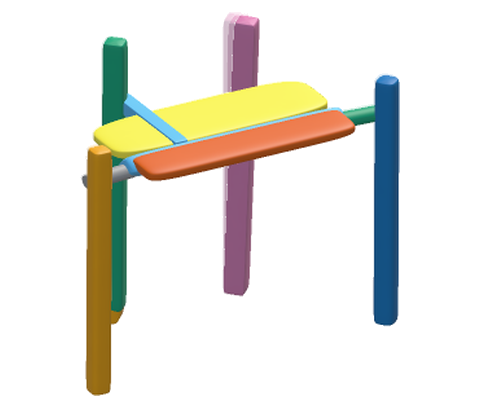} & \includegraphics[width=\gw]{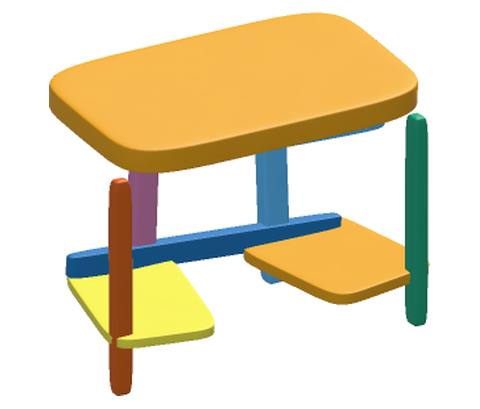} & \includegraphics[width=\gw]{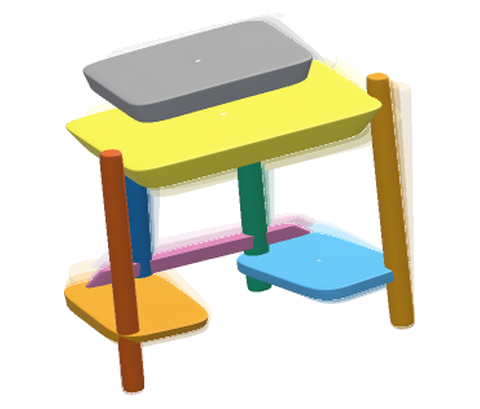} & \includegraphics[width=\gw]{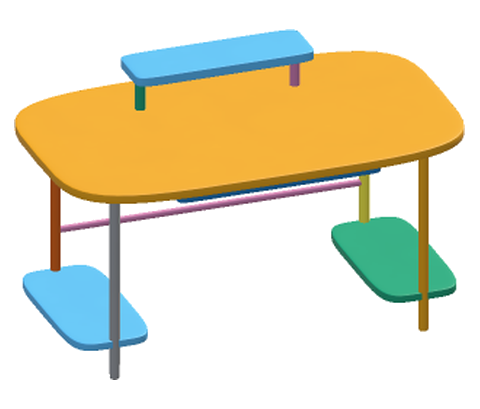} & \includegraphics[width=\gw]{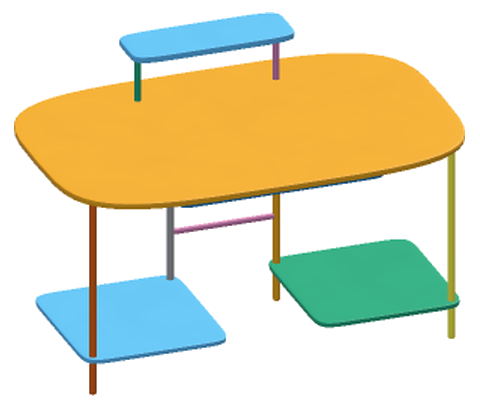} \\[0.4ex]
\scriptsize table & \scriptsize \textcolor{failc}{falls} & \scriptsize \textcolor{failc}{falls} & \scriptsize \textcolor{failc}{2 pieces} & \scriptsize \textcolor{passc}{stands} & \scriptsize \textcolor{passc}{stands} & \scriptsize \textcolor{passc}{stands} & \scriptsize \textcolor{passc}{stands} \\[1.4ex]
\includegraphics[width=\gw]{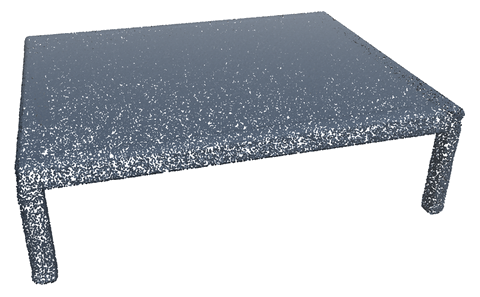} & \includegraphics[width=\gw]{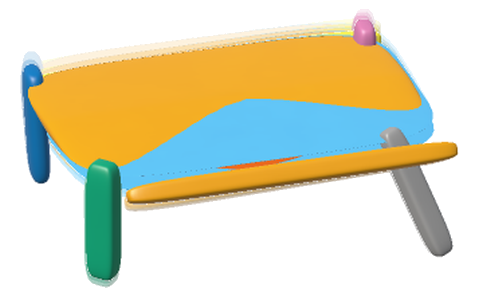} & \includegraphics[width=\gw]{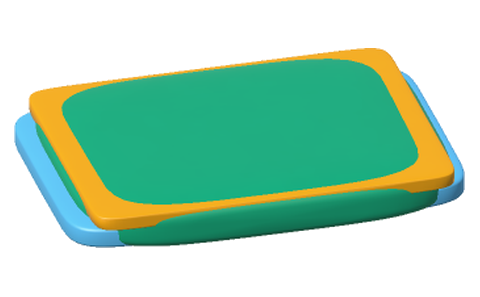} & \includegraphics[width=\gw]{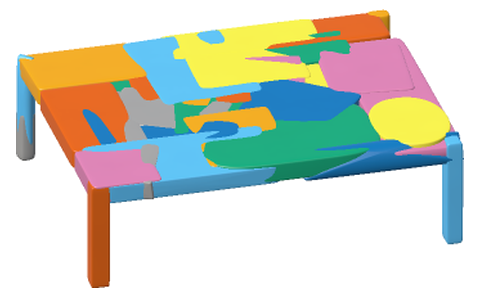} & \includegraphics[width=\gw]{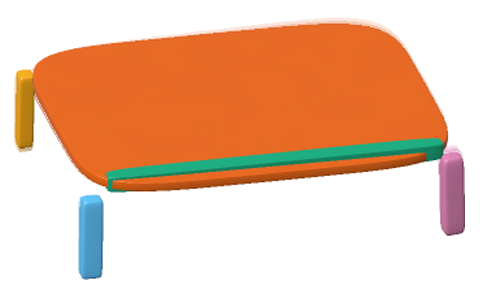} & \includegraphics[width=\gw]{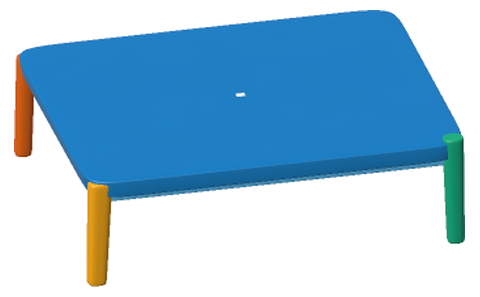} & \includegraphics[width=\gw]{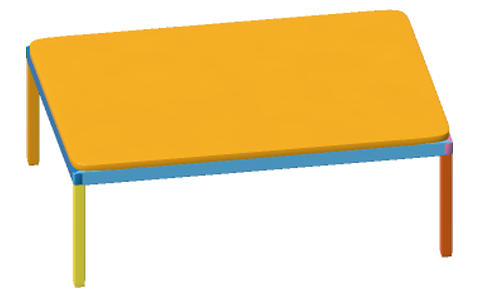} & \includegraphics[width=\gw]{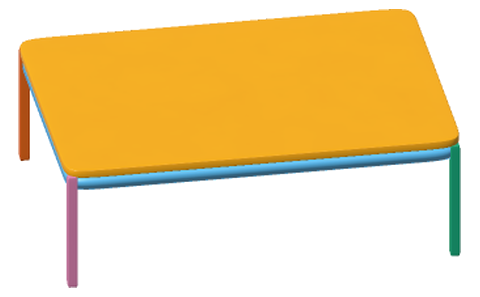} \\[0.4ex]
\scriptsize table & \scriptsize \textcolor{passc}{stands} & \scriptsize \textcolor{passc}{stands} & \scriptsize \textcolor{passc}{stands} & \scriptsize \textcolor{passc}{stands} & \scriptsize \textcolor{passc}{stands} & \scriptsize \textcolor{passc}{stands} & \scriptsize \textcolor{passc}{stands} \\[1.4ex]
\end{tabular}

  \caption{\textbf{Random objects in simulation} (rows 9--16 of 16). The free-standing objects of \crefrange{fig:s_more_visual_1}{fig:s_more_visual_4} after $4$\,s of the drop test, each with its own materials (the baselines, which have none, with one uniform density). Faint overlays trace motion. Under each result is the physical-check verdict, in green if it passes and in red if it fails.}
  \label{fig:s_more_physics_2}
\end{figure}

\Cref{fig:s_more_physics_1,fig:s_more_physics_2} drop the 16 free-standing objects among them for $4$\,s, which is the test of their physical check.
PhysDec-Loop and PhysDec-Evolved pass on all 16.
SuperFlex passes on 15, SuperDec on 14, SQ on 13, Marching-Primitives on 11 and EMS on 10.
The other ten objects are judged by checks that are not a drop on the floor: resting on a side, contact with a wall, or balance when hung (\cref{sec:s_evaluation}).
A drop would not show what was judged, so they are not simulated here.


\clearpage
\section{Additional manipulation results}
\label{sec:s_manipulation}

We evaluate pick-and-place manipulation in MuJoCo using a KUKA LBR iiwa 14 with an Allegro hand~\cite{menagerie2022github}.
We compare with SuperDec and SuperFlex using the same grasp planner on each method's superquadrics~\cite{vezzani2017grasping}, with object masses matched across methods.
The robot lifts, carries and places each object; success requires moving all its parts and leaving it stable, without disturbing surrounding objects in scenes.
A reconstruction that falls apart or topples when released has already failed, but the robot still plans on its pieces where they lie.

\begin{figure}[H]
  \def\robotlabel{fig:s_robot}
  \def\robotplanelabel{fig:s_robot_plane}
  \centering
  \begin{minipage}[t]{0.485\textwidth}
    \vspace{0pt}
    
  \end{minipage}\hfill
  \begin{minipage}[t]{0.485\textwidth}
    \vspace{0pt}
    
  \end{minipage}
\end{figure}

\paragraph{Objects in reconstructed scenes.}
The bin example uses a $10$\,cm lift. Our reconstruction provides handles for grasping. SuperDec's reconstruction comes apart during the initial release; the robot plans on the pieces where they lie and lifts one of them, $36\%$ of the mass.

The chair example uses a $26$\,cm lift while grasping the backrest. The seat contacts the table in all three reconstructed scenes. SuperDec's table consists of four disconnected pieces and collapses, while SuperFlex's chair displaces a neighboring chair. The illustrated pull was selected after testing multiple motions.

\paragraph{Objects on a ground plane.}
Each reconstruction is released for $2$\,s before the robot lifts it and places it $0.8$\,m away. Our lamp and chair remain connected and are successfully placed in all five tested carry directions. SuperDec's lamp and chair contain three and four disconnected pieces, respectively. All four baseline reconstructions fall apart during release, and the robot lifts only $4$--$66\%$ of their total mass across the tested directions.

\clearpage
\twocolumn
{\small
\bibliographystyle{ieeenat_fullname}
\bibliography{main}
}
\end{document}